\documentclass[10pt,twocolumn,letterpaper]{article}

\usepackage[pagenumbers]{wacv}              

\usepackage{graphicx}
\usepackage{amsmath}
\usepackage{amssymb}
\usepackage{booktabs}

\usepackage[accsupp]{axessibility} 

\usepackage[pagebackref,breaklinks,colorlinks]{hyperref}

\usepackage[capitalize]{cleveref}
\crefname{section}{Sec.}{Secs.}
\Crefname{section}{Section}{Sections}
\Crefname{table}{Table}{Tables}
\crefname{table}{Tab.}{Tabs.}

\usepackage{microtype}      
\usepackage[T1]{fontenc}
\usepackage{romannum}
\usepackage{multicol}
\usepackage{adjustbox}
\usepackage{multirow}
\usepackage{tabularx}
\usepackage{makecell}
\usepackage{float}
\usepackage{textcomp}
\usepackage{subcaption}

\usepackage{enumitem}
\usepackage[utf8]{inputenc}

\newcommand{\dataset}{ConeQuest}
\newcommand{\hyphen}[1]{\kern-.#1em }

\def\wacvPaperID{772} 
\def\confName{WACV}
\def\confYear{2024}

\begin{document}

\title{ConeQuest: A Benchmark for Cone Segmentation on Mars}

\author{
Mirali Purohit \quad \quad Jacob Adler \quad \quad Hannah Kerner
\\
Arizona State University
\\
\small{\texttt{\{mpurohi3, jbadler2, hkerner\}@asu.edu}}
}


\maketitle

\begin{abstract}
   Over the years, space scientists have collected terabytes of Mars data from satellites and rovers. One important set of features identified in Mars orbital images is pitted cones, which are interpreted to be mud volcanoes believed to form in regions that were once saturated in water (i.e., a lake or ocean). Identifying pitted cones globally on Mars would be of great importance, but expert geologists are unable to sort through the massive orbital image archives to identify all examples. However, this task is well suited for computer vision. Although several computer vision datasets exist for various Mars-related tasks, there is currently no open-source dataset available for cone detection/segmentation. Furthermore, previous studies trained models using data from a single region, which limits their applicability for global detection and mapping. Motivated by this, we introduce \dataset, the first expert-annotated public dataset to identify cones on Mars. \dataset{} consists of >$13k$ samples from 3 different regions of Mars. We propose two benchmark tasks using \dataset{}: (i) Spatial Generalization and (ii) Cone-size Generalization. We finetune and evaluate widely-used segmentation models on both benchmark tasks. Results indicate that cone segmentation is a challenging open problem not solved by existing segmentation models, which achieve an average IoU of $52.52\%$ and $42.55\%$ on in-distribution data for tasks (i) and (ii), respectively. We believe this new benchmark dataset will facilitate the development of more accurate and robust models for cone segmentation. Data and code are available at {\small \url{https://github.com/kerner-lab/ConeQuest}}.

\end{abstract}


\section{Introduction}
\label{sec:introduction}

\begin{figure}
  \centering

  \begin{subfigure}{\columnwidth}
    \centering
    \includegraphics[width=\linewidth]{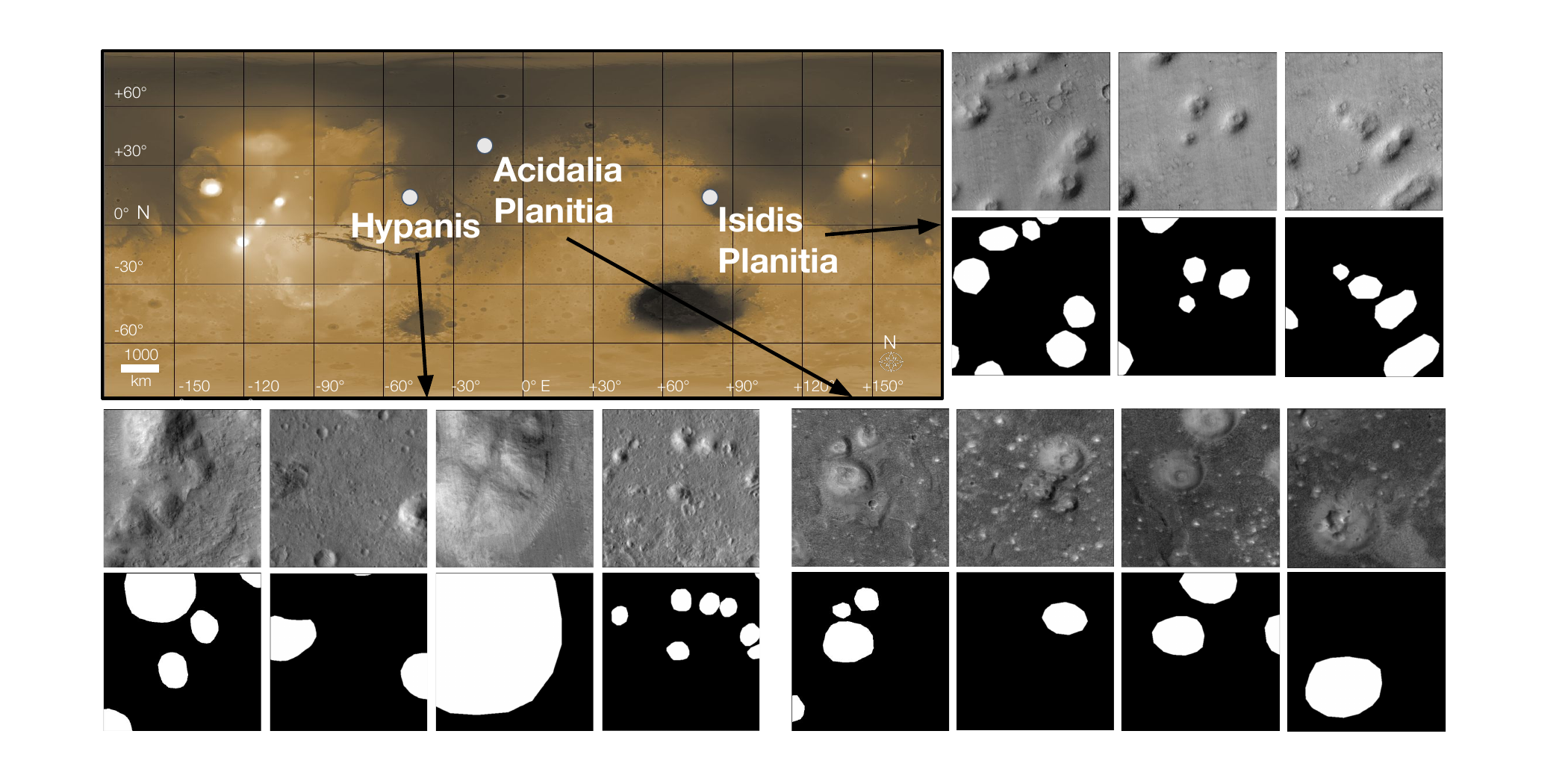}
    \caption{Spatial Generalization benchmark (BM-1)}
    \label{fig:teaser_bm_1}
  \end{subfigure}

  \medskip

  \begin{subfigure}{\columnwidth}
    \centering
    \includegraphics[width=\linewidth]{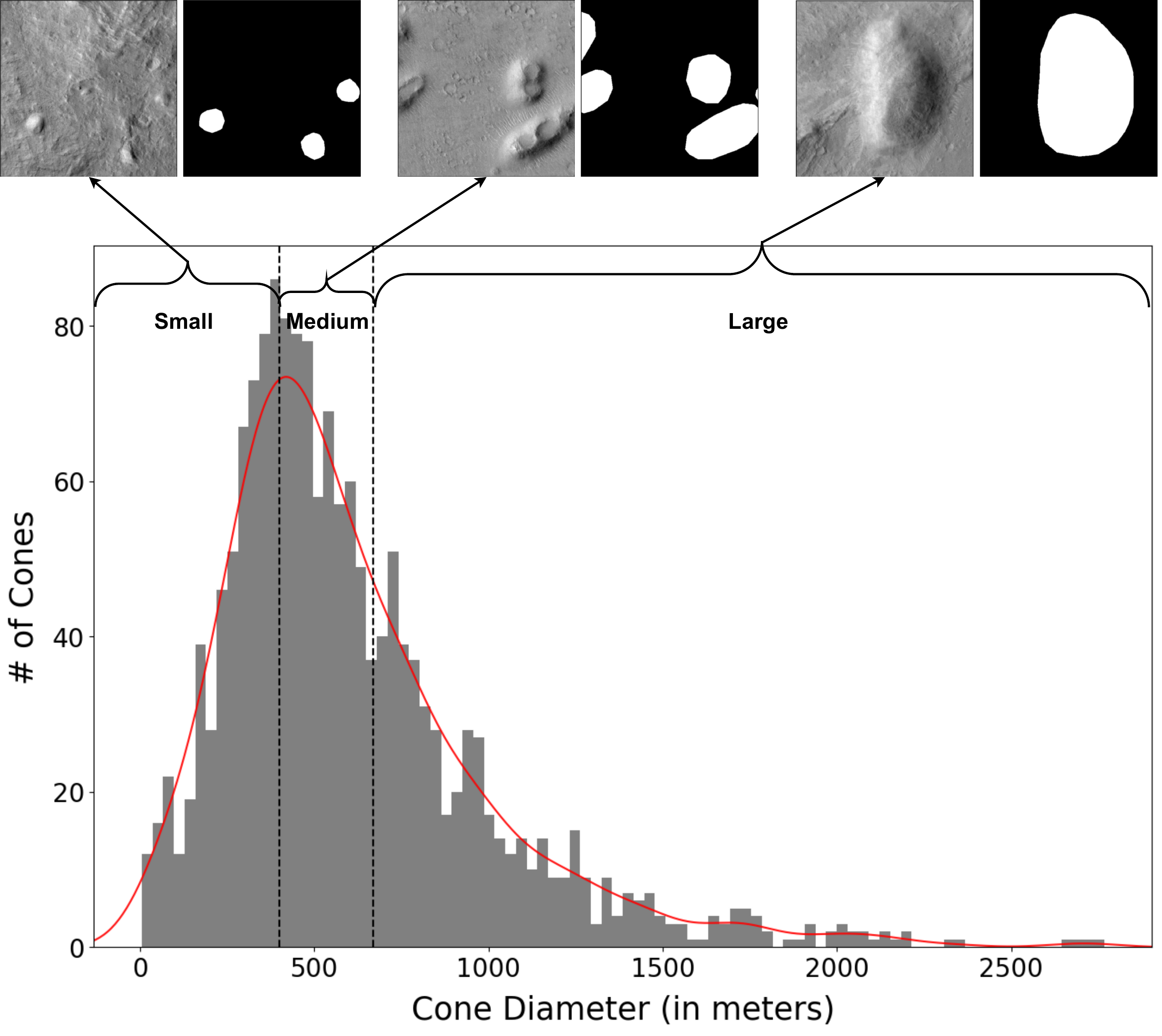}
    \caption{Cone-size Generalization benchmark (BM-2)}
    \label{fig:teaser_bm_2}
  \end{subfigure}

  \caption{Illustrative examples corresponding to both benchmarks}
  \label{fig:teaser_figure}
\end{figure}


With the advancement of camera technology and as data downlinking rates have improved, the entirety of the Martian surface has been imaged by multiple instruments that have acquired terabytes of data. Analyzing the data returned by spacecraft instruments is the only current way to gain insights into Mars surface processes as humans have yet to land on a planetary body other than the Moon. Finding evidence of past water is a top goal of the Mars science community \cite{bada2009seeking, board2012vision, national2022origins}. Mapping the presence of water-related features improves scientists' understanding of the planet's past climate and its potential to have oceans or localized habitable environments, and helps identify key sites to send a future rover or human mission. One important set of features on the martian surface is pitted \textit{cones} that are believed to be mud volcanoes \cite{skinner2009martian, oehler2010evidence, hemmi2018high, adler2022regional, brovz2019subsurface, davis1995curvilinear, farrand2005pitted, ivanov2014mud, mcgowan2011utopia}. These rounded mound-shaped features range in diameter from meter-sized to a few kilometers and are believed to form in regions that were once saturated in water (i.e., lakes or oceans). Three billion years ago, this water receded underground and turned to ice, or was buried by other deposits, only to emerge later as a mud volcano when tectonic compression, impact events, and other processes squeezed buried mud back to the surface along cracks and faults \cite{skinner2009martian}.


Manually reviewing the growing volumes of high-resolution remote sensing data of Mars to identify and characterize cones is prohibitively time-consuming and labor-intensive. Hence, an automated process is necessary to detect/segment and analyze the characteristics of cones. Machine learning methods provide a promising solution for developing an automation pipeline. However, training these models requires a substantial amount of data. Despite the availability of numerous Mars-related datasets, there is currently no open-source dataset for cone detection/segmentation.


Our study introduces a novel dataset called \dataset{}, which has been annotated by experts enabling ML methods to identify cones on Mars. This dataset comprises more than $13k$ samples from the Isidis Planitia, Acidalia Planitia, and Hypanis regions of the martian surface. Additionally, we provide metadata for every data sample, such as latitude-longitude, area, and bounding box. We formulated cone detection as a binary segmentation problem and developed two benchmarks (BMs) tasks based on \dataset{}: Spatial Generalization (BM-1) and Cone-size Generalization (BM-2). Figure \ref{fig:teaser_figure} shows the overview of both BMs. The evaluation of spatial generalization examines the model's performance across different regions, in which test data is from a region not used for training. Similarly, cone-size generalization assesses the model's performance based on variations in cone size within the data, involving training on specific size ranges and combining different size ranges.


We conduct training on commonly used segmentation-based models which include U-Net \cite{ronneberger2015u}, FPN \cite{lin2017feature}, DeepLab \cite{chen2017rethinking}, and MA-Net \cite{fan2020ma} for both BMs. We evaluate the model on in-distribution (id) as well as out-of-distribution (ood) data (i.e., data from region/size model has seen and unseen during training, respectively) to assess the efficiency of models. The average IoU for the id category is $52.52\%$ and $42.55\%$ for BM-1 and BM-2, respectively. Additionally, the average IoU on ood data is $15.04\%$ (BM-1) and $26.92\%$ (BM-2). The results obtained from the evaluation of ood data suggest that the model struggles to generalize on ood data and the evaluation of id data indicates that the model performs poorly in segmenting cones. These outcomes show that the cone segmentation task is not solved by existing segmentation models and there is a need for new solutions to segment cones accurately in future work. In summary, our contributions are as follows:

\vspace{2cm}


\begin{enumerate}[noitemsep]

    \item We introduce \dataset, the first expert-annotated publicly available dataset for cone segmentation across three different regions on Mars, along with metadata for each sample.

    \item We designed two benchmarks based on \dataset{}: (i) Spatial Generalization and (ii) Cone-size Generalization, and assessed the effectiveness of various segmentation-based models in segmenting cones.

    \item Evaluation of models indicates that existing models struggle to perform well on \dataset{}, highlighting the need for specialized models that can effectively capture the unique characteristics of cones.

\end{enumerate}

\section{Related Work}
\label{sec:related_work}

Deep learning has enabled researchers to develop models for a wide range of tasks in order to gain insights into data properties, improve labeling processes, and facilitate annotation. However, the success of these models heavily relies on the availability of large training datasets. The following sections provide an overview of the existing datasets for various Mars-related tasks created for training deep learning models (\textsection \ref{subsec:mars_dataset}) and past research on cone detection on Mars (\textsection \ref{subsec:cone_dataset}).

\subsection{Mars Datasets}
\label{subsec:mars_dataset}

In Mars research, recognition of geological landforms and terrain classification are commonly explored tasks. Among these tasks, crater detection has been the most prominent. A few widely used datasets for Mars crater detection are \cite{robbins2012newa, robbins2012newb, lee2019automated, wilhelm2020domars16k, wagstaff2018deep, wagstaff2021mars, marslunarcraterdataset}. In addition to crater detection, researchers have also generated datasets for other geological features, such as dunes, streaks, and ridges. For instance, \cite{wilhelm2020domars16k} introduced a dataset encompassing 15 classes and $\sim1k$ data samples per class across five distinct landform categories: aeolian bedforms, topographic landforms, slope feature landforms, impact landforms, and basic terrain landforms. Further, Wagstaff et al. have contributed landform datasets comprising $\sim3k$ and $\sim10k$ data samples from six classes \cite{wagstaff2018deep, wagstaff2021mars}. These datasets also include around $\sim2.9k$ and $\sim7k$ images representing over 20 classes of rover parts. Datasets are also available for terrain segmentation, encompassing classes such as rock, soil, sand, bedrock, and more. Two notable datasets in this domain are AI4MARS ($\sim326k$ samples from 5 classes) \cite{swan2021ai4mars} and $S^5$Mars ($\sim5k$  samples from 9 classes) \cite{zhang2022s}. Another dataset focuses on martian frost, classifying images as either containing frost or representing background scenery \cite{diniega_serina_2022_6561242}. Additionally, there are datasets tailored for change detection, comprising pairs of images that capture changes or the absence of changes over the same location at different times \cite{kerner2019toward}. Furthermore, novelty detection and outlier detection datasets have been formulated to identify novel and anomalous samples within Mars datasets \cite{kerner2021domain, kerner2020comparison}. Notably, there exists a dataset designed for classifying dusty versus non-dusty images \cite{dorangary34950682019gray}.

\begin{table*}
  \begin{center}
    {
        \small
        {
        \resizebox{0.96\linewidth}{!}{
            \begin{tabular}{ccccccc}
            \toprule[1.5pt]
            \textbf{Region} & \makecell{\textbf{CTX Mosaic Folder}\\\textbf{(4$^{\circ}$}$\times$ \textbf{4$^{\circ}$)}}  & \makecell{\textbf{CTX Mosaic Tile ID}\\\textbf{(2$^{\circ}$}$\times$ \textbf{2$^{\circ}$)}} & \makecell{\textbf{Area of Masked}\\\textbf{Region (\textbf{N$^{\circ}$}$\times$ \textbf{E$^{\circ}$})}} & \textbf{Resolution} & \textbf{Latitude} & \textbf{Longitude} \\
            \midrule[1pt]

            Isidis Planitia & beta01\_E084\_N12 & beta01\_E084\_N12.tif & Partial (0.5$^{\circ}$ $\times$ $0.5^{\circ}$) & 5927 $\times$ 5927 & 13.5$^{\circ}$ & 85.5$^{\circ}$ \\
            \midrule
            Acidalia Planitia & beta01\_E-016\_N36 & beta01\_E-014\_N38.tif & Partial (1$^{\circ}$ $\times$ $0.5^{\circ}$) & 11855 $\times$ 5927 & 39.5$^{\circ}$ & -13$^{\circ}$ \\
            \midrule
            \multirow{6}{*}{Hypanis} & beta01\_E-044\_N08 & beta01\_E-044\_N10.tif & Full (2$^{\circ}$ $\times$ $2^{\circ}$) & 23710 $\times$ 23710 & 10$^{\circ}$ & -44$^{\circ}$ \\
                                     & beta01\_E-044\_N12 & beta01\_E-044\_N12.tif & Full (2$^{\circ}$ $\times$ $2^{\circ}$) & 23710 $\times$ 23710 & 12$^{\circ}$ & -44$^{\circ}$ \\
                                     & beta01\_E-048\_N08 & beta01\_E-046\_N10.tif & Full (2$^{\circ}$ $\times$ $2^{\circ}$) & 23710 $\times$ 23710 & 10$^{\circ}$ & -46$^{\circ}$ \\
                                     & beta01\_E-048\_N08 & beta01\_E-048\_N10.tif & Full (2$^{\circ}$ $\times$ $2^{\circ}$) & 23710 $\times$ 23710 & 10$^{\circ}$ & -48$^{\circ}$ \\
                                     & beta01\_E-048\_N12 & beta01\_E-046\_N12.tif & Full (2$^{\circ}$ $\times$ $2^{\circ}$) & 23710 $\times$ 23710 & 12$^{\circ}$ & -46$^{\circ}$ \\
                                     & beta01\_E-048\_N12 & beta01\_E-048\_N12.tif & Full (2$^{\circ}$ $\times$ $2^{\circ}$) & 23710 $\times$ 23710 & 12$^{\circ}$ & -48$^{\circ}$ \\
            \bottomrule
            \end{tabular}
        }
    }
    }
    \end{center}
    \caption{Metadata of each CTX tile across three regions used in creation of \dataset}
    \label{tab:metadata}
\end{table*}

\subsection{Dataset on Cone Detection}
\label{subsec:cone_dataset}

Despite the wide range of crater and other feature databases for Mars, there is a scarcity of cone detection/segmentation studies. Palafox et al. introduced MarsNet, a CNN-based classifier, for the identification of volcanic rootless cones and transverse aeolian ridges \cite{palafox2017automated}. Pieterek et al. proposed a short study on pitted cones and crater detection by comparing a CNN with SVM \cite{pieterek2022automated}. Both of these studies lack detailed information about the annotation process, including whether the data were annotated by experts. Jiang et al. proposed a Single Shot MultiBox Detector model for cone and crater detection on Mars \cite{jiang2022automated}. One limitation of their work is that they did not use experts for annotation but instead relied on an unspecified Wikipedia definition of cones and example HiRISE images found online. Furthermore, their annotations were performed at the box level, rather than accurately masking the region of interest, which does not fully align with the expectations of planetary scientists. One common weakness among all previous studies is that they train models using data from a single region, which may limit the ability of the model to generalize on another region and hinder global mapping and detection, as studies have shown that cones have unique characteristics specific to their region on Mars \cite{skinner2009martian}.

\section{\dataset{}}
\label{sec:dataset}

This section provides information about the data source, annotation process, and overview of \dataset{} in detail.

\subsection{Source Imagery}
\label{subsec:source_imagery}

The Mars Reconnaissance Orbiter (MRO) Context Camera (CTX) acquires high-resolution images of the martian surface and has been operational since 2006 \cite{bell2013calibration}. To build \dataset{}, we used open-source CTX data from the Murray Lab \cite{murray_lab}. The dataset is a seam-corrected global image mosaic of Mars rendered at $\sim5.0$ meters/pixel \cite{malin2007context, dickson2023release}. Data covers the entirety of the martian surface (> 99.5\%). The global image data is divided into 3960 tiles (4$^{\circ}$ $\times$ 4$^{\circ}$) from 88$^{\circ}$S to 88$^{\circ}$N \cite{dickson2018global, dickson2023release}. Each tile is subdivided into 4 subtiles (2$^{\circ}$ $\times$ 2$^{\circ}$). This is the highest resolution complete-coverage global image data for Mars and is freely accessible at \cite{murray_lab}.


\subsection{Data Annotation}
\label{subsec:data_collection}

There have been at least six fields of pitted cones on Mars identified so far (where a leading hypotheses is formation by mud volcanism), with each field containing hundreds to tens of thousands of cones \cite{skinner2009martian}. However, a labeled dataset of cone annotations did not exist before our study. To create the labeled dataset, a planetary geologist annotated eight CTX subtiles spanning three different regions of Mars with known pitted cone fields (Table \ref{tab:metadata}). Using the CTX mosaic subtiles as a basemap, a shapefile was created with polygons outlining the shape of each cone. The shapefile was then converted to a bitmask of the same resolution and dimensions as the CTX basemap (where pixels inside a cone shape were 1, and those not on a cone were set to 0). The labeled dataset thus contained eight CTX subtile images and eight corresponding bitmasks. The total number of annotated cones was 163 in Acidalia Planitia, 325 in Isidis Planitia, and 1691 in the Hypanis region of Southern Chryse Planitia. For Isidis Planitia and Acidalia Planitia there was a high density of pitted cones in the subtile, so only part of the CTX subtile was mapped (`Partial' indicated in Table \ref{tab:metadata}), and the CTX image and bitmask were cropped to the mapped extent. The latitude and longitude in Table \ref{tab:metadata} correspond to the bottom-left corner of each annotated tile.


A planetary geologist with expertise in the morphology of cones in the dataset regions (co-author Adler) created all annotations of Isidis Planitia (IP) and Acidalia Planitia (AP) for this work. The Hypanis annotation shapes were sourced from a previously published peer-reviewed journal article \cite{adler2022regional} and thus have gone through quality control by other planetary mapping experts. While IP and AP regions have not been peer-reviewed, we deem such review unnecessary because 1) the same cone fields have been previously published in figures \cite{mcgowan2011utopia, davis1995curvilinear, hemmi2018high, farrand2005pitted, oehler2010evidence} (but were not digitized at the individual cone level), 2) the geologic setting within a uniform background unit makes identification obvious to experts, and 3) annotation was performed by a Mars mud volcano expert trained in geologic mapping. While we cannot be 100\% certain about the annotations without ground-truth confirmation, we think it is unlikely there are false positives or false negatives at the object-level. In the Hypanis region, erroneous labels may be more likely because this region is more varied and geologically complex with compound features (many that erode into a rounded shape) that could be misinterpreted \cite{adler2022regional}. We estimate that no more than 5-20\% of the annotations may be erroneous.


\subsection{Data Preparation}
\label{subsec:data_processing}

The original data provided by Murray Lab are large CTX subtiles of size $23,710$ $\times$ $23,710$ pixels ($\sim300$ MB). To prepare these images for deep learning models, it is necessary to create smaller patches of the data, that can be compatible with the DL-based models and the system on which the model will be trained. We generated input samples by dividing each subtile into chunks measuring $512$ $\times$ $512$ pixels (which covers $\sim2.5km^2$ area). The patches are generated in a column-wise manner. All the generated patches are distinct, ensuring that there is no overlap. This approach ensures there is no data leakage between training and test partitions. The resulting \dataset{} dataset has a total of 13,686 patches from 8 different subtiles across 3 regions.

\subsection{Data Overview}
\label{subsec:data_overview}

The \dataset{} dataset includes input image and target segmentation mask pairs as well as metadata for every sample pair. Figure \ref{fig:teaser_figure} shows the example of input data and their corresponding ground truth mask from each region. Each input-output pair has a unique name prefixed by its CTX Mosaic Tile ID. Generated input-output pairs can be used to train and evaluate segmentation or object detection models.

Additionally, \dataset{} provides the metadata of every CTX Mosaic Tile used in its creation (as displayed in Table \ref{tab:metadata}), and a set of attributes about each patch, which can be further used for model training and evaluation, and mapping cones on Mars. These attributes are crucial as they record the specifics of \textit{every cone} found across all the patches. The description of each attribute is as follows:

\begin{itemize}[noitemsep]

    \item \textbf{Patch Id:} Unique name of each input-output pair (e.g., \textit{E-044\_N10\_00516.tif}). For Isidis Planitia and Acidalia Planitia name follows $\_P$ which indicates a patch from a partially annotated tile (e.g., \textit{E084\_N12\_00068\_P.tif}).


    \item \textbf{Region:} Denotes the region name to which each patch belongs (e.g., Isidis Planitia).


    \item \textbf{CTX Mosaic Folder:} CTX Mosaic folder from where subtile of CTX Mosaic Tile is taken (e.g., \textit{beta01\_E-044\_N08}).


    \item \textbf{CTX Mosaic Tile ID:} CTX Mosaic Tile ID which was annotated and patch created from (e.g., \textit{beta01\_E-044\_N10.tif}).


    \item \textbf{Latitude-Longitude Bounding Box:} This attribute provides the latitude-longitude coordinates of the bounding box for each cone present in the patch in the below format:

    \begin{quote}
        \textit{[Polygon ((left-top, right-top, left-bottom, right-bottom)), Polygon ((left-top, right-top, left-bottom, right-bottom)), ...]}
    \end{quote} 

    Here, each \textit{Polygon} element represents the coordinates of the bounding box of a single cone, where each coordinate represents \textit{(longitude, latitude)} pair.


    \item \textbf{Latitude-Longitude Perimeter:} This attribute provides the latitude-longitude vertices of the perimeter for each cone present in the patch in the below format:
    \begin{quote}
        \textit{[Polygon (($v_0$, $v_1$ $v_2$, $v_3$, ...)), Polygon (($v_0$, $v_1$ $v_2$, $v_3$, ...)), ...]}
    \end{quote}

    Here, each \textit{Polygon} element represents the vertices of the perimeter of a single cone, where each vertex ($v_i$) represents \textit{(longitude, latitude)} pair.


    \item \textbf{Bounding Box:} Lists of bounding box list (e.g., [[$x_{min}$, $y_{min}$, $width$, $height$], [$x_{min}$, $y_{min}$, $width$, $height$], ...], here, each element in the list is a bounding box of a single cone).


    \item \textbf{Perimeter:} Lists of polygon vertices list (e.g., [[$x_1$, $y_1$, $x_2$, $y_2$, ...], [$x_1$, $y_1$, $x_2$, $y_2$, ...], ...], here, each element in the list is a polygon of a single cone).


    \item \textbf{Area:} Area of every cone in the patch\footnote{calculated using \textit{contourarea} function from OpenCV} (e.g., [$A_1$, $A_2$, ...], here, each $A_i$ is a float value).


    \item \textbf{Average Cone Diameter:} This attribute shows the average cone diameter of all cones in the patch. To calculate this, we have used the following formula which assumes the cone has a round shape:

    \begin{equation}
        D = \frac{1}{N} \hspace{1mm} \sum_{i=1}^{N} \hspace{1mm} 2 * \sqrt{\frac{A_i * 25}{\pi}}
    \end{equation}

    Where $N$ is the number of cones in the patch; $A_i$ is the area of cone $i$; and we multiplied the area by $25$ to convert the area into $meter^2$ (as 1 pixel covers an area of $5\times$ meters).


    \item \textbf{Number of Cones:} A total number of cones in the patch ($N$).

\end{itemize}

Latitude and Longitude for the bounding box and perimeter follow a standard format, which simplifies the process for users to plot individual cones or clusters of cones from any patch on Mars using any Geographic Information Systems (GIS) software. Bounding Box, Area, and Perimeter are given in a similar format as the COCO dataset \cite{lin2014microsoft}.

It is crucial to note that \dataset{} includes samples that do not contain any cones, and these are referred to as \textit{negative samples} or \textit{non-cone patches}. Samples that contain cone/s are referred as \textit{positive samples} or \textit{cone-patches}. It is essential to recognize the significance of negative samples as they help to identify characteristics in the data that do \textit{not} represent cones. For non-cone patches, attribute values contain empty list values or 0 accordingly.

\section{Benchmarks}
\label{sec:benchmarks}

As discussed in \textsection \ref{subsec:data_overview}, the \dataset task is binary segmentation where the goal is to segment the cones in the input data and to mask (segment) the particular region where the cone is present (as shown in Figure \ref{fig:teaser_figure}). Cone segmentation is a very difficult task as shadows, contrast, and lighting change due to the time of day and season the images were acquired which strongly affects whether cones stand out from the background terrain and have similar shadow angles and lengths. Also, cones present in each of the three regions exhibit variations in terms of size, shape, and other characteristics \cite{skinner2009martian}. For example, the morphology of cones in the Hypanis region is highly variable (small-large, bright-dark, circular-elongated-clustered) \cite{adler2022regional}. These characteristics make global segmentation of cones challenging, similar to other remote sensing tasks with high intra-class variance (such as building damage detection \cite{benson2020assessing}). Based on this, we defined two benchmark tasks using \dataset: (i) Spatial Generalization (BM-1) and (ii) Cone-size Generalization (BM-2).


\subsection{Spatial Generalization}
\label{subsec:bm_1}

In this benchmark, our objective is to assess model performance across the regional variability of cones (Figure \ref{fig:teaser_bm_1}). Table \ref{tab:num_patches} shows the total number of patches and the cone-patches in each region. For experiments, we train the model on two configurations: (i) single-region: model is trained using data from each region individually, and (ii) multi-region: model is trained on a combination of 2 or all 3 regions. For both configurations, we evaluate the model on each region separately where we denote regions included in the training as \textit{in-distribution}, and excluded from the training as \textit{out-of-distribution}.

\begin{table}[H]
  \begin{center}
    {
        \small
        {
            \begin{tabular}{ccc}
            \toprule
            \textbf{Region} & \makecell{\textbf{\# of patches}\\\textbf{created}}  & \makecell{\textbf{\# of cone-}\\\textbf{patches}} \\
            \midrule
            Isidis Planitia (IP) & 144 & 131\\
            Acidalia Planitia (AP) & 288 & 135\\
            Hypanis (HP) & 13,254 & 1,392\\
            \midrule
            Total & 13,686 & 1,658\\
            \bottomrule
            \end{tabular}
        }
    }
    \end{center}
    \caption{Number of patches and cone-patches across three regions}
    \label{tab:num_patches}
\end{table}

\subsection{Cone-size Generalization}
\label{subsec:bm_2}


This task evaluates the capability of a model to segment cones of different sizes which can identify any model biases in terms of cone size, which would substantially impact the downstream use of the model. Figure \ref{fig:teaser_bm_2} shows the histogram of cone size in terms of \textit{Average Cone Diameter} of patches and shows the variations in cone size across different patches. The peak of the distribution is approximately 400 m. We split all cone-patches into 3 categories: i) small, ii) medium, and iii) large. Table \ref{tab:size_info} gives statistics of size range and number of samples in each category. We create almost equal splits to fairly compare a model's performance when trained on different size categories.

\begin{table}[H]
  \begin{center}
    {
        \small
        {
            \begin{tabular}{ccc}
            \toprule
            \textbf{Category} & \makecell{\textbf{Size range}\\\textbf{(Cone Diameter)}} & \textbf{\# of Patches} \\
            \midrule
            Small (S) & 5m < D $\le$ 400 & 537\\
            Medium (M) & 400m < D $\le$ 670m & 569\\
            Large (L) & 670m < D & 550\\
            \bottomrule
            \end{tabular}
        }
    }
    \end{center}
    \caption{Size range and number of patches across three categories}
    \label{tab:size_info}
\end{table}

For experiments, we train the model on two configurations: (i) single-category (i.e., a model trained on each size category), and (ii) multi-category (i.e., a model trained on a combination of two size categories). For both configurations, we evaluate the model on each category separately where we denote the category included in the training as \textit{in-distribution} (id), and excluded from the training as \textit{out-of-distribution} (ood). id and ood data used for evaluation in both tasks will be denoted as $\mathcal{D}_{id}$ and $\mathcal{D}_{ood}$, respectively.

\begin{figure*}
  \centering
  \begin{subfigure}{\columnwidth}
    \centering
    \includegraphics[width=1.06\linewidth]{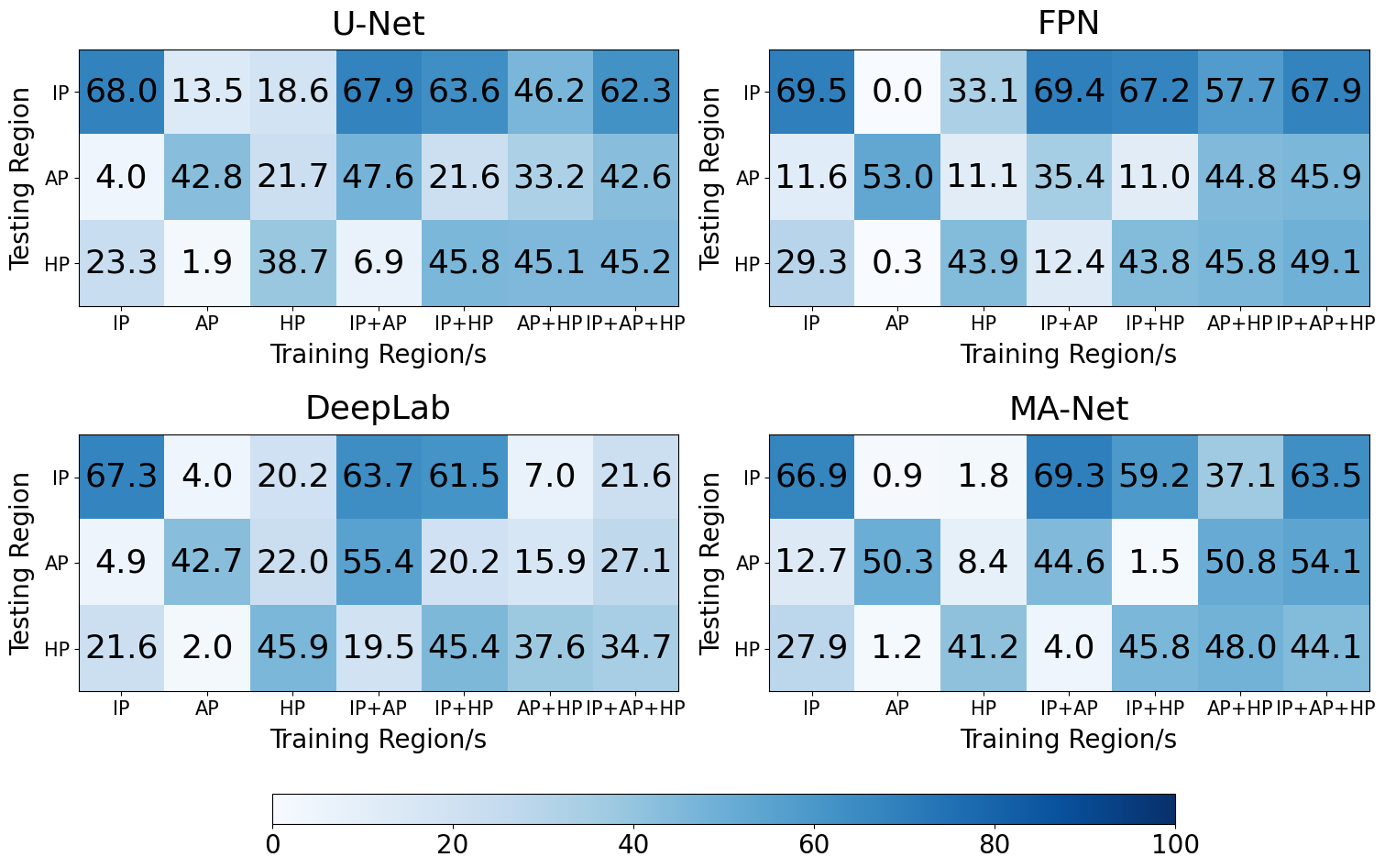}
    \caption{Pixel IoU for BM-1}
    \label{fig:iou_bm_1}
  \end{subfigure}
  \hspace{2mm}
  \begin{subfigure}{\columnwidth}
    \centering
    \includegraphics[width=0.94\linewidth]{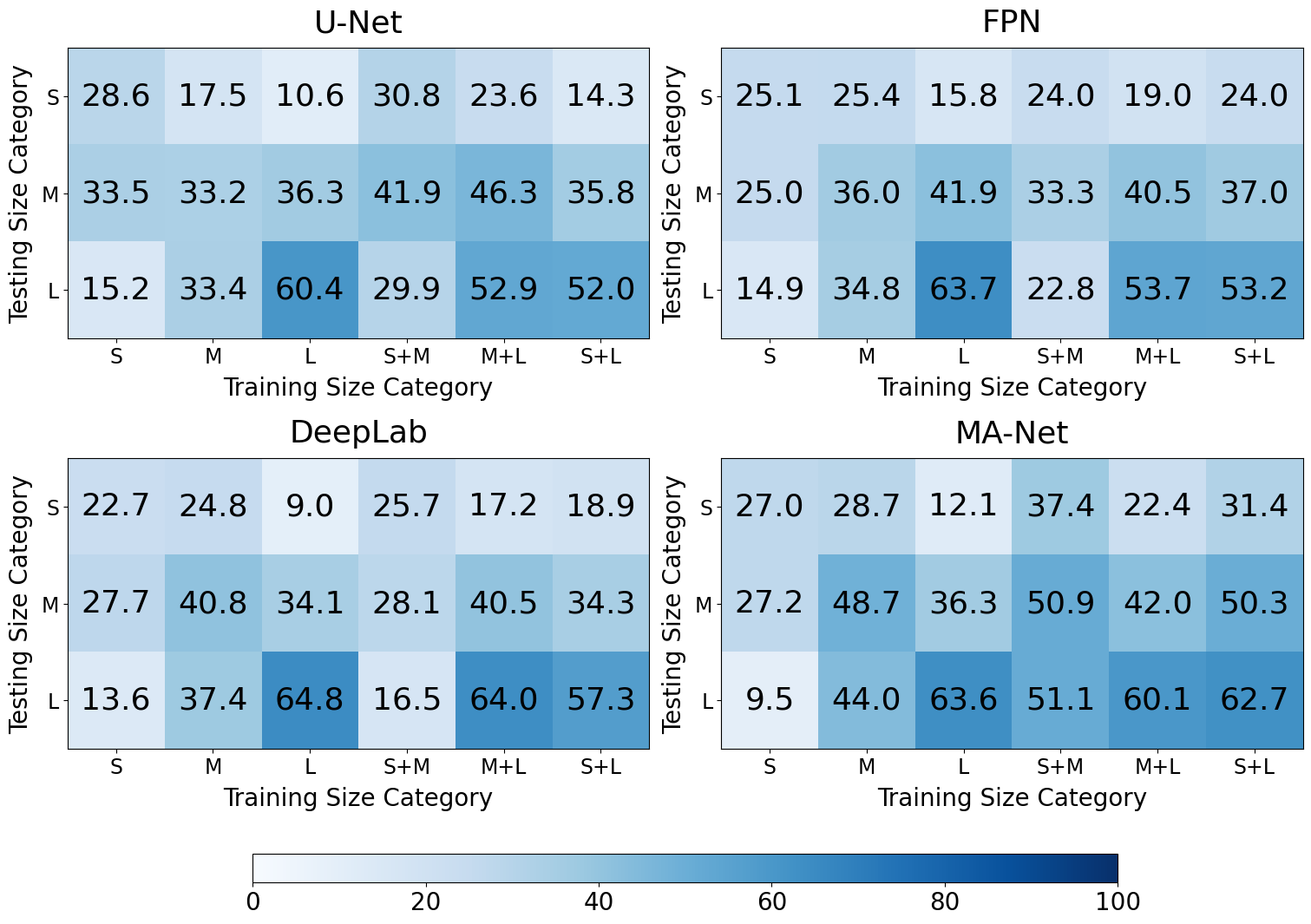}
    \caption{Pixel IoU for BM-2}
    \label{fig:iou_bm_2}
  \end{subfigure}
  \caption{Results on $\mathcal{D}_{id}$ and $\mathcal{D}_{ood}$ for both BMs based on Pixel IoU (not a confusion matrix). IP: Isidis Planitia, AP: Acidalia Planitia, HP: Hypanis, S: Small, M: Medium, L: Large.}
  \label{fig:ious}
\end{figure*}

\section{Experiments Setup}
\label{sec:exp_results}

\paragraph{Data split:} As stated in \textsection \ref{subsec:data_overview}, \dataset{} has negative samples, and it is necessary to include those in the training. Table \ref{tab:num_patches} shows that for BM-1 in Acidalia Planitia and Hypanis, there are equal and fewer negative samples compared to positive samples, respectively. Hence, for BM-1, we use balanced data of positive and negative samples to ensure that the model does not overfit the majority class except for Isidis Planitia (IP) region. Also, we have stratified positive and negative samples across dataset splits. In BM-2, we added negative samples to balance the data for each category. To evaluate the effectiveness of training with negative patches, we performed an ablation experiment in which models were trained only on positive samples.

For all experiments, we split the data into training, validation, and testing sets with a ratio of 7:1:2. To do a fair comparison across experiments, the same dataset splits for all regions and size categories are used for single and multi configurations training. These splits are also provided as part of our public benchmark dataset. As mentioned in \textsection \ref{subsec:data_processing}, non-overlapping samples were created with a stride of 512, which is equivalent to the patch height/width. This ensures that all samples in the data are distinct from each other.

\vspace{-4mm}

\paragraph{Models:} For training and evaluation, we selected commonly-used state-of-the-art segmentation-based models: U-Net \cite{ronneberger2015u}, FPN \cite{lin2017feature}, DeepLab \cite{chen2017rethinking}, and MA-Net \cite{fan2020ma}. For the encoder of each model, we used ResNet-101 \cite{he2016deep} as the backbone pre-trained on ImageNet \cite{ILSVRC15}.

\vspace{-4mm}

\paragraph{Training Configuration:} All models were trained for 200 epochs with a batch size of 8. Soft binary cross-entropy with logits was used as the loss function with the Adam optimizer \cite{kingma2014adam}. We used early stopping to avoid overfitting and all models were evaluated on the model from the epoch with the lowest validation loss. All the experiments were conducted on Tesla V100-SXM2 with 16 GB GPU RAM.

\vspace{-4mm}

\paragraph{Metrics:} We report standard pixel-wise segmentation metrics and object-wise metrics for evaluation. We use pixel-Intersection over Union (IoU) \cite{everingham2015pascal} and mask IoU \cite{lin2014microsoft}. Pixel accuracy, pixel precision, and pixel recall are calculated by using all 4 quadrants of the confusion matrix as defined in \cite{iakubovskii_2019}. Models are also evaluated on mean Average Precision (mAP) \cite{lin2014microsoft} and Panoptic Quality \cite{kirillov2019panoptic}. Since planetary scientists are also interested in instance-level metrics, we have shown evaluation based on object-wise metrics. For object IoU, we ran Hungarian matching \cite{wang2007study} between all ground truth and predicted bounding boxes with the threshold of 0.5. To calculate object-wise accuracy, precision, and recall, the object is considered as True Positive (TP) if IoU is above 0.5.

As discussed in \textsection \ref{subsec:data_overview}, \dataset{} includes non-cone patches and it is important to analyze the model's performance when it incorrectly tries to segment the object. Most previous research on segmentation does not consider this scenario. To incorporate this, we evaluated the model's performance by computing pixel-wise area segmented as a cone which is defined as follows:

\begin{equation}
\label{eqn:area-fp}
    A_{FP} = \frac{\text{FP} \;\; (\text{\# pixels segmented as cone})}{\text{FP + TN} \;\;  (\text{total \# pixels in image})}*100
\end{equation}

In Equation \ref{eqn:area-fp}, \textit{lower} $A_{FP}$ indicates better model performance, i.e., lower false positive area in negative patches.

\begin{table*}[!ht]
\setlength\tabcolsep{4.0pt}
\setlength{\belowcaptionskip}{-10pt}
\centering
\footnotesize
\resizebox{0.96\linewidth}{!}{
\begin{tabular}{c|c|ccccccccccc|c}
\toprule[1.5pt]
\multirow{2}{*}{\textbf{\begin{tabular}[c]{@{}c@{}}Training\\Region\end{tabular}}} &
  \multirow{2}{*}{\textbf{\begin{tabular}[c]{@{}c@{}}Testing\\Region\end{tabular}}} &
  \multicolumn{11}{c|}{\textbf{Cone}} &
  \textbf{Non-Cone} \\ \cmidrule{3-14} 
 &
   &
  \multicolumn{1}{c|}{\textbf{\begin{tabular}[c]{@{}c@{}}Mask\\IoU\end{tabular}}} &
  \multicolumn{1}{c|}{\textbf{\begin{tabular}[c]{@{}c@{}}Pixel\\IoU\end{tabular}}} &
  \multicolumn{1}{c|}{\textbf{\begin{tabular}[c]{@{}c@{}}Pixel\\Accuracy\end{tabular}}} &
  \multicolumn{1}{c|}{\textbf{\begin{tabular}[c]{@{}c@{}}Pixel\\Precision\end{tabular}}} &
  \multicolumn{1}{c|}{\textbf{\begin{tabular}[c]{@{}c@{}}Pixel\\Recall\end{tabular}}} &
  \multicolumn{1}{c|}{\textbf{\begin{tabular}[c]{@{}c@{}}Panoptic\\Quality\end{tabular}}} &
  \multicolumn{1}{c|}{\textbf{mAP}} &
  \multicolumn{1}{c|}{\textbf{\begin{tabular}[c]{@{}c@{}}Object\\IoU\end{tabular}}} &
  \multicolumn{1}{c|}{\textbf{\begin{tabular}[c]{@{}c@{}}Object\\Accuracy\end{tabular}}} &
  \multicolumn{1}{c|}{\textbf{\begin{tabular}[c]{@{}c@{}}Object\\Precision\end{tabular}}} &
  \textbf{\begin{tabular}[c]{@{}c@{}}Object\\ Recall\end{tabular}} &
  \textbf{$A_{FP}$} \\ \midrule[1pt]
IP &
  IP &
  \multicolumn{1}{c|}{64.81} &
  \multicolumn{1}{c|}{67.92} &
  \multicolumn{1}{c|}{96.66} &
  \multicolumn{1}{c|}{83.54} &
  \multicolumn{1}{c|}{79.09} &
  \multicolumn{1}{c|}{54.87} &
  \multicolumn{1}{c|}{33.15} &
  \multicolumn{1}{c|}{74.29} &
  \multicolumn{1}{c|}{60.71} &
  \multicolumn{1}{c|}{71.77} &
  75.51 &
  0.00 \\ \midrule
AP &
  AP &
  \multicolumn{1}{c|}{45.13} &
  \multicolumn{1}{c|}{45.83} &
  \multicolumn{1}{c|}{96.21} &
  \multicolumn{1}{c|}{82.97} &
  \multicolumn{1}{c|}{52.77} &
  \multicolumn{1}{c|}{38.91} &
  \multicolumn{1}{c|}{19.44} &
  \multicolumn{1}{c|}{49.30} &
  \multicolumn{1}{c|}{44.45} &
  \multicolumn{1}{c|}{52.25} &
  50.87 &
  0.23 \\ \midrule
HP &
  HP &
  \multicolumn{1}{c|}{41.39} &
  \multicolumn{1}{c|}{42.43} &
  \multicolumn{1}{c|}{92.09} &
  \multicolumn{1}{c|}{82.71} &
  \multicolumn{1}{c|}{49.99} &
  \multicolumn{1}{c|}{31.41} &
  \multicolumn{1}{c|}{14.39} &
  \multicolumn{1}{c|}{43.41} &
  \multicolumn{1}{c|}{36.97} &
  \multicolumn{1}{c|}{43.72} &
  44.31 &
  0.50 \\ \midrule
\multirow{2}{*}{IP + AP} &
  IP &
  \multicolumn{1}{c|}{64.19} &
  \multicolumn{1}{c|}{67.58} &
  \multicolumn{1}{c|}{96.68} &
  \multicolumn{1}{c|}{85.24} &
  \multicolumn{1}{c|}{76.54} &
  \multicolumn{1}{c|}{54.96} &
  \multicolumn{1}{c|}{34.15} &
  \multicolumn{1}{c|}{74.81} &
  \multicolumn{1}{c|}{60.89} &
  \multicolumn{1}{c|}{72.06} &
  72.68 &
  0.00 \\
 &
  AP &
  \multicolumn{1}{c|}{45.34} &
  \multicolumn{1}{c|}{45.76} &
  \multicolumn{1}{c|}{96.36} &
  \multicolumn{1}{c|}{90.16} &
  \multicolumn{1}{c|}{50.35} &
  \multicolumn{1}{c|}{36.10} &
  \multicolumn{1}{c|}{20.92} &
  \multicolumn{1}{c|}{50.52} &
  \multicolumn{1}{c|}{39.67} &
  \multicolumn{1}{c|}{47.15} &
  50.42 &
  0.31 \\ \midrule
\multirow{2}{*}{IP + HP} &
  IP &
  \multicolumn{1}{c|}{61.25} &
  \multicolumn{1}{c|}{62.82} &
  \multicolumn{1}{c|}{96.24} &
  \multicolumn{1}{c|}{83.32} &
  \multicolumn{1}{c|}{74.58} &
  \multicolumn{1}{c|}{50.14} &
  \multicolumn{1}{c|}{28.69} &
  \multicolumn{1}{c|}{71.68} &
  \multicolumn{1}{c|}{54.70} &
  \multicolumn{1}{c|}{69.81} &
  63.05 &
  0.00 \\
 &
  HP &
  \multicolumn{1}{c|}{42.82} &
  \multicolumn{1}{c|}{43.95} &
  \multicolumn{1}{c|}{92.28} &
  \multicolumn{1}{c|}{79.47} &
  \multicolumn{1}{c|}{53.16} &
  \multicolumn{1}{c|}{32.20} &
  \multicolumn{1}{c|}{14.26} &
  \multicolumn{1}{c|}{45.14} &
  \multicolumn{1}{c|}{37.34} &
  \multicolumn{1}{c|}{44.29} &
  45.63 &
  0.66 \\ \midrule
\multirow{2}{*}{AP + HP} &
  AP &
  \multicolumn{1}{c|}{36.53} &
  \multicolumn{1}{c|}{36.17} &
  \multicolumn{1}{c|}{95.52} &
  \multicolumn{1}{c|}{89.52} &
  \multicolumn{1}{c|}{40.14} &
  \multicolumn{1}{c|}{28.32} &
  \multicolumn{1}{c|}{11.55} &
  \multicolumn{1}{c|}{37.71} &
  \multicolumn{1}{c|}{32.88} &
  \multicolumn{1}{c|}{39.80} &
  38.03 &
  1.46 \\
 &
  HP &
  \multicolumn{1}{c|}{42.66} &
  \multicolumn{1}{c|}{44.12} &
  \multicolumn{1}{c|}{92.25} &
  \multicolumn{1}{c|}{81.40} &
  \multicolumn{1}{c|}{52.92} &
  \multicolumn{1}{c|}{32.51} &
  \multicolumn{1}{c|}{14.01} &
  \multicolumn{1}{c|}{47.10} &
  \multicolumn{1}{c|}{37.35} &
  \multicolumn{1}{c|}{44.78} &
  46.48 &
  1.08 \\ \midrule
\multirow{3}{*}{IP + AP + HP} &
  IP &
  \multicolumn{1}{c|}{55.85} &
  \multicolumn{1}{c|}{56.36} &
  \multicolumn{1}{c|}{95.76} &
  \multicolumn{1}{c|}{89.19} &
  \multicolumn{1}{c|}{63.82} &
  \multicolumn{1}{c|}{45.55} &
  \multicolumn{1}{c|}{26.91} &
  \multicolumn{1}{c|}{63.78} &
  \multicolumn{1}{c|}{51.40} &
  \multicolumn{1}{c|}{65.94} &
  56.95 &
  0.00 \\
 &
  AP &
  \multicolumn{1}{c|}{45.95} &
  \multicolumn{1}{c|}{45.76} &
  \multicolumn{1}{c|}{96.40} &
  \multicolumn{1}{c|}{85.09} &
  \multicolumn{1}{c|}{50.85} &
  \multicolumn{1}{c|}{40.53} &
  \multicolumn{1}{c|}{16.36} &
  \multicolumn{1}{c|}{53.68} &
  \multicolumn{1}{c|}{45.48} &
  \multicolumn{1}{c|}{53.84} &
  52.50 &
  2.01 \\
 &
  HP &
  \multicolumn{1}{c|}{43.01} &
  \multicolumn{1}{c|}{44.58} &
  \multicolumn{1}{c|}{92.30} &
  \multicolumn{1}{c|}{81.60} &
  \multicolumn{1}{c|}{53.56} &
  \multicolumn{1}{c|}{32.85} &
  \multicolumn{1}{c|}{13.46} &
  \multicolumn{1}{c|}{45.99} &
  \multicolumn{1}{c|}{38.50} &
  \multicolumn{1}{c|}{44.94} &
  47.60 &
  0.79 \\ \midrule
\end{tabular}
}
\caption{Results for BM-1 for all metrics on $\mathcal{D}_{id}$. Here, the results in each row are the average across 4 models. See Appendix B (Table 1 and 2) for individual model results.}
\label{tab:bm_1_results}
\end{table*}

\section{Analysis and Discussion}
\label{sec:evaluation}

In the following two sections, we describe quantitative and qualitative analysis of the results for both BMs.

\subsection{Quantitative Analysis}
\label{subsec:quantitative_analysis}

Figure \ref{fig:iou_bm_1} and \ref{fig:iou_bm_2} show pixel IoU for all 4 models on $\mathcal{D}_{id}$ and $\mathcal{D}_{ood}$ for BM-1 and BM-2, respectively. For BM-1, the average IoU across all models on $\mathcal{D}_{id}$ is $52.52\%$ for single-region and $49.03\%$ for multi-region training. For BM-2, the average IoU across all models on single-category and multi-category is $42.88\%$ and $41.08\%$, respectively. Table \ref{tab:bm_1_results} and \ref{tab:bm_2_results} report results on $\mathcal{D}_{id}$ for all evaluation metrics for BM-1 and BM-2, respectively. Results for each model on $\mathcal{D}_{ood}$ are reported in the Appendix. From Table \ref{tab:bm_1_results} and \ref{tab:bm_2_results}, it can be observed that the maximum mAP is $33.16\%$ for BM-1 and $22.54\%$ for BM-2. These results indicate that cone segmentation is an open challenging problem, even on $\mathcal{D}_{id}$, for both BMs. From Figure \ref{fig:ious}, it can be observed that an average pixel IoU is $8.3\%$ higher for BM-1 compared to BM-2. This indicates that current models generalize better across different regions than different size categories of cones. Comparing model-wise performance, MA-Net outperformed all 3 models for single and multi-group training for both benchmarks. Moreover, other metrics shown in Table \ref{tab:bm_1_results} and \ref{tab:bm_2_results} show similar observations.

\subsection{Analysis}
\label{subsec:qualitative_analysis}

\paragraph{Models fail to generalize on $\mathcal{D}_{ood}$:} Figure \ref{fig:iou_bm_1} shows that all models do not generalize well to  $\mathcal{D}_{ood}$ for single-region or multi-region training. For example, U-Net trained on HP achieves $38.7$-pixel IoU on HP ($\mathcal{D}_{id}$), but drops substantially to $18.6$ for IP and $21.7$ for AP ($\mathcal{D}_{ood}$). This discrepancy in performance could be attributed to the variations in cone characteristics among the three regions. Models show the same generalization gap on $\mathcal{D}_{ood}$ for BM-2.

\vspace{-4mm}

\begin{figure}[!htbp]

  \centering
  \begin{subfigure}{0.155\columnwidth}
    \centering
    \caption{Input}
    \includegraphics[width=\linewidth]{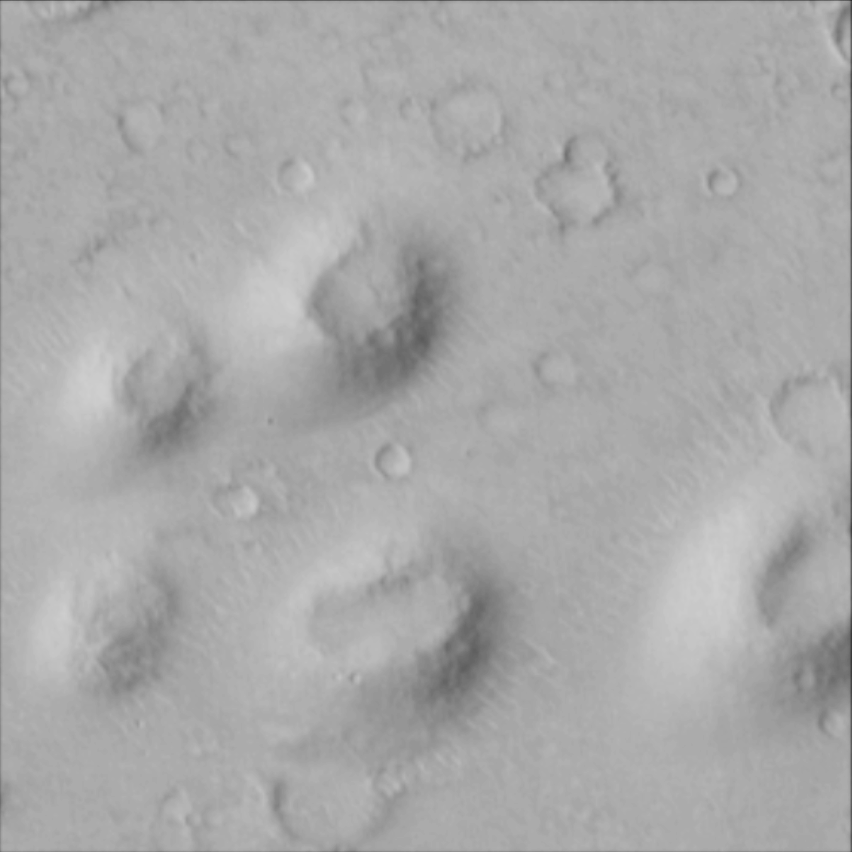}
  \end{subfigure}
  \begin{subfigure}{0.155\columnwidth}
    \centering
    \caption{GT}
    \includegraphics[width=\linewidth]{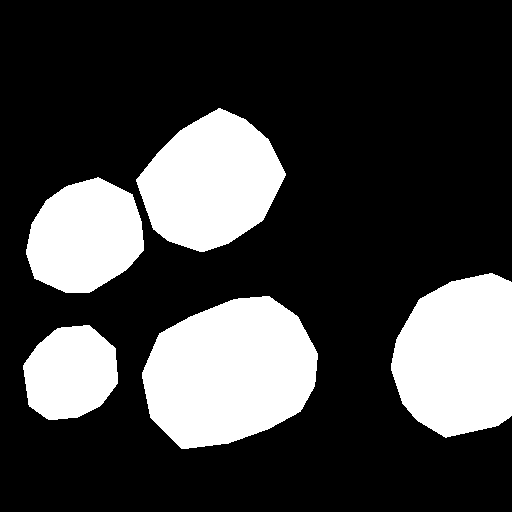}
  \end{subfigure}
  \begin{subfigure}{0.155\columnwidth}
    \centering
    \caption{U-Net}
    \includegraphics[width=\linewidth]{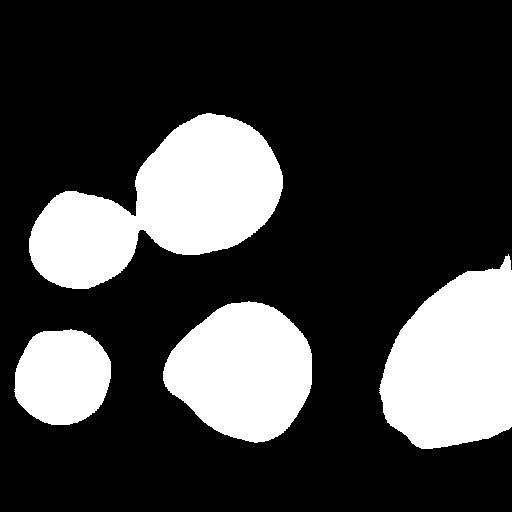}
  \end{subfigure}
  \begin{subfigure}{0.155\columnwidth}
    \centering
    \caption{FPN}
    \includegraphics[width=\linewidth]{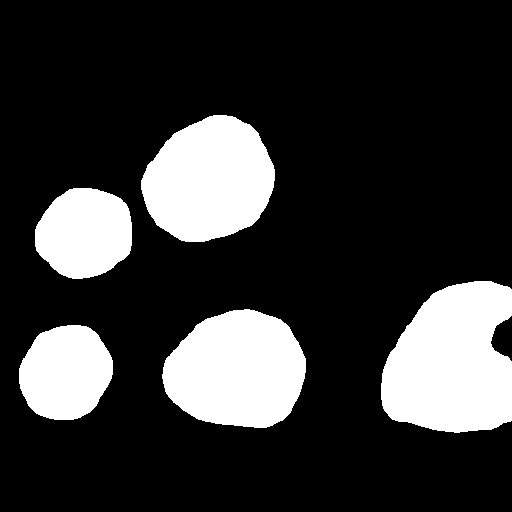}
  \end{subfigure}
  \begin{subfigure}{0.155\columnwidth}
    \centering
    \caption{DeepLab}
    \includegraphics[width=\linewidth]{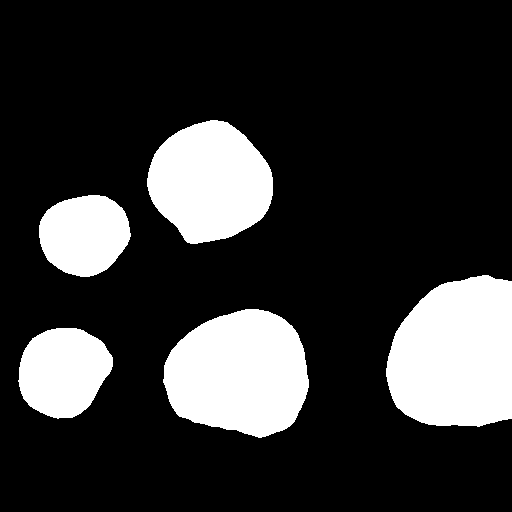}
  \end{subfigure}
  \begin{subfigure}{0.155\columnwidth}
    \centering
    \caption{MA-Net}
    \includegraphics[width=\linewidth]{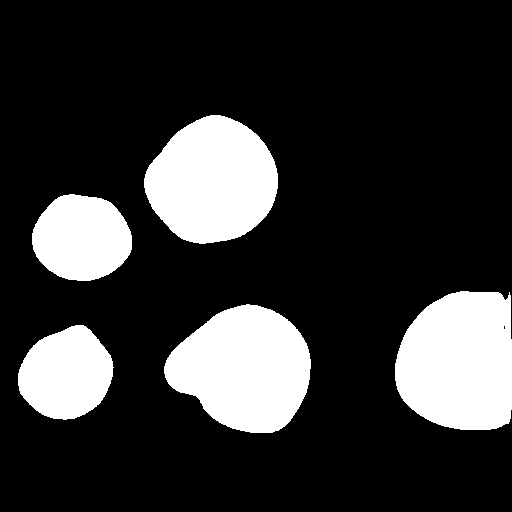}
  \end{subfigure}
    
  \begin{subfigure}{0.155\columnwidth}
    \centering
    \includegraphics[width=\linewidth]{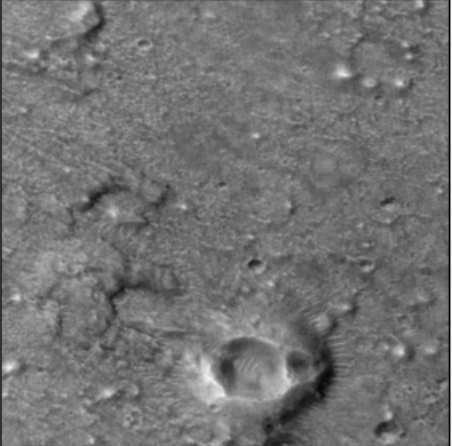}
  \end{subfigure}
  \begin{subfigure}{0.155\columnwidth}
    \centering
    \includegraphics[width=\linewidth]{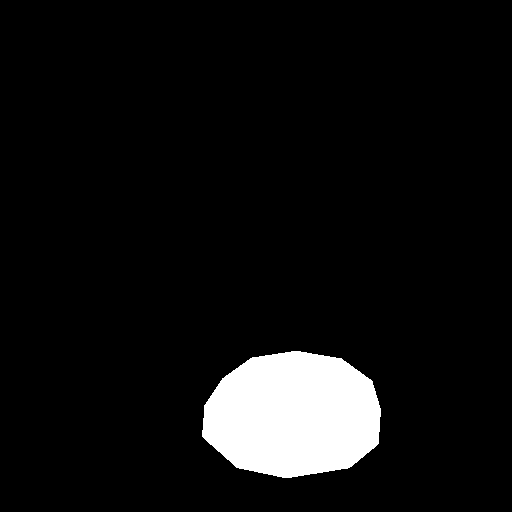}
  \end{subfigure}
  \begin{subfigure}{0.155\columnwidth}
    \centering
    \includegraphics[width=\linewidth]{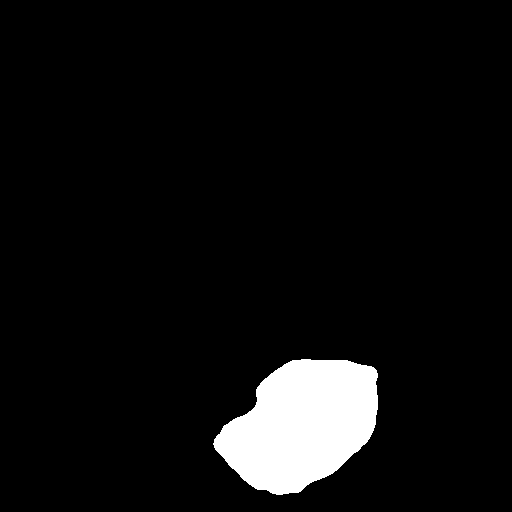}
  \end{subfigure}
  \begin{subfigure}{0.155\columnwidth}
    \centering
    \includegraphics[width=\linewidth]{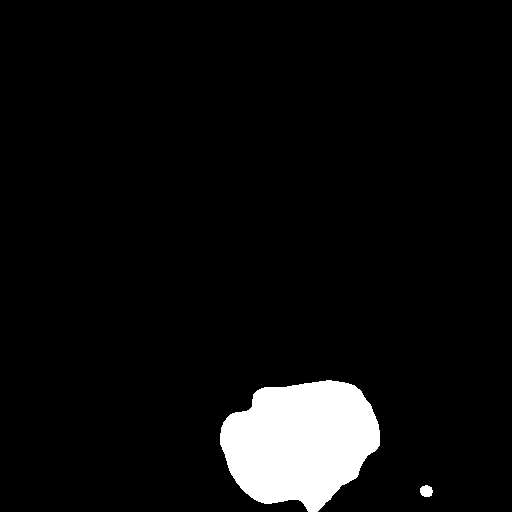}
  \end{subfigure}
  \begin{subfigure}{0.155\columnwidth}
    \centering
    \includegraphics[width=\linewidth]{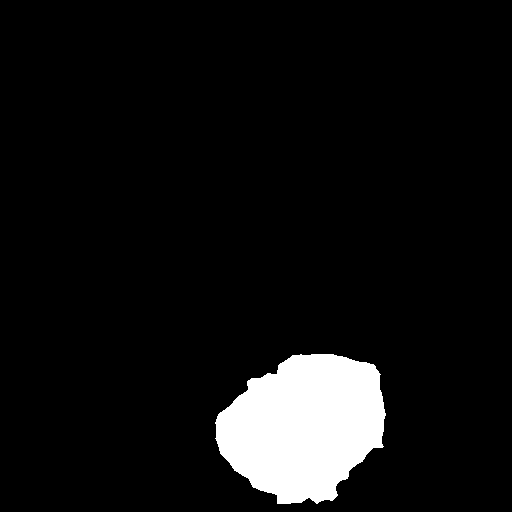}
  \end{subfigure}
  \begin{subfigure}{0.155\columnwidth}
    \centering
    \includegraphics[width=\linewidth]{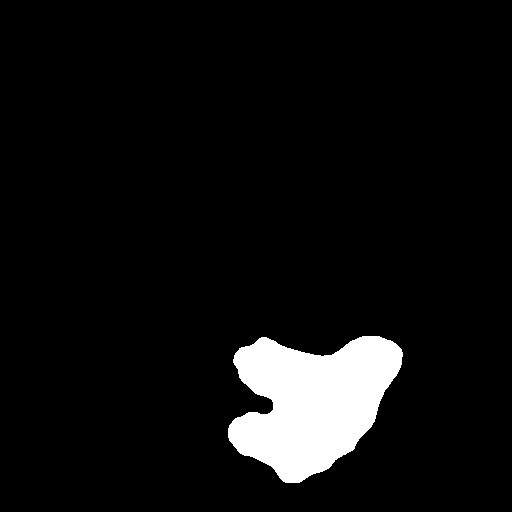}
  \end{subfigure}

  \begin{subfigure}{0.155\columnwidth}
    \centering
    \includegraphics[width=\linewidth]{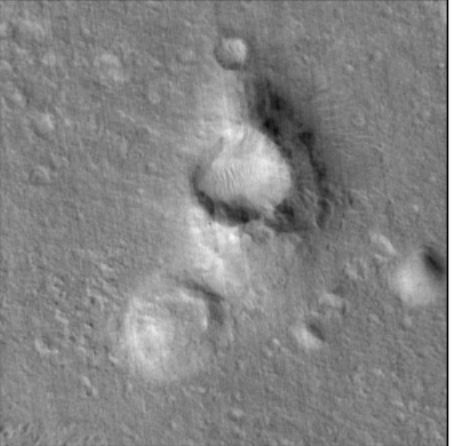}
  \end{subfigure}
  \begin{subfigure}{0.155\columnwidth}
    \centering
    \includegraphics[width=\linewidth]{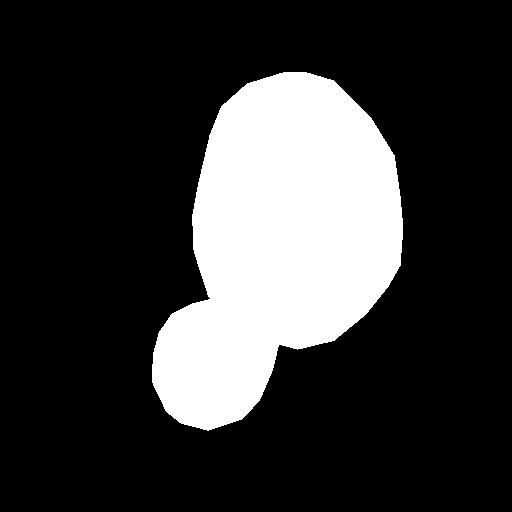}
  \end{subfigure}
  \begin{subfigure}{0.155\columnwidth}
    \centering
    \includegraphics[width=\linewidth]{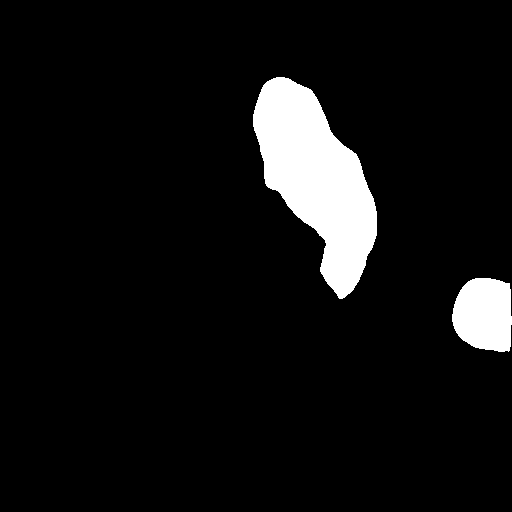}
  \end{subfigure}
  \begin{subfigure}{0.155\columnwidth}
    \centering
    \includegraphics[width=\linewidth]{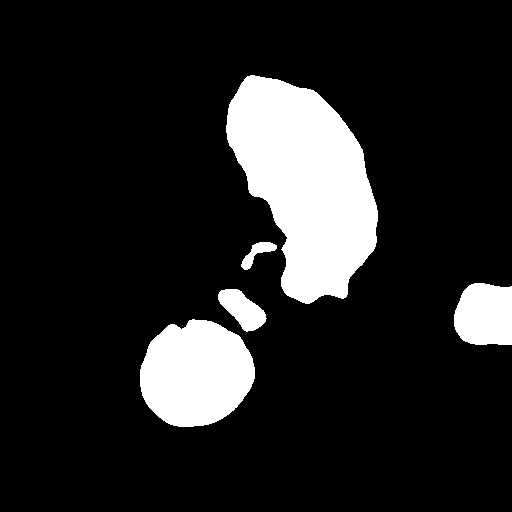}
  \end{subfigure}
  \begin{subfigure}{0.155\columnwidth}
    \centering
    \includegraphics[width=\linewidth]{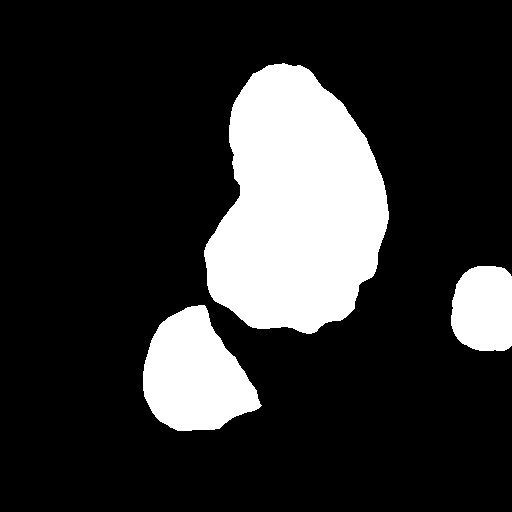}
  \end{subfigure}
  \begin{subfigure}{0.155\columnwidth}
    \centering
    \includegraphics[width=\linewidth]{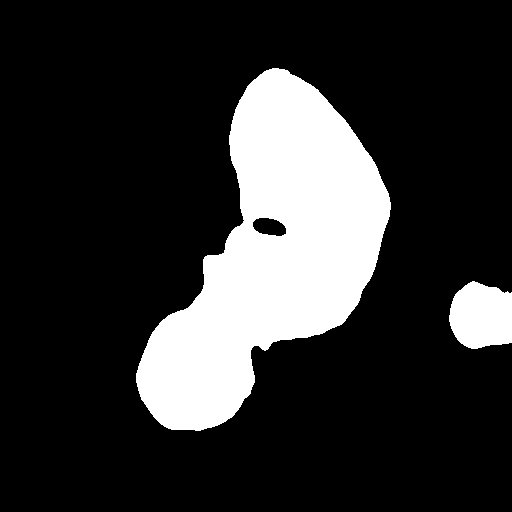}
  \end{subfigure}

  \caption{Illustration of predictions of in-distribution evaluation for single-region training (BM-1). Row 1, row 2, and row 3 represent test data samples from IP, AP, and HP, respectively. Columns c, d, e, and f show predictions from models with their corresponding Input (column a) and Ground Truth (GT) (column b).}
  \vspace{-2mm}
  \label{fig:prediction}
\end{figure}

\paragraph{Segmenting all cones:} Table \ref{tab:bm_1_results} and \ref{tab:bm_2_results} show that pixel precision is higher compared to pixel recall. This indicates that the models are better at reducing false positives than false negatives, which means all true cone pixels are not accurately captured. Object precision and object recall exhibit similar patterns across most cases. Moreover, in BM-1 for IP, the disparity between pixel precision and pixel recall is minimal, however, AP and HP show a larger discrepancy. Figure \ref{fig:prediction} shows similar observations for AP and HP\footnote{Detailed results of BM-1 and BM-2 are shown in Appendix A.}. Identical trend for BM-2, the small and medium categories have higher discrepancies between pixel precision and pixel recall compared to the large cone category.


\vspace{-4mm}

\paragraph{Non-cone patches:} To assess the significance of negative samples, we trained all models for both BMs only on positive samples. Results indicate that models perform poorly for non-cone patches when \textit{trained solely on positive samples} ($\mathcal{T}_p$) compared to those \textit{trained on both positive and negative samples} ($\mathcal{T}_{pn}$). Results show that $\mathcal{T}_p$ performs relatively lower compared to $\mathcal{T}_{pn}$ by $3.94\%$ (for BM-1) and $2.14\%$ (for BM-2) in terms of $A_{FP}$ for non-cone patches. Additionally, performance for $\mathcal{T}_p$ does not show any significant improvement for positive samples, showing relative improvement of $0.09\%$ (for BM-1) and $0.15\%$ (for BM-2) on $A_{FP}$. Hence, we can conclude that negative samples help in training to improve the performance of non-cone patches while maintaining the performance of cone-patches. Detailed results of $\mathcal{T}_p$ are shown in Appendix C (Tables $5$, $6$, $7$, and $8$).

\subsubsection{Benchmark - 1}
\label{subsubsec:bm_1}

\begin{table*}[!ht]
\setlength\tabcolsep{4.0pt}
\setlength{\belowcaptionskip}{-10pt}
\centering
\footnotesize
\resizebox{0.96\linewidth}{!}{
\begin{tabular}{c|c|ccccccccccc|c}
\toprule[1.5pt]
\multirow{2}{*}{\textbf{\begin{tabular}[c]{@{}c@{}}Training Size\\Category\end{tabular}}} &
  \multirow{2}{*}{\textbf{\begin{tabular}[c]{@{}c@{}}Testing Size\\Category\end{tabular}}} &
  \multicolumn{11}{c|}{\textbf{Cone}} &
  \textbf{Non-Cone} \\ \cmidrule{3-14} 
 &
   &
  \multicolumn{1}{c|}{\textbf{\begin{tabular}[c]{@{}c@{}}Mask\\ IoU\end{tabular}}} &
  \multicolumn{1}{c|}{\textbf{\begin{tabular}[c]{@{}c@{}}Pixel\\ IoU\end{tabular}}} &
  \multicolumn{1}{c|}{\textbf{\begin{tabular}[c]{@{}c@{}}Pixel\\ Accuracy\end{tabular}}} &
  \multicolumn{1}{c|}{\textbf{\begin{tabular}[c]{@{}c@{}}Pixel\\ Precision\end{tabular}}} &
  \multicolumn{1}{c|}{\textbf{\begin{tabular}[c]{@{}c@{}}Pixel\\ Recall\end{tabular}}} &
  \multicolumn{1}{c|}{\textbf{\begin{tabular}[c]{@{}c@{}}Panoptic\\ Quality\end{tabular}}} &
  \multicolumn{1}{c|}{\textbf{mAP}} &
  \multicolumn{1}{c|}{\textbf{\begin{tabular}[c]{@{}c@{}}Object\\ IoU\end{tabular}}} &
  \multicolumn{1}{c|}{\textbf{\begin{tabular}[c]{@{}c@{}}Object\\ Accuracy\end{tabular}}} &
  \multicolumn{1}{c|}{\textbf{\begin{tabular}[c]{@{}c@{}}Object\\ Precision\end{tabular}}} &
  \textbf{\begin{tabular}[c]{@{}c@{}}Object\\ Recall\end{tabular}} &
  \textbf{$A_{FP}$} \\ \midrule[1pt]
S &
  S &
  \multicolumn{1}{c|}{26.65} &
  \multicolumn{1}{c|}{25.85} &
  \multicolumn{1}{c|}{97.61} &
  \multicolumn{1}{c|}{76.50} &
  \multicolumn{1}{c|}{32.40} &
  \multicolumn{1}{c|}{16.64} &
  \multicolumn{1}{c|}{10.26} &
  \multicolumn{1}{c|}{25.21} &
  \multicolumn{1}{c|}{19.34} &
  \multicolumn{1}{c|}{25.20} &
  24.73 &
  0.12 \\ \midrule
M &
  M &
  \multicolumn{1}{c|}{39.84} &
  \multicolumn{1}{c|}{39.66} &
  \multicolumn{1}{c|}{92.90} &
  \multicolumn{1}{c|}{81.51} &
  \multicolumn{1}{c|}{45.40} &
  \multicolumn{1}{c|}{27.67} &
  \multicolumn{1}{c|}{12.65} &
  \multicolumn{1}{c|}{43.15} &
  \multicolumn{1}{c|}{32.65} &
  \multicolumn{1}{c|}{42.42} &
  39.00 &
  0.35 \\ \midrule
L &
  L &
  \multicolumn{1}{c|}{59.40} &
  \multicolumn{1}{c|}{63.13} &
  \multicolumn{1}{c|}{90.68} &
  \multicolumn{1}{c|}{84.50} &
  \multicolumn{1}{c|}{72.65} &
  \multicolumn{1}{c|}{47.89} &
  \multicolumn{1}{c|}{22.55} &
  \multicolumn{1}{c|}{62.32} &
  \multicolumn{1}{c|}{55.03} &
  \multicolumn{1}{c|}{62.36} &
  66.80 &
  0.88 \\ \midrule
\multirow{2}{*}{S + M} &
  S &
  \multicolumn{1}{c|}{30.40} &
  \multicolumn{1}{c|}{29.47} &
  \multicolumn{1}{c|}{97.67} &
  \multicolumn{1}{c|}{77.65} &
  \multicolumn{1}{c|}{37.56} &
  \multicolumn{1}{c|}{20.56} &
  \multicolumn{1}{c|}{11.86} &
  \multicolumn{1}{c|}{30.41} &
  \multicolumn{1}{c|}{24.02} &
  \multicolumn{1}{c|}{31.81} &
  29.30 &
  0.21 \\
 &
  M &
  \multicolumn{1}{c|}{40.31} &
  \multicolumn{1}{c|}{38.54} &
  \multicolumn{1}{c|}{92.84} &
  \multicolumn{1}{c|}{85.79} &
  \multicolumn{1}{c|}{44.01} &
  \multicolumn{1}{c|}{29.72} &
  \multicolumn{1}{c|}{15.57} &
  \multicolumn{1}{c|}{43.12} &
  \multicolumn{1}{c|}{35.64} &
  \multicolumn{1}{c|}{44.69} &
  42.09 &
  0.35 \\ \midrule
\multirow{2}{*}{M + L} &
  M &
  \multicolumn{1}{c|}{42.78} &
  \multicolumn{1}{c|}{42.31} &
  \multicolumn{1}{c|}{92.69} &
  \multicolumn{1}{c|}{79.04} &
  \multicolumn{1}{c|}{49.82} &
  \multicolumn{1}{c|}{29.38} &
  \multicolumn{1}{c|}{13.90} &
  \multicolumn{1}{c|}{45.12} &
  \multicolumn{1}{c|}{34.50} &
  \multicolumn{1}{c|}{46.72} &
  39.94 &
  0.51 \\
 &
  L &
  \multicolumn{1}{c|}{54.91} &
  \multicolumn{1}{c|}{57.67} &
  \multicolumn{1}{c|}{89.90} &
  \multicolumn{1}{c|}{90.19} &
  \multicolumn{1}{c|}{62.63} &
  \multicolumn{1}{c|}{45.05} &
  \multicolumn{1}{c|}{20.58} &
  \multicolumn{1}{c|}{59.02} &
  \multicolumn{1}{c|}{52.29} &
  \multicolumn{1}{c|}{59.08} &
  63.00 &
  2.38 \\ \midrule
\multirow{2}{*}{S + L} &
  S &
  \multicolumn{1}{c|}{22.58} &
  \multicolumn{1}{c|}{22.19} &
  \multicolumn{1}{c|}{97.17} &
  \multicolumn{1}{c|}{78.60} &
  \multicolumn{1}{c|}{29.71} &
  \multicolumn{1}{c|}{13.66} &
  \multicolumn{1}{c|}{6.22} &
  \multicolumn{1}{c|}{21.43} &
  \multicolumn{1}{c|}{16.00} &
  \multicolumn{1}{c|}{21.45} &
  20.63 &
  0.89 \\
 &
  L &
  \multicolumn{1}{c|}{52.08} &
  \multicolumn{1}{c|}{56.31} &
  \multicolumn{1}{c|}{89.48} &
  \multicolumn{1}{c|}{87.50} &
  \multicolumn{1}{c|}{63.24} &
  \multicolumn{1}{c|}{41.37} &
  \multicolumn{1}{c|}{17.83} &
  \multicolumn{1}{c|}{54.19} &
  \multicolumn{1}{c|}{47.37} &
  \multicolumn{1}{c|}{52.86} &
  59.5 &
  2.12 \\ \midrule
\end{tabular}
}
\caption{Results for BM-2 for all metrics on $\mathcal{D}_{id}$. Here, the results in each row are the average across 4 models. See Appendix B (Table 3 and 4) for individual model results.}
\label{tab:bm_2_results}
\end{table*}

\paragraph{Multi-Region training:} In natural language processing and general-domain computer vision, it has been established that models outperform multi-domain learning compared to single-task learning \cite{liu2019multi, crawshaw2020multi}. However, comparing results on $\mathcal{D}_{id}$ for single-region and multi-region training (from Figure \ref{fig:iou_bm_1}) shows lower performance for multi-region. This suggests that tested models struggle to learn the multiple data distributions in multi-region scenarios, making global mapping a challenging task.

\vspace{-4mm}

\paragraph{Performance differences across regions:} From Figure \ref{fig:iou_bm_1} and Table \ref{tab:bm_1_results}, it can be observed that test metrics are higher for the IP region compared to AP and HP. The average number of cones per patch for IP, AP, and HP are 3.52, 1.92, and 2.08, respectively. The higher cone density within IP compared to the other two regions, even though the number of training patches is smaller in IP, may explain the higher performance. 

Figure \ref{fig:iou_bm_1} and Table \ref{tab:bm_1_results} show performance is worst in HP for the single and multi-region training even on $\mathcal{D}_{id}$. Figure \ref{fig:teaser_bm_1} shows data samples from all 3 regions. This may be due to the greater variability in cone appearance in HP compared to the other regions, which can be seen in the example images in Figure \ref{fig:teaser_bm_1}. As discussed in \textsection \ref{sec:benchmarks} and \textsection \ref{subsec:data_collection}, cones in HP have greater diversity in terms of size, shape (circular, elongated, and clustered cones), and brightness/appearance, and it is possible a small percentage of annotations have errors. This makes HP a challenging region for cone segmentation. 




\subsubsection{Benchmark - 2}
\label{subsubsec:bm_2}

\paragraph{Model behavior \textit{vs.} cone size:} From Figure \ref{fig:iou_bm_2}, we can observe that performance for all 4 models increases as cone size increases for single-category and multi-category training. This effect is likely a result of the larger size category providing the model with more information about the cone morphology compared to other smaller categories.

\vspace{-4mm}

\paragraph{Multi-category Training:} Similar to BM-1, in BM-2 multi-category performs worse than single-category training, except for the combination of small and medium categories. When training with small and medium combined, there is a significant improvement in object-wise metrics for small cones, while pixel-wise metrics show slight improvement. This indicates that the inclusion of the medium category has a positive impact on the performance of the small category. Although the model may not be able to precisely mask the cone at the pixel level, it is able to accurately localize the cone compared to training with only the small category.

\vspace{-4mm}

\paragraph{Results on Non-cone patches:} From Table \ref{tab:bm_2_results}, we can observe that $A_{FP}$ increases for non-cone patches, as we go from a small to large category. This is due to the fact that when cone size in the patches increases across all training data, the model encounters more samples with larger cone areas. For instance, a model trained on the small category exhibits a lower false positive area ($0.12$) in non-cone patches compared to the one trained on the large category ($0.88$).






\section{Conclusion}
\label{sec:conclusion}


Despite the importance of cone segmentation in planetary science and the potential for computer vision techniques to facilitate this task, cone segmentation is under-explored and no publicly available dataset exists. In this research, we introduced \dataset{}, a benchmark for cone segmentation in Mars orbital images. We proposed two benchmark tasks based on \dataset{}: (i) Spatial Generalization and (ii) Cone-size Generalization. We evaluated four widely-used segmentation-based models for these tasks. Results show that for both benchmark tasks, existing models struggle in segmenting cones accurately for both in-distribution and out-of-distribution sub-groups. Furthermore, the evaluation of multi-region and multi-category training shows that models do not generalize for multi-domain learning. To enhance the model's performance, various techniques can be employed: ($1$) Employing a pixel-wise ensemble by combining the outputs of multiple models can benefit. ($2$) Histogram Matching can be used to improve the results of multi-region training as different lightning and brightness across regions can affect the model's performance. In the future, we plan to expand \dataset{} to include additional regions on Mars and evaluate model performance in these additional geologic settings. The dataset and metadata provided in \dataset{} enable researchers to develop customized models that may also incorporate context from metadata in model learning. We hope that \dataset{} will facilitate the development of models for cone segmentation and global mapping of cones and other important features on Mars and ultimately improve scientists' understanding of the Red Planet.


{
    \small
    \bibliographystyle{ieee_fullname}
    \bibliography{egbib}
}

\end{document}


\title{ConeQuest: A Benchmark for Cone Segmentation on Mars\\(\textit{Supplementary Material})}

\author{
Mirali Purohit \quad \quad Jacob Adler \quad \quad Hannah Kerner
\\
Arizona State University
\\
\small{\texttt{\{mpurohi3, jbadler2, hkerner\}@asu.edu}}
}


\maketitle



\appendix

\section{Visualization of Predictions}
\label{sec:visualization}

This section discusses the visualization of predictions from models and compares inter-region and inter-size category results.

\subsection{Single-region Evaluation}
\label{subsec:bm_1}

\begin{figure*}

  \centering
  \begin{subfigure}{0.26\columnwidth}
    \centering
    \caption{\textbf{Input}}
  \end{subfigure}
  \begin{subfigure}{0.26\columnwidth}
    \centering
    \caption{\textbf{GT}}
  \end{subfigure}
  \begin{subfigure}{0.26\columnwidth}
    \centering
    \caption{\textbf{U-Net}}
  \end{subfigure}
  \begin{subfigure}{0.26\columnwidth}
    \centering
    \caption{\textbf{FPN}}
  \end{subfigure}
  \begin{subfigure}{0.26\columnwidth}
    \centering
    \caption{\textbf{DeepLab}}
  \end{subfigure}
  \begin{subfigure}{0.26\columnwidth}
    \centering
    \caption{\textbf{MANet}}
  \end{subfigure}

  \centering
  \begin{subfigure}{2\columnwidth}
    \centering
    \includegraphics[width=0.13\columnwidth]{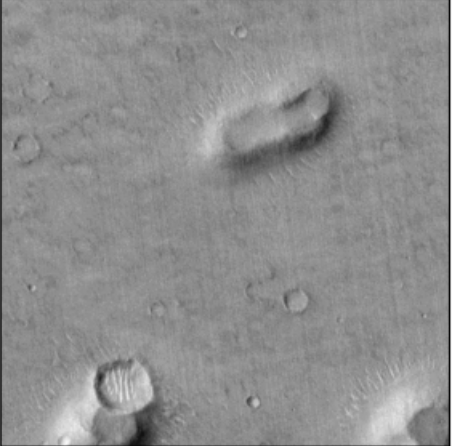}
    \includegraphics[width=0.13\columnwidth]{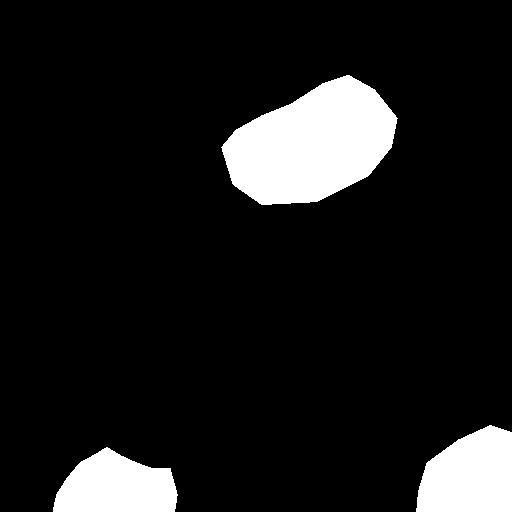}
    \includegraphics[width=0.13\columnwidth]{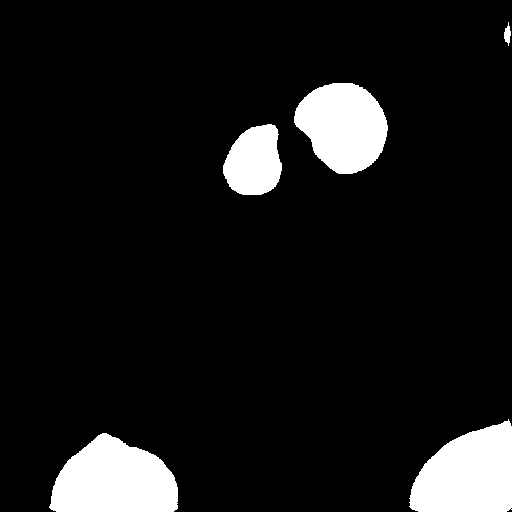}
    \includegraphics[width=0.13\columnwidth]{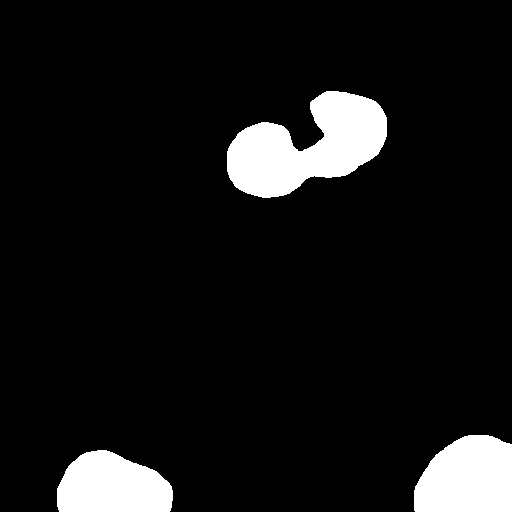}
    \includegraphics[width=0.13\columnwidth]{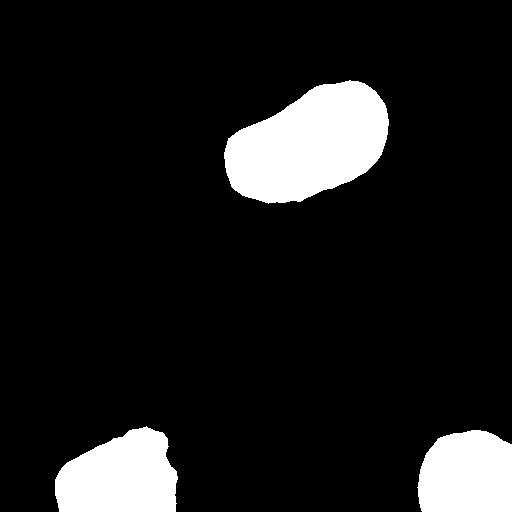}
    \includegraphics[width=0.13\columnwidth]{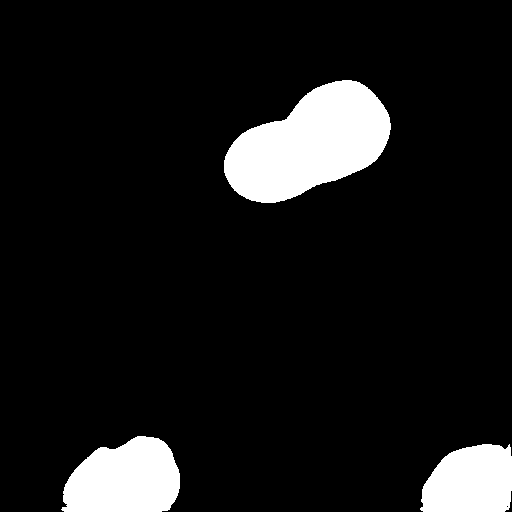}
  \end{subfigure}

  \centering
  \begin{subfigure}{2\columnwidth}
    \centering
    \includegraphics[width=0.13\columnwidth]{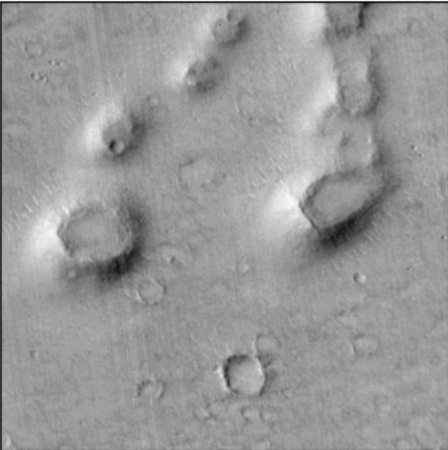}
    \includegraphics[width=0.13\columnwidth]{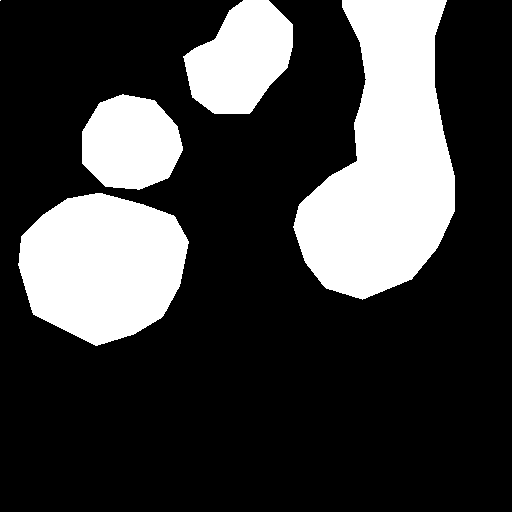}
    \includegraphics[width=0.13\columnwidth]{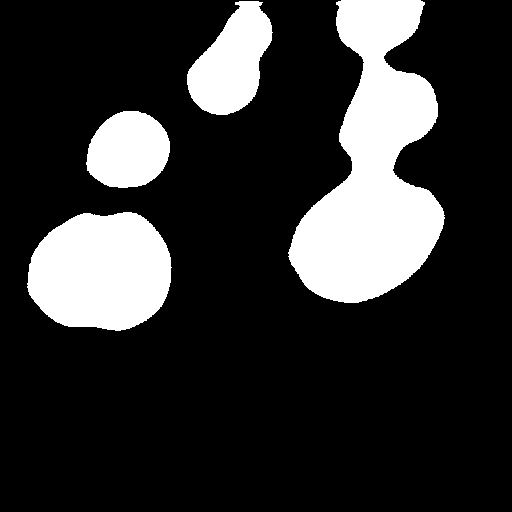}
    \includegraphics[width=0.13\columnwidth]{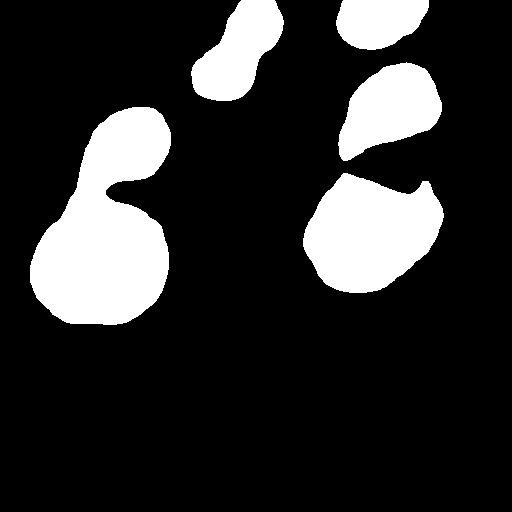}
    \includegraphics[width=0.13\columnwidth]{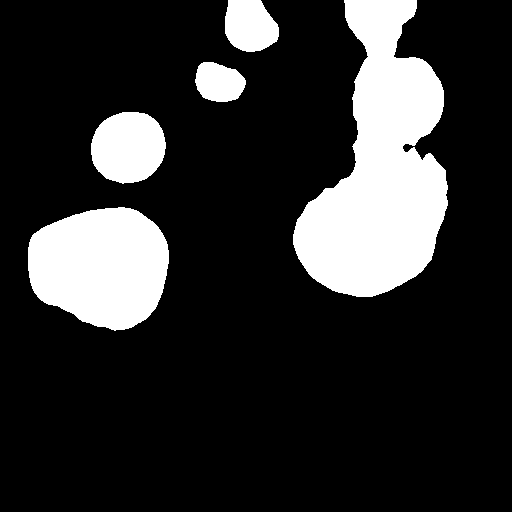}
    \includegraphics[width=0.13\columnwidth]{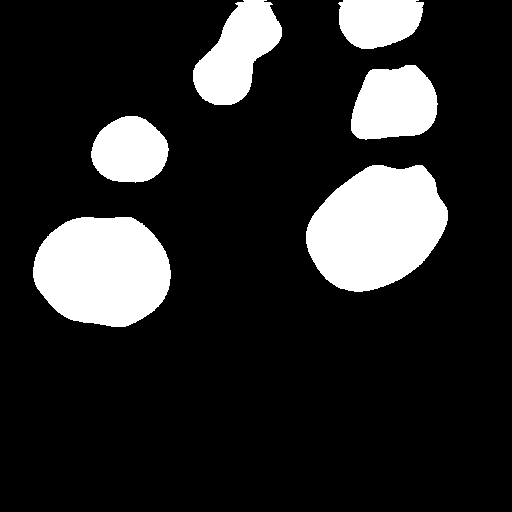}
  \end{subfigure}

  \centering
  \begin{subfigure}{2\columnwidth}
    \centering
    \includegraphics[width=0.13\columnwidth]{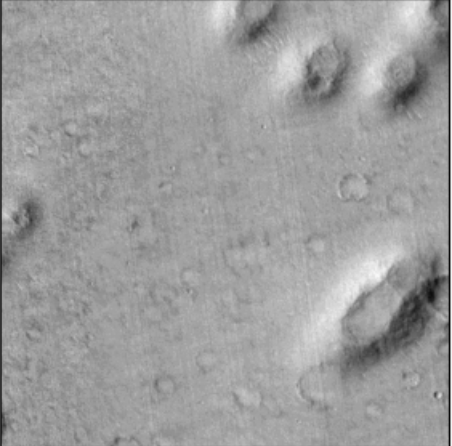}
    \includegraphics[width=0.13\columnwidth]{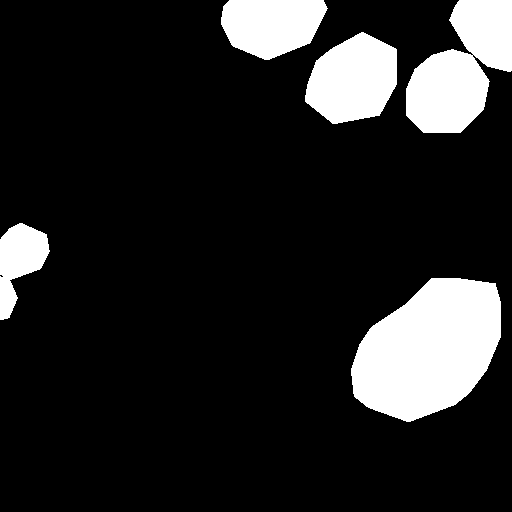}
    \includegraphics[width=0.13\columnwidth]{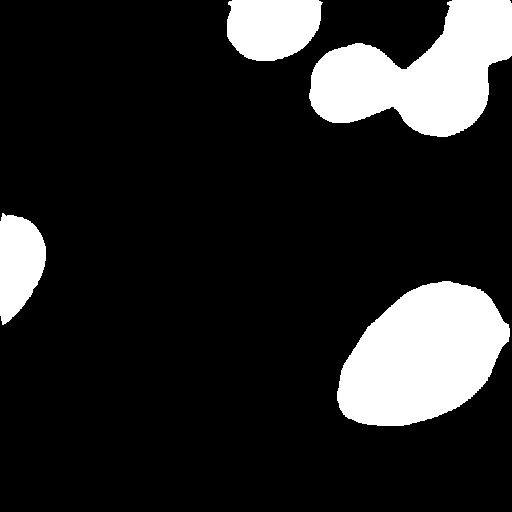}
    \includegraphics[width=0.13\columnwidth]{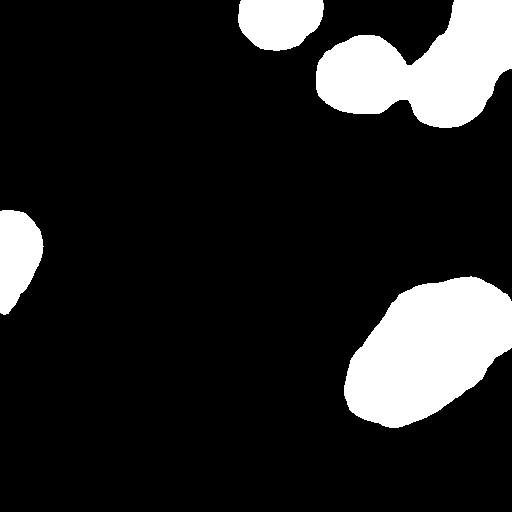}
    \includegraphics[width=0.13\columnwidth]{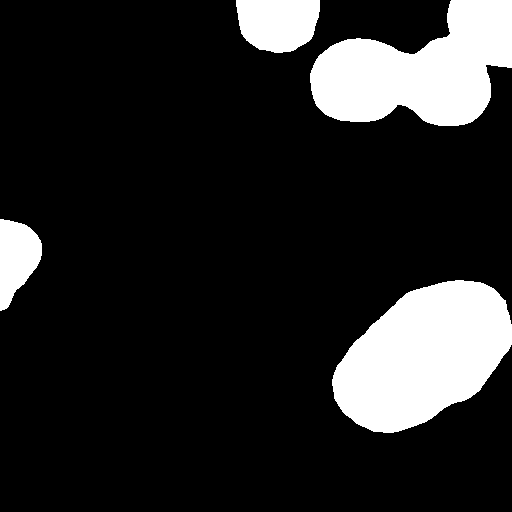}
    \includegraphics[width=0.13\columnwidth]{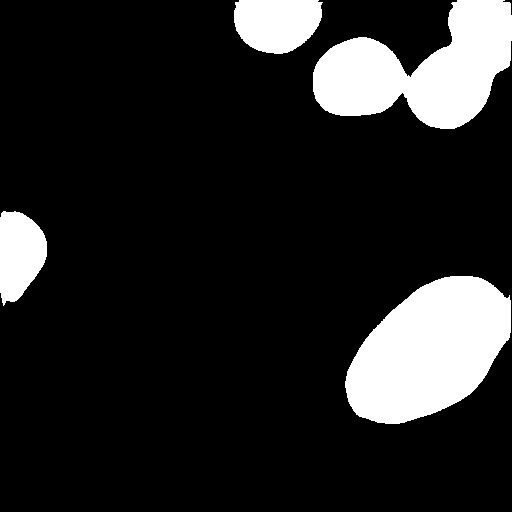}
    \caption*{\textbf{Isidis Planitia}}
  \end{subfigure}

  \centering
  \begin{subfigure}{2\columnwidth}
    \centering
    \includegraphics[width=0.13\columnwidth]{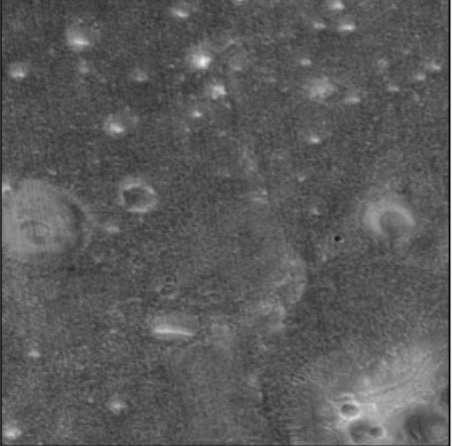}
    \includegraphics[width=0.13\columnwidth]{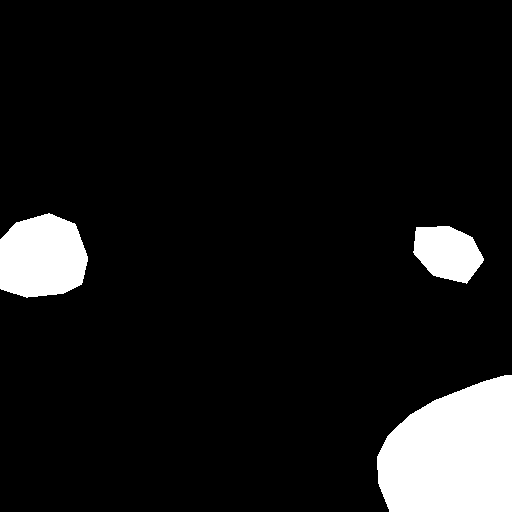}
    \includegraphics[width=0.13\columnwidth]{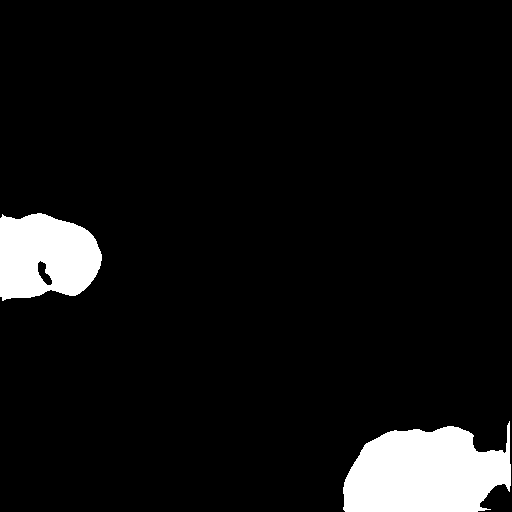}
    \includegraphics[width=0.13\columnwidth]{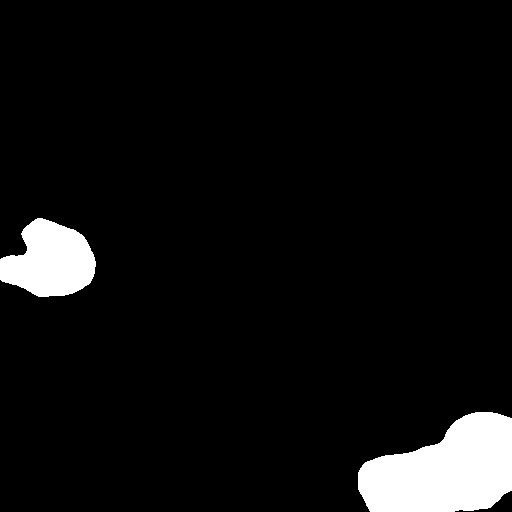}
    \includegraphics[width=0.13\columnwidth]{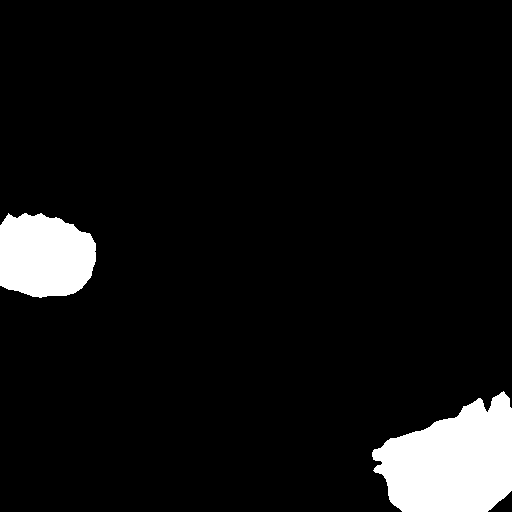}
    \includegraphics[width=0.13\columnwidth]{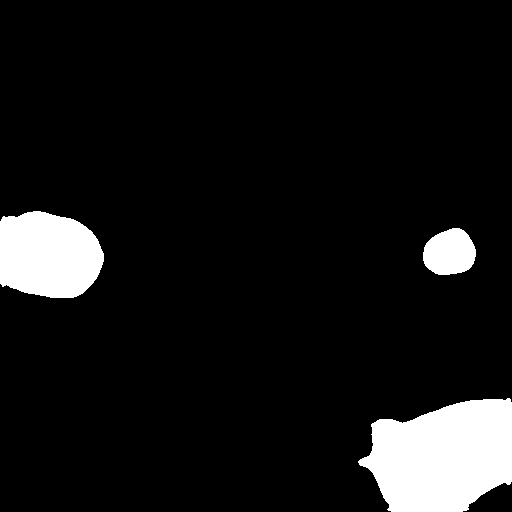}
  \end{subfigure}

  \centering
  \begin{subfigure}{2\columnwidth}
    \centering
    \includegraphics[width=0.13\columnwidth]{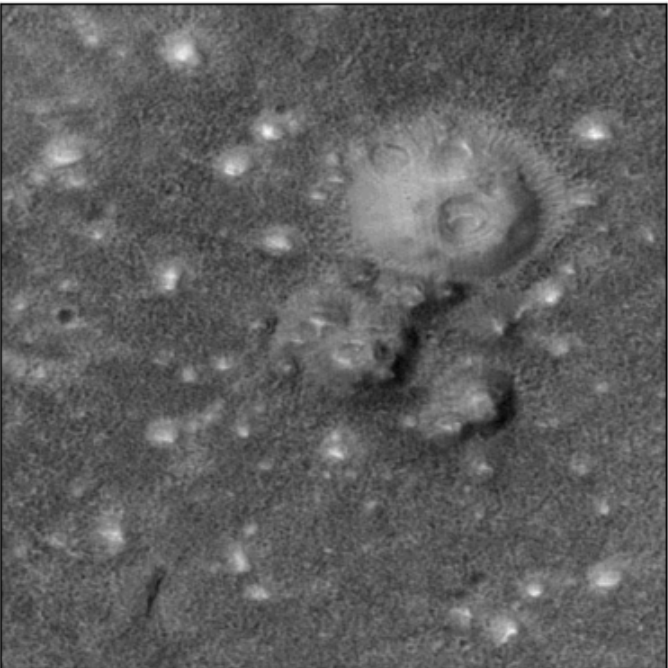}
    \includegraphics[width=0.13\columnwidth]{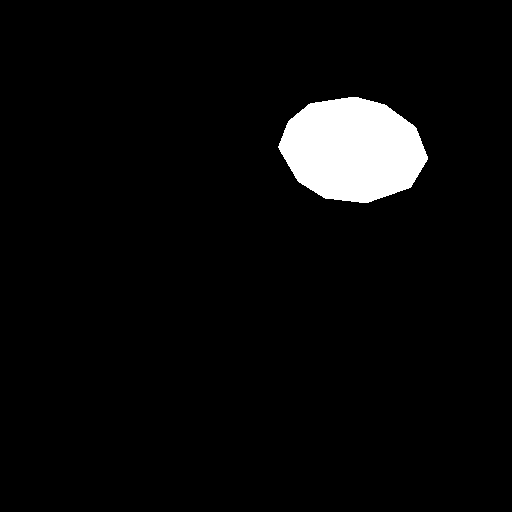}
    \includegraphics[width=0.13\columnwidth]{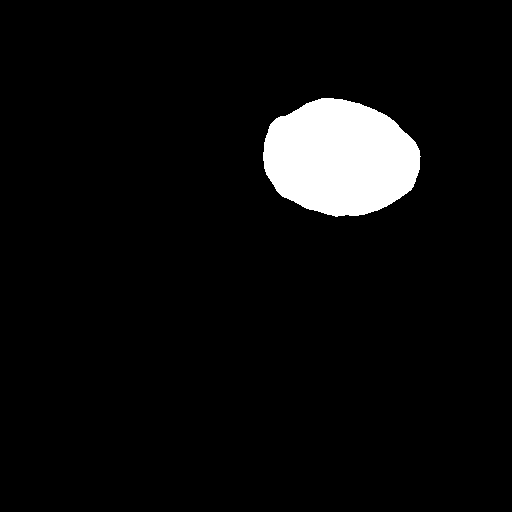}
    \includegraphics[width=0.13\columnwidth]{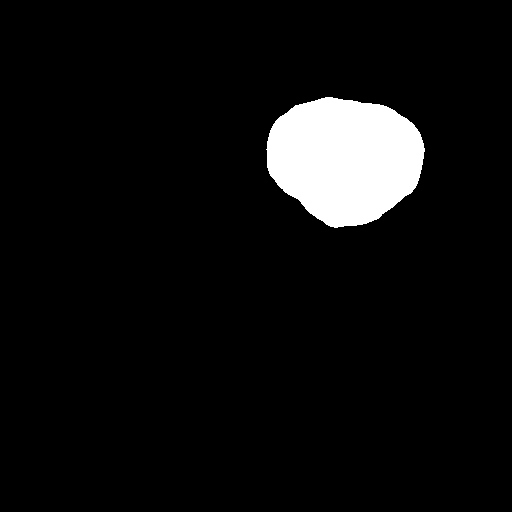}
    \includegraphics[width=0.13\columnwidth]{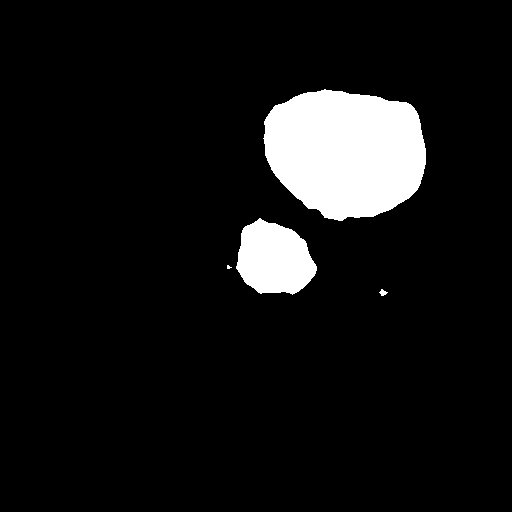}
    \includegraphics[width=0.13\columnwidth]{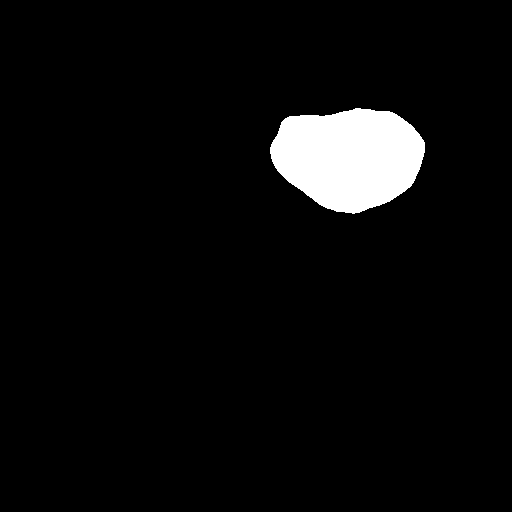}
  \end{subfigure}

  \centering
  \begin{subfigure}{2\columnwidth}
    \centering
    \includegraphics[width=0.13\columnwidth]{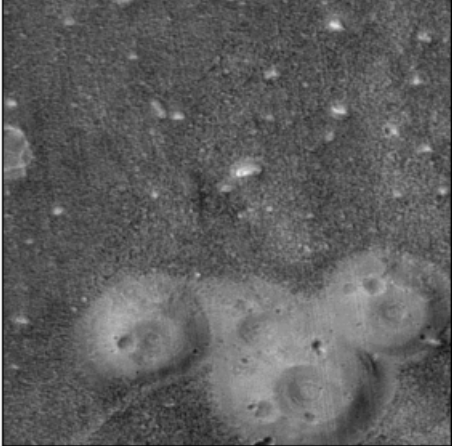}
    \includegraphics[width=0.13\columnwidth]{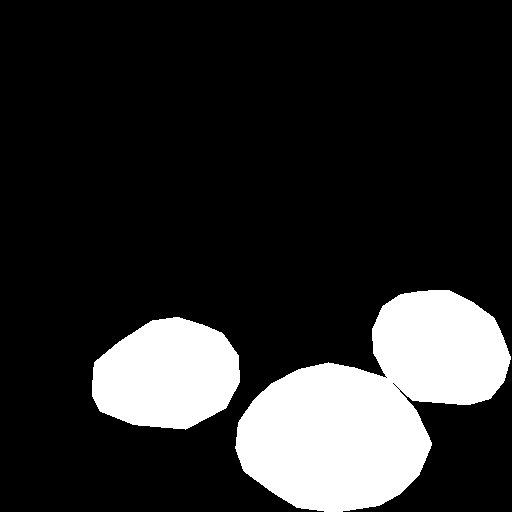}
    \includegraphics[width=0.13\columnwidth]{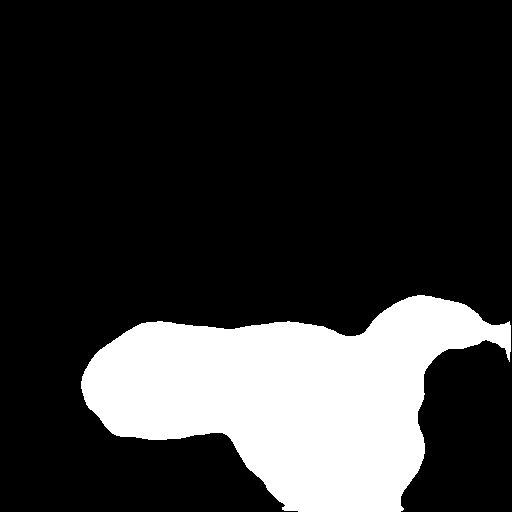}
    \includegraphics[width=0.13\columnwidth]{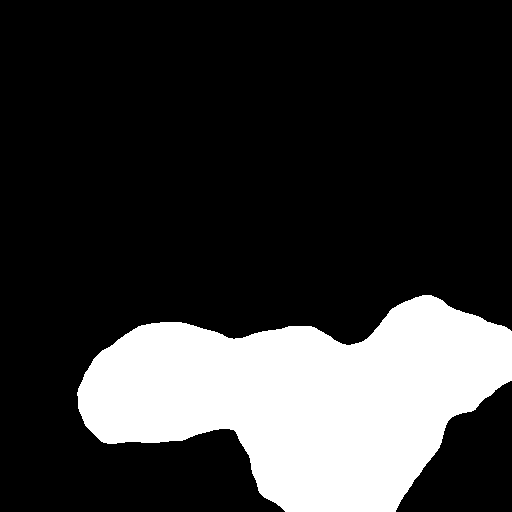}
    \includegraphics[width=0.13\columnwidth]{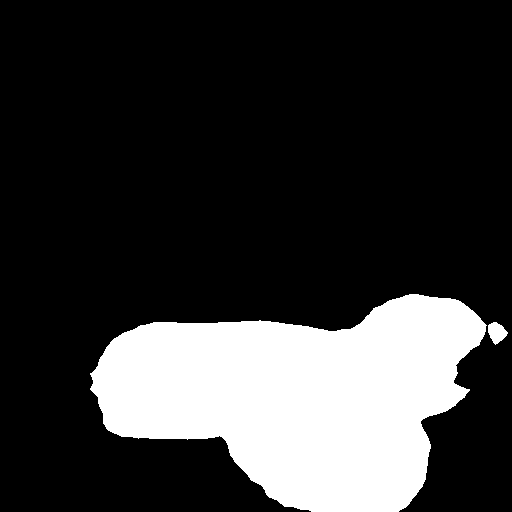}
    \includegraphics[width=0.13\columnwidth]{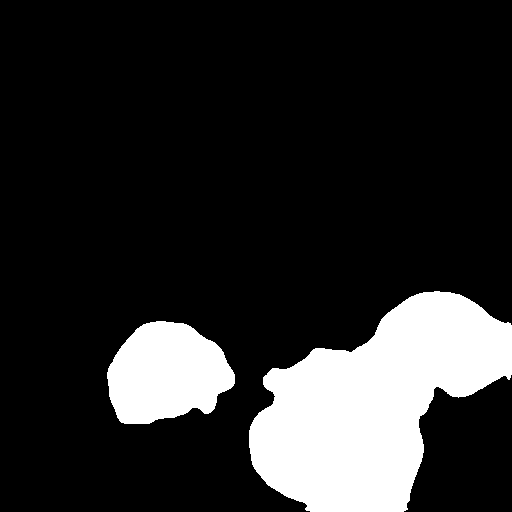}
    \caption*{\textbf{Acidalia Planitia}}
  \end{subfigure}

  \centering
  \begin{subfigure}{2\columnwidth}
    \centering
    \includegraphics[width=0.13\columnwidth]{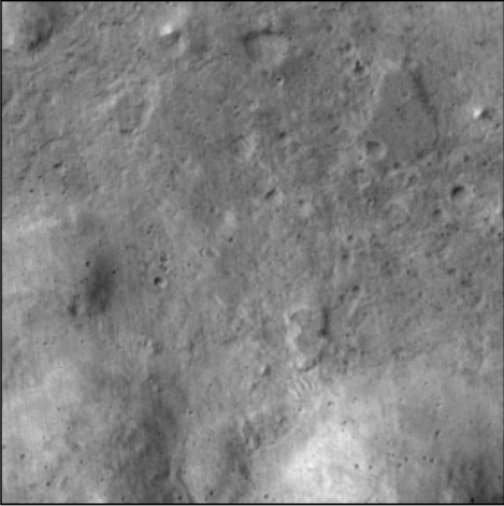}
    \includegraphics[width=0.13\columnwidth]{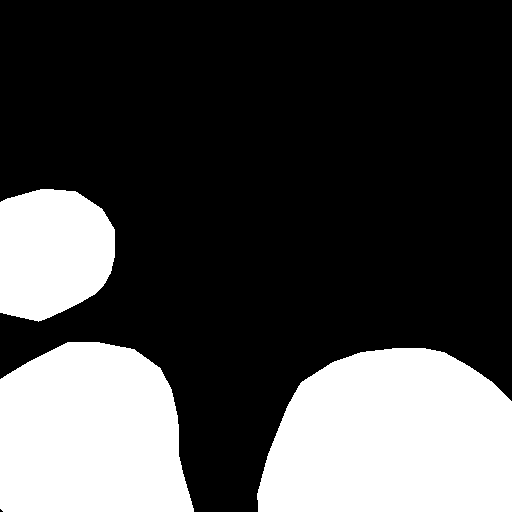}
    \includegraphics[width=0.13\columnwidth]{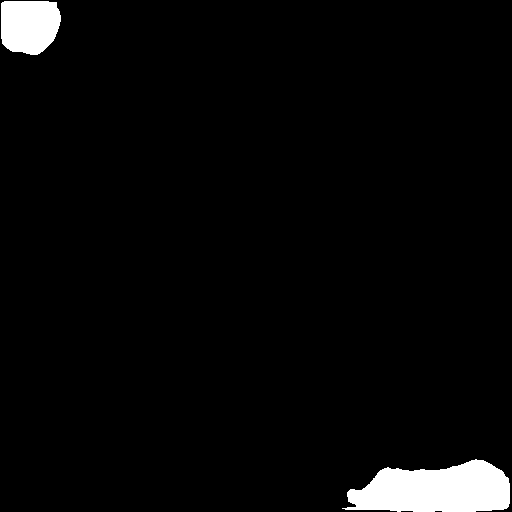}
    \includegraphics[width=0.13\columnwidth]{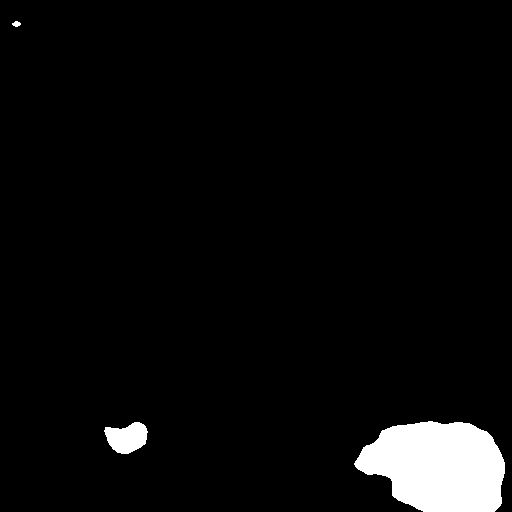}
    \includegraphics[width=0.13\columnwidth]{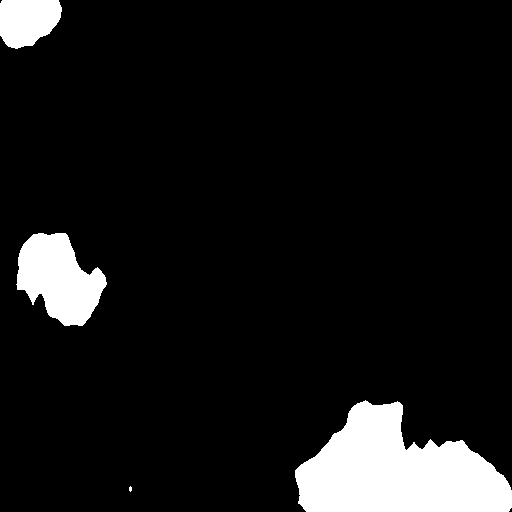}
    \includegraphics[width=0.13\columnwidth]{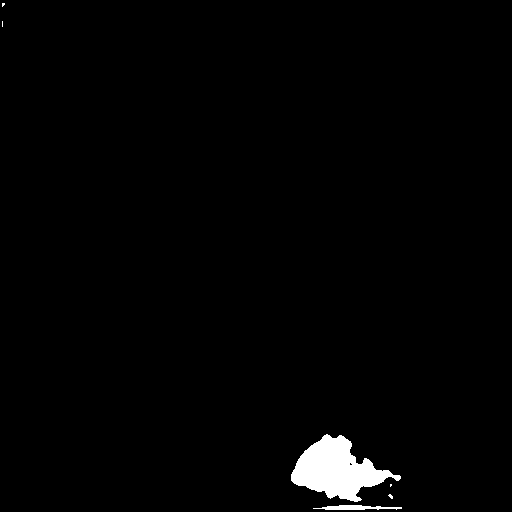}
  \end{subfigure}

  \centering
  \begin{subfigure}{2\columnwidth}
    \centering
    \includegraphics[width=0.13\columnwidth]{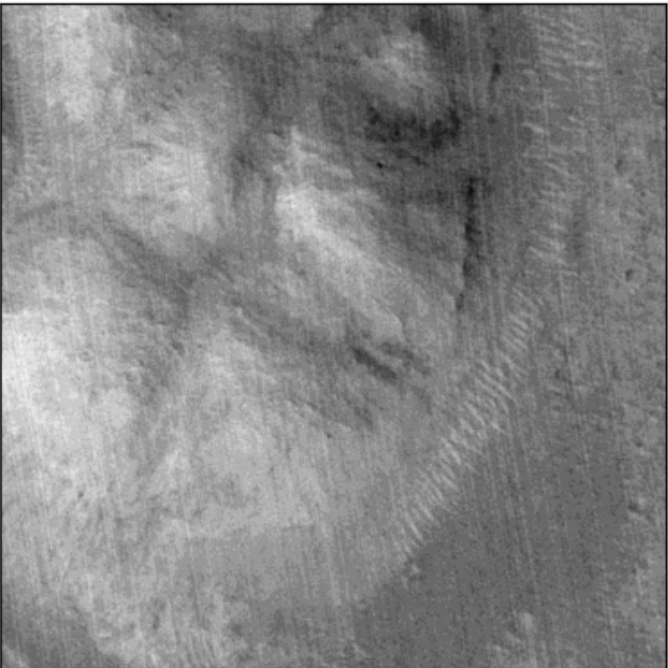}
    \includegraphics[width=0.13\columnwidth]{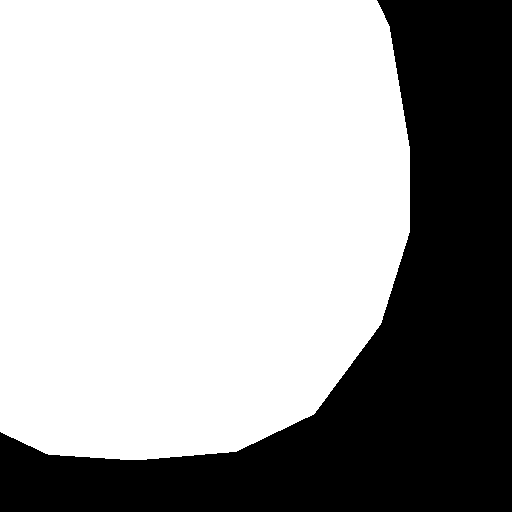}
    \includegraphics[width=0.13\columnwidth]{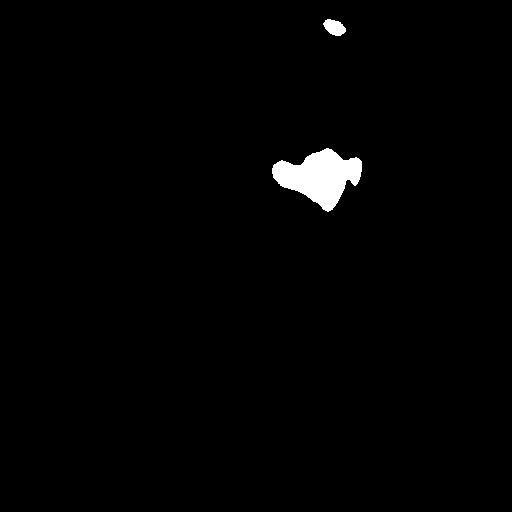}
    \includegraphics[width=0.13\columnwidth]{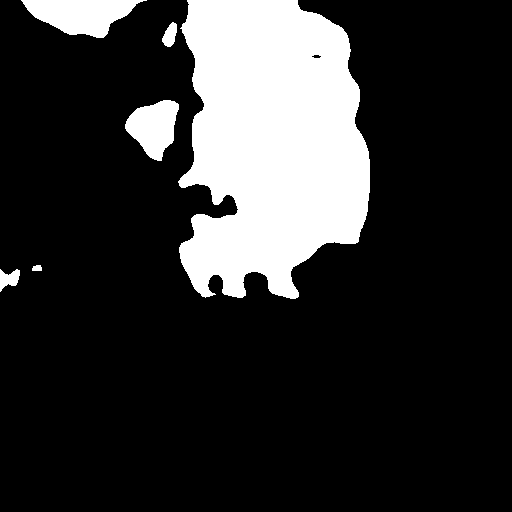}
    \includegraphics[width=0.13\columnwidth]{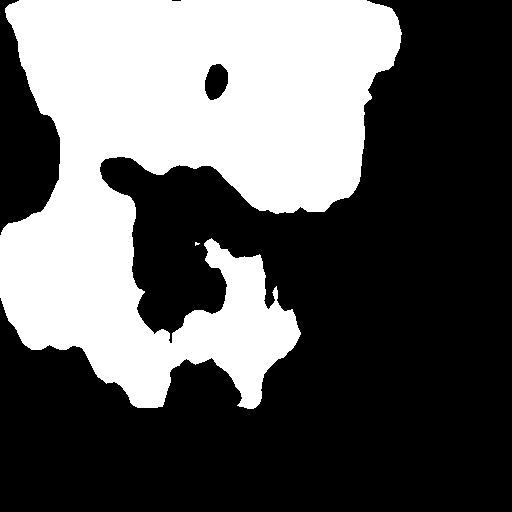}
    \includegraphics[width=0.13\columnwidth]{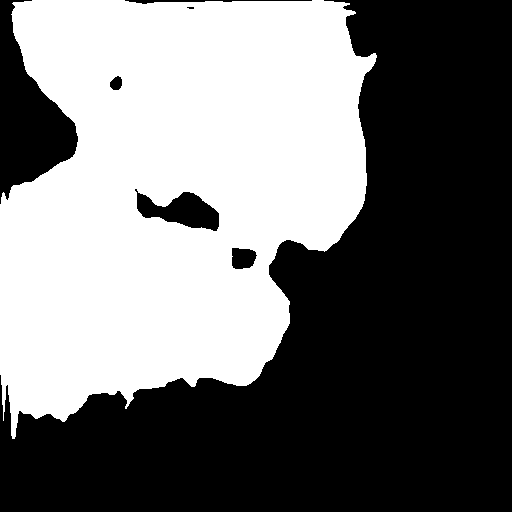}
  \end{subfigure}

  \centering
  \begin{subfigure}{2\columnwidth}
    \centering
    \includegraphics[width=0.13\columnwidth]{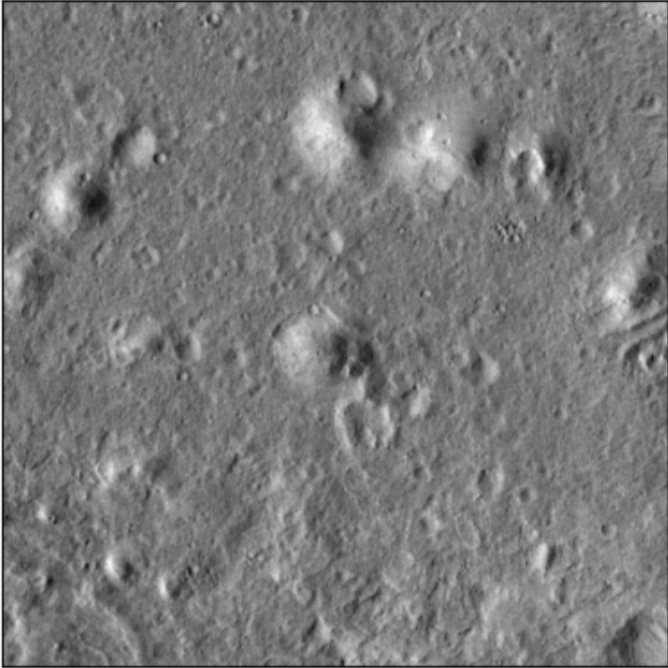}
    \includegraphics[width=0.13\columnwidth]{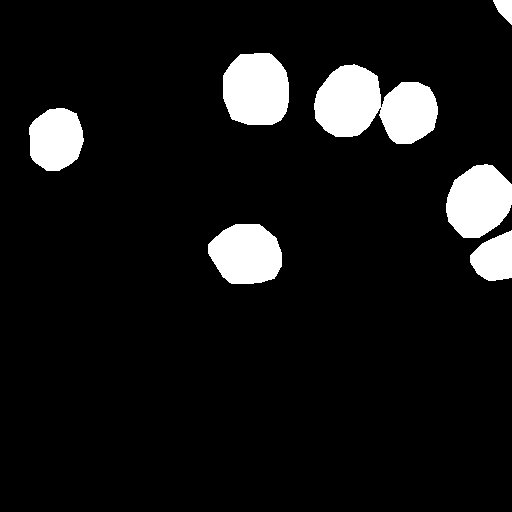}
    \includegraphics[width=0.13\columnwidth]{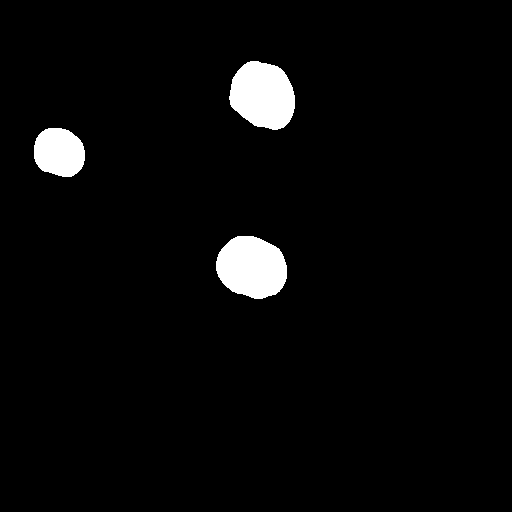}
    \includegraphics[width=0.13\columnwidth]{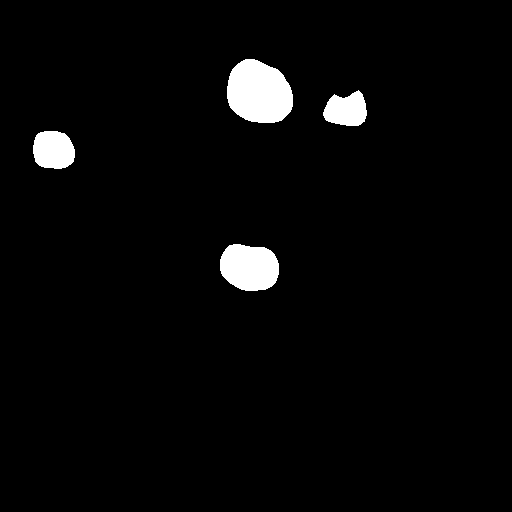}
    \includegraphics[width=0.13\columnwidth]{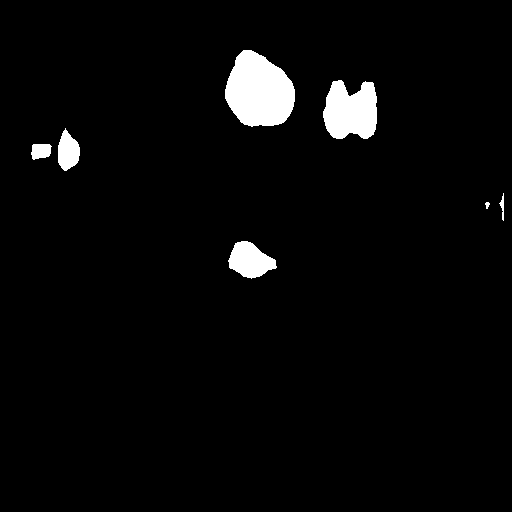}
    \includegraphics[width=0.13\columnwidth]{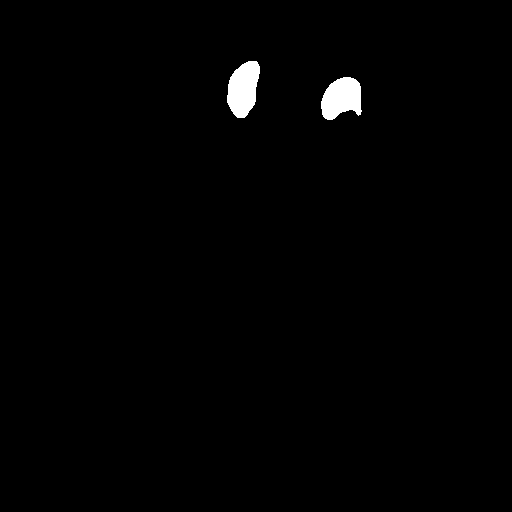}
    \caption*{\textbf{Hypanis}}
  \end{subfigure}

  \caption{Illustration of predictions of in-distribution evaluation for single-region training (BM-1). Columns c, d, e, and f show prediction from all models with their corresponding Input (column a) and Ground Truth (GT) (column b).}
  \label{fig:example_data_bm_1}

\end{figure*}

Figure \ref{fig:example_data_bm_1} shows the visualization of prediction for the evaluation of in-distribution data for single-region training. From Figure \ref{fig:example_data_bm_1}, it can be observed that Isidis Planitia (IP) performs very well compared to Acidalia Planitia (AP) and Hypanis (HP). Predictions of HP are distorted and the model is not able to predict all the cones from the input sample. As discussed in Section $6.2.1$ (main paper), this can be explained by the fact that IP has very dense cones per patch, i.e., an average of $3.52$ cones per patch while AP and HP have $1.92$ and $2.08$ cones, respectively. Also, IP cones have a very different appearance than their surface, i.e., the distinct appearance of cones on the IP surface contributes to easier cone segmentation in this region. For AP and HP, it can be seen from Figure \ref{fig:example_data_bm_1} that cones in these two regions are difficult to distinguish from their surface and hence it results in low performance. Also, as discussed in Section $4$ and $6.2.1$ (main paper) HP cones have more variability which is the cause for low performance.

\subsection{Single-category Evaluation}
\label{subsec:bm_2}

\begin{figure*}

  \centering
  \begin{subfigure}{0.26\columnwidth}
    \centering
    \caption{\textbf{Input}}
  \end{subfigure}
  \begin{subfigure}{0.26\columnwidth}
    \centering
    \caption{\textbf{GT}}
  \end{subfigure}
  \begin{subfigure}{0.26\columnwidth}
    \centering
    \caption{\textbf{U-Net}}
  \end{subfigure}
  \begin{subfigure}{0.26\columnwidth}
    \centering
    \caption{\textbf{FPN}}
  \end{subfigure}
  \begin{subfigure}{0.26\columnwidth}
    \centering
    \caption{\textbf{DeepLab}}
  \end{subfigure}
  \begin{subfigure}{0.26\columnwidth}
    \centering
    \caption{\textbf{MANet}}
  \end{subfigure}

  \centering
  \begin{subfigure}{2\columnwidth}
    \centering
    \includegraphics[width=0.13\columnwidth]{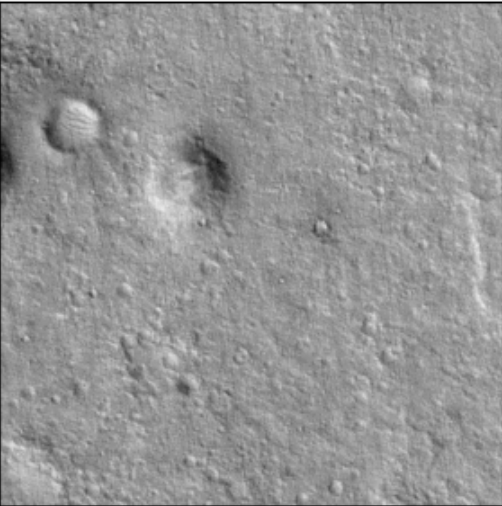}
    \includegraphics[width=0.13\columnwidth]{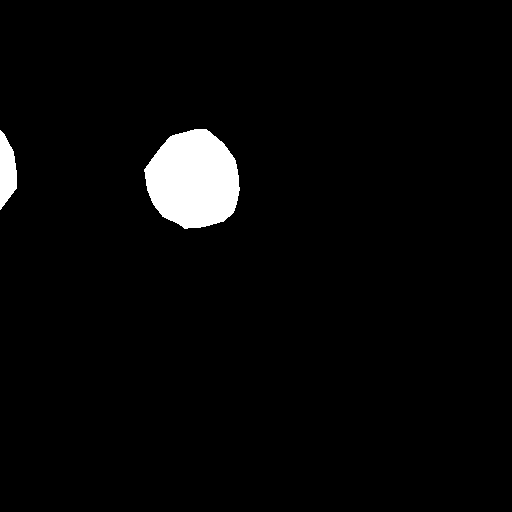}
    \includegraphics[width=0.13\columnwidth]{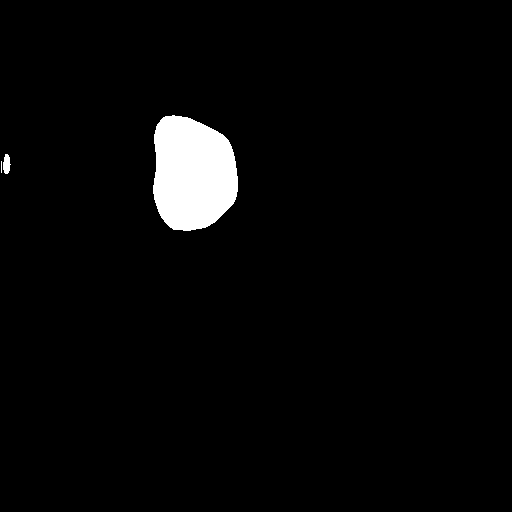}
    \includegraphics[width=0.13\columnwidth]{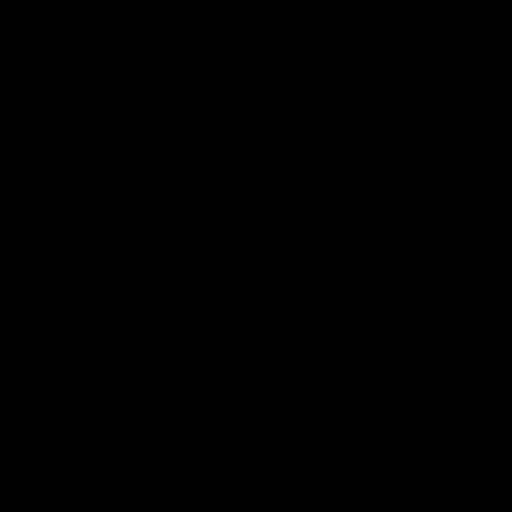}
    \includegraphics[width=0.13\columnwidth]{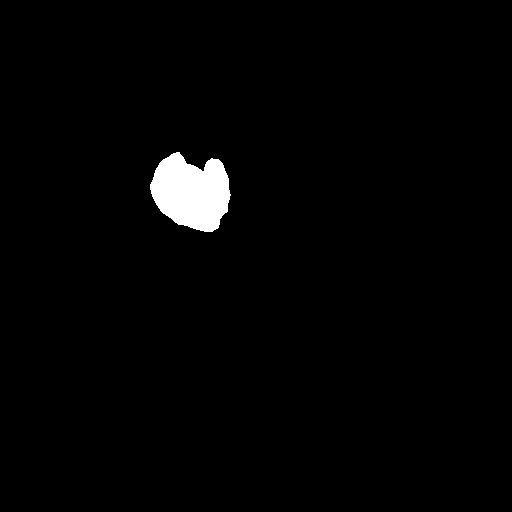}
    \includegraphics[width=0.13\columnwidth]{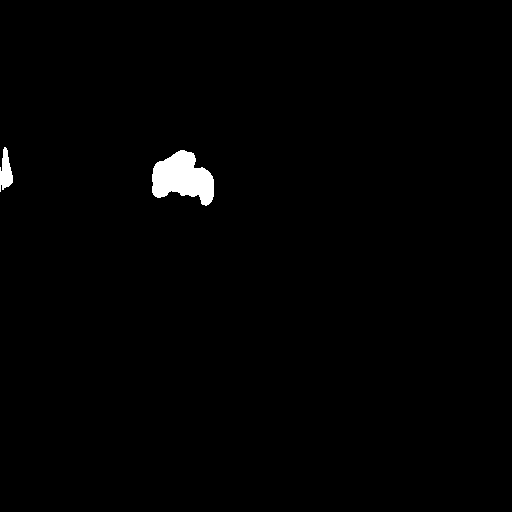}
  \end{subfigure}

  \centering
  \begin{subfigure}{2\columnwidth}
    \centering
    \includegraphics[width=0.13\columnwidth]{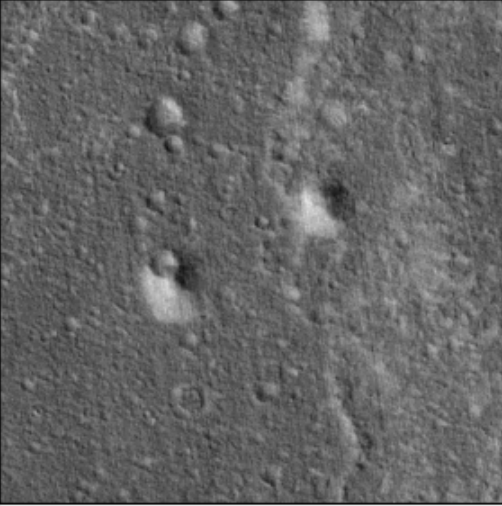}
    \includegraphics[width=0.13\columnwidth]{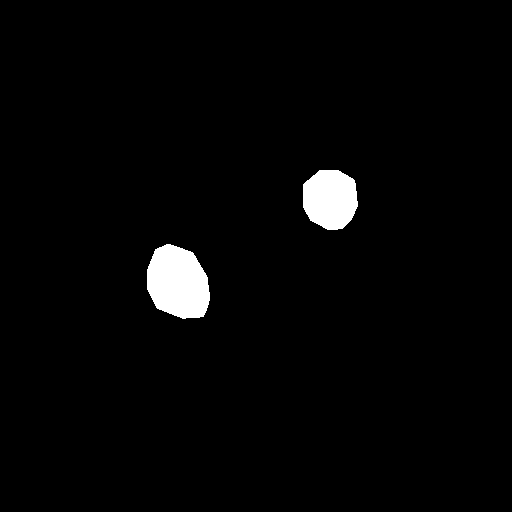}
    \includegraphics[width=0.13\columnwidth]{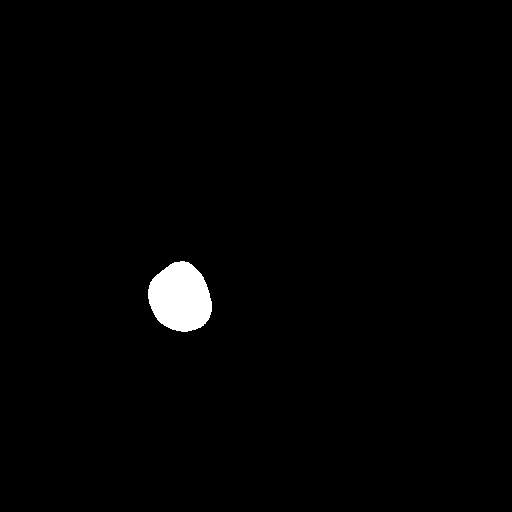}
    \includegraphics[width=0.13\columnwidth]{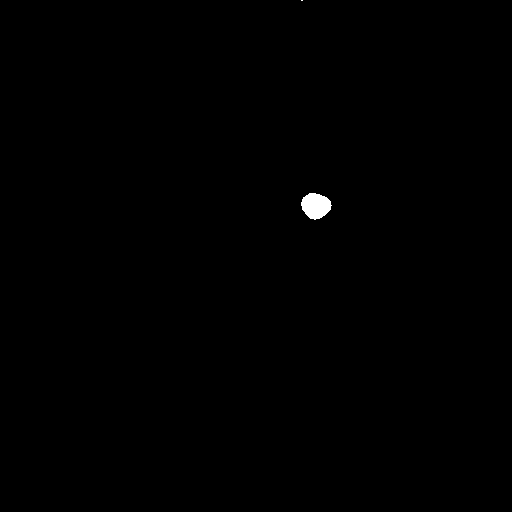}
    \includegraphics[width=0.13\columnwidth]{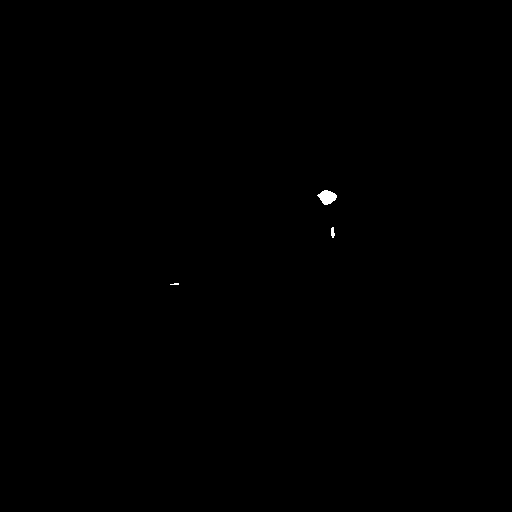}
    \includegraphics[width=0.13\columnwidth]{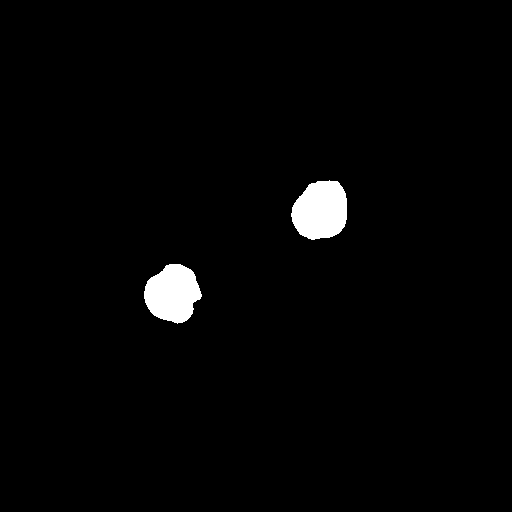}
  \end{subfigure}

  \centering
  \begin{subfigure}{2\columnwidth}
    \centering
    \includegraphics[width=0.13\columnwidth]{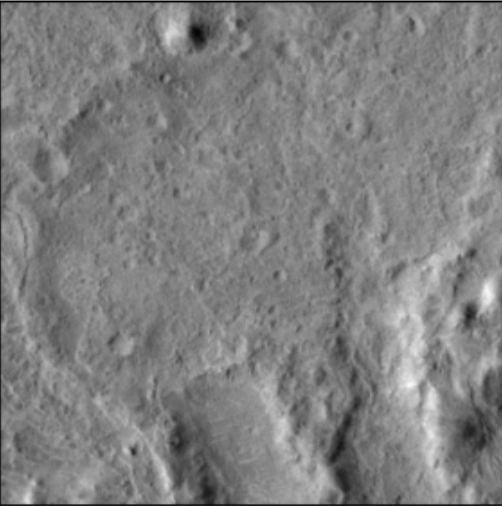}
    \includegraphics[width=0.13\columnwidth]{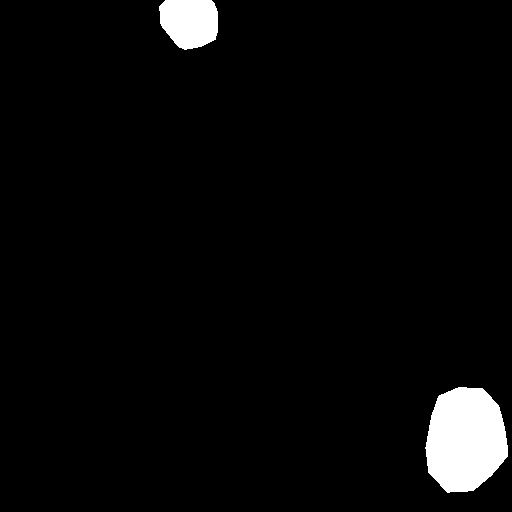}
    \includegraphics[width=0.13\columnwidth]{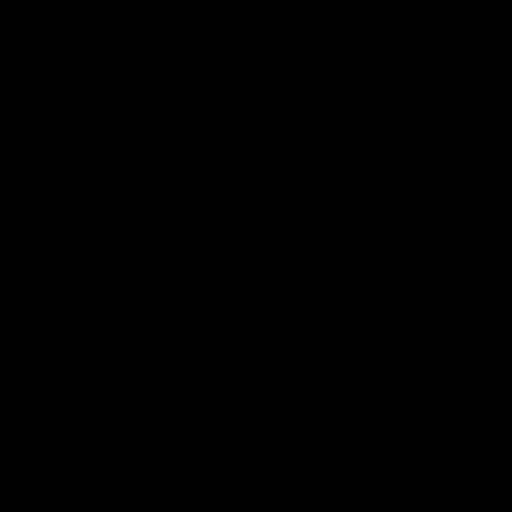}
    \includegraphics[width=0.13\columnwidth]{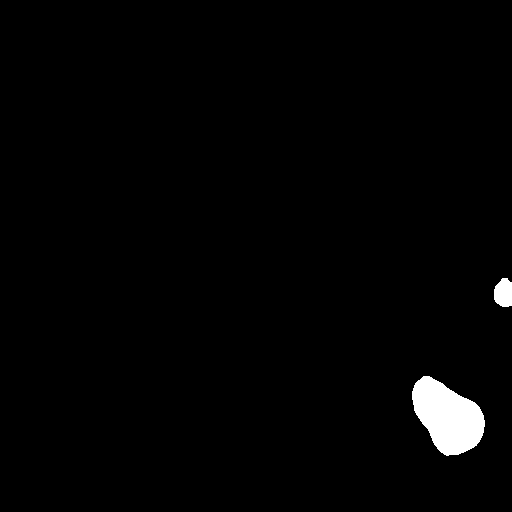}
    \includegraphics[width=0.13\columnwidth]{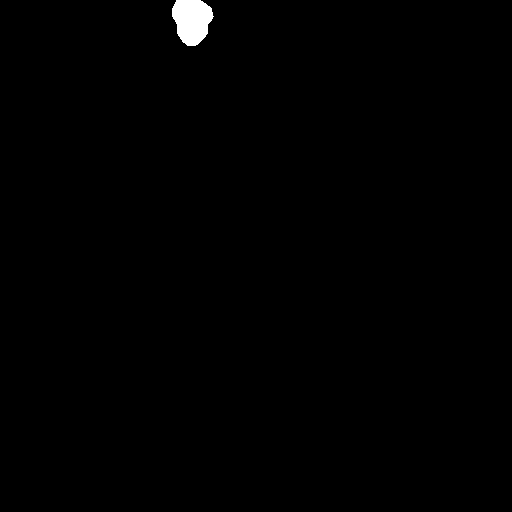}
    \includegraphics[width=0.13\columnwidth]{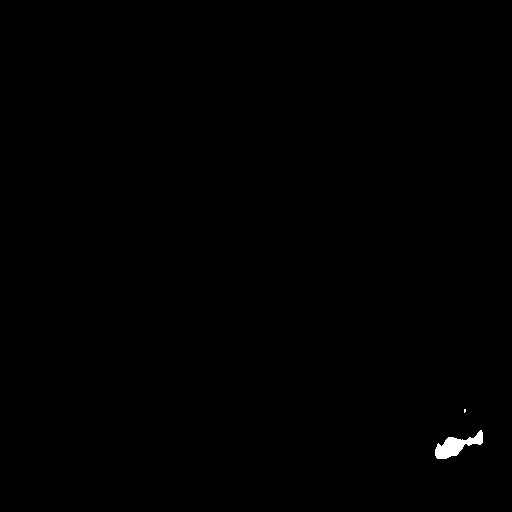}
    \caption*{\textbf{Small}}
  \end{subfigure}

  \centering
  \begin{subfigure}{2\columnwidth}
    \centering
    \includegraphics[width=0.13\columnwidth]{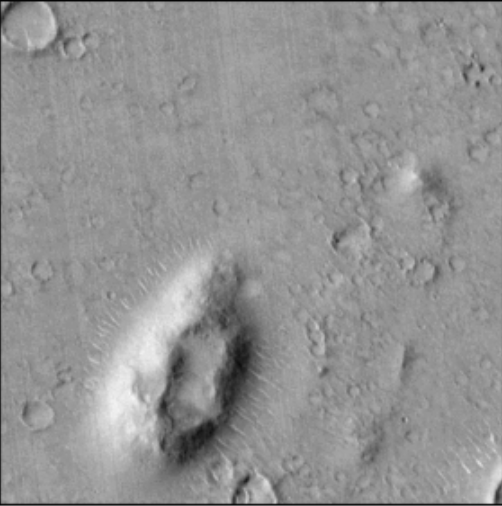}
    \includegraphics[width=0.13\columnwidth]{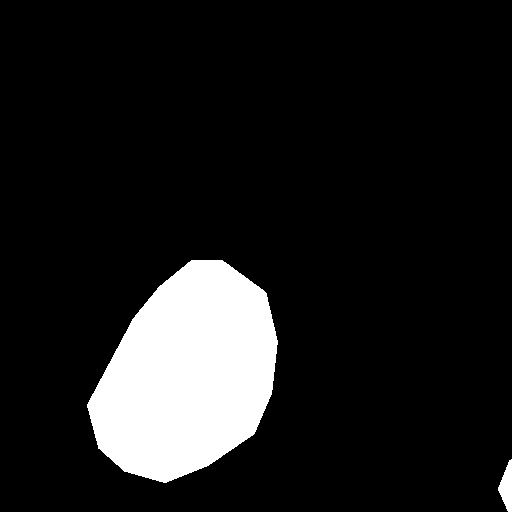}
    \includegraphics[width=0.13\columnwidth]{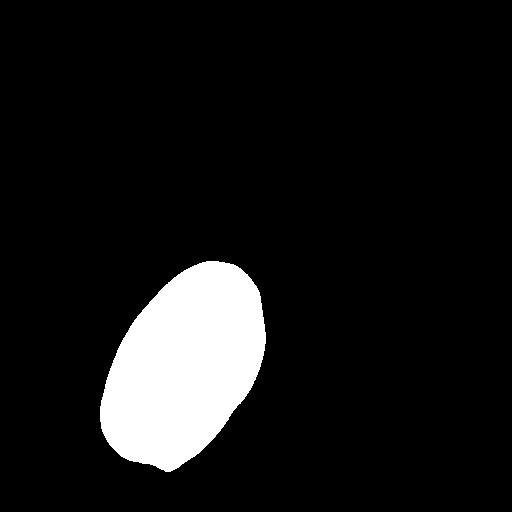}
    \includegraphics[width=0.13\columnwidth]{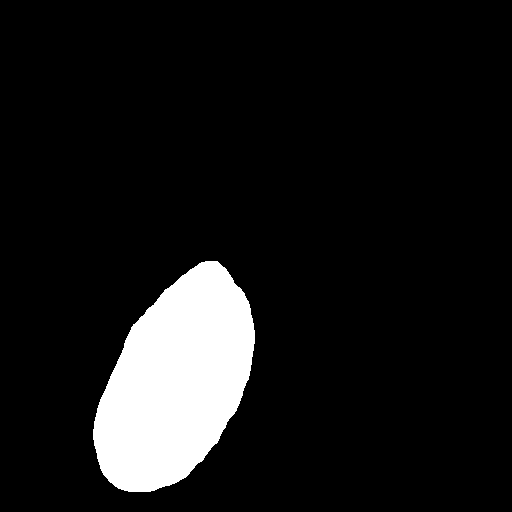}
    \includegraphics[width=0.13\columnwidth]{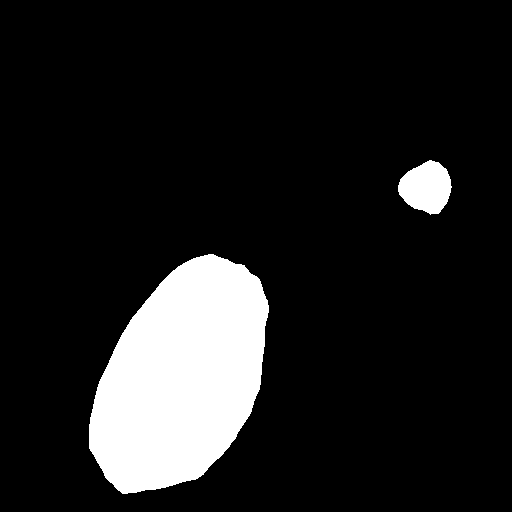}
    \includegraphics[width=0.13\columnwidth]{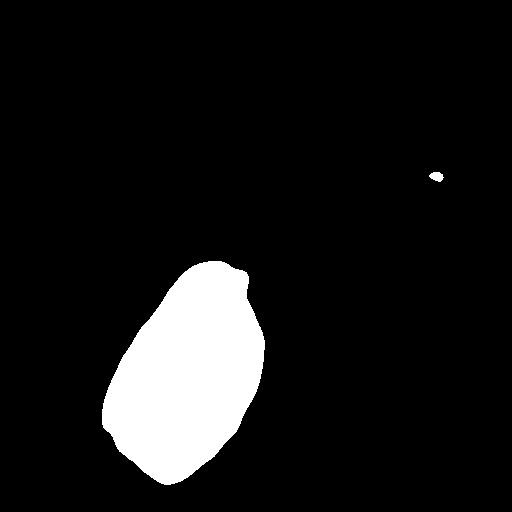}
  \end{subfigure}

  \centering
  \begin{subfigure}{2\columnwidth}
    \centering
    \includegraphics[width=0.13\columnwidth]{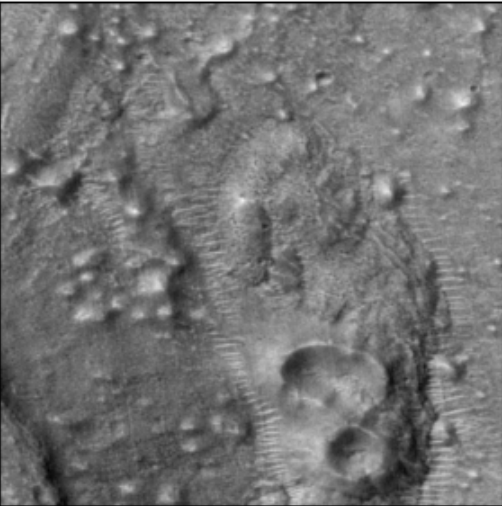}
    \includegraphics[width=0.13\columnwidth]{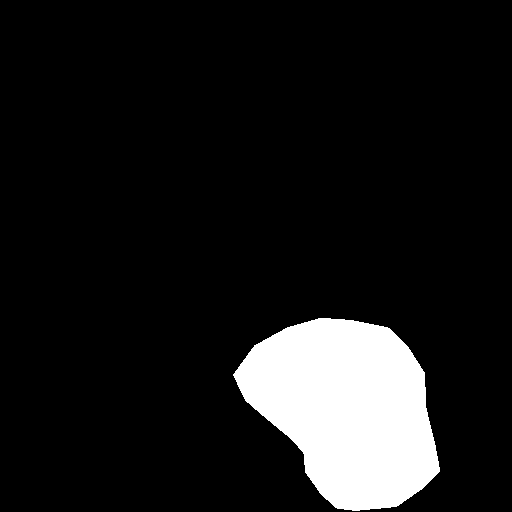}
    \includegraphics[width=0.13\columnwidth]{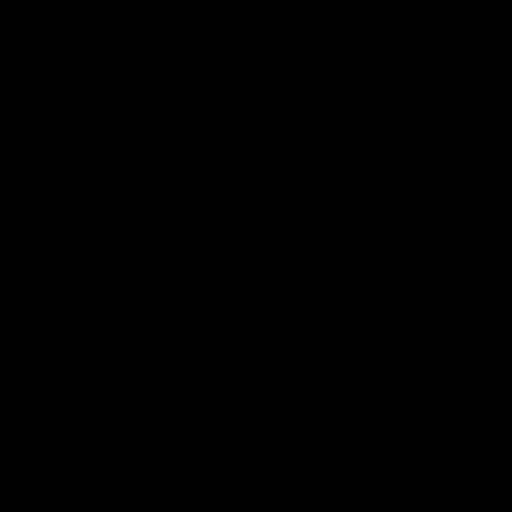}
    \includegraphics[width=0.13\columnwidth]{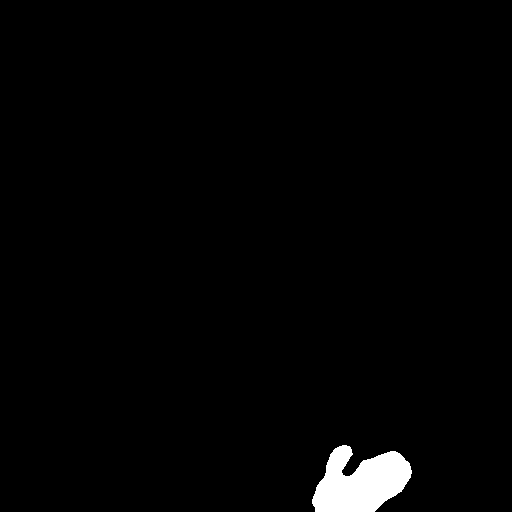}
    \includegraphics[width=0.13\columnwidth]{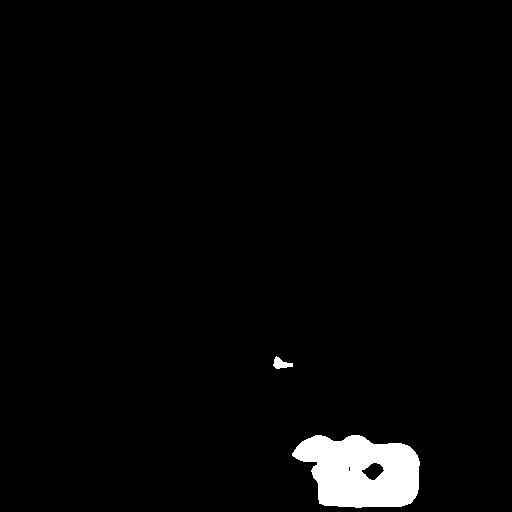}
    \includegraphics[width=0.13\columnwidth]{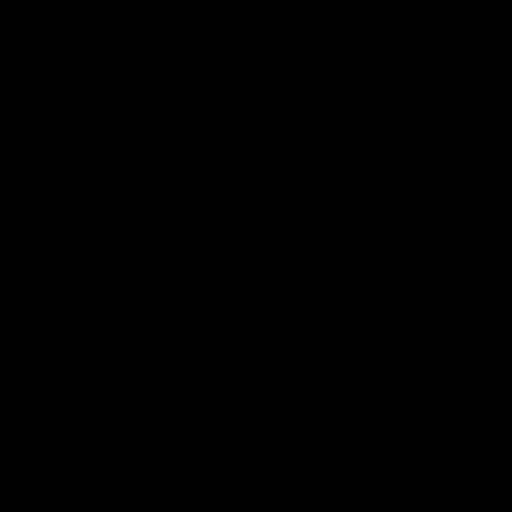}
  \end{subfigure}

  \centering
  \begin{subfigure}{2\columnwidth}
    \centering
    \includegraphics[width=0.13\columnwidth]{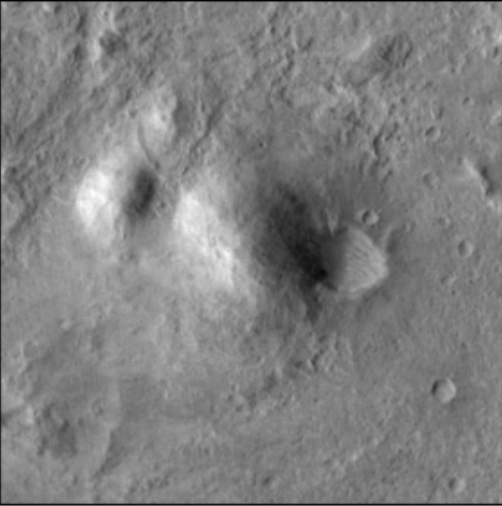}
    \includegraphics[width=0.13\columnwidth]{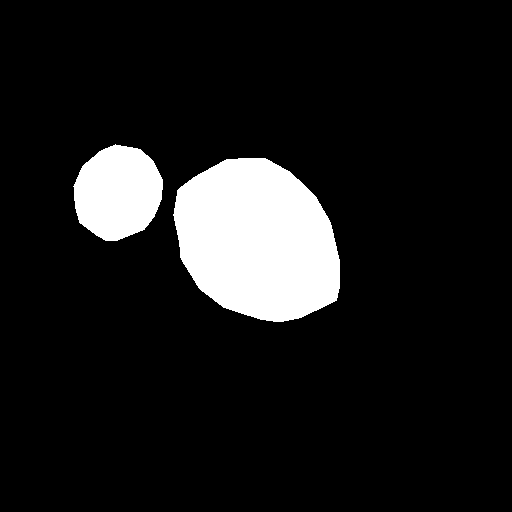}
    \includegraphics[width=0.13\columnwidth]{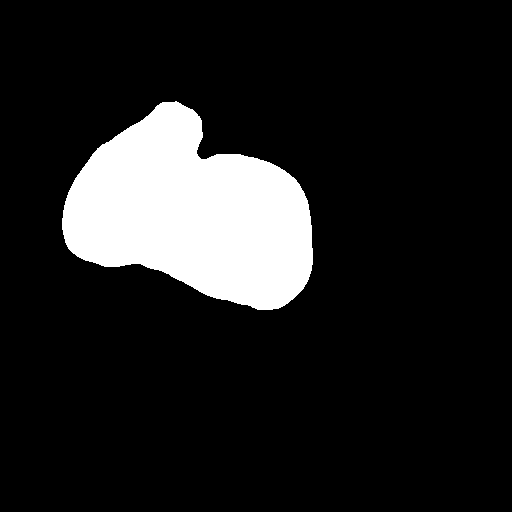}
    \includegraphics[width=0.13\columnwidth]{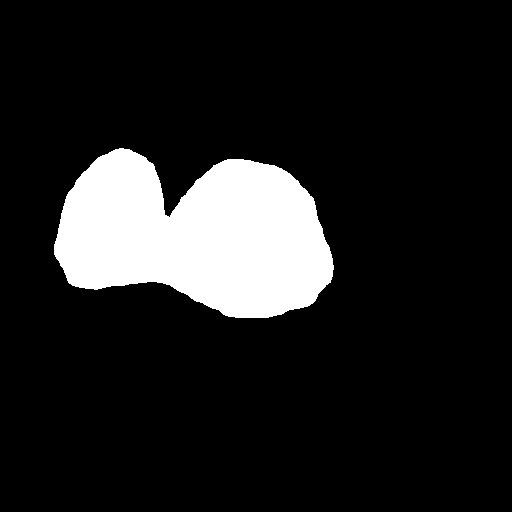}
    \includegraphics[width=0.13\columnwidth]{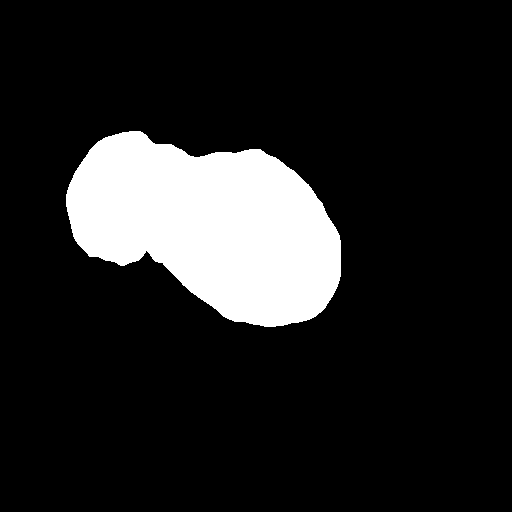}
    \includegraphics[width=0.13\columnwidth]{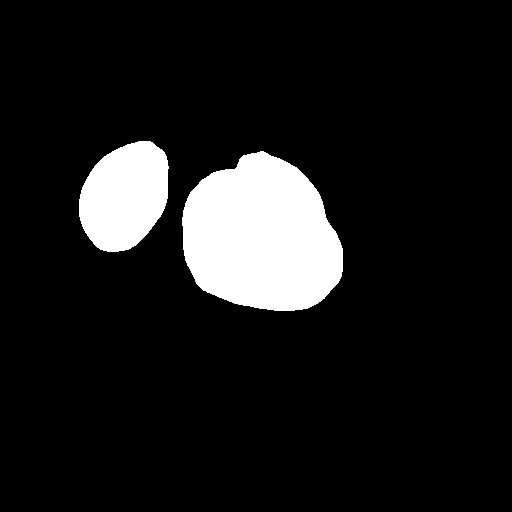}
    \caption*{\textbf{Medium}}
  \end{subfigure}

  \centering
  \begin{subfigure}{2\columnwidth}
    \centering
    \includegraphics[width=0.13\columnwidth]{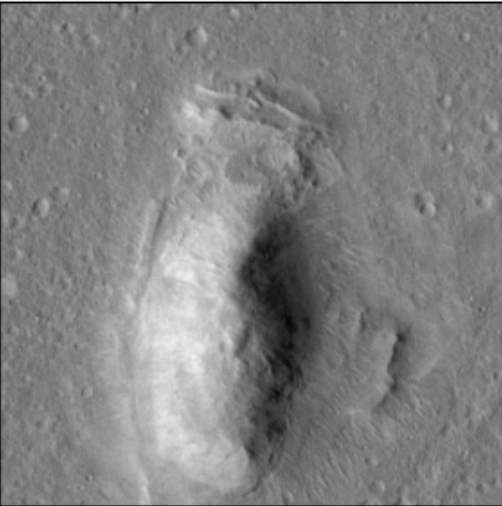}
    \includegraphics[width=0.13\columnwidth]{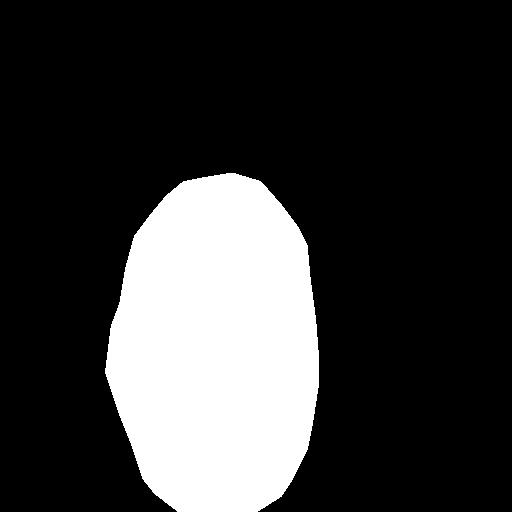}
    \includegraphics[width=0.13\columnwidth]{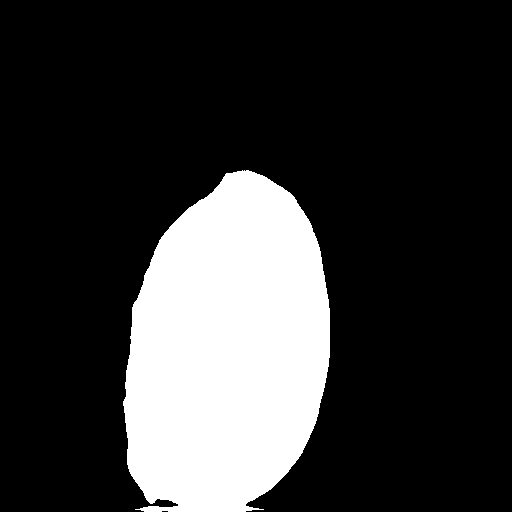}
    \includegraphics[width=0.13\columnwidth]{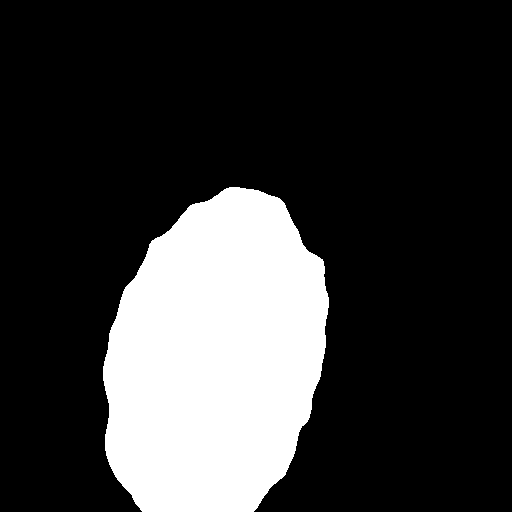}
    \includegraphics[width=0.13\columnwidth]{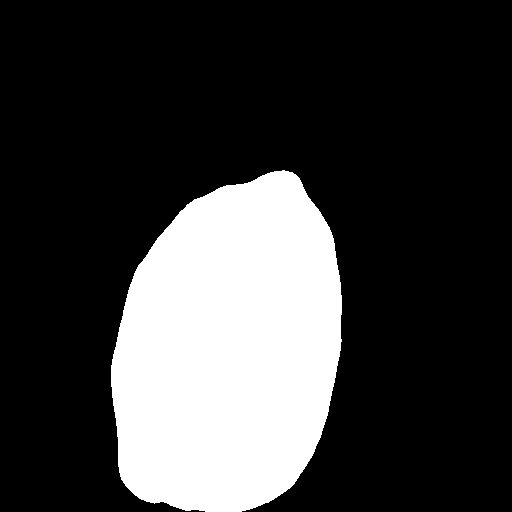}
    \includegraphics[width=0.13\columnwidth]{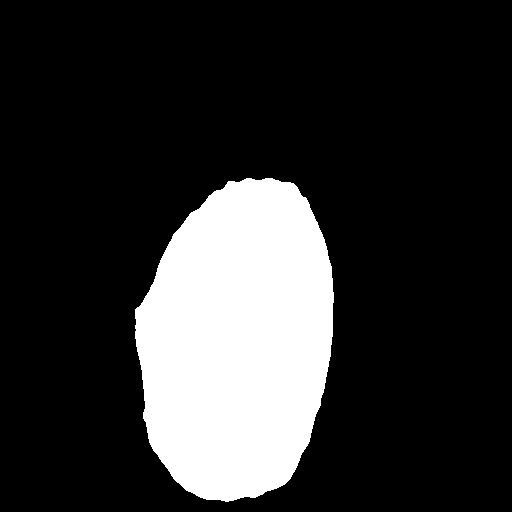}
  \end{subfigure}

  \centering
  \begin{subfigure}{2\columnwidth}
    \centering
    \includegraphics[width=0.13\columnwidth]{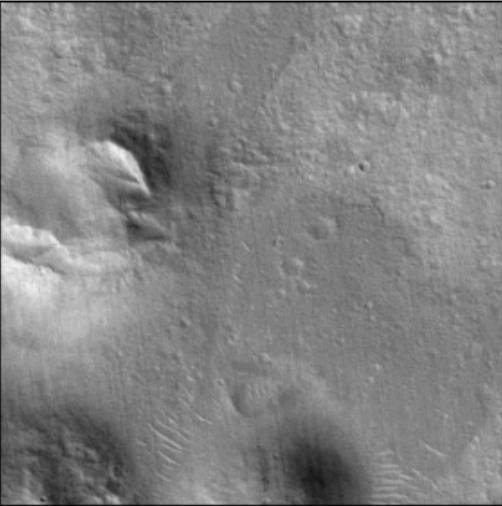}
    \includegraphics[width=0.13\columnwidth]{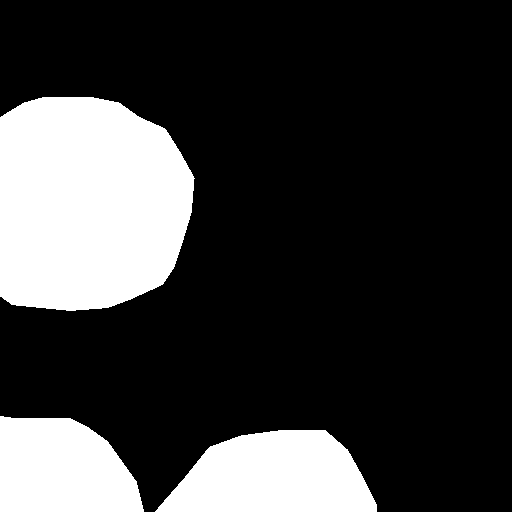}
    \includegraphics[width=0.13\columnwidth]{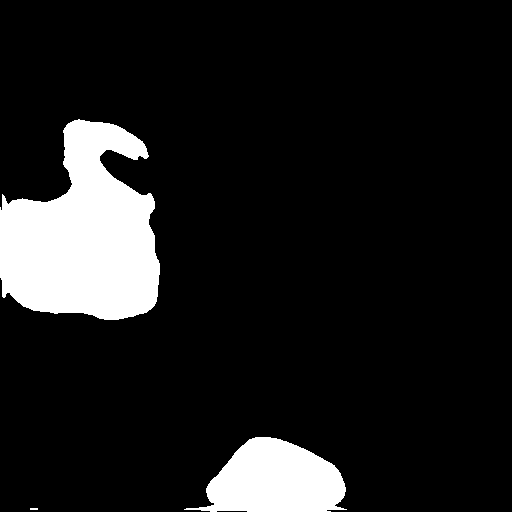}
    \includegraphics[width=0.13\columnwidth]{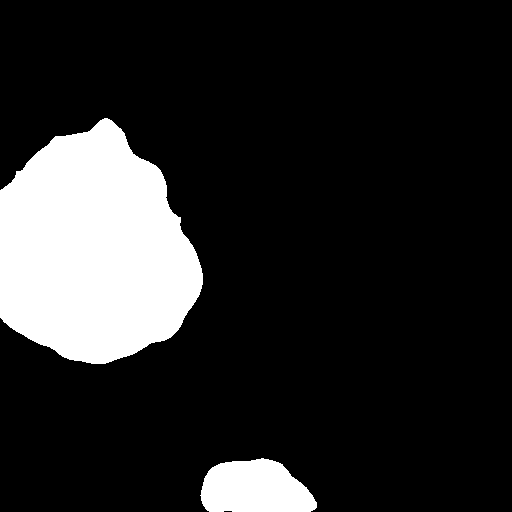}
    \includegraphics[width=0.13\columnwidth]{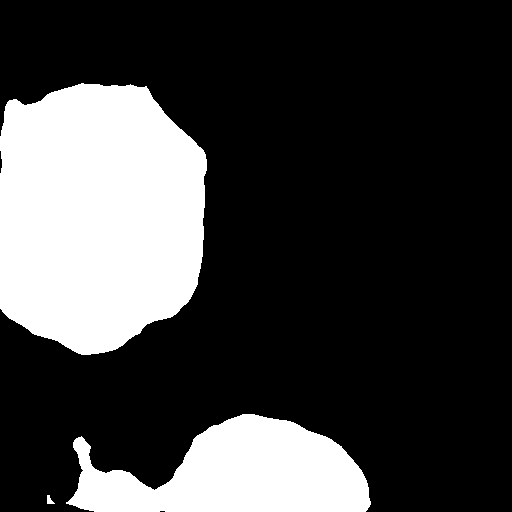}
    \includegraphics[width=0.13\columnwidth]{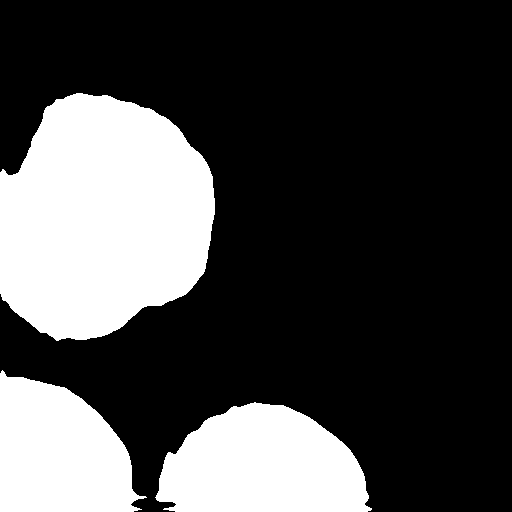}
  \end{subfigure}

  \centering
  \begin{subfigure}{2\columnwidth}
    \centering
    \includegraphics[width=0.13\columnwidth]{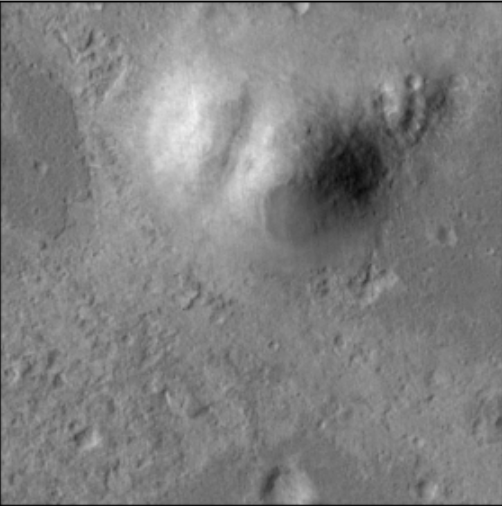}
    \includegraphics[width=0.13\columnwidth]{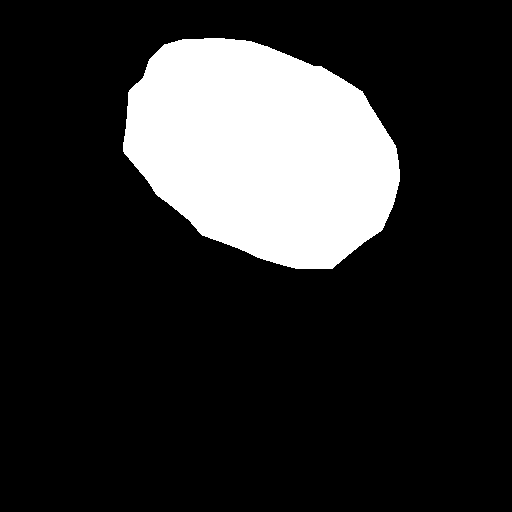}
    \includegraphics[width=0.13\columnwidth]{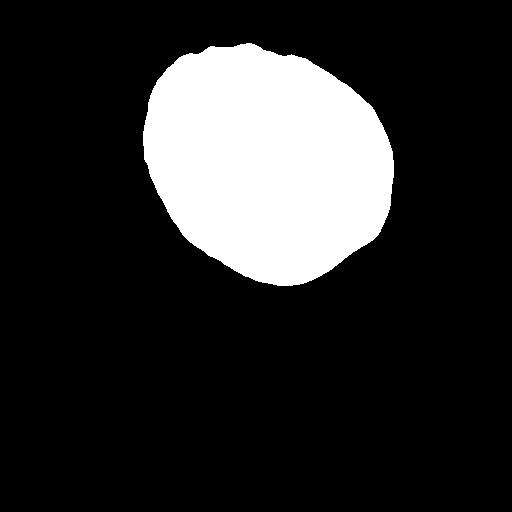}
    \includegraphics[width=0.13\columnwidth]{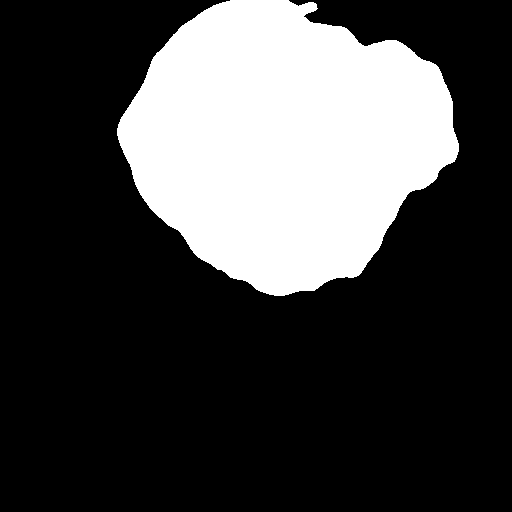}
    \includegraphics[width=0.13\columnwidth]{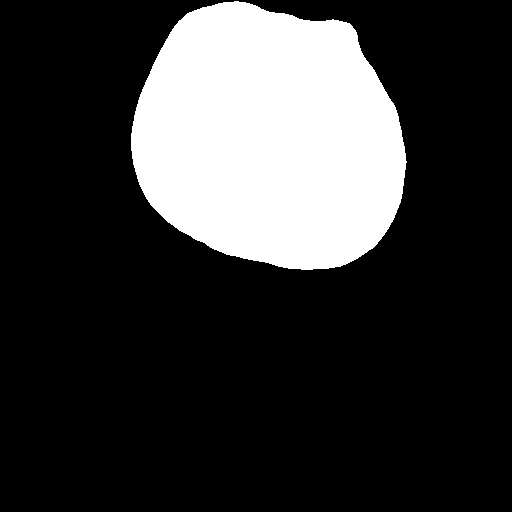}
    \includegraphics[width=0.13\columnwidth]{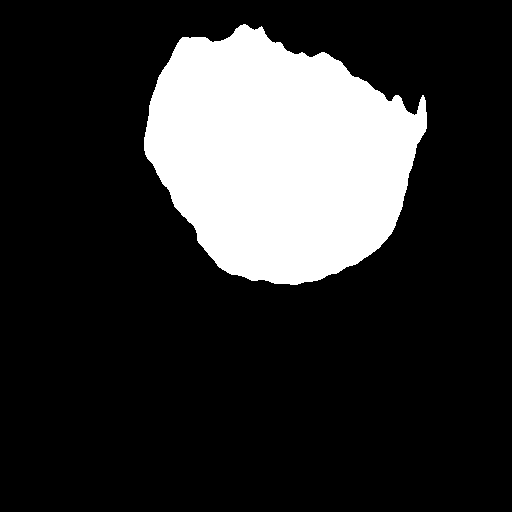}
    \caption*{\textbf{Large}}
  \end{subfigure}

  \caption{Illustration of predictions of in-distribution evaluation for single-size category training (BM-2). Columns c, d, e, and f show prediction from all models with their corresponding Input (column a) and Ground Truth (GT) (column b).}
  \label{fig:example_data_bm_2}

\end{figure*}

Figure \ref{fig:example_data_bm_2} shows the visualization of prediction for evaluating in-distribution data for single-size category training. From Figure \ref{fig:example_data_bm_2}, it is evident that all models are working better for Large size compared to Medium and Small. Moreover, predictions show improvement as we transition from small to large size. The reason behind this is that the model gets very little information while training only for the small-size category which leads to more FN pixels in prediction. Also, an increase in FN pixels does not only impact pixel-based metrics but highly affects instance-level metrics. This scenario can be observed from Table \ref{tab:bm_2_single_all_results}, the range of Object accuracy and Object IoU for small are $16$-$22$ and $22$-$29$, respectively; while for large, this range is $59$-$64$ and $50$-$60$. This indicates although models are trained on an equal amount of data for all 3 size categories, models struggle on smaller size.

\section{Additional Results and Analysis}
\label{sec:additional_results}

In this section, we present all the additional results encompassing the performance of each model on in-distribution ($\mathcal{D}_{id}$), and the detailed analysis of results on out-of-distribution ($\mathcal{D}_{ood}$). All results are for $\mathcal{T}_{pn}$ (when the model is trained on positive as well as negative samples), denoting the scenario where a model is trained on combined positive and negative samples.

\subsection{Results on \texorpdfstring{$\mathcal{D}_{id}$}{dseen} and \texorpdfstring{$\mathcal{D}_{ood}$}{dunseen}}
\label{subsec:results_seen_unseen}

For BM-1, Table \ref{tab:bm_1_single_all_results} and \ref{tab:bm_1_multi_all_results} provide detailed results for each model for single-region and multi-region training, respectively. For BM-2, Table \ref{tab:bm_2_single_all_results} and \ref{tab:bm_2_multi_all_results} provide detailed results for each model for single-category and multi-category training, respectively. In these tables, the \textit{highlighted} rows indicate results on $\mathcal{D}_{id}$, while the \textit{non-highlighted} rows indicate results on the $\mathcal{D}_{ood}$ for both BMs. Detailed analysis on $\mathcal{D}_{id}$ for both BMs can be found in Section $6$ (main paper).

\subsection{Analysis on \texorpdfstring{$\mathcal{D}_{ood}$}{dunseen}}
\label{subsec:analysis_unseen}

Here, we provide a detailed analysis of the results on $\mathcal{D}_{ood}$ for both BMs. As discussed in Section $6.2$ (main paper), and as observed from Table \ref{tab:bm_1_single_all_results}, the models demonstrate lower performance on $\mathcal{D}_{ood}$ compared to $\mathcal{D}_{id}$, with an average pixel IoU of $12.67$ and $21.11$ across all models for BM-1 using single-region and multi-region training, respectively. In contrast, the performance on $\mathcal{D}_{id}$ yields an average IoU of $52.06$ and $49.68$ for the respective configurations. These results indicate that the models do not generalize well to $\mathcal{D}_{ood}$ regions. A similar observation can be made for BM-2 on $\mathcal{D}_{ood}$, where the average pixel IoU across all models is $25.37$ and $30.00$ for single-category and multi-category training, respectively. Meanwhile, the performance on $\mathcal{D}_{id}$ yields an average IoU of $42.88$ and $41.08$ for the corresponding configurations. This further demonstrates that the models struggle to generalize to $\mathcal{D}_{ood}$ categories. Overall, this indicates that the model's generalization ability to both out-of-distribution regions and categories is limited, as shown by the lower IoU compared to performance on $\mathcal{D}_{id}$.

Interestingly, for BM-2, the medium category exhibits relatively better performance when the model is trained on either small or large categories, or trained on a combination of small and large categories (denoted as $S + L$ in tables). This can be attributed to the cone size in the medium category being closer in scale to both small and large categories. As a result, the model is better at handling the medium category when it is unseen during evaluation.

\section{\texorpdfstring{$\mathcal{T}_{pn}$}{tpn} \textit{vs.} \texorpdfstring{$\mathcal{T}_{p}$}{tp}}
\label{app_subsec:tpn_vs_tp}

In this section, we aim to provide a detailed discussion on having an advantage of incorporating negative samples during training. To accomplish this, we provide the performance of the model on $\mathcal{T}_{p}$ (when the model is trained only on positive samples), denoting the scenario where the model is exclusively trained on positive samples.

\subsection{Evaluation on BM-1}
\label{app_subsec:training_2_bm_1}

Table \ref{tab:bm_1_single_negative_all_results} and \ref{tab:bm_1_multi_negative_all_results} show the results of $\mathcal{T}{p}$ for BM-1 on single-region and multi-region training, respectively. Similarly to $\mathcal{T}{pn}$, the \textit{highlighted} rows in these tables represent results on $\mathcal{D}_{id}$, while the \textit{non-highlighted} rows represent results on $\mathcal{D}_{ood}$.

\paragraph{Performance on Single-region training:} 
When evaluating cone-patches for $\mathcal{T}_{p}$ on single-region training, we observe an average increase in pixel IoU of $4.18$ on $\mathcal{D}_{id}$ and $10.65$ on $\mathcal{D}_{ood}$ (across all models) compared to $\mathcal{T}_{pn}$. However, we do not observe significant improvements in other metrics. For instance, pixel precision shows a performance drop of $4.87$ on $\mathcal{D}_{id}$ and $23.43$ on $\mathcal{D}_{ood}$ for $\mathcal{T}_{p}$ compared to $\mathcal{T}_{pn}$. Furthermore, we observe that the performance on non-cone patches for $\mathcal{T}_{p}$ on $\mathcal{D}_{id}$ shows the degradation of $1.85$ in $A_{FP}$ compared to $\mathcal{T}_{pn}$. Similarly, on $\mathcal{D}_{ood}$, the performance drop for $A_{FP}$ is $8.26$ for $\mathcal{T}_{p}$ compared to $\mathcal{T}_{pn}$.

\paragraph{Performance on Multi-region training:} Evaluation of model performance on cone-patches for single-region training, we can observe an average increase of $4.08$ in pixel IoU on $\mathcal{D}_{id}$ and $14.21$ on $\mathcal{D}_{ood}$ for $\mathcal{T}_{p}$ compared to $\mathcal{T}_{pn}$. However, other metrics do not show significant improvements. Whereas, pixel precision decreases by $7.54$ on $\mathcal{D}_{id}$ and $18.07$ on $\mathcal{D}_{ood}$ for $\mathcal{T}_{p}$ compared $\mathcal{T}_{pn}$. Additionally, the performance of non-cone patches deteriorates for $\mathcal{T}_{p}$, with $A_{FP}$ showing a degradation of $2.35$ on $\mathcal{D}_{id}$ and $4.03$ on $\mathcal{D}_{ood}$ compared to $\mathcal{T}_{pn}$. From this, we can see that although cone-patches benefit from improved pixel IoU on both $\mathcal{D}_{id}$ and $\mathcal{D}_{ood}$, there is a noticeable decline in pixel precision, and the performance of non-cone patches.

\subsection{Evaluation on BM-2}
\label{app_subsec:training_2_bm_2}

Here, we can see similar observations as BM-1. Table \ref{tab:bm_2_single_negative_all_results} and \ref{tab:bm_2_multi_negative_all_results} show results of $\mathcal{T}_{p}$ for BM-2 on single-category and multi-category training, respectively.

\paragraph{Performance on Single-category training:} Similar to the single-region training, evaluation of models on cone-patches for $\mathcal{T}_{p}$ show an increase in an average pixel IoU by $4.86$ (across all models) on $\mathcal{D}_{id}$ and $6.66$ (average across all models) on $\mathcal{D}_{ood}$ compared to $\mathcal{T}_{pn}$ for single-category training. However, there is no such significant improvement for other metrics. For instance, pixel precision shows performance drop by $7.08$ on $\mathcal{D}_{id}$ and $9.12$ on $\mathcal{D}_{ood}$ for $\mathcal{T}_{p}$ compared to $\mathcal{T}_{pn}$. Moreover, evaluation of models on non-cone patches exhibits performance deterioration of $A_{FP}$ by $1.44$ and $1.39$ on $\mathcal{T}_{p}$ compared to $\mathcal{T}_{pn}$ on $\mathcal{D}_{id}$ and $\mathcal{D}_{ood}$, respectively.

\paragraph{Performance on Multi-category training:}

Upon evaluating models on cone-patches for single-category training for $\mathcal{T}_{p}$, we observe an increase in pixel IoU of $6.62$ (average across all models) on $\mathcal{D}_{id}$ and $8.11$ on $\mathcal{D}_{ood}$ when compared to $\mathcal{T}_{pn}$. However, we do not observe significant improvements in other metrics. Even, pixel precision demonstrates a performance drop of $7.15$ on $\mathcal{D}_{id}$ and $9.93$ on $\mathcal{D}_{ood}$ for $\mathcal{T}_{p}$ compared to $\mathcal{T}_{pn}$.
Furthermore, considering the performance on non-cone patches, we observe a degradation of $2.06$ in $A_{FP}$ compared to $\mathcal{T}_{pn}$ on $\mathcal{D}_{id}$. Similarly, on $\mathcal{D}_{ood}$, the deterioration in performance for $A_{FP}$ is $1.3$ for $\mathcal{T}_{p}$ compared to $\mathcal{T}_{pn}$.

Based on the aforementioned analysis in \ref{app_subsec:training_2_bm_1} and \ref{app_subsec:training_2_bm_1}, it can be concluded that excluding negative samples has a positive impact on the performance of positive samples (cone-patches), but it adversely affects the performance of non-cone patches. This is evident from the degraded pixel precision, indicating that the model generates more false positives due to insufficient information about ``\textit{what is not a cone}''. Furthermore, the performance drop on non-cone patches is not desirable as it increases the workload for planetary scientists, since they would need to filter out the increased number of falsely annotated cones generated by the model.

\begin{table*}[!ht]
\setlength\tabcolsep{4.0pt}
\setlength{\belowcaptionskip}{-10pt}
\centering
\footnotesize
\resizebox{0.96\linewidth}{!}{
\begin{tabular}{c|c|c|ccccccccccc|c}
\toprule[1.5pt]
\multirow{2}{*}{\textbf{\begin{tabular}[c]{@{}c@{}}Training\\Region\end{tabular}}} &
  \multirow{2}{*}{\textbf{\begin{tabular}[c]{@{}c@{}}Testing\\Region\end{tabular}}} &
  \multirow{2}{*}{\textbf{\begin{tabular}[c]{@{}c@{}}Training\\Model\end{tabular}}} &
  \multicolumn{11}{c|}{\textbf{Cone}} &
  \textbf{Non-cone} \\ \cmidrule{4-15} 
 &
   &
   &
  \multicolumn{1}{c|}{\textbf{\begin{tabular}[c]{@{}c@{}}Mask\\IoU\end{tabular}}} &
  \multicolumn{1}{c|}{\textbf{\begin{tabular}[c]{@{}c@{}}Pixel\\IoU\end{tabular}}} &
  \multicolumn{1}{c|}{\textbf{\begin{tabular}[c]{@{}c@{}}Pixel\\Accuracy\end{tabular}}} &
  \multicolumn{1}{c|}{\textbf{\begin{tabular}[c]{@{}c@{}}Pixel\\Precision\end{tabular}}} &
  \multicolumn{1}{c|}{\textbf{\begin{tabular}[c]{@{}c@{}}Pixel\\Recall\end{tabular}}} &
  \multicolumn{1}{c|}{\textbf{\begin{tabular}[c]{@{}c@{}}Panoptic\\Quality\end{tabular}}} &
  \multicolumn{1}{c|}{\textbf{mAP}} &
  \multicolumn{1}{c|}{\textbf{\begin{tabular}[c]{@{}c@{}}Object\\IoU\end{tabular}}} &
  \multicolumn{1}{c|}{\textbf{\begin{tabular}[c]{@{}c@{}}Object\\Accuracy\end{tabular}}} &
  \multicolumn{1}{c|}{\textbf{\begin{tabular}[c]{@{}c@{}}Object\\Precision\end{tabular}}} &
  \textbf{\begin{tabular}[c]{@{}c@{}}Object\\ Recall\end{tabular}} &
  \textbf{$A_{FP}$} \\ \midrule[1pt]
 &
  \cellcolor[HTML]{FBB982} &
  U-Net &
  \multicolumn{1}{c|}{63.90} &
  \multicolumn{1}{c|}{67.99} &
  \multicolumn{1}{c|}{96.66} &
  \multicolumn{1}{c|}{79.56} &
  \multicolumn{1}{c|}{83.69} &
  \multicolumn{1}{c|}{54.38} &
  \multicolumn{1}{c|}{36.67} &
  \multicolumn{1}{c|}{74.17} &
  \multicolumn{1}{c|}{59.83} &
  \multicolumn{1}{c|}{69.13} &
  76.07 &
  0.00 \\
 &
  \cellcolor[HTML]{FBB982} &
  FPN &
  \multicolumn{1}{c|}{69.06} &
  \multicolumn{1}{c|}{69.49} &
  \multicolumn{1}{c|}{96.41} &
  \multicolumn{1}{c|}{83.35} &
  \multicolumn{1}{c|}{81.44} &
  \multicolumn{1}{c|}{58.47} &
  \multicolumn{1}{c|}{31.85} &
  \multicolumn{1}{c|}{77.04} &
  \multicolumn{1}{c|}{65.34} &
  \multicolumn{1}{c|}{76.97} &
  78.76 &
  0.00 \\
 &
  \cellcolor[HTML]{FBB982} &
  DeepLab &
  \multicolumn{1}{c|}{62.11} &
  \multicolumn{1}{c|}{67.28} &
  \multicolumn{1}{c|}{96.80} &
  \multicolumn{1}{c|}{85.41} &
  \multicolumn{1}{c|}{76.98} &
  \multicolumn{1}{c|}{53.50} &
  \multicolumn{1}{c|}{32.29} &
  \multicolumn{1}{c|}{73.79} &
  \multicolumn{1}{c|}{58.58} &
  \multicolumn{1}{c|}{70.08} &
  72.61 &
  0.00 \\
 &
  \multirow{-4}{*}{\cellcolor[HTML]{FBB982}IP} &
  MA-Net &
  \multicolumn{1}{c|}{64.17} &
  \multicolumn{1}{c|}{66.90} &
  \multicolumn{1}{c|}{96.77} &
  \multicolumn{1}{c|}{85.82} &
  \multicolumn{1}{c|}{74.24} &
  \multicolumn{1}{c|}{53.12} &
  \multicolumn{1}{c|}{31.80} &
  \multicolumn{1}{c|}{72.17} &
  \multicolumn{1}{c|}{59.08} &
  \multicolumn{1}{c|}{70.91} &
  74.59 &
  0.00 \\ \cmidrule{2-15} 
 &
   &
  U-Net &
  \multicolumn{1}{c|}{6.72} &
  \multicolumn{1}{c|}{3.99} &
  \multicolumn{1}{c|}{89.58} &
  \multicolumn{1}{c|}{28.60} &
  \multicolumn{1}{c|}{9.04} &
  \multicolumn{1}{c|}{1.75} &
  \multicolumn{1}{c|}{0.67} &
  \multicolumn{1}{c|}{2.19} &
  \multicolumn{1}{c|}{2.47} &
  \multicolumn{1}{c|}{3.70} &
  2.47 &
  5.70 \\
 &
   &
  FPN &
  \multicolumn{1}{c|}{15.03} &
  \multicolumn{1}{c|}{11.60} &
  \multicolumn{1}{c|}{79.84} &
  \multicolumn{1}{c|}{16.97} &
  \multicolumn{1}{c|}{34.43} &
  \multicolumn{1}{c|}{1.80} &
  \multicolumn{1}{c|}{0.96} &
  \multicolumn{1}{c|}{4.39} &
  \multicolumn{1}{c|}{1.98} &
  \multicolumn{1}{c|}{2.16} &
  5.56 &
  19.60 \\
 &
   &
  DeepLab &
  \multicolumn{1}{c|}{7.82} &
  \multicolumn{1}{c|}{4.87} &
  \multicolumn{1}{c|}{92.75} &
  \multicolumn{1}{c|}{74.96} &
  \multicolumn{1}{c|}{5.20} &
  \multicolumn{1}{c|}{1.46} &
  \multicolumn{1}{c|}{0.54} &
  \multicolumn{1}{c|}{2.93} &
  \multicolumn{1}{c|}{1.23} &
  \multicolumn{1}{c|}{3.70} &
  1.23 &
  7.45 \\
 &
  \multirow{-4}{*}{AP} &
  MA-Net &
  \multicolumn{1}{c|}{16.24} &
  \multicolumn{1}{c|}{12.69} &
  \multicolumn{1}{c|}{89.86} &
  \multicolumn{1}{c|}{35.54} &
  \multicolumn{1}{c|}{24.30} &
  \multicolumn{1}{c|}{1.97} &
  \multicolumn{1}{c|}{0.77} &
  \multicolumn{1}{c|}{8.93} &
  \multicolumn{1}{c|}{1.97} &
  \multicolumn{1}{c|}{3.06} &
  8.02 &
  10.33 \\ \cmidrule{2-15} 
 &
   &
  U-Net &
  \multicolumn{1}{c|}{25.02} &
  \multicolumn{1}{c|}{23.29} &
  \multicolumn{1}{c|}{88.14} &
  \multicolumn{1}{c|}{69.08} &
  \multicolumn{1}{c|}{29.01} &
  \multicolumn{1}{c|}{11.99} &
  \multicolumn{1}{c|}{3.83} &
  \multicolumn{1}{c|}{23.37} &
  \multicolumn{1}{c|}{13.56} &
  \multicolumn{1}{c|}{16.18} &
  22.41 &
  1.09 \\
 &
   &
  FPN &
  \multicolumn{1}{c|}{34.35} &
  \multicolumn{1}{c|}{29.32} &
  \multicolumn{1}{c|}{87.03} &
  \multicolumn{1}{c|}{53.51} &
  \multicolumn{1}{c|}{48.34} &
  \multicolumn{1}{c|}{13.76} &
  \multicolumn{1}{c|}{3.75} &
  \multicolumn{1}{c|}{34.96} &
  \multicolumn{1}{c|}{13.4} &
  \multicolumn{1}{c|}{15.41} &
  34.22 &
  5.13 \\
 &
   &
  DeepLab &
  \multicolumn{1}{c|}{24.01} &
  \multicolumn{1}{c|}{21.57} &
  \multicolumn{1}{c|}{87.36} &
  \multicolumn{1}{c|}{70.05} &
  \multicolumn{1}{c|}{28.67} &
  \multicolumn{1}{c|}{11.47} &
  \multicolumn{1}{c|}{3.34} &
  \multicolumn{1}{c|}{24.51} &
  \multicolumn{1}{c|}{12.79} &
  \multicolumn{1}{c|}{15.94} &
  23.72 &
  1.11 \\
\multirow{-12}{*}{IP} &
  \multirow{-4}{*}{HP} &
  MA-Net &
  \multicolumn{1}{c|}{28.40} &
  \multicolumn{1}{c|}{27.95} &
  \multicolumn{1}{c|}{89.07} &
  \multicolumn{1}{c|}{72.11} &
  \multicolumn{1}{c|}{35.54} &
  \multicolumn{1}{c|}{14.46} &
  \multicolumn{1}{c|}{4.45} &
  \multicolumn{1}{c|}{30.82} &
  \multicolumn{1}{c|}{15.95} &
  \multicolumn{1}{c|}{18.64} &
  30.86 &
  1.55 \\ \midrule
 &
   &
  U-Net &
  \multicolumn{1}{c|}{8.37} &
  \multicolumn{1}{c|}{11.21} &
  \multicolumn{1}{c|}{89.53} &
  \multicolumn{1}{c|}{46.7} &
  \multicolumn{1}{c|}{19.71} &
  \multicolumn{1}{c|}{4.75} &
  \multicolumn{1}{c|}{2.16} &
  \multicolumn{1}{c|}{11.28} &
  \multicolumn{1}{c|}{5.02} &
  \multicolumn{1}{c|}{7.65} &
  8.80 &
  0.70 \\
 &
   &
  FPN &
  \multicolumn{1}{c|}{11.89} &
  \multicolumn{1}{c|}{10.83} &
  \multicolumn{1}{c|}{91.32} &
  \multicolumn{1}{c|}{70.53} &
  \multicolumn{1}{c|}{12.1} &
  \multicolumn{1}{c|}{5.42} &
  \multicolumn{1}{c|}{3.21} &
  \multicolumn{1}{c|}{13.73} &
  \multicolumn{1}{c|}{5.93} &
  \multicolumn{1}{c|}{11.67} &
  8.06 &
  0.00 \\
 &
   &
  DeepLab &
  \multicolumn{1}{c|}{0.69} &
  \multicolumn{1}{c|}{1.53} &
  \multicolumn{1}{c|}{90.36} &
  \multicolumn{1}{c|}{95.16} &
  \multicolumn{1}{c|}{1.86} &
  \multicolumn{1}{c|}{0.87} &
  \multicolumn{1}{c|}{0.22} &
  \multicolumn{1}{c|}{2.18} &
  \multicolumn{1}{c|}{0.93} &
  \multicolumn{1}{c|}{1.23} &
  1.85 &
  0.00 \\
 &
  \multirow{-4}{*}{IP} &
  MA-Net &
  \multicolumn{1}{c|}{0.00} &
  \multicolumn{1}{c|}{0.00} &
  \multicolumn{1}{c|}{90.65} &
  \multicolumn{1}{c|}{100.00} &
  \multicolumn{1}{c|}{0.00} &
  \multicolumn{1}{c|}{0.00} &
  \multicolumn{1}{c|}{0.00} &
  \multicolumn{1}{c|}{0.00} &
  \multicolumn{1}{c|}{0.00} &
  \multicolumn{1}{c|}{0.00} &
  0.00 &
  0.00 \\ \cmidrule{2-15} 
 &
  \cellcolor[HTML]{FBB982} &
  U-Net &
  \multicolumn{1}{c|}{50.28} &
  \multicolumn{1}{c|}{49.77} &
  \multicolumn{1}{c|}{96.32} &
  \multicolumn{1}{c|}{85.50} &
  \multicolumn{1}{c|}{57.78} &
  \multicolumn{1}{c|}{44.83} &
  \multicolumn{1}{c|}{19.82} &
  \multicolumn{1}{c|}{58.10} &
  \multicolumn{1}{c|}{51.45} &
  \multicolumn{1}{c|}{62.16} &
  57.35 &
  0.08 \\
 &
  \cellcolor[HTML]{FBB982} &
  FPN &
  \multicolumn{1}{c|}{47.96} &
  \multicolumn{1}{c|}{46.85} &
  \multicolumn{1}{c|}{96.22} &
  \multicolumn{1}{c|}{80.55} &
  \multicolumn{1}{c|}{56.05} &
  \multicolumn{1}{c|}{38.62} &
  \multicolumn{1}{c|}{20.43} &
  \multicolumn{1}{c|}{50.68} &
  \multicolumn{1}{c|}{45.49} &
  \multicolumn{1}{c|}{54.94} &
  50.56 &
  0.38 \\
 &
  \cellcolor[HTML]{FBB982} &
  DeepLab &
  \multicolumn{1}{c|}{47.85} &
  \multicolumn{1}{c|}{49.9} &
  \multicolumn{1}{c|}{96.58} &
  \multicolumn{1}{c|}{79.57} &
  \multicolumn{1}{c|}{58.10} &
  \multicolumn{1}{c|}{40.45} &
  \multicolumn{1}{c|}{16.68} &
  \multicolumn{1}{c|}{51.67} &
  \multicolumn{1}{c|}{42.59} &
  \multicolumn{1}{c|}{51.17} &
  52.10 &
  0.38 \\
 &
  \multirow{-4}{*}{\cellcolor[HTML]{FBB982}AP} &
  MA-Net &
  \multicolumn{1}{c|}{34.42} &
  \multicolumn{1}{c|}{36.81} &
  \multicolumn{1}{c|}{95.72} &
  \multicolumn{1}{c|}{86.24} &
  \multicolumn{1}{c|}{39.15} &
  \multicolumn{1}{c|}{31.74} &
  \multicolumn{1}{c|}{20.84} &
  \multicolumn{1}{c|}{36.73} &
  \multicolumn{1}{c|}{38.27} &
  \multicolumn{1}{c|}{40.74} &
  43.46 &
  0.06 \\ \cmidrule{2-15} 
 &
   &
  U-Net &
  \multicolumn{1}{c|}{2.34} &
  \multicolumn{1}{c|}{1.63} &
  \multicolumn{1}{c|}{86.52} &
  \multicolumn{1}{c|}{75.76} &
  \multicolumn{1}{c|}{1.93} &
  \multicolumn{1}{c|}{0.67} &
  \multicolumn{1}{c|}{1.28} &
  \multicolumn{1}{c|}{1.69} &
  \multicolumn{1}{c|}{0.73} &
  \multicolumn{1}{c|}{1.79} &
  0.98 &
  0.35 \\
 &
   &
  FPN &
  \multicolumn{1}{c|}{4.27} &
  \multicolumn{1}{c|}{3.39} &
  \multicolumn{1}{c|}{86.79} &
  \multicolumn{1}{c|}{86.01} &
  \multicolumn{1}{c|}{3.81} &
  \multicolumn{1}{c|}{0.86} &
  \multicolumn{1}{c|}{1.61} &
  \multicolumn{1}{c|}{1.19} &
  \multicolumn{1}{c|}{0.93} &
  \multicolumn{1}{c|}{1.61} &
  1.11 &
  0.13 \\
 &
   &
  DeepLab &
  \multicolumn{1}{c|}{1.63} &
  \multicolumn{1}{c|}{1.62} &
  \multicolumn{1}{c|}{86.47} &
  \multicolumn{1}{c|}{91.31} &
  \multicolumn{1}{c|}{2.73} &
  \multicolumn{1}{c|}{0.49} &
  \multicolumn{1}{c|}{0.70} &
  \multicolumn{1}{c|}{0.73} &
  \multicolumn{1}{c|}{0.60} &
  \multicolumn{1}{c|}{0.78} &
  0.90 &
  0.07 \\
\multirow{-12}{*}{AP} &
  \multirow{-4}{*}{HP} &
  MA-Net &
  \multicolumn{1}{c|}{1.50} &
  \multicolumn{1}{c|}{1.67} &
  \multicolumn{1}{c|}{86.42} &
  \multicolumn{1}{c|}{95.25} &
  \multicolumn{1}{c|}{2.90} &
  \multicolumn{1}{c|}{0.56} &
  \multicolumn{1}{c|}{1.73} &
  \multicolumn{1}{c|}{1.30} &
  \multicolumn{1}{c|}{0.50} &
  \multicolumn{1}{c|}{0.55} &
  1.33 &
  0.05 \\ \midrule
 &
   &
  U-Net &
  \multicolumn{1}{c|}{25.77} &
  \multicolumn{1}{c|}{18.64} &
  \multicolumn{1}{c|}{92.21} &
  \multicolumn{1}{c|}{97.30} &
  \multicolumn{1}{c|}{19.69} &
  \multicolumn{1}{c|}{15.67} &
  \multicolumn{1}{c|}{9.72} &
  \multicolumn{1}{c|}{25.68} &
  \multicolumn{1}{c|}{17.64} &
  \multicolumn{1}{c|}{28.46} &
  19.81 &
  0.00 \\
 &
   &
  FPN &
  \multicolumn{1}{c|}{38.32} &
  \multicolumn{1}{c|}{33.07} &
  \multicolumn{1}{c|}{93.60} &
  \multicolumn{1}{c|}{97.69} &
  \multicolumn{1}{c|}{34.49} &
  \multicolumn{1}{c|}{26.41} &
  \multicolumn{1}{c|}{15.83} &
  \multicolumn{1}{c|}{39.63} &
  \multicolumn{1}{c|}{28.72} &
  \multicolumn{1}{c|}{41.05} &
  31.31 &
  0.00 \\
 &
   &
  DeepLab &
  \multicolumn{1}{c|}{27.18} &
  \multicolumn{1}{c|}{20.23} &
  \multicolumn{1}{c|}{92.26} &
  \multicolumn{1}{c|}{92.17} &
  \multicolumn{1}{c|}{21.28} &
  \multicolumn{1}{c|}{14.65} &
  \multicolumn{1}{c|}{9.74} &
  \multicolumn{1}{c|}{25.64} &
  \multicolumn{1}{c|}{15.40} &
  \multicolumn{1}{c|}{28.70} &
  17.35 &
  0.00 \\
 &
  \multirow{-4}{*}{IP} &
  MA-Net &
  \multicolumn{1}{c|}{5.00} &
  \multicolumn{1}{c|}{1.80} &
  \multicolumn{1}{c|}{88.07} &
  \multicolumn{1}{c|}{93.15} &
  \multicolumn{1}{c|}{6.34} &
  \multicolumn{1}{c|}{2.45} &
  \multicolumn{1}{c|}{1.63} &
  \multicolumn{1}{c|}{6.6} &
  \multicolumn{1}{c|}{2.44} &
  \multicolumn{1}{c|}{3.00} &
  6.79 &
  0.00 \\ \cmidrule{2-15} 
 &
   &
  U-Net &
  \multicolumn{1}{c|}{24.79} &
  \multicolumn{1}{c|}{21.68} &
  \multicolumn{1}{c|}{93.18} &
  \multicolumn{1}{c|}{79.08} &
  \multicolumn{1}{c|}{24.15} &
  \multicolumn{1}{c|}{16.27} &
  \multicolumn{1}{c|}{2.17} &
  \multicolumn{1}{c|}{22.93} &
  \multicolumn{1}{c|}{20.80} &
  \multicolumn{1}{c|}{25.31} &
  24.07 &
  11.36 \\
 &
   &
  FPN &
  \multicolumn{1}{c|}{11.29} &
  \multicolumn{1}{c|}{11.08} &
  \multicolumn{1}{c|}{92.80} &
  \multicolumn{1}{c|}{91.06} &
  \multicolumn{1}{c|}{11.42} &
  \multicolumn{1}{c|}{5.44} &
  \multicolumn{1}{c|}{1.29} &
  \multicolumn{1}{c|}{9.14} &
  \multicolumn{1}{c|}{7.22} &
  \multicolumn{1}{c|}{8.64} &
  9.88 &
  7.56 \\
 &
   &
  DeepLab &
  \multicolumn{1}{c|}{20.26} &
  \multicolumn{1}{c|}{22.00} &
  \multicolumn{1}{c|}{94.27} &
  \multicolumn{1}{c|}{81.52} &
  \multicolumn{1}{c|}{24.92} &
  \multicolumn{1}{c|}{13.23} &
  \multicolumn{1}{c|}{3.32} &
  \multicolumn{1}{c|}{21.51} &
  \multicolumn{1}{c|}{16.17} &
  \multicolumn{1}{c|}{19.14} &
  23.46 &
  7.06 \\
 &
  \multirow{-4}{*}{AP} &
  MA-Net &
  \multicolumn{1}{c|}{8.00} &
  \multicolumn{1}{c|}{8.44} &
  \multicolumn{1}{c|}{89.93} &
  \multicolumn{1}{c|}{70.04} &
  \multicolumn{1}{c|}{10.36} &
  \multicolumn{1}{c|}{2.06} &
  \multicolumn{1}{c|}{0.68} &
  \multicolumn{1}{c|}{4.08} &
  \multicolumn{1}{c|}{2.59} &
  \multicolumn{1}{c|}{3.09} &
  4.94 &
  7.21 \\ \cmidrule{2-15} 
 &
  \cellcolor[HTML]{FBB982} &
  U-Net &
  \multicolumn{1}{c|}{39.66} &
  \multicolumn{1}{c|}{38.70} &
  \multicolumn{1}{c|}{91.22} &
  \multicolumn{1}{c|}{83.16} &
  \multicolumn{1}{c|}{45.07} &
  \multicolumn{1}{c|}{29.48} &
  \multicolumn{1}{c|}{12.49} &
  \multicolumn{1}{c|}{39.69} &
  \multicolumn{1}{c|}{34.82} &
  \multicolumn{1}{c|}{42.93} &
  39.57 &
  0.43 \\
 &
  \cellcolor[HTML]{FBB982} &
  FPN &
  \multicolumn{1}{c|}{43.08} &
  \multicolumn{1}{c|}{43.87} &
  \multicolumn{1}{c|}{92.31} &
  \multicolumn{1}{c|}{84.70} &
  \multicolumn{1}{c|}{50.32} &
  \multicolumn{1}{c|}{33.50} &
  \multicolumn{1}{c|}{17.35} &
  \multicolumn{1}{c|}{45.47} &
  \multicolumn{1}{c|}{39.92} &
  \multicolumn{1}{c|}{47.88} &
  46.11 &
  0.55 \\
 &
  \cellcolor[HTML]{FBB982} &
  DeepLab &
  \multicolumn{1}{c|}{43.86} &
  \multicolumn{1}{c|}{45.94} &
  \multicolumn{1}{c|}{92.63} &
  \multicolumn{1}{c|}{79.34} &
  \multicolumn{1}{c|}{56.01} &
  \multicolumn{1}{c|}{33.11} &
  \multicolumn{1}{c|}{14.07} &
  \multicolumn{1}{c|}{47.30} &
  \multicolumn{1}{c|}{38.82} &
  \multicolumn{1}{c|}{44.29} &
  48.95 &
  0.65 \\
\multirow{-12}{*}{HP} &
  \multirow{-4}{*}{\cellcolor[HTML]{FBB982}HP} &
  MA-Net &
  \multicolumn{1}{c|}{38.94} &
  \multicolumn{1}{c|}{41.20} &
  \multicolumn{1}{c|}{92.18} &
  \multicolumn{1}{c|}{83.65} &
  \multicolumn{1}{c|}{48.54} &
  \multicolumn{1}{c|}{29.56} &
  \multicolumn{1}{c|}{13.66} &
  \multicolumn{1}{c|}{41.19} &
  \multicolumn{1}{c|}{34.30} &
  \multicolumn{1}{c|}{39.79} &
  42.60 &
  0.35 \\ \bottomrule
\end{tabular}
}
\caption{Results for BM-1 on $\mathcal{D}_{id}$ (\textit{highlighted}) and $\mathcal{D}_{ood}$ for single-region $\mathcal{T}_{pn}$.}
\label{tab:bm_1_single_all_results}
\end{table*}

\begin{table*}[!ht]
\setlength\tabcolsep{4.0pt}
\setlength{\belowcaptionskip}{-10pt}
\centering
\footnotesize
\resizebox{0.96\linewidth}{!}{
\begin{tabular}{c|c|c|ccccccccccc|c}
\toprule[1.5pt]
\multirow{2}{*}{\textbf{\begin{tabular}[c]{@{}c@{}}Training\\Region\end{tabular}}} &
  \multirow{2}{*}{\textbf{\begin{tabular}[c]{@{}c@{}}Testing\\Region\end{tabular}}} &
  \multirow{2}{*}{\textbf{\begin{tabular}[c]{@{}c@{}}Training\\Model\end{tabular}}} &
  \multicolumn{11}{c|}{\textbf{Cone}} &
  \textbf{Non-cone} \\ \cmidrule{4-15} 
 &
   &
   &
  \multicolumn{1}{c|}{\textbf{\begin{tabular}[c]{@{}c@{}}Mask\\IoU\end{tabular}}} &
  \multicolumn{1}{c|}{\textbf{\begin{tabular}[c]{@{}c@{}}Pixel\\IoU\end{tabular}}} &
  \multicolumn{1}{c|}{\textbf{\begin{tabular}[c]{@{}c@{}}Pixel\\Accuracy\end{tabular}}} &
  \multicolumn{1}{c|}{\textbf{\begin{tabular}[c]{@{}c@{}}Pixel\\Precision\end{tabular}}} &
  \multicolumn{1}{c|}{\textbf{\begin{tabular}[c]{@{}c@{}}Pixel\\Recall\end{tabular}}} &
  \multicolumn{1}{c|}{\textbf{\begin{tabular}[c]{@{}c@{}}Panoptic\\Quality\end{tabular}}} &
  \multicolumn{1}{c|}{\textbf{mAP}} &
  \multicolumn{1}{c|}{\textbf{\begin{tabular}[c]{@{}c@{}}Object\\IoU\end{tabular}}} &
  \multicolumn{1}{c|}{\textbf{\begin{tabular}[c]{@{}c@{}}Object\\Accuracy\end{tabular}}} &
  \multicolumn{1}{c|}{\textbf{\begin{tabular}[c]{@{}c@{}}Object\\Precision\end{tabular}}} &
  \textbf{\begin{tabular}[c]{@{}c@{}}Object\\ Recall\end{tabular}} &
  \textbf{$A_{FP}$} \\ \midrule[1pt]
 &
  \cellcolor[HTML]{FBB982} &
  U-Net &
  \multicolumn{1}{c|}{63.30} &
  \multicolumn{1}{c|}{67.90} &
  \multicolumn{1}{c|}{96.81} &
  \multicolumn{1}{c|}{91.35} &
  \multicolumn{1}{c|}{72.81} &
  \multicolumn{1}{c|}{52.40} &
  \multicolumn{1}{c|}{32.59} &
  \multicolumn{1}{c|}{74.20} &
  \multicolumn{1}{c|}{57.41} &
  \multicolumn{1}{c|}{68.76} &
  70.24 &
  0.00 \\
 &
  \cellcolor[HTML]{FBB982} &
  FPN &
  \multicolumn{1}{c|}{66.92} &
  \multicolumn{1}{c|}{69.44} &
  \multicolumn{1}{c|}{96.70} &
  \multicolumn{1}{c|}{82.85} &
  \multicolumn{1}{c|}{80.91} &
  \multicolumn{1}{c|}{59.10} &
  \multicolumn{1}{c|}{36.76} &
  \multicolumn{1}{c|}{76.35} &
  \multicolumn{1}{c|}{67.03} &
  \multicolumn{1}{c|}{76.81} &
  81.1 &
  0.00 \\
 &
  \cellcolor[HTML]{FBB982} &
  DeepLab &
  \multicolumn{1}{c|}{59.87} &
  \multicolumn{1}{c|}{63.70} &
  \multicolumn{1}{c|}{96.29} &
  \multicolumn{1}{c|}{83.85} &
  \multicolumn{1}{c|}{72.98} &
  \multicolumn{1}{c|}{48.93} &
  \multicolumn{1}{c|}{28.91} &
  \multicolumn{1}{c|}{72.69} &
  \multicolumn{1}{c|}{53.60} &
  \multicolumn{1}{c|}{65.31} &
  61.99 &
  0.00 \\
 &
  \multirow{-4}{*}{\cellcolor[HTML]{FBB982}IP} &
  MA-Net &
  \multicolumn{1}{c|}{66.66} &
  \multicolumn{1}{c|}{69.29} &
  \multicolumn{1}{c|}{96.91} &
  \multicolumn{1}{c|}{82.92} &
  \multicolumn{1}{c|}{79.46} &
  \multicolumn{1}{c|}{59.41} &
  \multicolumn{1}{c|}{38.32} &
  \multicolumn{1}{c|}{76.01} &
  \multicolumn{1}{c|}{65.50} &
  \multicolumn{1}{c|}{77.34} &
  77.40 &
  0.00 \\ \cmidrule{2-15} 
 &
  \cellcolor[HTML]{FBB982} &
  U-Net &
  \multicolumn{1}{c|}{48.10} &
  \multicolumn{1}{c|}{47.60} &
  \multicolumn{1}{c|}{96.26} &
  \multicolumn{1}{c|}{90.48} &
  \multicolumn{1}{c|}{52.34} &
  \multicolumn{1}{c|}{38.05} &
  \multicolumn{1}{c|}{20.69} &
  \multicolumn{1}{c|}{54.90} &
  \multicolumn{1}{c|}{42.07} &
  \multicolumn{1}{c|}{51.05} &
  55.00 &
  0.09 \\
 &
  \cellcolor[HTML]{FBB982} &
  FPN &
  \multicolumn{1}{c|}{36.08} &
  \multicolumn{1}{c|}{35.38} &
  \multicolumn{1}{c|}{95.46} &
  \multicolumn{1}{c|}{94.01} &
  \multicolumn{1}{c|}{38.52} &
  \multicolumn{1}{c|}{28.75} &
  \multicolumn{1}{c|}{19.82} &
  \multicolumn{1}{c|}{38.11} &
  \multicolumn{1}{c|}{33.58} &
  \multicolumn{1}{c|}{39.20} &
  40.37 &
  0.23 \\
 &
  \cellcolor[HTML]{FBB982} &
  DeepLab &
  \multicolumn{1}{c|}{52.99} &
  \multicolumn{1}{c|}{55.44} &
  \multicolumn{1}{c|}{97.25} &
  \multicolumn{1}{c|}{89.79} &
  \multicolumn{1}{c|}{61.68} &
  \multicolumn{1}{c|}{43.66} &
  \multicolumn{1}{c|}{22.56} &
  \multicolumn{1}{c|}{63.99} &
  \multicolumn{1}{c|}{45.62} &
  \multicolumn{1}{c|}{53.33} &
  60.25 &
  0.88 \\
 &
  \multirow{-4}{*}{\cellcolor[HTML]{FBB982}AP} &
  MA-Net &
  \multicolumn{1}{c|}{44.18} &
  \multicolumn{1}{c|}{44.61} &
  \multicolumn{1}{c|}{96.45} &
  \multicolumn{1}{c|}{86.35} &
  \multicolumn{1}{c|}{48.84} &
  \multicolumn{1}{c|}{33.92} &
  \multicolumn{1}{c|}{20.61} &
  \multicolumn{1}{c|}{45.09} &
  \multicolumn{1}{c|}{37.41} &
  \multicolumn{1}{c|}{45.00} &
  46.05 &
  0.03 \\ \cmidrule{2-15} 
 &
   &
  U-Net &
  \multicolumn{1}{c|}{8.75} &
  \multicolumn{1}{c|}{6.92} &
  \multicolumn{1}{c|}{87.11} &
  \multicolumn{1}{c|}{84.86} &
  \multicolumn{1}{c|}{7.28} &
  \multicolumn{1}{c|}{4.04} &
  \multicolumn{1}{c|}{2.54} &
  \multicolumn{1}{c|}{7.03} &
  \multicolumn{1}{c|}{5.35} &
  \multicolumn{1}{c|}{8.02} &
  6.56 &
  0.19 \\
 &
   &
  FPN &
  \multicolumn{1}{c|}{14.42} &
  \multicolumn{1}{c|}{12.35} &
  \multicolumn{1}{c|}{87.46} &
  \multicolumn{1}{c|}{84.76} &
  \multicolumn{1}{c|}{13.28} &
  \multicolumn{1}{c|}{7.15} &
  \multicolumn{1}{c|}{3.66} &
  \multicolumn{1}{c|}{10.83} &
  \multicolumn{1}{c|}{9.71} &
  \multicolumn{1}{c|}{12.02} &
  11.48 &
  0.49 \\
 &
   &
  DeepLab &
  \multicolumn{1}{c|}{18.25} &
  \multicolumn{1}{c|}{19.55} &
  \multicolumn{1}{c|}{88.28} &
  \multicolumn{1}{c|}{84.25} &
  \multicolumn{1}{c|}{20.93} &
  \multicolumn{1}{c|}{11.43} &
  \multicolumn{1}{c|}{4.62} &
  \multicolumn{1}{c|}{19.16} &
  \multicolumn{1}{c|}{13.76} &
  \multicolumn{1}{c|}{16.49} &
  18.69 &
  0.35 \\
\multirow{-12}{*}{IP + AP} &
  \multirow{-4}{*}{HP} &
  MA-Net &
  \multicolumn{1}{c|}{4.26} &
  \multicolumn{1}{c|}{3.97} &
  \multicolumn{1}{c|}{86.82} &
  \multicolumn{1}{c|}{97.59} &
  \multicolumn{1}{c|}{4.24} &
  \multicolumn{1}{c|}{2.13} &
  \multicolumn{1}{c|}{6.29} &
  \multicolumn{1}{c|}{3.15} &
  \multicolumn{1}{c|}{2.72} &
  \multicolumn{1}{c|}{3.22} &
  3.33 &
  0.14 \\ \midrule
 &
  \cellcolor[HTML]{FBB982} &
  U-Net &
  \multicolumn{1}{c|}{58.90} &
  \multicolumn{1}{c|}{63.58} &
  \multicolumn{1}{c|}{96.16} &
  \multicolumn{1}{c|}{78.02} &
  \multicolumn{1}{c|}{79.80} &
  \multicolumn{1}{c|}{47.24} &
  \multicolumn{1}{c|}{25.91} &
  \multicolumn{1}{c|}{70.93} &
  \multicolumn{1}{c|}{50.80} &
  \multicolumn{1}{c|}{65.79} &
  58.76 &
  0.00 \\
 &
  \cellcolor[HTML]{FBB982} &
  FPN &
  \multicolumn{1}{c|}{68.33} &
  \multicolumn{1}{c|}{67.24} &
  \multicolumn{1}{c|}{96.56} &
  \multicolumn{1}{c|}{83.23} &
  \multicolumn{1}{c|}{78.00} &
  \multicolumn{1}{c|}{57.53} &
  \multicolumn{1}{c|}{34.56} &
  \multicolumn{1}{c|}{75.30} &
  \multicolumn{1}{c|}{63.97} &
  \multicolumn{1}{c|}{78.40} &
  71.6 &
  0.00 \\
 &
  \cellcolor[HTML]{FBB982} &
  DeepLab &
  \multicolumn{1}{c|}{59.32} &
  \multicolumn{1}{c|}{61.55} &
  \multicolumn{1}{c|}{96.09} &
  \multicolumn{1}{c|}{83.42} &
  \multicolumn{1}{c|}{73.72} &
  \multicolumn{1}{c|}{47.08} &
  \multicolumn{1}{c|}{22.96} &
  \multicolumn{1}{c|}{71.52} &
  \multicolumn{1}{c|}{50.38} &
  \multicolumn{1}{c|}{65.95} &
  61.28 &
  0.00 \\
 &
  \multirow{-4}{*}{\cellcolor[HTML]{FBB982}IP} &
  MA-Net &
  \multicolumn{1}{c|}{58.44} &
  \multicolumn{1}{c|}{58.90} &
  \multicolumn{1}{c|}{96.15} &
  \multicolumn{1}{c|}{88.61} &
  \multicolumn{1}{c|}{66.80} &
  \multicolumn{1}{c|}{48.71} &
  \multicolumn{1}{c|}{31.32} &
  \multicolumn{1}{c|}{68.97} &
  \multicolumn{1}{c|}{53.64} &
  \multicolumn{1}{c|}{69.11} &
  60.55 &
  0.00 \\ \cmidrule{2-15} 
 &
   &
  U-Net &
  \multicolumn{1}{c|}{24.45} &
  \multicolumn{1}{c|}{21.61} &
  \multicolumn{1}{c|}{94.28} &
  \multicolumn{1}{c|}{87.44} &
  \multicolumn{1}{c|}{22.02} &
  \multicolumn{1}{c|}{12.93} &
  \multicolumn{1}{c|}{4.08} &
  \multicolumn{1}{c|}{22.16} &
  \multicolumn{1}{c|}{16.89} &
  \multicolumn{1}{c|}{20.49} &
  21.11 &
  6.41 \\
 &
   &
  FPN &
  \multicolumn{1}{c|}{12.97} &
  \multicolumn{1}{c|}{11.01} &
  \multicolumn{1}{c|}{93.31} &
  \multicolumn{1}{c|}{89.34} &
  \multicolumn{1}{c|}{12.31} &
  \multicolumn{1}{c|}{10.50} &
  \multicolumn{1}{c|}{1.82} &
  \multicolumn{1}{c|}{15.01} &
  \multicolumn{1}{c|}{12.65} &
  \multicolumn{1}{c|}{16.67} &
  15.43 &
  4.30 \\
 &
   &
  DeepLab &
  \multicolumn{1}{c|}{20.19} &
  \multicolumn{1}{c|}{20.21} &
  \multicolumn{1}{c|}{93.28} &
  \multicolumn{1}{c|}{79.97} &
  \multicolumn{1}{c|}{22.91} &
  \multicolumn{1}{c|}{15.90} &
  \multicolumn{1}{c|}{2.29} &
  \multicolumn{1}{c|}{25.80} &
  \multicolumn{1}{c|}{19.60} &
  \multicolumn{1}{c|}{27.38} &
  24.69 &
  10.32 \\
 &
  \multirow{-4}{*}{AP} &
  MA-Net &
  \multicolumn{1}{c|}{9.58} &
  \multicolumn{1}{c|}{9.63} &
  \multicolumn{1}{c|}{93.18} &
  \multicolumn{1}{c|}{83.88} &
  \multicolumn{1}{c|}{10.36} &
  \multicolumn{1}{c|}{1.86} &
  \multicolumn{1}{c|}{1.52} &
  \multicolumn{1}{c|}{4.65} &
  \multicolumn{1}{c|}{1.85} &
  \multicolumn{1}{c|}{2.78} &
  4.94 &
  8.10 \\ \cmidrule{2-15} 
 &
  \cellcolor[HTML]{FBB982} &
  U-Net &
  \multicolumn{1}{c|}{46.02} &
  \multicolumn{1}{c|}{45.84} &
  \multicolumn{1}{c|}{92.5} &
  \multicolumn{1}{c|}{78.06} &
  \multicolumn{1}{c|}{55.89} &
  \multicolumn{1}{c|}{33.59} &
  \multicolumn{1}{c|}{15.40} &
  \multicolumn{1}{c|}{46.38} &
  \multicolumn{1}{c|}{39.58} &
  \multicolumn{1}{c|}{48.59} &
  46.29 &
  1.19 \\
 &
  \cellcolor[HTML]{FBB982} &
  FPN &
  \multicolumn{1}{c|}{44.16} &
  \multicolumn{1}{c|}{43.84} &
  \multicolumn{1}{c|}{92.28} &
  \multicolumn{1}{c|}{80.92} &
  \multicolumn{1}{c|}{52.14} &
  \multicolumn{1}{c|}{33.77} &
  \multicolumn{1}{c|}{16.23} &
  \multicolumn{1}{c|}{46.02} &
  \multicolumn{1}{c|}{38.89} &
  \multicolumn{1}{c|}{46.49} &
  46.27 &
  0.41 \\
 &
  \cellcolor[HTML]{FBB982} &
  DeepLab &
  \multicolumn{1}{c|}{41.75} &
  \multicolumn{1}{c|}{45.38} &
  \multicolumn{1}{c|}{92.34} &
  \multicolumn{1}{c|}{77.02} &
  \multicolumn{1}{c|}{56.27} &
  \multicolumn{1}{c|}{32.49} &
  \multicolumn{1}{c|}{11.93} &
  \multicolumn{1}{c|}{46.51} &
  \multicolumn{1}{c|}{37.34} &
  \multicolumn{1}{c|}{42.37} &
  47.45 &
  0.77 \\
\multirow{-12}{*}{HP + IP} &
  \multirow{-4}{*}{\cellcolor[HTML]{FBB982}HP} &
  MA-Net &
  \multicolumn{1}{c|}{39.35} &
  \multicolumn{1}{c|}{40.72} &
  \multicolumn{1}{c|}{91.99} &
  \multicolumn{1}{c|}{81.88} &
  \multicolumn{1}{c|}{48.35} &
  \multicolumn{1}{c|}{28.94} &
  \multicolumn{1}{c|}{13.46} &
  \multicolumn{1}{c|}{41.63} &
  \multicolumn{1}{c|}{33.56} &
  \multicolumn{1}{c|}{39.69} &
  42.51 &
  0.28 \\ \midrule
 &
   &
  U-Net &
  \multicolumn{1}{c|}{47.66} &
  \multicolumn{1}{c|}{46.23} &
  \multicolumn{1}{c|}{94.76} &
  \multicolumn{1}{c|}{89.93} &
  \multicolumn{1}{c|}{51.45} &
  \multicolumn{1}{c|}{32.25} &
  \multicolumn{1}{c|}{19.72} &
  \multicolumn{1}{c|}{52.97} &
  \multicolumn{1}{c|}{35.36} &
  \multicolumn{1}{c|}{56.42} &
  41.07 &
  0.00 \\
 &
   &
  FPN &
  \multicolumn{1}{c|}{58.25} &
  \multicolumn{1}{c|}{57.70} &
  \multicolumn{1}{c|}{95.14} &
  \multicolumn{1}{c|}{81.41} &
  \multicolumn{1}{c|}{71.36} &
  \multicolumn{1}{c|}{44.79} &
  \multicolumn{1}{c|}{24.25} &
  \multicolumn{1}{c|}{69.93} &
  \multicolumn{1}{c|}{48.20} &
  \multicolumn{1}{c|}{70.31} &
  53.92 &
  0.00 \\
 &
   &
  DeepLab &
  \multicolumn{1}{c|}{8.63} &
  \multicolumn{1}{c|}{7.02} &
  \multicolumn{1}{c|}{91.40} &
  \multicolumn{1}{c|}{96.10} &
  \multicolumn{1}{c|}{7.04} &
  \multicolumn{1}{c|}{3.58} &
  \multicolumn{1}{c|}{5.73} &
  \multicolumn{1}{c|}{6.82} &
  \multicolumn{1}{c|}{4.32} &
  \multicolumn{1}{c|}{8.89} &
  4.88 &
  0.00 \\
 &
  \multirow{-4}{*}{IP} &
  MA-Net &
  \multicolumn{1}{c|}{36.53} &
  \multicolumn{1}{c|}{37.11} &
  \multicolumn{1}{c|}{93.56} &
  \multicolumn{1}{c|}{90.40} &
  \multicolumn{1}{c|}{41.35} &
  \multicolumn{1}{c|}{24.22} &
  \multicolumn{1}{c|}{15.05} &
  \multicolumn{1}{c|}{45.95} &
  \multicolumn{1}{c|}{22.82} &
  \multicolumn{1}{c|}{36.62} &
  30.63 &
  0.00 \\ \cmidrule{2-15} 
 &
  \cellcolor[HTML]{FBB982} &
  U-Net &
  \multicolumn{1}{c|}{35.61} &
  \multicolumn{1}{c|}{33.17} &
  \multicolumn{1}{c|}{95.29} &
  \multicolumn{1}{c|}{89.55} &
  \multicolumn{1}{c|}{35.46} &
  \multicolumn{1}{c|}{27.31} &
  \multicolumn{1}{c|}{8.50} &
  \multicolumn{1}{c|}{39.26} &
  \multicolumn{1}{c|}{32.97} &
  \multicolumn{1}{c|}{41.67} &
  37.90 &
  2.65 \\
 &
  \cellcolor[HTML]{FBB982} &
  FPN &
  \multicolumn{1}{c|}{45.38} &
  \multicolumn{1}{c|}{44.81} &
  \multicolumn{1}{c|}{96.22} &
  \multicolumn{1}{c|}{90.60} &
  \multicolumn{1}{c|}{49.74} &
  \multicolumn{1}{c|}{38.21} &
  \multicolumn{1}{c|}{17.40} &
  \multicolumn{1}{c|}{47.96} &
  \multicolumn{1}{c|}{44.81} &
  \multicolumn{1}{c|}{56.30} &
  46.67 &
  0.70 \\
 &
  \cellcolor[HTML]{FBB982} &
  DeepLab &
  \multicolumn{1}{c|}{16.92} &
  \multicolumn{1}{c|}{15.93} &
  \multicolumn{1}{c|}{93.82} &
  \multicolumn{1}{c|}{94.27} &
  \multicolumn{1}{c|}{16.66} &
  \multicolumn{1}{c|}{12.66} &
  \multicolumn{1}{c|}{2.44} &
  \multicolumn{1}{c|}{17.09} &
  \multicolumn{1}{c|}{17.20} &
  \multicolumn{1}{c|}{21.11} &
  18.52 &
  1.20 \\
 &
  \multirow{-4}{*}{\cellcolor[HTML]{FBB982}AP} &
  MA-Net &
  \multicolumn{1}{c|}{48.19} &
  \multicolumn{1}{c|}{50.78} &
  \multicolumn{1}{c|}{96.75} &
  \multicolumn{1}{c|}{83.64} &
  \multicolumn{1}{c|}{58.69} &
  \multicolumn{1}{c|}{35.11} &
  \multicolumn{1}{c|}{17.87} &
  \multicolumn{1}{c|}{46.51} &
  \multicolumn{1}{c|}{36.55} &
  \multicolumn{1}{c|}{40.12} &
  49.01 &
  1.29 \\ \cmidrule{2-15} 
 &
  \cellcolor[HTML]{FBB982} &
  U-Net &
  \multicolumn{1}{c|}{44.36} &
  \multicolumn{1}{c|}{45.08} &
  \multicolumn{1}{c|}{92.58} &
  \multicolumn{1}{c|}{81.46} &
  \multicolumn{1}{c|}{53.53} &
  \multicolumn{1}{c|}{33.73} &
  \multicolumn{1}{c|}{13.71} &
  \multicolumn{1}{c|}{47.84} &
  \multicolumn{1}{c|}{38.91} &
  \multicolumn{1}{c|}{48.31} &
  46.44 &
  1.59 \\
 &
  \cellcolor[HTML]{FBB982} &
  FPN &
  \multicolumn{1}{c|}{44.71} &
  \multicolumn{1}{c|}{45.79} &
  \multicolumn{1}{c|}{92.44} &
  \multicolumn{1}{c|}{81.59} &
  \multicolumn{1}{c|}{55.21} &
  \multicolumn{1}{c|}{34.30} &
  \multicolumn{1}{c|}{16.53} &
  \multicolumn{1}{c|}{48.14} &
  \multicolumn{1}{c|}{39.63} &
  \multicolumn{1}{c|}{47.80} &
  48.00 &
  1.10 \\
 &
  \cellcolor[HTML]{FBB982} &
  DeepLab &
  \multicolumn{1}{c|}{35.33} &
  \multicolumn{1}{c|}{37.64} &
  \multicolumn{1}{c|}{91.48} &
  \multicolumn{1}{c|}{84.88} &
  \multicolumn{1}{c|}{42.92} &
  \multicolumn{1}{c|}{26.73} &
  \multicolumn{1}{c|}{13.08} &
  \multicolumn{1}{c|}{39.63} &
  \multicolumn{1}{c|}{31.39} &
  \multicolumn{1}{c|}{37.54} &
  38.22 &
  0.52 \\
\multirow{-12}{*}{AP + HP} &
  \multirow{-4}{*}{\cellcolor[HTML]{FBB982}HP} &
  MA-Net &
  \multicolumn{1}{c|}{46.22} &
  \multicolumn{1}{c|}{47.96} &
  \multicolumn{1}{c|}{92.48} &
  \multicolumn{1}{c|}{77.66} &
  \multicolumn{1}{c|}{60.00} &
  \multicolumn{1}{c|}{35.27} &
  \multicolumn{1}{c|}{12.72} &
  \multicolumn{1}{c|}{52.77} &
  \multicolumn{1}{c|}{39.46} &
  \multicolumn{1}{c|}{45.47} &
  53.26 &
  1.11 \\ \midrule
 &
  \cellcolor[HTML]{FBB982} &
  U-Net &
  \multicolumn{1}{c|}{63.46} &
  \multicolumn{1}{c|}{62.33} &
  \multicolumn{1}{c|}{96.27} &
  \multicolumn{1}{c|}{81.92} &
  \multicolumn{1}{c|}{76.44} &
  \multicolumn{1}{c|}{53.00} &
  \multicolumn{1}{c|}{31.88} &
  \multicolumn{1}{c|}{71.54} &
  \multicolumn{1}{c|}{59.01} &
  \multicolumn{1}{c|}{76.71} &
  64.98 &
  0.00 \\
 &
  \cellcolor[HTML]{FBB982} &
  FPN &
  \multicolumn{1}{c|}{65.94} &
  \multicolumn{1}{c|}{67.91} &
  \multicolumn{1}{c|}{96.62} &
  \multicolumn{1}{c|}{85.96} &
  \multicolumn{1}{c|}{78.00} &
  \multicolumn{1}{c|}{56.60} &
  \multicolumn{1}{c|}{33.54} &
  \multicolumn{1}{c|}{76.45} &
  \multicolumn{1}{c|}{62.42} &
  \multicolumn{1}{c|}{78.40} &
  66.03 &
  0.00 \\
 &
  \cellcolor[HTML]{FBB982} &
  DeepLab &
  \multicolumn{1}{c|}{34.08} &
  \multicolumn{1}{c|}{31.70} &
  \multicolumn{1}{c|}{93.72} &
  \multicolumn{1}{c|}{97.37} &
  \multicolumn{1}{c|}{32.93} &
  \multicolumn{1}{c|}{23.20} &
  \multicolumn{1}{c|}{12.38} &
  \multicolumn{1}{c|}{38.05} &
  \multicolumn{1}{c|}{27.17} &
  \multicolumn{1}{c|}{39.51} &
  31.80 &
  0.00 \\
 &
  \multirow{-4}{*}{\cellcolor[HTML]{FBB982}IP} &
  MA-Net &
  \multicolumn{1}{c|}{59.90} &
  \multicolumn{1}{c|}{63.50} &
  \multicolumn{1}{c|}{96.44} &
  \multicolumn{1}{c|}{91.52} &
  \multicolumn{1}{c|}{67.92} &
  \multicolumn{1}{c|}{49.38} &
  \multicolumn{1}{c|}{29.83} &
  \multicolumn{1}{c|}{69.09} &
  \multicolumn{1}{c|}{56.98} &
  \multicolumn{1}{c|}{69.14} &
  65.00 &
  0.00 \\ \cmidrule{2-15}
 &
  \cellcolor[HTML]{FBB982} &
  U-Net &
  \multicolumn{1}{c|}{47.82} &
  \multicolumn{1}{c|}{42.56} &
  \multicolumn{1}{c|}{95.99} &
  \multicolumn{1}{c|}{82.38} &
  \multicolumn{1}{c|}{47.09} &
  \multicolumn{1}{c|}{42.49} &
  \multicolumn{1}{c|}{12.88} &
  \multicolumn{1}{c|}{57.74} &
  \multicolumn{1}{c|}{48.89} &
  \multicolumn{1}{c|}{62.04} &
  56.23 &
  2.27 \\
 &
  \cellcolor[HTML]{FBB982} &
  FPN &
  \multicolumn{1}{c|}{48.26} &
  \multicolumn{1}{c|}{45.91} &
  \multicolumn{1}{c|}{96.51} &
  \multicolumn{1}{c|}{87.12} &
  \multicolumn{1}{c|}{51.16} &
  \multicolumn{1}{c|}{40.66} &
  \multicolumn{1}{c|}{17.63} &
  \multicolumn{1}{c|}{52.17} &
  \multicolumn{1}{c|}{45.56} &
  \multicolumn{1}{c|}{54.57} &
  50.37 &
  0.76 \\
 &
  \cellcolor[HTML]{FBB982} &
  DeepLab &
  \multicolumn{1}{c|}{37.78} &
  \multicolumn{1}{c|}{40.50} &
  \multicolumn{1}{c|}{95.91} &
  \multicolumn{1}{c|}{86.91} &
  \multicolumn{1}{c|}{45.30} &
  \multicolumn{1}{c|}{33.36} &
  \multicolumn{1}{c|}{11.56} &
  \multicolumn{1}{c|}{45.32} &
  \multicolumn{1}{c|}{38.27} &
  \multicolumn{1}{c|}{43.74} &
  45.93 &
  4.55 \\
 &
  \multirow{-4}{*}{\cellcolor[HTML]{FBB982}AP} &
  MA-Net &
  \multicolumn{1}{c|}{49.94} &
  \multicolumn{1}{c|}{54.05} &
  \multicolumn{1}{c|}{97.20} &
  \multicolumn{1}{c|}{83.96} &
  \multicolumn{1}{c|}{59.85} &
  \multicolumn{1}{c|}{45.62} &
  \multicolumn{1}{c|}{23.38} &
  \multicolumn{1}{c|}{59.47} &
  \multicolumn{1}{c|}{49.20} &
  \multicolumn{1}{c|}{55.00} &
  57.47 &
  0.45 \\ \cmidrule{2-15}
 &
  \cellcolor[HTML]{FBB982} &
  U-Net &
  \multicolumn{1}{c|}{43.28} &
  \multicolumn{1}{c|}{45.23} &
  \multicolumn{1}{c|}{92.14} &
  \multicolumn{1}{c|}{79.73} &
  \multicolumn{1}{c|}{55.20} &
  \multicolumn{1}{c|}{33.74} &
  \multicolumn{1}{c|}{11.50} &
  \multicolumn{1}{c|}{48.29} &
  \multicolumn{1}{c|}{40.17} &
  \multicolumn{1}{c|}{46.34} &
  49.30 &
  0.93 \\
 &
  \cellcolor[HTML]{FBB982} &
  FPN &
  \multicolumn{1}{c|}{48.39} &
  \multicolumn{1}{c|}{49.13} &
  \multicolumn{1}{c|}{92.98} &
  \multicolumn{1}{c|}{82.66} &
  \multicolumn{1}{c|}{58.38} &
  \multicolumn{1}{c|}{37.96} &
  \multicolumn{1}{c|}{17.77} &
  \multicolumn{1}{c|}{50.42} &
  \multicolumn{1}{c|}{44.69} &
  \multicolumn{1}{c|}{53.45} &
  52.18 &
  1.36 \\
 &
  \cellcolor[HTML]{FBB982} &
  DeepLab &
  \multicolumn{1}{c|}{37.93} &
  \multicolumn{1}{c|}{39.83} &
  \multicolumn{1}{c|}{91.63} &
  \multicolumn{1}{c|}{81.11} &
  \multicolumn{1}{c|}{49.48} &
  \multicolumn{1}{c|}{27.14} &
  \multicolumn{1}{c|}{11.14} &
  \multicolumn{1}{c|}{39.53} &
  \multicolumn{1}{c|}{30.97} &
  \multicolumn{1}{c|}{36.07} &
  40.91 &
  0.50 \\
\multirow{-12}{*}{IP + AP + HP} &
  \multirow{-4}{*}{\cellcolor[HTML]{FBB982}HP} &
  MA-Net &
  \multicolumn{1}{c|}{42.44} &
  \multicolumn{1}{c|}{44.13} &
  \multicolumn{1}{c|}{92.46} &
  \multicolumn{1}{c|}{82.88} &
  \multicolumn{1}{c|}{51.18} &
  \multicolumn{1}{c|}{32.57} &
  \multicolumn{1}{c|}{13.42} &
  \multicolumn{1}{c|}{45.70} &
  \multicolumn{1}{c|}{38.17} &
  \multicolumn{1}{c|}{43.90} &
  48.02 &
  0.35 \\ \bottomrule
\end{tabular}
}
\caption{Results for BM-1 on $\mathcal{D}_{id}$ (\textit{highlighted}) and $\mathcal{D}_{ood}$ for multi-region $\mathcal{T}_{pn}$.}
\label{tab:bm_1_multi_all_results}
\end{table*}

\begin{table*}[!ht]
\setlength\tabcolsep{4.0pt}
\setlength{\belowcaptionskip}{-10pt}
\centering
\footnotesize
\resizebox{0.96\linewidth}{!}{
\begin{tabular}{c|c|c|ccccccccccc|c}
\toprule[1.5pt]
\multirow{2}{*}{\textbf{\begin{tabular}[c]{@{}c@{}}Training\\Category\end{tabular}}} &
  \multirow{2}{*}{\textbf{\begin{tabular}[c]{@{}c@{}}Testing\\Category\end{tabular}}} &
  \multirow{2}{*}{\textbf{\begin{tabular}[c]{@{}c@{}}Training\\Model\end{tabular}}} &
  \multicolumn{11}{c|}{\textbf{Cone}} &
  \textbf{Non-cone} \\ \cmidrule{4-15} 
 &
   &
   &
  \multicolumn{1}{c|}{\textbf{\begin{tabular}[c]{@{}c@{}}Mask\\IoU\end{tabular}}} &
  \multicolumn{1}{c|}{\textbf{\begin{tabular}[c]{@{}c@{}}Pixel\\IoU\end{tabular}}} &
  \multicolumn{1}{c|}{\textbf{\begin{tabular}[c]{@{}c@{}}Pixel\\Accuracy\end{tabular}}} &
  \multicolumn{1}{c|}{\textbf{\begin{tabular}[c]{@{}c@{}}Pixel\\Precision\end{tabular}}} &
  \multicolumn{1}{c|}{\textbf{\begin{tabular}[c]{@{}c@{}}Pixel\\Recall\end{tabular}}} &
  \multicolumn{1}{c|}{\textbf{\begin{tabular}[c]{@{}c@{}}Panoptic\\Quality\end{tabular}}} &
  \multicolumn{1}{c|}{\textbf{mAP}} &
  \multicolumn{1}{c|}{\textbf{\begin{tabular}[c]{@{}c@{}}Object\\IoU\end{tabular}}} &
  \multicolumn{1}{c|}{\textbf{\begin{tabular}[c]{@{}c@{}}Object\\Accuracy\end{tabular}}} &
  \multicolumn{1}{c|}{\textbf{\begin{tabular}[c]{@{}c@{}}Object\\Precision\end{tabular}}} &
  \textbf{\begin{tabular}[c]{@{}c@{}}Object\\ Recall\end{tabular}} &
  \textbf{$A_{FP}$} \\ \midrule[1pt]
 &
  \cellcolor[HTML]{FBB982} &
  U-Net &
  \multicolumn{1}{c|}{31.28} &
  \multicolumn{1}{c|}{28.64} &
  \multicolumn{1}{c|}{97.59} &
  \multicolumn{1}{c|}{74.69} &
  \multicolumn{1}{c|}{37.39} &
  \multicolumn{1}{c|}{19.43} &
  \multicolumn{1}{c|}{11.08} &
  \multicolumn{1}{c|}{29.46} &
  \multicolumn{1}{c|}{22.53} &
  \multicolumn{1}{c|}{31.36} &
  27.35 &
  0.29 \\
 &
  \cellcolor[HTML]{FBB982} &
  FPN &
  \multicolumn{1}{c|}{25.06} &
  \multicolumn{1}{c|}{25.06} &
  \multicolumn{1}{c|}{97.51} &
  \multicolumn{1}{c|}{76.26} &
  \multicolumn{1}{c|}{31.68} &
  \multicolumn{1}{c|}{15.64} &
  \multicolumn{1}{c|}{8.27} &
  \multicolumn{1}{c|}{23.31} &
  \multicolumn{1}{c|}{17.89} &
  \multicolumn{1}{c|}{21.62} &
  22.65 &
  0.05 \\
 &
  \cellcolor[HTML]{FBB982} &
  DeepLab &
  \multicolumn{1}{c|}{22.86} &
  \multicolumn{1}{c|}{22.70} &
  \multicolumn{1}{c|}{97.64} &
  \multicolumn{1}{c|}{78.85} &
  \multicolumn{1}{c|}{26.56} &
  \multicolumn{1}{c|}{13.93} &
  \multicolumn{1}{c|}{9.58} &
  \multicolumn{1}{c|}{22.13} &
  \multicolumn{1}{c|}{16.39} &
  \multicolumn{1}{c|}{22.06} &
  21.55 &
  0.12 \\
 &
  \multirow{-4}{*}{\cellcolor[HTML]{FBB982}S} &
  MA-Net &
  \multicolumn{1}{c|}{27.41} &
  \multicolumn{1}{c|}{26.99} &
  \multicolumn{1}{c|}{97.70} &
  \multicolumn{1}{c|}{76.21} &
  \multicolumn{1}{c|}{33.96} &
  \multicolumn{1}{c|}{17.56} &
  \multicolumn{1}{c|}{12.09} &
  \multicolumn{1}{c|}{25.93} &
  \multicolumn{1}{c|}{20.55} &
  \multicolumn{1}{c|}{25.74} &
  27.38 &
  0.03 \\\cmidrule{2-15}
 &
   &
  U-Net &
  \multicolumn{1}{c|}{37.26} &
  \multicolumn{1}{c|}{33.50} &
  \multicolumn{1}{c|}{92.08} &
  \multicolumn{1}{c|}{77.37} &
  \multicolumn{1}{c|}{40.56} &
  \multicolumn{1}{c|}{25.43} &
  \multicolumn{1}{c|}{12.36} &
  \multicolumn{1}{c|}{39.01} &
  \multicolumn{1}{c|}{31.46} &
  \multicolumn{1}{c|}{40.73} &
  35.74 &
  0.15 \\
 &
   &
  FPN &
  \multicolumn{1}{c|}{27.98} &
  \multicolumn{1}{c|}{24.97} &
  \multicolumn{1}{c|}{91.47} &
  \multicolumn{1}{c|}{88.86} &
  \multicolumn{1}{c|}{26.86} &
  \multicolumn{1}{c|}{17.51} &
  \multicolumn{1}{c|}{12.50} &
  \multicolumn{1}{c|}{25.61} &
  \multicolumn{1}{c|}{21.62} &
  \multicolumn{1}{c|}{27.61} &
  25.46 &
  0.03 \\
 &
   &
  DeepLab &
  \multicolumn{1}{c|}{28.97} &
  \multicolumn{1}{c|}{27.74} &
  \multicolumn{1}{c|}{91.90} &
  \multicolumn{1}{c|}{91.70} &
  \multicolumn{1}{c|}{29.37} &
  \multicolumn{1}{c|}{20.02} &
  \multicolumn{1}{c|}{10.37} &
  \multicolumn{1}{c|}{27.71} &
  \multicolumn{1}{c|}{25.98} &
  \multicolumn{1}{c|}{31.30} &
  29.45 &
  0.05 \\
 &
  \multirow{-4}{*}{M} &
  MA-Net &
  \multicolumn{1}{c|}{27.14} &
  \multicolumn{1}{c|}{27.18} &
  \multicolumn{1}{c|}{91.76} &
  \multicolumn{1}{c|}{93.03} &
  \multicolumn{1}{c|}{29.01} &
  \multicolumn{1}{c|}{20.33} &
  \multicolumn{1}{c|}{10.09} &
  \multicolumn{1}{c|}{30.69} &
  \multicolumn{1}{c|}{23.22} &
  \multicolumn{1}{c|}{26.82} &
  29.64 &
  0.02 \\\cmidrule{2-15}
 &
   &
  U-Net &
  \multicolumn{1}{c|}{17.72} &
  \multicolumn{1}{c|}{15.21} &
  \multicolumn{1}{c|}{79.38} &
  \multicolumn{1}{c|}{84.34} &
  \multicolumn{1}{c|}{15.92} &
  \multicolumn{1}{c|}{9.93} &
  \multicolumn{1}{c|}{7.22} &
  \multicolumn{1}{c|}{15.97} &
  \multicolumn{1}{c|}{11.72} &
  \multicolumn{1}{c|}{14.02} &
  14.30 &
  0.11 \\
 &
   &
  FPN &
  \multicolumn{1}{c|}{15.51} &
  \multicolumn{1}{c|}{14.86} &
  \multicolumn{1}{c|}{79.70} &
  \multicolumn{1}{c|}{91.64} &
  \multicolumn{1}{c|}{15.33} &
  \multicolumn{1}{c|}{9.43} &
  \multicolumn{1}{c|}{5.93} &
  \multicolumn{1}{c|}{13.24} &
  \multicolumn{1}{c|}{12.78} &
  \multicolumn{1}{c|}{15.02} &
  14.91 &
  0.01 \\
 &
   &
  DeepLab &
  \multicolumn{1}{c|}{14.93} &
  \multicolumn{1}{c|}{13.61} &
  \multicolumn{1}{c|}{78.72} &
  \multicolumn{1}{c|}{89.78} &
  \multicolumn{1}{c|}{13.99} &
  \multicolumn{1}{c|}{8.92} &
  \multicolumn{1}{c|}{7.08} &
  \multicolumn{1}{c|}{13.22} &
  \multicolumn{1}{c|}{11.02} &
  \multicolumn{1}{c|}{13.23} &
  14.14 &
  0.10 \\
\multirow{-12}{*}{S} &
  \multirow{-4}{*}{L} &
  MA-Net &
  \multicolumn{1}{c|}{9.81} &
  \multicolumn{1}{c|}{9.46} &
  \multicolumn{1}{c|}{78.01} &
  \multicolumn{1}{c|}{93.44} &
  \multicolumn{1}{c|}{9.97} &
  \multicolumn{1}{c|}{6.96} &
  \multicolumn{1}{c|}{4.66} &
  \multicolumn{1}{c|}{9.46} &
  \multicolumn{1}{c|}{9.08} &
  \multicolumn{1}{c|}{9.54} &
  12.39 &
  0.05 \\ \midrule
 &
   &
  U-Net &
  \multicolumn{1}{c|}{18.73} &
  \multicolumn{1}{c|}{17.50} &
  \multicolumn{1}{c|}{96.96} &
  \multicolumn{1}{c|}{75.53} &
  \multicolumn{1}{c|}{22.02} &
  \multicolumn{1}{c|}{9.95} &
  \multicolumn{1}{c|}{7.84} &
  \multicolumn{1}{c|}{15.16} &
  \multicolumn{1}{c|}{12.15} &
  \multicolumn{1}{c|}{16.67} &
  14.69 &
  0.57 \\
 &
   &
  FPN &
  \multicolumn{1}{c|}{26.27} &
  \multicolumn{1}{c|}{25.42} &
  \multicolumn{1}{c|}{97.30} &
  \multicolumn{1}{c|}{69.99} &
  \multicolumn{1}{c|}{33.05} &
  \multicolumn{1}{c|}{16.84} &
  \multicolumn{1}{c|}{10.43} &
  \multicolumn{1}{c|}{24.24} &
  \multicolumn{1}{c|}{20.04} &
  \multicolumn{1}{c|}{28.44} &
  21.97 &
  0.22 \\
 &
   &
  DeepLab &
  \multicolumn{1}{c|}{25.9} &
  \multicolumn{1}{c|}{24.83} &
  \multicolumn{1}{c|}{97.01} &
  \multicolumn{1}{c|}{66.84} &
  \multicolumn{1}{c|}{32.77} &
  \multicolumn{1}{c|}{16.56} &
  \multicolumn{1}{c|}{11.03} &
  \multicolumn{1}{c|}{28.60} &
  \multicolumn{1}{c|}{18.60} &
  \multicolumn{1}{c|}{28.65} &
  23.45 &
  0.48 \\
 &
  \multirow{-4}{*}{S} &
  MA-Net &
  \multicolumn{1}{c|}{29.04} &
  \multicolumn{1}{c|}{28.68} &
  \multicolumn{1}{c|}{97.35} &
  \multicolumn{1}{c|}{70.84} &
  \multicolumn{1}{c|}{37.84} &
  \multicolumn{1}{c|}{17.77} &
  \multicolumn{1}{c|}{11.37} &
  \multicolumn{1}{c|}{28.71} &
  \multicolumn{1}{c|}{19.98} &
  \multicolumn{1}{c|}{28.59} &
  25.22 &
  0.31 \\\cmidrule{2-15}
 &
  \cellcolor[HTML]{FBB982} &
  U-Net &
  \multicolumn{1}{c|}{36.26} &
  \multicolumn{1}{c|}{33.19} &
  \multicolumn{1}{c|}{92.43} &
  \multicolumn{1}{c|}{81.92} &
  \multicolumn{1}{c|}{37.17} &
  \multicolumn{1}{c|}{20.92} &
  \multicolumn{1}{c|}{12.54} &
  \multicolumn{1}{c|}{36.28} &
  \multicolumn{1}{c|}{24.23} &
  \multicolumn{1}{c|}{36.05} &
  28.58 &
  0.35 \\
 &
  \cellcolor[HTML]{FBB982} &
  FPN &
  \multicolumn{1}{c|}{37.53} &
  \multicolumn{1}{c|}{36.01} &
  \multicolumn{1}{c|}{92.45} &
  \multicolumn{1}{c|}{81.41} &
  \multicolumn{1}{c|}{41.07} &
  \multicolumn{1}{c|}{25.14} &
  \multicolumn{1}{c|}{13.28} &
  \multicolumn{1}{c|}{37.48} &
  \multicolumn{1}{c|}{30.31} &
  \multicolumn{1}{c|}{39.71} &
  34.58 &
  0.23 \\
 &
  \cellcolor[HTML]{FBB982} &
  DeepLab &
  \multicolumn{1}{c|}{38.29} &
  \multicolumn{1}{c|}{40.80} &
  \multicolumn{1}{c|}{92.82} &
  \multicolumn{1}{c|}{80.68} &
  \multicolumn{1}{c|}{48.21} &
  \multicolumn{1}{c|}{29.42} &
  \multicolumn{1}{c|}{10.96} &
  \multicolumn{1}{c|}{44.83} &
  \multicolumn{1}{c|}{35.23} &
  \multicolumn{1}{c|}{43.75} &
  43.03 &
  0.31 \\
 &
  \multirow{-4}{*}{\cellcolor[HTML]{FBB982}M} &
  MA-Net &
  \multicolumn{1}{c|}{47.28} &
  \multicolumn{1}{c|}{48.65} &
  \multicolumn{1}{c|}{93.89} &
  \multicolumn{1}{c|}{82.04} &
  \multicolumn{1}{c|}{55.15} &
  \multicolumn{1}{c|}{35.21} &
  \multicolumn{1}{c|}{13.83} &
  \multicolumn{1}{c|}{54.01} &
  \multicolumn{1}{c|}{40.83} &
  \multicolumn{1}{c|}{50.15} &
  49.81 &
  0.49 \\\cmidrule{2-15}
 &
   &
  U-Net &
  \multicolumn{1}{c|}{34.11} &
  \multicolumn{1}{c|}{33.40} &
  \multicolumn{1}{c|}{83.53} &
  \multicolumn{1}{c|}{90.37} &
  \multicolumn{1}{c|}{34.82} &
  \multicolumn{1}{c|}{21.58} &
  \multicolumn{1}{c|}{14.30} &
  \multicolumn{1}{c|}{29.02} &
  \multicolumn{1}{c|}{28.02} &
  \multicolumn{1}{c|}{32.43} &
  32.11 &
  0.35 \\
 &
   &
  FPN &
  \multicolumn{1}{c|}{36.97} &
  \multicolumn{1}{c|}{34.83} &
  \multicolumn{1}{c|}{83.11} &
  \multicolumn{1}{c|}{90.43} &
  \multicolumn{1}{c|}{36.62} &
  \multicolumn{1}{c|}{26.31} &
  \multicolumn{1}{c|}{15.41} &
  \multicolumn{1}{c|}{33.31} &
  \multicolumn{1}{c|}{32.99} &
  \multicolumn{1}{c|}{38.30} &
  36.62 &
  0.31 \\
 &
   &
  DeepLab &
  \multicolumn{1}{c|}{37.42} &
  \multicolumn{1}{c|}{37.38} &
  \multicolumn{1}{c|}{83.88} &
  \multicolumn{1}{c|}{92.04} &
  \multicolumn{1}{c|}{39.86} &
  \multicolumn{1}{c|}{27.29} &
  \multicolumn{1}{c|}{14.41} &
  \multicolumn{1}{c|}{39.94} &
  \multicolumn{1}{c|}{32.95} &
  \multicolumn{1}{c|}{36.53} &
  43.27 &
  0.36 \\
\multirow{-12}{*}{M} &
  \multirow{-4}{*}{L} &
  MA-Net &
  \multicolumn{1}{c|}{43.82} &
  \multicolumn{1}{c|}{44.02} &
  \multicolumn{1}{c|}{85.44} &
  \multicolumn{1}{c|}{91.21} &
  \multicolumn{1}{c|}{47.17} &
  \multicolumn{1}{c|}{34.27} &
  \multicolumn{1}{c|}{18.82} &
  \multicolumn{1}{c|}{48.67} &
  \multicolumn{1}{c|}{39.69} &
  \multicolumn{1}{c|}{44.15} &
  53.21 &
  0.30 \\ \midrule
 &
   &
  U-Net &
  \multicolumn{1}{c|}{11.56} &
  \multicolumn{1}{c|}{10.63} &
  \multicolumn{1}{c|}{96.44} &
  \multicolumn{1}{c|}{74.30} &
  \multicolumn{1}{c|}{15.40} &
  \multicolumn{1}{c|}{6.57} &
  \multicolumn{1}{c|}{1.97} &
  \multicolumn{1}{c|}{11.81} &
  \multicolumn{1}{c|}{7.94} &
  \multicolumn{1}{c|}{11.86} &
  10.27 &
  1.50 \\
 &
   &
  FPN &
  \multicolumn{1}{c|}{16.76} &
  \multicolumn{1}{c|}{15.81} &
  \multicolumn{1}{c|}{95.30} &
  \multicolumn{1}{c|}{49.56} &
  \multicolumn{1}{c|}{27.48} &
  \multicolumn{1}{c|}{7.47} &
  \multicolumn{1}{c|}{4.84} &
  \multicolumn{1}{c|}{13.84} &
  \multicolumn{1}{c|}{8.58} &
  \multicolumn{1}{c|}{13.21} &
  11.44 &
  1.39 \\
 &
   &
  DeepLab &
  \multicolumn{1}{c|}{8.84} &
  \multicolumn{1}{c|}{9.01} &
  \multicolumn{1}{c|}{95.54} &
  \multicolumn{1}{c|}{65.53} &
  \multicolumn{1}{c|}{14.45} &
  \multicolumn{1}{c|}{3.99} &
  \multicolumn{1}{c|}{1.92} &
  \multicolumn{1}{c|}{6.75} &
  \multicolumn{1}{c|}{4.72} &
  \multicolumn{1}{c|}{6.57} &
  5.86 &
  1.35 \\
 &
  \multirow{-4}{*}{S} &
  MA-Net &
  \multicolumn{1}{c|}{12.73} &
  \multicolumn{1}{c|}{12.13} &
  \multicolumn{1}{c|}{93.70} &
  \multicolumn{1}{c|}{58.47} &
  \multicolumn{1}{c|}{25.67} &
  \multicolumn{1}{c|}{6.16} &
  \multicolumn{1}{c|}{1.81} &
  \multicolumn{1}{c|}{12.07} &
  \multicolumn{1}{c|}{7.57} &
  \multicolumn{1}{c|}{10.08} &
  11.09 &
  0.88 \\\cmidrule{2-15}
 &
   &
  U-Net &
  \multicolumn{1}{c|}{36.38} &
  \multicolumn{1}{c|}{36.29} &
  \multicolumn{1}{c|}{91.79} &
  \multicolumn{1}{c|}{77.73} &
  \multicolumn{1}{c|}{45.80} &
  \multicolumn{1}{c|}{21.86} &
  \multicolumn{1}{c|}{7.67} &
  \multicolumn{1}{c|}{39.88} &
  \multicolumn{1}{c|}{24.59} &
  \multicolumn{1}{c|}{34.69} &
  31.55 &
  0.78 \\
 &
   &
  FPN &
  \multicolumn{1}{c|}{42.22} &
  \multicolumn{1}{c|}{41.93} &
  \multicolumn{1}{c|}{91.84} &
  \multicolumn{1}{c|}{63.61} &
  \multicolumn{1}{c|}{59.94} &
  \multicolumn{1}{c|}{27.04} &
  \multicolumn{1}{c|}{13.08} &
  \multicolumn{1}{c|}{43.91} &
  \multicolumn{1}{c|}{33.15} &
  \multicolumn{1}{c|}{48.81} &
  39.00 &
  1.20 \\
 &
   &
  DeepLab &
  \multicolumn{1}{c|}{31.74} &
  \multicolumn{1}{c|}{34.10} &
  \multicolumn{1}{c|}{91.04} &
  \multicolumn{1}{c|}{69.21} &
  \multicolumn{1}{c|}{46.22} &
  \multicolumn{1}{c|}{19.36} &
  \multicolumn{1}{c|}{6.51} &
  \multicolumn{1}{c|}{36.39} &
  \multicolumn{1}{c|}{20.26} &
  \multicolumn{1}{c|}{30.12} &
  26.26 &
  1.22 \\
 &
  \multirow{-4}{*}{M} &
  MA-Net &
  \multicolumn{1}{c|}{34.41} &
  \multicolumn{1}{c|}{36.35} &
  \multicolumn{1}{c|}{90.84} &
  \multicolumn{1}{c|}{71.19} &
  \multicolumn{1}{c|}{54.10} &
  \multicolumn{1}{c|}{22.09} &
  \multicolumn{1}{c|}{6.97} &
  \multicolumn{1}{c|}{40.42} &
  \multicolumn{1}{c|}{25.27} &
  \multicolumn{1}{c|}{34.41} &
  32.21 &
  0.44 \\\cmidrule{2-15}
 &
  \cellcolor[HTML]{FBB982} &
  U-Net &
  \multicolumn{1}{c|}{56.87} &
  \multicolumn{1}{c|}{60.42} &
  \multicolumn{1}{c|}{90.61} &
  \multicolumn{1}{c|}{89.93} &
  \multicolumn{1}{c|}{66.75} &
  \multicolumn{1}{c|}{44.57} &
  \multicolumn{1}{c|}{21.02} &
  \multicolumn{1}{c|}{62.86} &
  \multicolumn{1}{c|}{50.09} &
  \multicolumn{1}{c|}{57.21} &
  67.81 &
  1.08 \\
 &
  \cellcolor[HTML]{FBB982} &
  FPN &
  \multicolumn{1}{c|}{59.49} &
  \multicolumn{1}{c|}{63.69} &
  \multicolumn{1}{c|}{90.40} &
  \multicolumn{1}{c|}{81.33} &
  \multicolumn{1}{c|}{75.75} &
  \multicolumn{1}{c|}{47.49} &
  \multicolumn{1}{c|}{21.45} &
  \multicolumn{1}{c|}{59.68} &
  \multicolumn{1}{c|}{56.28} &
  \multicolumn{1}{c|}{63.49} &
  64.45 &
  1.00 \\
 &
  \cellcolor[HTML]{FBB982} &
  DeepLab &
  \multicolumn{1}{c|}{61.76} &
  \multicolumn{1}{c|}{64.81} &
  \multicolumn{1}{c|}{90.80} &
  \multicolumn{1}{c|}{82.05} &
  \multicolumn{1}{c|}{74.87} &
  \multicolumn{1}{c|}{51.98} &
  \multicolumn{1}{c|}{24.86} &
  \multicolumn{1}{c|}{64.94} &
  \multicolumn{1}{c|}{60.06} &
  \multicolumn{1}{c|}{68.78} &
  68.43 &
  1.05 \\
\multirow{-12}{*}{L} &
  \multirow{-4}{*}{\cellcolor[HTML]{FBB982}L} &
  MA-Net &
  \multicolumn{1}{c|}{59.47} &
  \multicolumn{1}{c|}{63.61} &
  \multicolumn{1}{c|}{90.89} &
  \multicolumn{1}{c|}{84.67} &
  \multicolumn{1}{c|}{73.21} &
  \multicolumn{1}{c|}{47.50} &
  \multicolumn{1}{c|}{22.85} &
  \multicolumn{1}{c|}{61.79} &
  \multicolumn{1}{c|}{53.68} &
  \multicolumn{1}{c|}{59.97} &
  66.51 &
  0.38 \\ \bottomrule
\end{tabular}
}
\caption{Results for BM-2 on $\mathcal{D}_{id}$ (\textit{highlighted}) and $\mathcal{D}_{ood}$ for single-size category $\mathcal{T}_{pn}$.}
\label{tab:bm_2_single_all_results}
\end{table*}

\begin{table*}[!ht]
\setlength\tabcolsep{4.0pt}
\setlength{\belowcaptionskip}{-10pt}
\centering
\footnotesize
\resizebox{0.96\linewidth}{!}{
\begin{tabular}{c|c|c|ccccccccccc|c}
\toprule[1.5pt]
\multirow{2}{*}{\textbf{\begin{tabular}[c]{@{}c@{}}Training\\Category\end{tabular}}} &
  \multirow{2}{*}{\textbf{\begin{tabular}[c]{@{}c@{}}Testing\\Category\end{tabular}}} &
  \multirow{2}{*}{\textbf{\begin{tabular}[c]{@{}c@{}}Training\\Model\end{tabular}}} &
  \multicolumn{11}{c|}{\textbf{Cone}} &
  \textbf{Non-cone} \\ \cmidrule{4-15} 
 &
   &
   &
  \multicolumn{1}{c|}{\textbf{\begin{tabular}[c]{@{}c@{}}Mask\\IoU\end{tabular}}} &
  \multicolumn{1}{c|}{\textbf{\begin{tabular}[c]{@{}c@{}}Pixel\\IoU\end{tabular}}} &
  \multicolumn{1}{c|}{\textbf{\begin{tabular}[c]{@{}c@{}}Pixel\\Accuracy\end{tabular}}} &
  \multicolumn{1}{c|}{\textbf{\begin{tabular}[c]{@{}c@{}}Pixel\\Precision\end{tabular}}} &
  \multicolumn{1}{c|}{\textbf{\begin{tabular}[c]{@{}c@{}}Pixel\\Recall\end{tabular}}} &
  \multicolumn{1}{c|}{\textbf{\begin{tabular}[c]{@{}c@{}}Panoptic\\Quality\end{tabular}}} &
  \multicolumn{1}{c|}{\textbf{mAP}} &
  \multicolumn{1}{c|}{\textbf{\begin{tabular}[c]{@{}c@{}}Object\\IoU\end{tabular}}} &
  \multicolumn{1}{c|}{\textbf{\begin{tabular}[c]{@{}c@{}}Object\\Accuracy\end{tabular}}} &
  \multicolumn{1}{c|}{\textbf{\begin{tabular}[c]{@{}c@{}}Object\\Precision\end{tabular}}} &
  \textbf{\begin{tabular}[c]{@{}c@{}}Object\\ Recall\end{tabular}} &
  \textbf{$A_{FP}$} \\ \midrule[1pt]
 &
  \cellcolor[HTML]{FBB982} &
  U-Net &
  \multicolumn{1}{c|}{32.50} &
  \multicolumn{1}{c|}{30.85} &
  \multicolumn{1}{c|}{97.84} &
  \multicolumn{1}{c|}{79.57} &
  \multicolumn{1}{c|}{37.36} &
  \multicolumn{1}{c|}{21.67} &
  \multicolumn{1}{c|}{14.14} &
  \multicolumn{1}{c|}{30.52} &
  \multicolumn{1}{c|}{25.34} &
  \multicolumn{1}{c|}{34.08} &
  28.33 &
  0.25 \\
 &
  \cellcolor[HTML]{FBB982} &
  FPN &
  \multicolumn{1}{c|}{24.86} &
  \multicolumn{1}{c|}{23.97} &
  \multicolumn{1}{c|}{97.53} &
  \multicolumn{1}{c|}{82.85} &
  \multicolumn{1}{c|}{28.39} &
  \multicolumn{1}{c|}{16.36} &
  \multicolumn{1}{c|}{7.42} &
  \multicolumn{1}{c|}{24.44} &
  \multicolumn{1}{c|}{19.53} &
  \multicolumn{1}{c|}{27.48} &
  22.46 &
  0.13 \\
 &
  \cellcolor[HTML]{FBB982} &
  DeepLab &
  \multicolumn{1}{c|}{25.06} &
  \multicolumn{1}{c|}{25.68} &
  \multicolumn{1}{c|}{97.77} &
  \multicolumn{1}{c|}{85.34} &
  \multicolumn{1}{c|}{30.06} &
  \multicolumn{1}{c|}{17.65} &
  \multicolumn{1}{c|}{11.67} &
  \multicolumn{1}{c|}{24.85} &
  \multicolumn{1}{c|}{20.63} &
  \multicolumn{1}{c|}{27.25} &
  24.85 &
  0.03 \\
 &
  \multirow{-4}{*}{\cellcolor[HTML]{FBB982}S} &
  MA-Net &
  \multicolumn{1}{c|}{39.17} &
  \multicolumn{1}{c|}{37.39} &
  \multicolumn{1}{c|}{97.52} &
  \multicolumn{1}{c|}{62.84} &
  \multicolumn{1}{c|}{54.41} &
  \multicolumn{1}{c|}{26.54} &
  \multicolumn{1}{c|}{14.21} &
  \multicolumn{1}{c|}{41.84} &
  \multicolumn{1}{c|}{30.58} &
  \multicolumn{1}{c|}{38.42} &
  41.57 &
  0.42 \\ \cmidrule{2-15}
 &
  \cellcolor[HTML]{FBB982} &
  U-Net &
  \multicolumn{1}{c|}{45.19} &
  \multicolumn{1}{c|}{41.88} &
  \multicolumn{1}{c|}{93.27} &
  \multicolumn{1}{c|}{85.45} &
  \multicolumn{1}{c|}{47.18} &
  \multicolumn{1}{c|}{34.49} &
  \multicolumn{1}{c|}{19.15} &
  \multicolumn{1}{c|}{48.00} &
  \multicolumn{1}{c|}{41.69} &
  \multicolumn{1}{c|}{54.05} &
  46.08 &
  0.50 \\
 &
  \cellcolor[HTML]{FBB982} &
  FPN &
  \multicolumn{1}{c|}{36.24} &
  \multicolumn{1}{c|}{33.26} &
  \multicolumn{1}{c|}{92.22} &
  \multicolumn{1}{c|}{84.74} &
  \multicolumn{1}{c|}{37.58} &
  \multicolumn{1}{c|}{26.20} &
  \multicolumn{1}{c|}{13.15} &
  \multicolumn{1}{c|}{39.57} &
  \multicolumn{1}{c|}{32.25} &
  \multicolumn{1}{c|}{42.61} &
  37.31 &
  0.26 \\
 &
  \cellcolor[HTML]{FBB982} &
  DeepLab &
  \multicolumn{1}{c|}{29.96} &
  \multicolumn{1}{c|}{28.14} &
  \multicolumn{1}{c|}{91.80} &
  \multicolumn{1}{c|}{91.75} &
  \multicolumn{1}{c|}{30.26} &
  \multicolumn{1}{c|}{20.13} &
  \multicolumn{1}{c|}{11.96} &
  \multicolumn{1}{c|}{29.31} &
  \multicolumn{1}{c|}{24.94} &
  \multicolumn{1}{c|}{31.90} &
  28.41 &
  0.07 \\
 &
  \multirow{-4}{*}{\cellcolor[HTML]{FBB982}M} &
  MA-Net &
  \multicolumn{1}{c|}{49.86} &
  \multicolumn{1}{c|}{50.88} &
  \multicolumn{1}{c|}{94.06} &
  \multicolumn{1}{c|}{81.20} &
  \multicolumn{1}{c|}{61.02} &
  \multicolumn{1}{c|}{38.05} &
  \multicolumn{1}{c|}{18.00} &
  \multicolumn{1}{c|}{55.60} &
  \multicolumn{1}{c|}{43.67} &
  \multicolumn{1}{c|}{50.20} &
  56.56 &
  0.55 \\ \cmidrule{2-15}
 &
   &
  U-Net &
  \multicolumn{1}{c|}{32.82} &
  \multicolumn{1}{c|}{29.90} &
  \multicolumn{1}{c|}{82.65} &
  \multicolumn{1}{c|}{92.01} &
  \multicolumn{1}{c|}{31.07} &
  \multicolumn{1}{c|}{24.13} &
  \multicolumn{1}{c|}{14.53} &
  \multicolumn{1}{c|}{33.21} &
  \multicolumn{1}{c|}{29.35} &
  \multicolumn{1}{c|}{34.88} &
  34.56 &
  0.05 \\
 &
   &
  FPN &
  \multicolumn{1}{c|}{24.17} &
  \multicolumn{1}{c|}{22.83} &
  \multicolumn{1}{c|}{81.06} &
  \multicolumn{1}{c|}{93.26} &
  \multicolumn{1}{c|}{23.51} &
  \multicolumn{1}{c|}{16.70} &
  \multicolumn{1}{c|}{10.72} &
  \multicolumn{1}{c|}{22.56} &
  \multicolumn{1}{c|}{22.63} &
  \multicolumn{1}{c|}{25.61} &
  25.61 &
  0.09 \\
 &
   &
  DeepLab &
  \multicolumn{1}{c|}{18.60} &
  \multicolumn{1}{c|}{16.48} &
  \multicolumn{1}{c|}{79.34} &
  \multicolumn{1}{c|}{93.39} &
  \multicolumn{1}{c|}{16.83} &
  \multicolumn{1}{c|}{13.61} &
  \multicolumn{1}{c|}{8.42} &
  \multicolumn{1}{c|}{19.94} &
  \multicolumn{1}{c|}{16.45} &
  \multicolumn{1}{c|}{20.44} &
  19.95 &
  0.05 \\
\multirow{-12}{*}{S + M} &
  \multirow{-4}{*}{L} &
  MA-Net &
  \multicolumn{1}{c|}{50.00} &
  \multicolumn{1}{c|}{51.11} &
  \multicolumn{1}{c|}{86.35} &
  \multicolumn{1}{c|}{90.39} &
  \multicolumn{1}{c|}{56.31} &
  \multicolumn{1}{c|}{39.35} &
  \multicolumn{1}{c|}{18.89} &
  \multicolumn{1}{c|}{53.78} &
  \multicolumn{1}{c|}{44.56} &
  \multicolumn{1}{c|}{48.19} &
  63.07 &
  0.33 \\ \midrule
 &
   &
  U-Net &
  \multicolumn{1}{c|}{24.94} &
  \multicolumn{1}{c|}{23.63} &
  \multicolumn{1}{c|}{96.83} &
  \multicolumn{1}{c|}{69.32} &
  \multicolumn{1}{c|}{30.29} &
  \multicolumn{1}{c|}{14.14} &
  \multicolumn{1}{c|}{7.37} &
  \multicolumn{1}{c|}{21.23} &
  \multicolumn{1}{c|}{16.79} &
  \multicolumn{1}{c|}{24.06} &
  18.68 &
  1.00 \\
 &
   &
  FPN &
  \multicolumn{1}{c|}{19.34} &
  \multicolumn{1}{c|}{19.01} &
  \multicolumn{1}{c|}{96.60} &
  \multicolumn{1}{c|}{73.26} &
  \multicolumn{1}{c|}{25.05} &
  \multicolumn{1}{c|}{8.21} &
  \multicolumn{1}{c|}{7.23} &
  \multicolumn{1}{c|}{15.24} &
  \multicolumn{1}{c|}{8.94} &
  \multicolumn{1}{c|}{14.33} &
  11.87 &
  0.65 \\
 &
   &
  DeepLab &
  \multicolumn{1}{c|}{16.52} &
  \multicolumn{1}{c|}{17.24} &
  \multicolumn{1}{c|}{96.95} &
  \multicolumn{1}{c|}{80.52} &
  \multicolumn{1}{c|}{21.44} &
  \multicolumn{1}{c|}{9.82} &
  \multicolumn{1}{c|}{8.26} &
  \multicolumn{1}{c|}{15.60} &
  \multicolumn{1}{c|}{11.16} &
  \multicolumn{1}{c|}{16.74} &
  12.87 &
  1.06 \\
 &
  \multirow{-4}{*}{S} &
  MA-Net &
  \multicolumn{1}{c|}{24.41} &
  \multicolumn{1}{c|}{22.42} &
  \multicolumn{1}{c|}{97.02} &
  \multicolumn{1}{c|}{76.47} &
  \multicolumn{1}{c|}{28.04} &
  \multicolumn{1}{c|}{13.51} &
  \multicolumn{1}{c|}{6.36} &
  \multicolumn{1}{c|}{21.96} &
  \multicolumn{1}{c|}{15.71} &
  \multicolumn{1}{c|}{23.10} &
  17.55 &
  0.81 \\ \cmidrule{2-15}
 &
  \cellcolor[HTML]{FBB982} &
  U-Net &
  \multicolumn{1}{c|}{47.06} &
  \multicolumn{1}{c|}{46.28} &
  \multicolumn{1}{c|}{92.98} &
  \multicolumn{1}{c|}{75.76} &
  \multicolumn{1}{c|}{55.25} &
  \multicolumn{1}{c|}{32.41} &
  \multicolumn{1}{c|}{14.89} &
  \multicolumn{1}{c|}{48.40} &
  \multicolumn{1}{c|}{37.85} &
  \multicolumn{1}{c|}{50.29} &
  43.68 &
  0.70 \\
 &
  \cellcolor[HTML]{FBB982} &
  FPN &
  \multicolumn{1}{c|}{40.06} &
  \multicolumn{1}{c|}{40.52} &
  \multicolumn{1}{c|}{92.56} &
  \multicolumn{1}{c|}{76.16} &
  \multicolumn{1}{c|}{48.22} &
  \multicolumn{1}{c|}{26.19} &
  \multicolumn{1}{c|}{15.70} &
  \multicolumn{1}{c|}{41.77} &
  \multicolumn{1}{c|}{31.19} &
  \multicolumn{1}{c|}{42.90} &
  36.84 &
  0.48 \\
 &
  \cellcolor[HTML]{FBB982} &
  DeepLab &
  \multicolumn{1}{c|}{40.07} &
  \multicolumn{1}{c|}{40.49} &
  \multicolumn{1}{c|}{92.49} &
  \multicolumn{1}{c|}{81.37} &
  \multicolumn{1}{c|}{47.06} &
  \multicolumn{1}{c|}{27.67} &
  \multicolumn{1}{c|}{10.12} &
  \multicolumn{1}{c|}{44.56} &
  \multicolumn{1}{c|}{31.98} &
  \multicolumn{1}{c|}{43.67} &
  37.41 &
  0.51 \\
 &
  \multirow{-4}{*}{\cellcolor[HTML]{FBB982}M} &
  MA-Net &
  \multicolumn{1}{c|}{43.92} &
  \multicolumn{1}{c|}{41.96} &
  \multicolumn{1}{c|}{92.71} &
  \multicolumn{1}{c|}{82.85} &
  \multicolumn{1}{c|}{48.76} &
  \multicolumn{1}{c|}{31.25} &
  \multicolumn{1}{c|}{14.89} &
  \multicolumn{1}{c|}{45.75} &
  \multicolumn{1}{c|}{36.98} &
  \multicolumn{1}{c|}{50.01} &
  41.83 &
  0.33 \\ \cmidrule{2-15}
 &
  \cellcolor[HTML]{FBB982} &
  U-Net &
  \multicolumn{1}{c|}{49.92} &
  \multicolumn{1}{c|}{52.92} &
  \multicolumn{1}{c|}{88.97} &
  \multicolumn{1}{c|}{88.46} &
  \multicolumn{1}{c|}{58.01} &
  \multicolumn{1}{c|}{39.83} &
  \multicolumn{1}{c|}{17.20} &
  \multicolumn{1}{c|}{53.63} &
  \multicolumn{1}{c|}{47.05} &
  \multicolumn{1}{c|}{53.07} &
  58.56 &
  2.57 \\
 &
  \cellcolor[HTML]{FBB982} &
  FPN &
  \multicolumn{1}{c|}{51.83} &
  \multicolumn{1}{c|}{53.70} &
  \multicolumn{1}{c|}{88.67} &
  \multicolumn{1}{c|}{88.55} &
  \multicolumn{1}{c|}{59.06} &
  \multicolumn{1}{c|}{41.37} &
  \multicolumn{1}{c|}{17.94} &
  \multicolumn{1}{c|}{54.81} &
  \multicolumn{1}{c|}{48.26} &
  \multicolumn{1}{c|}{56.02} &
  57.57 &
  2.88 \\
 &
  \cellcolor[HTML]{FBB982} &
  DeepLab &
  \multicolumn{1}{c|}{60.37} &
  \multicolumn{1}{c|}{64.01} &
  \multicolumn{1}{c|}{91.55} &
  \multicolumn{1}{c|}{91.58} &
  \multicolumn{1}{c|}{69.28} &
  \multicolumn{1}{c|}{50.19} &
  \multicolumn{1}{c|}{23.65} &
  \multicolumn{1}{c|}{65.89} &
  \multicolumn{1}{c|}{56.61} &
  \multicolumn{1}{c|}{63.98} &
  68.73 &
  2.31 \\
\multirow{-12}{*}{M + L} &
  \multirow{-4}{*}{\cellcolor[HTML]{FBB982}L} &
  MA-Net &
  \multicolumn{1}{c|}{57.52} &
  \multicolumn{1}{c|}{60.06} &
  \multicolumn{1}{c|}{90.42} &
  \multicolumn{1}{c|}{92.17} &
  \multicolumn{1}{c|}{64.18} &
  \multicolumn{1}{c|}{48.82} &
  \multicolumn{1}{c|}{23.54} &
  \multicolumn{1}{c|}{61.74} &
  \multicolumn{1}{c|}{57.25} &
  \multicolumn{1}{c|}{63.24} &
  67.13 &
  1.76 \\ \midrule
 &
  \cellcolor[HTML]{FBB982} &
  U-Net &
  \multicolumn{1}{c|}{15.61} &
  \multicolumn{1}{c|}{14.35} &
  \multicolumn{1}{c|}{97.02} &
  \multicolumn{1}{c|}{84.55} &
  \multicolumn{1}{c|}{18.09} &
  \multicolumn{1}{c|}{7.80} &
  \multicolumn{1}{c|}{5.10} &
  \multicolumn{1}{c|}{12.39} &
  \multicolumn{1}{c|}{9.14} &
  \multicolumn{1}{c|}{14.34} &
  10.32 &
  0.93 \\
 &
  \cellcolor[HTML]{FBB982} &
  FPN &
  \multicolumn{1}{c|}{23.82} &
  \multicolumn{1}{c|}{24.01} &
  \multicolumn{1}{c|}{97.28} &
  \multicolumn{1}{c|}{81.13} &
  \multicolumn{1}{c|}{32.41} &
  \multicolumn{1}{c|}{16.12} &
  \multicolumn{1}{c|}{6.93} &
  \multicolumn{1}{c|}{23.82} &
  \multicolumn{1}{c|}{18.47} &
  \multicolumn{1}{c|}{24.34} &
  23.08 &
  0.48 \\
 &
  \cellcolor[HTML]{FBB982} &
  DeepLab &
  \multicolumn{1}{c|}{18.62} &
  \multicolumn{1}{c|}{18.95} &
  \multicolumn{1}{c|}{97.22} &
  \multicolumn{1}{c|}{82.84} &
  \multicolumn{1}{c|}{22.90} &
  \multicolumn{1}{c|}{11.15} &
  \multicolumn{1}{c|}{5.53} &
  \multicolumn{1}{c|}{17.85} &
  \multicolumn{1}{c|}{13.49} &
  \multicolumn{1}{c|}{19.31} &
  16.04 &
  1.19 \\
 &
  \multirow{-4}{*}{\cellcolor[HTML]{FBB982}S} &
  MA-Net &
  \multicolumn{1}{c|}{32.27} &
  \multicolumn{1}{c|}{31.44} &
  \multicolumn{1}{c|}{97.16} &
  \multicolumn{1}{c|}{65.87} &
  \multicolumn{1}{c|}{45.44} &
  \multicolumn{1}{c|}{19.56} &
  \multicolumn{1}{c|}{7.33} &
  \multicolumn{1}{c|}{31.66} &
  \multicolumn{1}{c|}{22.91} &
  \multicolumn{1}{c|}{27.81} &
  33.09 &
  0.95 \\ \cmidrule{2-15}
 &
   &
  U-Net &
  \multicolumn{1}{c|}{34.43} &
  \multicolumn{1}{c|}{35.82} &
  \multicolumn{1}{c|}{92.78} &
  \multicolumn{1}{c|}{86.03} &
  \multicolumn{1}{c|}{40.87} &
  \multicolumn{1}{c|}{22.62} &
  \multicolumn{1}{c|}{9.83} &
  \multicolumn{1}{c|}{35.77} &
  \multicolumn{1}{c|}{26.82} &
  \multicolumn{1}{c|}{37.47} &
  29.91 &
  0.38 \\
 &
   &
  FPN &
  \multicolumn{1}{c|}{37.92} &
  \multicolumn{1}{c|}{37.05} &
  \multicolumn{1}{c|}{92.63} &
  \multicolumn{1}{c|}{85.55} &
  \multicolumn{1}{c|}{42.08} &
  \multicolumn{1}{c|}{28.02} &
  \multicolumn{1}{c|}{12.42} &
  \multicolumn{1}{c|}{40.55} &
  \multicolumn{1}{c|}{34.11} &
  \multicolumn{1}{c|}{44.38} &
  39.48 &
  0.20 \\
 &
   &
  DeepLab &
  \multicolumn{1}{c|}{31.96} &
  \multicolumn{1}{c|}{34.30} &
  \multicolumn{1}{c|}{92.10} &
  \multicolumn{1}{c|}{81.33} &
  \multicolumn{1}{c|}{40.50} &
  \multicolumn{1}{c|}{20.40} &
  \multicolumn{1}{c|}{8.11} &
  \multicolumn{1}{c|}{36.56} &
  \multicolumn{1}{c|}{23.43} &
  \multicolumn{1}{c|}{30.83} &
  29.79 &
  0.31 \\
 &
  \multirow{-4}{*}{M} &
  MA-Net &
  \multicolumn{1}{c|}{49.29} &
  \multicolumn{1}{c|}{50.29} &
  \multicolumn{1}{c|}{92.94} &
  \multicolumn{1}{c|}{75.99} &
  \multicolumn{1}{c|}{63.65} &
  \multicolumn{1}{c|}{34.61} &
  \multicolumn{1}{c|}{16.41} &
  \multicolumn{1}{c|}{55.66} &
  \multicolumn{1}{c|}{37.56} &
  \multicolumn{1}{c|}{45.75} &
  53.47 &
  0.23 \\ \cmidrule{2-15}
 &
  \cellcolor[HTML]{FBB982} &
  U-Net &
  \multicolumn{1}{c|}{50.51} &
  \multicolumn{1}{c|}{52.01} &
  \multicolumn{1}{c|}{89.13} &
  \multicolumn{1}{c|}{91.55} &
  \multicolumn{1}{c|}{56.26} &
  \multicolumn{1}{c|}{42.93} &
  \multicolumn{1}{c|}{21.49} &
  \multicolumn{1}{c|}{52.84} &
  \multicolumn{1}{c|}{49.93} &
  \multicolumn{1}{c|}{58.67} &
  56.80 &
  2.22 \\
 &
  \cellcolor[HTML]{FBB982} &
  FPN &
  \multicolumn{1}{c|}{49.39} &
  \multicolumn{1}{c|}{53.22} &
  \multicolumn{1}{c|}{88.22} &
  \multicolumn{1}{c|}{87.40} &
  \multicolumn{1}{c|}{60.37} &
  \multicolumn{1}{c|}{40.01} &
  \multicolumn{1}{c|}{20.24} &
  \multicolumn{1}{c|}{51.47} &
  \multicolumn{1}{c|}{45.58} &
  \multicolumn{1}{c|}{50.81} &
  55.73 &
  1.40 \\
 &
  \cellcolor[HTML]{FBB982} &
  DeepLab &
  \multicolumn{1}{c|}{49.69} &
  \multicolumn{1}{c|}{57.35} &
  \multicolumn{1}{c|}{89.76} &
  \multicolumn{1}{c|}{86.88} &
  \multicolumn{1}{c|}{64.73} &
  \multicolumn{1}{c|}{39.20} &
  \multicolumn{1}{c|}{14.86} &
  \multicolumn{1}{c|}{52.72} &
  \multicolumn{1}{c|}{44.78} &
  \multicolumn{1}{c|}{49.30} &
  57.57 &
  2.36 \\
\multirow{-12}{*}{L + S} &
  \multirow{-4}{*}{\cellcolor[HTML]{FBB982}L} &
  MA-Net &
  \multicolumn{1}{c|}{58.73} &
  \multicolumn{1}{c|}{62.65} &
  \multicolumn{1}{c|}{90.80} &
  \multicolumn{1}{c|}{84.17} &
  \multicolumn{1}{c|}{71.61} &
  \multicolumn{1}{c|}{43.32} &
  \multicolumn{1}{c|}{14.74} &
  \multicolumn{1}{c|}{59.74} &
  \multicolumn{1}{c|}{49.18} &
  \multicolumn{1}{c|}{52.64} &
  67.89 &
  2.50 \\ \bottomrule
\end{tabular}
}
\caption{Results for BM-2 on $\mathcal{D}_{id}$ (\textit{highlighted}) and $\mathcal{D}_{ood}$ for multi-size category $\mathcal{T}_{pn}$.}
\label{tab:bm_2_multi_all_results}
\end{table*}

\begin{table*}[!ht]
\setlength\tabcolsep{4.0pt}
\setlength{\belowcaptionskip}{-10pt}
\centering
\footnotesize
\resizebox{0.96\linewidth}{!}{
\begin{tabular}{c|c|c|ccccccccccc|c}
\toprule[1.5pt]
\multirow{2}{*}{\textbf{\begin{tabular}[c]{@{}c@{}}Training\\Region\end{tabular}}} &
  \multirow{2}{*}{\textbf{\begin{tabular}[c]{@{}c@{}}Testing\\Region\end{tabular}}} &
  \multirow{2}{*}{\textbf{\begin{tabular}[c]{@{}c@{}}Training\\Model\end{tabular}}} &
  \multicolumn{11}{c|}{\textbf{Cone}} &
  \textbf{Non-cone} \\ \cmidrule{4-15} 
 &
   &
   &
  \multicolumn{1}{c|}{\textbf{\begin{tabular}[c]{@{}c@{}}Mask\\IoU\end{tabular}}} &
  \multicolumn{1}{c|}{\textbf{\begin{tabular}[c]{@{}c@{}}Pixel\\IoU\end{tabular}}} &
  \multicolumn{1}{c|}{\textbf{\begin{tabular}[c]{@{}c@{}}Pixel\\Accuracy\end{tabular}}} &
  \multicolumn{1}{c|}{\textbf{\begin{tabular}[c]{@{}c@{}}Pixel\\Precision\end{tabular}}} &
  \multicolumn{1}{c|}{\textbf{\begin{tabular}[c]{@{}c@{}}Pixel\\Recall\end{tabular}}} &
  \multicolumn{1}{c|}{\textbf{\begin{tabular}[c]{@{}c@{}}Panoptic\\Quality\end{tabular}}} &
  \multicolumn{1}{c|}{\textbf{mAP}} &
  \multicolumn{1}{c|}{\textbf{\begin{tabular}[c]{@{}c@{}}Object\\IoU\end{tabular}}} &
  \multicolumn{1}{c|}{\textbf{\begin{tabular}[c]{@{}c@{}}Object\\Accuracy\end{tabular}}} &
  \multicolumn{1}{c|}{\textbf{\begin{tabular}[c]{@{}c@{}}Object\\Precision\end{tabular}}} &
  \textbf{\begin{tabular}[c]{@{}c@{}}Object\\ Recall\end{tabular}} &
  \textbf{$A_{FP}$} \\ \midrule[1pt]
 &
  \cellcolor[HTML]{FBB982} &
  U-Net &
  \multicolumn{1}{c|}{62.62} &
  \multicolumn{1}{c|}{65.60} &
  \multicolumn{1}{c|}{96.75} &
  \multicolumn{1}{c|}{89.51} &
  \multicolumn{1}{c|}{71.99} &
  \multicolumn{1}{c|}{52.27} &
  \multicolumn{1}{c|}{31.84} &
  \multicolumn{1}{c|}{73.62} &
  \multicolumn{1}{c|}{58.50} &
  \multicolumn{1}{c|}{68.63} &
  72.12 &
  0.00 \\
 &
  \cellcolor[HTML]{FBB982} &
  FPN &
  \multicolumn{1}{c|}{57.42} &
  \multicolumn{1}{c|}{59.32} &
  \multicolumn{1}{c|}{96.33} &
  \multicolumn{1}{c|}{88.83} &
  \multicolumn{1}{c|}{65.23} &
  \multicolumn{1}{c|}{44.91} &
  \multicolumn{1}{c|}{25.66} &
  \multicolumn{1}{c|}{63.75} &
  \multicolumn{1}{c|}{51.05} &
  \multicolumn{1}{c|}{64.82} &
  57.57 &
  0.00 \\
 &
  \cellcolor[HTML]{FBB982} &
  DeepLab &
  \multicolumn{1}{c|}{57.41} &
  \multicolumn{1}{c|}{64.37} &
  \multicolumn{1}{c|}{96.61} &
  \multicolumn{1}{c|}{85.27} &
  \multicolumn{1}{c|}{72.34} &
  \multicolumn{1}{c|}{45.88} &
  \multicolumn{1}{c|}{29.38} &
  \multicolumn{1}{c|}{69.45} &
  \multicolumn{1}{c|}{49.98} &
  \multicolumn{1}{c|}{57.94} &
  68.32 &
  0.00 \\
 &
  \multirow{-4}{*}{\cellcolor[HTML]{FBB982}IP} &
  MA-Net &
  \multicolumn{1}{c|}{59.32} &
  \multicolumn{1}{c|}{64.96} &
  \multicolumn{1}{c|}{96.45} &
  \multicolumn{1}{c|}{84.37} &
  \multicolumn{1}{c|}{71.98} &
  \multicolumn{1}{c|}{47.67} &
  \multicolumn{1}{c|}{29.37} &
  \multicolumn{1}{c|}{72.73} &
  \multicolumn{1}{c|}{53.55} &
  \multicolumn{1}{c|}{62.79} &
  67.8 &
  0.00 \\ \cmidrule{2-15}
 &
   &
  U-Net &
  \multicolumn{1}{c|}{17.36} &
  \multicolumn{1}{c|}{14.36} &
  \multicolumn{1}{c|}{91.47} &
  \multicolumn{1}{c|}{62.60} &
  \multicolumn{1}{c|}{17.80} &
  \multicolumn{1}{c|}{6.71} &
  \multicolumn{1}{c|}{5.60} &
  \multicolumn{1}{c|}{17.73} &
  \multicolumn{1}{c|}{6.04} &
  \multicolumn{1}{c|}{10.54} &
  12.20 &
  9.96 \\
 &
   &
  FPN &
  \multicolumn{1}{c|}{19.00} &
  \multicolumn{1}{c|}{14.60} &
  \multicolumn{1}{c|}{90.13} &
  \multicolumn{1}{c|}{43.44} &
  \multicolumn{1}{c|}{21.25} &
  \multicolumn{1}{c|}{8.89} &
  \multicolumn{1}{c|}{4.98} &
  \multicolumn{1}{c|}{14.74} &
  \multicolumn{1}{c|}{10.06} &
  \multicolumn{1}{c|}{12.74} &
  16.07 &
  8.52 \\
 &
   &
  DeepLab &
  \multicolumn{1}{c|}{21.22} &
  \multicolumn{1}{c|}{14.94} &
  \multicolumn{1}{c|}{90.07} &
  \multicolumn{1}{c|}{37.72} &
  \multicolumn{1}{c|}{26.57} &
  \multicolumn{1}{c|}{4.63} &
  \multicolumn{1}{c|}{3.05} &
  \multicolumn{1}{c|}{17.99} &
  \multicolumn{1}{c|}{4.42} &
  \multicolumn{1}{c|}{4.90} &
  19.17 &
  7.46 \\
 &
  \multirow{-4}{*}{AP} &
  MA-Net &
  \multicolumn{1}{c|}{30.22} &
  \multicolumn{1}{c|}{24.97} &
  \multicolumn{1}{c|}{88.52} &
  \multicolumn{1}{c|}{35.44} &
  \multicolumn{1}{c|}{47.87} &
  \multicolumn{1}{c|}{6.75} &
  \multicolumn{1}{c|}{5.89} &
  \multicolumn{1}{c|}{28.39} &
  \multicolumn{1}{c|}{5.66} &
  \multicolumn{1}{c|}{6.36} &
  25.60 &
  19.08 \\ \cmidrule{2-15}
 &
   &
  U-Net &
  \multicolumn{1}{c|}{30.05} &
  \multicolumn{1}{c|}{27.92} &
  \multicolumn{1}{c|}{89.74} &
  \multicolumn{1}{c|}{72.56} &
  \multicolumn{1}{c|}{35.22} &
  \multicolumn{1}{c|}{16.87} &
  \multicolumn{1}{c|}{7.11} &
  \multicolumn{1}{c|}{28.85} &
  \multicolumn{1}{c|}{19.65} &
  \multicolumn{1}{c|}{23.59} &
  30.59 &
  1.31 \\
 &
   &
  FPN &
  \multicolumn{1}{c|}{22.14} &
  \multicolumn{1}{c|}{20.64} &
  \multicolumn{1}{c|}{89.14} &
  \multicolumn{1}{c|}{74.23} &
  \multicolumn{1}{c|}{25.80} &
  \multicolumn{1}{c|}{11.06} &
  \multicolumn{1}{c|}{5.81} &
  \multicolumn{1}{c|}{18.29} &
  \multicolumn{1}{c|}{13.49} &
  \multicolumn{1}{c|}{15.81} &
  19.34 &
  0.94 \\
 &
   &
  DeepLab &
  \multicolumn{1}{c|}{24.54} &
  \multicolumn{1}{c|}{24.03} &
  \multicolumn{1}{c|}{89.55} &
  \multicolumn{1}{c|}{68.04} &
  \multicolumn{1}{c|}{29.27} &
  \multicolumn{1}{c|}{11.81} &
  \multicolumn{1}{c|}{4.29} &
  \multicolumn{1}{c|}{25.86} &
  \multicolumn{1}{c|}{12.85} &
  \multicolumn{1}{c|}{14.26} &
  27.89 &
  0.95 \\
\multirow{-12}{*}{IP} &
  \multirow{-4}{*}{HP} &
  MA-Net &
  \multicolumn{1}{c|}{34.04} &
  \multicolumn{1}{c|}{30.93} &
  \multicolumn{1}{c|}{85.25} &
  \multicolumn{1}{c|}{50.10} &
  \multicolumn{1}{c|}{54.18} &
  \multicolumn{1}{c|}{12.36} &
  \multicolumn{1}{c|}{4.71} &
  \multicolumn{1}{c|}{33.21} &
  \multicolumn{1}{c|}{12.02} &
  \multicolumn{1}{c|}{14.36} &
  35.86 &
  4.73 \\ \midrule
 &
   &
  U-Net &
  \multicolumn{1}{c|}{7.00} &
  \multicolumn{1}{c|}{11.82} &
  \multicolumn{1}{c|}{32.22} &
  \multicolumn{1}{c|}{12.95} &
  \multicolumn{1}{c|}{86.48} &
  \multicolumn{1}{c|}{0.79} &
  \multicolumn{1}{c|}{1.81} &
  \multicolumn{1}{c|}{4.15} &
  \multicolumn{1}{c|}{0.78} &
  \multicolumn{1}{c|}{1.15} &
  1.85 &
  85.26 \\
 &
   &
  FPN &
  \multicolumn{1}{c|}{13.89} &
  \multicolumn{1}{c|}{11.20} &
  \multicolumn{1}{c|}{82.17} &
  \multicolumn{1}{c|}{17.89} &
  \multicolumn{1}{c|}{27.63} &
  \multicolumn{1}{c|}{3.04} &
  \multicolumn{1}{c|}{3.02} &
  \multicolumn{1}{c|}{11.50} &
  \multicolumn{1}{c|}{2.84} &
  \multicolumn{1}{c|}{6.36} &
  5.56 &
  10.84 \\
 &
   &
  DeepLab &
  \multicolumn{1}{c|}{8.24} &
  \multicolumn{1}{c|}{6.53} &
  \multicolumn{1}{c|}{78.24} &
  \multicolumn{1}{c|}{10.92} &
  \multicolumn{1}{c|}{17.76} &
  \multicolumn{1}{c|}{0.22} &
  \multicolumn{1}{c|}{1.11} &
  \multicolumn{1}{c|}{1.89} &
  \multicolumn{1}{c|}{0.23} &
  \multicolumn{1}{c|}{0.25} &
  1.85 &
  10.71 \\
 &
  \multirow{-4}{*}{IP} &
  MA-Net &
  \multicolumn{1}{c|}{9.89} &
  \multicolumn{1}{c|}{8.24} &
  \multicolumn{1}{c|}{62.47} &
  \multicolumn{1}{c|}{12.10} &
  \multicolumn{1}{c|}{42.58} &
  \multicolumn{1}{c|}{0.61} &
  \multicolumn{1}{c|}{2.16} &
  \multicolumn{1}{c|}{4.21} &
  \multicolumn{1}{c|}{0.59} &
  \multicolumn{1}{c|}{0.73} &
  2.16 &
  26.54 \\ \cmidrule{2-15}
 &
  \cellcolor[HTML]{FBB982} &
  U-Net &
  \multicolumn{1}{c|}{61.30} &
  \multicolumn{1}{c|}{60.29} &
  \multicolumn{1}{c|}{96.48} &
  \multicolumn{1}{c|}{70.42} &
  \multicolumn{1}{c|}{74.78} &
  \multicolumn{1}{c|}{50.23} &
  \multicolumn{1}{c|}{27.90} &
  \multicolumn{1}{c|}{68.59} &
  \multicolumn{1}{c|}{52.69} &
  \multicolumn{1}{c|}{60.54} &
  73.21 &
  3.30 \\
 &
  \cellcolor[HTML]{FBB982} &
  FPN &
  \multicolumn{1}{c|}{52.90} &
  \multicolumn{1}{c|}{58.17} &
  \multicolumn{1}{c|}{96.95} &
  \multicolumn{1}{c|}{77.54} &
  \multicolumn{1}{c|}{65.10} &
  \multicolumn{1}{c|}{48.34} &
  \multicolumn{1}{c|}{26.53} &
  \multicolumn{1}{c|}{63.48} &
  \multicolumn{1}{c|}{54.21} &
  \multicolumn{1}{c|}{60.83} &
  65.00 &
  2.33 \\
 &
  \cellcolor[HTML]{FBB982} &
  DeepLab &
  \multicolumn{1}{c|}{53.78} &
  \multicolumn{1}{c|}{59.87} &
  \multicolumn{1}{c|}{96.82} &
  \multicolumn{1}{c|}{76.65} &
  \multicolumn{1}{c|}{69.44} &
  \multicolumn{1}{c|}{48.19} &
  \multicolumn{1}{c|}{25.25} &
  \multicolumn{1}{c|}{62.99} &
  \multicolumn{1}{c|}{54.44} &
  \multicolumn{1}{c|}{60.71} &
  66.19 &
  1.34 \\
 &
  \multirow{-4}{*}{\cellcolor[HTML]{FBB982}AP} &
  MA-Net &
  \multicolumn{1}{c|}{48.63} &
  \multicolumn{1}{c|}{51.89} &
  \multicolumn{1}{c|}{96.19} &
  \multicolumn{1}{c|}{65.00} &
  \multicolumn{1}{c|}{65.06} &
  \multicolumn{1}{c|}{41.65} &
  \multicolumn{1}{c|}{18.18} &
  \multicolumn{1}{c|}{61.44} &
  \multicolumn{1}{c|}{43.57} &
  \multicolumn{1}{c|}{48.22} &
  59.94 &
  1.87 \\ \cmidrule{2-15}
 &
   &
  U-Net &
  \multicolumn{1}{c|}{26.04} &
  \multicolumn{1}{c|}{22.67} &
  \multicolumn{1}{c|}{71.75} &
  \multicolumn{1}{c|}{29.07} &
  \multicolumn{1}{c|}{66.11} &
  \multicolumn{1}{c|}{7.27} &
  \multicolumn{1}{c|}{2.69} &
  \multicolumn{1}{c|}{20.72} &
  \multicolumn{1}{c|}{7.09} &
  \multicolumn{1}{c|}{8.72} &
  22.32 &
  30.40 \\
 &
   &
  FPN &
  \multicolumn{1}{c|}{23.26} &
  \multicolumn{1}{c|}{21.69} &
  \multicolumn{1}{c|}{87.68} &
  \multicolumn{1}{c|}{55.24} &
  \multicolumn{1}{c|}{30.88} &
  \multicolumn{1}{c|}{9.68} &
  \multicolumn{1}{c|}{5.29} &
  \multicolumn{1}{c|}{19.44} &
  \multicolumn{1}{c|}{10.95} &
  \multicolumn{1}{c|}{13.80} &
  19.97 &
  4.38 \\
 &
   &
  DeepLab &
  \multicolumn{1}{c|}{19.10} &
  \multicolumn{1}{c|}{17.96} &
  \multicolumn{1}{c|}{86.75} &
  \multicolumn{1}{c|}{50.37} &
  \multicolumn{1}{c|}{26.18} &
  \multicolumn{1}{c|}{5.97} &
  \multicolumn{1}{c|}{3.90} &
  \multicolumn{1}{c|}{14.42} &
  \multicolumn{1}{c|}{6.69} &
  \multicolumn{1}{c|}{8.42} &
  15.02 &
  5.20 \\
\multirow{-12}{*}{AP} &
  \multirow{-4}{*}{HP} &
  MA-Net &
  \multicolumn{1}{c|}{13.49} &
  \multicolumn{1}{c|}{14.38} &
  \multicolumn{1}{c|}{86.87} &
  \multicolumn{1}{c|}{51.64} &
  \multicolumn{1}{c|}{19.95} &
  \multicolumn{1}{c|}{4.30} &
  \multicolumn{1}{c|}{2.02} &
  \multicolumn{1}{c|}{11.61} &
  \multicolumn{1}{c|}{4.75} &
  \multicolumn{1}{c|}{5.59} &
  12.06 &
  6.34 \\ \midrule
 &
   &
  U-Net &
  \multicolumn{1}{c|}{42.62} &
  \multicolumn{1}{c|}{44.23} &
  \multicolumn{1}{c|}{93.79} &
  \multicolumn{1}{c|}{74.00} &
  \multicolumn{1}{c|}{59.09} &
  \multicolumn{1}{c|}{24.89} &
  \multicolumn{1}{c|}{9.99} &
  \multicolumn{1}{c|}{54.69} &
  \multicolumn{1}{c|}{27.18} &
  \multicolumn{1}{c|}{45.99} &
  33.49 &
  0.00 \\
 &
   &
  FPN &
  \multicolumn{1}{c|}{40.02} &
  \multicolumn{1}{c|}{33.78} &
  \multicolumn{1}{c|}{85.67} &
  \multicolumn{1}{c|}{66.90} &
  \multicolumn{1}{c|}{44.59} &
  \multicolumn{1}{c|}{24.52} &
  \multicolumn{1}{c|}{10.73} &
  \multicolumn{1}{c|}{47.33} &
  \multicolumn{1}{c|}{25.82} &
  \multicolumn{1}{c|}{47.28} &
  30.31 &
  0.00 \\
 &
   &
  DeepLab &
  \multicolumn{1}{c|}{40.55} &
  \multicolumn{1}{c|}{38.44} &
  \multicolumn{1}{c|}{86.34} &
  \multicolumn{1}{c|}{72.64} &
  \multicolumn{1}{c|}{46.94} &
  \multicolumn{1}{c|}{27.37} &
  \multicolumn{1}{c|}{11.16} &
  \multicolumn{1}{c|}{52.15} &
  \multicolumn{1}{c|}{29.55} &
  \multicolumn{1}{c|}{50.43} &
  33.92 &
  0.00 \\
 &
  \multirow{-4}{*}{IP} &
  MA-Net &
  \multicolumn{1}{c|}{39.50} &
  \multicolumn{1}{c|}{43.19} &
  \multicolumn{1}{c|}{94.25} &
  \multicolumn{1}{c|}{73.50} &
  \multicolumn{1}{c|}{56.95} &
  \multicolumn{1}{c|}{24.60} &
  \multicolumn{1}{c|}{10.68} &
  \multicolumn{1}{c|}{48.44} &
  \multicolumn{1}{c|}{27.67} &
  \multicolumn{1}{c|}{47.22} &
  33.03 &
  0.00 \\ \cmidrule{2-15}
 &
   &
  U-Net &
  \multicolumn{1}{c|}{30.31} &
  \multicolumn{1}{c|}{33.29} &
  \multicolumn{1}{c|}{92.29} &
  \multicolumn{1}{c|}{67.55} &
  \multicolumn{1}{c|}{44.88} &
  \multicolumn{1}{c|}{17.64} &
  \multicolumn{1}{c|}{8.49} &
  \multicolumn{1}{c|}{26.81} &
  \multicolumn{1}{c|}{20.77} &
  \multicolumn{1}{c|}{25.89} &
  27.68 &
  9.29 \\
 &
   &
  FPN &
  \multicolumn{1}{c|}{26.80} &
  \multicolumn{1}{c|}{27.06} &
  \multicolumn{1}{c|}{87.92} &
  \multicolumn{1}{c|}{52.61} &
  \multicolumn{1}{c|}{44.57} &
  \multicolumn{1}{c|}{17.44} &
  \multicolumn{1}{c|}{7.43} &
  \multicolumn{1}{c|}{25.67} &
  \multicolumn{1}{c|}{23.15} &
  \multicolumn{1}{c|}{26.19} &
  28.87 &
  20.00 \\
 &
   &
  DeepLab &
  \multicolumn{1}{c|}{28.48} &
  \multicolumn{1}{c|}{27.79} &
  \multicolumn{1}{c|}{89.65} &
  \multicolumn{1}{c|}{67.22} &
  \multicolumn{1}{c|}{39.68} &
  \multicolumn{1}{c|}{18.86} &
  \multicolumn{1}{c|}{8.56} &
  \multicolumn{1}{c|}{31.81} &
  \multicolumn{1}{c|}{22.47} &
  \multicolumn{1}{c|}{27.56} &
  31.85 &
  13.29 \\
 &
  \multirow{-4}{*}{AP} &
  MA-Net &
  \multicolumn{1}{c|}{26.24} &
  \multicolumn{1}{c|}{25.18} &
  \multicolumn{1}{c|}{89.08} &
  \multicolumn{1}{c|}{62.47} &
  \multicolumn{1}{c|}{39.08} &
  \multicolumn{1}{c|}{13.45} &
  \multicolumn{1}{c|}{5.01} &
  \multicolumn{1}{c|}{20.77} &
  \multicolumn{1}{c|}{16.82} &
  \multicolumn{1}{c|}{21.90} &
  23.21 &
  9.44 \\ \cmidrule{2-15}
 &
  \cellcolor[HTML]{FBB982} &
  U-Net &
  \multicolumn{1}{c|}{48.45} &
  \multicolumn{1}{c|}{48.51} &
  \multicolumn{1}{c|}{93.41} &
  \multicolumn{1}{c|}{77.77} &
  \multicolumn{1}{c|}{58.63} &
  \multicolumn{1}{c|}{36.34} &
  \multicolumn{1}{c|}{18.13} &
  \multicolumn{1}{c|}{50.38} &
  \multicolumn{1}{c|}{42.28} &
  \multicolumn{1}{c|}{51.18} &
  49.52 &
  2.66 \\
 &
  \cellcolor[HTML]{FBB982} &
  FPN &
  \multicolumn{1}{c|}{47.44} &
  \multicolumn{1}{c|}{48.59} &
  \multicolumn{1}{c|}{90.74} &
  \multicolumn{1}{c|}{71.01} &
  \multicolumn{1}{c|}{64.21} &
  \multicolumn{1}{c|}{35.64} &
  \multicolumn{1}{c|}{16.47} &
  \multicolumn{1}{c|}{49.67} &
  \multicolumn{1}{c|}{42.14} &
  \multicolumn{1}{c|}{50.85} &
  50.47 &
  5.95 \\
 &
  \cellcolor[HTML]{FBB982} &
  DeepLab &
  \multicolumn{1}{c|}{47.14} &
  \multicolumn{1}{c|}{47.59} &
  \multicolumn{1}{c|}{91.34} &
  \multicolumn{1}{c|}{76.21} &
  \multicolumn{1}{c|}{56.85} &
  \multicolumn{1}{c|}{34.61} &
  \multicolumn{1}{c|}{15.07} &
  \multicolumn{1}{c|}{49.80} &
  \multicolumn{1}{c|}{39.01} &
  \multicolumn{1}{c|}{46.65} &
  50.54 &
  4.57 \\
\multirow{-12}{*}{HP} &
  \multirow{-4}{*}{\cellcolor[HTML]{FBB982}HP} &
  MA-Net &
  \multicolumn{1}{c|}{44.24} &
  \multicolumn{1}{c|}{45.65} &
  \multicolumn{1}{c|}{93.35} &
  \multicolumn{1}{c|}{75.80} &
  \multicolumn{1}{c|}{57.09} &
  \multicolumn{1}{c|}{31.12} &
  \multicolumn{1}{c|}{14.38} &
  \multicolumn{1}{c|}{44.66} &
  \multicolumn{1}{c|}{36.60} &
  \multicolumn{1}{c|}{45.22} &
  44.90 &
  3.06 \\ \bottomrule
\end{tabular}
}
\caption{Results for BM-1 on $\mathcal{D}_{id}$ (\textit{highlighted}) and $\mathcal{D}_{ood}$ for single-region $\mathcal{T}_{p}$.}
\label{tab:bm_1_single_negative_all_results}
\end{table*}

\begin{table*}[!ht]
\setlength\tabcolsep{4.0pt}
\setlength{\belowcaptionskip}{-10pt}
\centering
\footnotesize
\resizebox{0.96\linewidth}{!}{
\begin{tabular}{c|c|c|ccccccccccc|c}
\toprule[1.5pt]
\multirow{2}{*}{\textbf{\begin{tabular}[c]{@{}c@{}}Training\\Region\end{tabular}}} &
  \multirow{2}{*}{\textbf{\begin{tabular}[c]{@{}c@{}}Testing\\Region\end{tabular}}} &
  \multirow{2}{*}{\textbf{\begin{tabular}[c]{@{}c@{}}Training\\Model\end{tabular}}} &
  \multicolumn{11}{c|}{\textbf{Cone}} &
  \textbf{Non-cone} \\ \cmidrule{4-15} 
 &
   &
   &
  \multicolumn{1}{c|}{\textbf{\begin{tabular}[c]{@{}c@{}}Mask\\IoU\end{tabular}}} &
  \multicolumn{1}{c|}{\textbf{\begin{tabular}[c]{@{}c@{}}Pixel\\IoU\end{tabular}}} &
  \multicolumn{1}{c|}{\textbf{\begin{tabular}[c]{@{}c@{}}Pixel\\Accuracy\end{tabular}}} &
  \multicolumn{1}{c|}{\textbf{\begin{tabular}[c]{@{}c@{}}Pixel\\Precision\end{tabular}}} &
  \multicolumn{1}{c|}{\textbf{\begin{tabular}[c]{@{}c@{}}Pixel\\Recall\end{tabular}}} &
  \multicolumn{1}{c|}{\textbf{\begin{tabular}[c]{@{}c@{}}Panoptic\\Quality\end{tabular}}} &
  \multicolumn{1}{c|}{\textbf{mAP}} &
  \multicolumn{1}{c|}{\textbf{\begin{tabular}[c]{@{}c@{}}Object\\IoU\end{tabular}}} &
  \multicolumn{1}{c|}{\textbf{\begin{tabular}[c]{@{}c@{}}Object\\Accuracy\end{tabular}}} &
  \multicolumn{1}{c|}{\textbf{\begin{tabular}[c]{@{}c@{}}Object\\Precision\end{tabular}}} &
  \textbf{\begin{tabular}[c]{@{}c@{}}Object\\ Recall\end{tabular}} &
  \textbf{$A_{FP}$} \\ \midrule[1pt]
 &
  \cellcolor[HTML]{FBB982} &
  U-Net &
  \multicolumn{1}{c|}{61.50} &
  \multicolumn{1}{c|}{66.48} &
  \multicolumn{1}{c|}{97.01} &
  \multicolumn{1}{c|}{87.87} &
  \multicolumn{1}{c|}{74.32} &
  \multicolumn{1}{c|}{52.03} &
  \multicolumn{1}{c|}{32.51} &
  \multicolumn{1}{c|}{69.87} &
  \multicolumn{1}{c|}{59.15} &
  \multicolumn{1}{c|}{66.95} &
  71.23 &
  0.00 \\
 &
  \cellcolor[HTML]{FBB982} &
  FPN &
  \multicolumn{1}{c|}{56.59} &
  \multicolumn{1}{c|}{58.70} &
  \multicolumn{1}{c|}{96.07} &
  \multicolumn{1}{c|}{87.53} &
  \multicolumn{1}{c|}{64.58} &
  \multicolumn{1}{c|}{42.66} &
  \multicolumn{1}{c|}{26.58} &
  \multicolumn{1}{c|}{66.17} &
  \multicolumn{1}{c|}{47.27} &
  \multicolumn{1}{c|}{62.88} &
  54.58 &
  0.00 \\
 &
  \cellcolor[HTML]{FBB982} &
  DeepLab &
  \multicolumn{1}{c|}{60.90} &
  \multicolumn{1}{c|}{64.36} &
  \multicolumn{1}{c|}{96.64} &
  \multicolumn{1}{c|}{84.67} &
  \multicolumn{1}{c|}{70.95} &
  \multicolumn{1}{c|}{51.17} &
  \multicolumn{1}{c|}{31.75} &
  \multicolumn{1}{c|}{71.11} &
  \multicolumn{1}{c|}{55.63} &
  \multicolumn{1}{c|}{69.00} &
  66.96 &
  0.00 \\
 &
  \multirow{-4}{*}{\cellcolor[HTML]{FBB982}IP} &
  MA-Net &
  \multicolumn{1}{c|}{58.47} &
  \multicolumn{1}{c|}{63.56} &
  \multicolumn{1}{c|}{96.74} &
  \multicolumn{1}{c|}{86.47} &
  \multicolumn{1}{c|}{69.49} &
  \multicolumn{1}{c|}{49.84} &
  \multicolumn{1}{c|}{28.39} &
  \multicolumn{1}{c|}{68.69} &
  \multicolumn{1}{c|}{56.31} &
  \multicolumn{1}{c|}{67.38} &
  64.22 &
  0.00 \\\cmidrule{2-15}
 &
  \cellcolor[HTML]{FBB982} &
  U-Net &
  \multicolumn{1}{c|}{58.92} &
  \multicolumn{1}{c|}{61.85} &
  \multicolumn{1}{c|}{96.87} &
  \multicolumn{1}{c|}{81.31} &
  \multicolumn{1}{c|}{75.19} &
  \multicolumn{1}{c|}{52.98} &
  \multicolumn{1}{c|}{27.07} &
  \multicolumn{1}{c|}{69.80} &
  \multicolumn{1}{c|}{62.13} &
  \multicolumn{1}{c|}{67.54} &
  78.81 &
  1.72 \\
 &
  \cellcolor[HTML]{FBB982} &
  FPN &
  \multicolumn{1}{c|}{59.29} &
  \multicolumn{1}{c|}{60.35} &
  \multicolumn{1}{c|}{96.82} &
  \multicolumn{1}{c|}{77.09} &
  \multicolumn{1}{c|}{72.28} &
  \multicolumn{1}{c|}{51.50} &
  \multicolumn{1}{c|}{27.62} &
  \multicolumn{1}{c|}{62.39} &
  \multicolumn{1}{c|}{58.49} &
  \multicolumn{1}{c|}{65.48} &
  65.60 &
  1.74 \\
 &
  \cellcolor[HTML]{FBB982} &
  DeepLab &
  \multicolumn{1}{c|}{53.73} &
  \multicolumn{1}{c|}{54.45} &
  \multicolumn{1}{c|}{96.33} &
  \multicolumn{1}{c|}{74.81} &
  \multicolumn{1}{c|}{67.67} &
  \multicolumn{1}{c|}{45.62} &
  \multicolumn{1}{c|}{25.61} &
  \multicolumn{1}{c|}{61.44} &
  \multicolumn{1}{c|}{49.31} &
  \multicolumn{1}{c|}{56.33} &
  66.07 &
  1.67 \\
 &
  \multirow{-4}{*}{\cellcolor[HTML]{FBB982}AP} &
  MA-Net &
  \multicolumn{1}{c|}{51.51} &
  \multicolumn{1}{c|}{53.54} &
  \multicolumn{1}{c|}{96.17} &
  \multicolumn{1}{c|}{60.95} &
  \multicolumn{1}{c|}{72.16} &
  \multicolumn{1}{c|}{39.28} &
  \multicolumn{1}{c|}{17.02} &
  \multicolumn{1}{c|}{59.43} &
  \multicolumn{1}{c|}{42.01} &
  \multicolumn{1}{c|}{48.10} &
  63.69 &
  1.98 \\\cmidrule{2-15}
 &
   &
  U-Net &
  \multicolumn{1}{c|}{29.49} &
  \multicolumn{1}{c|}{26.65} &
  \multicolumn{1}{c|}{89.32} &
  \multicolumn{1}{c|}{72.37} &
  \multicolumn{1}{c|}{32.54} &
  \multicolumn{1}{c|}{16.56} &
  \multicolumn{1}{c|}{7.81} &
  \multicolumn{1}{c|}{27.71} &
  \multicolumn{1}{c|}{20.17} &
  \multicolumn{1}{c|}{23.87} &
  29.09 &
  0.96 \\
 &
   &
  FPN &
  \multicolumn{1}{c|}{33.10} &
  \multicolumn{1}{c|}{30.94} &
  \multicolumn{1}{c|}{89.74} &
  \multicolumn{1}{c|}{70.08} &
  \multicolumn{1}{c|}{39.60} &
  \multicolumn{1}{c|}{18.31} &
  \multicolumn{1}{c|}{8.77} &
  \multicolumn{1}{c|}{31.06} &
  \multicolumn{1}{c|}{20.91} &
  \multicolumn{1}{c|}{26.46} &
  30.47 &
  1.54 \\
 &
   &
  DeepLab &
  \multicolumn{1}{c|}{29.82} &
  \multicolumn{1}{c|}{29.64} &
  \multicolumn{1}{c|}{89.67} &
  \multicolumn{1}{c|}{70.29} &
  \multicolumn{1}{c|}{35.08} &
  \multicolumn{1}{c|}{16.69} &
  \multicolumn{1}{c|}{6.19} &
  \multicolumn{1}{c|}{28.98} &
  \multicolumn{1}{c|}{20.16} &
  \multicolumn{1}{c|}{23.59} &
  29.69 &
  1.60 \\
\multirow{-12}{*}{IP + AP} &
  \multirow{-4}{*}{HP} &
  MA-Net &
  \multicolumn{1}{c|}{27.39} &
  \multicolumn{1}{c|}{27.68} &
  \multicolumn{1}{c|}{89.68} &
  \multicolumn{1}{c|}{71.80} &
  \multicolumn{1}{c|}{32.66} &
  \multicolumn{1}{c|}{14.43} &
  \multicolumn{1}{c|}{6.94} &
  \multicolumn{1}{c|}{25.10} &
  \multicolumn{1}{c|}{16.61} &
  \multicolumn{1}{c|}{18.69} &
  27.65 &
  1.10 \\ \midrule
 &
  \cellcolor[HTML]{FBB982} &
  U-Net &
  \multicolumn{1}{c|}{58.85} &
  \multicolumn{1}{c|}{58.41} &
  \multicolumn{1}{c|}{96.07} &
  \multicolumn{1}{c|}{84.29} &
  \multicolumn{1}{c|}{65.90} &
  \multicolumn{1}{c|}{44.36} &
  \multicolumn{1}{c|}{26.65} &
  \multicolumn{1}{c|}{67.89} &
  \multicolumn{1}{c|}{46.35} &
  \multicolumn{1}{c|}{66.54} &
  52.91 &
  0.00 \\
 &
  \cellcolor[HTML]{FBB982} &
  FPN &
  \multicolumn{1}{c|}{57.66} &
  \multicolumn{1}{c|}{52.00} &
  \multicolumn{1}{c|}{88.61} &
  \multicolumn{1}{c|}{74.34} &
  \multicolumn{1}{c|}{63.95} &
  \multicolumn{1}{c|}{48.66} &
  \multicolumn{1}{c|}{26.92} &
  \multicolumn{1}{c|}{71.01} &
  \multicolumn{1}{c|}{53.49} &
  \multicolumn{1}{c|}{67.79} &
  61.74 &
  0.00 \\
 &
  \cellcolor[HTML]{FBB982} &
  DeepLab &
  \multicolumn{1}{c|}{43.94} &
  \multicolumn{1}{c|}{46.36} &
  \multicolumn{1}{c|}{94.62} &
  \multicolumn{1}{c|}{75.83} &
  \multicolumn{1}{c|}{49.39} &
  \multicolumn{1}{c|}{32.80} &
  \multicolumn{1}{c|}{15.98} &
  \multicolumn{1}{c|}{49.12} &
  \multicolumn{1}{c|}{37.55} &
  \multicolumn{1}{c|}{48.60} &
  41.23 &
  0.00 \\
 &
  \multirow{-4}{*}{\cellcolor[HTML]{FBB982}IP} &
  MA-Net &
  \multicolumn{1}{c|}{60.31} &
  \multicolumn{1}{c|}{66.09} &
  \multicolumn{1}{c|}{96.84} &
  \multicolumn{1}{c|}{79.02} &
  \multicolumn{1}{c|}{78.66} &
  \multicolumn{1}{c|}{49.03} &
  \multicolumn{1}{c|}{28.70} &
  \multicolumn{1}{c|}{65.66} &
  \multicolumn{1}{c|}{54.86} &
  \multicolumn{1}{c|}{67.78} &
  62.98 &
  0.00 \\\cmidrule{2-15}
 &
   &
  U-Net &
  \multicolumn{1}{c|}{28.83} &
  \multicolumn{1}{c|}{25.69} &
  \multicolumn{1}{c|}{86.98} &
  \multicolumn{1}{c|}{62.01} &
  \multicolumn{1}{c|}{44.35} &
  \multicolumn{1}{c|}{16.60} &
  \multicolumn{1}{c|}{5.66} &
  \multicolumn{1}{c|}{30.20} &
  \multicolumn{1}{c|}{20.29} &
  \multicolumn{1}{c|}{24.90} &
  29.46 &
  15.92 \\
 &
   &
  FPN &
  \multicolumn{1}{c|}{23.31} &
  \multicolumn{1}{c|}{18.45} &
  \multicolumn{1}{c|}{75.74} &
  \multicolumn{1}{c|}{40.23} &
  \multicolumn{1}{c|}{49.16} &
  \multicolumn{1}{c|}{9.05} &
  \multicolumn{1}{c|}{5.24} &
  \multicolumn{1}{c|}{18.89} &
  \multicolumn{1}{c|}{9.97} &
  \multicolumn{1}{c|}{12.50} &
  20.24 &
  28.93 \\
 &
   &
  DeepLab &
  \multicolumn{1}{c|}{26.28} &
  \multicolumn{1}{c|}{26.44} &
  \multicolumn{1}{c|}{91.28} &
  \multicolumn{1}{c|}{63.61} &
  \multicolumn{1}{c|}{36.22} &
  \multicolumn{1}{c|}{17.15} &
  \multicolumn{1}{c|}{6.33} &
  \multicolumn{1}{c|}{34.27} &
  \multicolumn{1}{c|}{18.98} &
  \multicolumn{1}{c|}{23.54} &
  35.42 &
  11.26 \\
 &
  \multirow{-4}{*}{AP} &
  MA-Net &
  \multicolumn{1}{c|}{36.88} &
  \multicolumn{1}{c|}{38.31} &
  \multicolumn{1}{c|}{88.87} &
  \multicolumn{1}{c|}{56.01} &
  \multicolumn{1}{c|}{65.51} &
  \multicolumn{1}{c|}{22.76} &
  \multicolumn{1}{c|}{6.88} &
  \multicolumn{1}{c|}{43.29} &
  \multicolumn{1}{c|}{24.17} &
  \multicolumn{1}{c|}{26.01} &
  50.30 &
  17.39 \\\cmidrule{2-15}
 &
  \cellcolor[HTML]{FBB982} &
  U-Net &
  \multicolumn{1}{c|}{49.65} &
  \multicolumn{1}{c|}{50.26} &
  \multicolumn{1}{c|}{93.21} &
  \multicolumn{1}{c|}{75.10} &
  \multicolumn{1}{c|}{62.85} &
  \multicolumn{1}{c|}{38.16} &
  \multicolumn{1}{c|}{18.04} &
  \multicolumn{1}{c|}{53.79} &
  \multicolumn{1}{c|}{43.24} &
  \multicolumn{1}{c|}{51.10} &
  53.90 &
  3.63 \\
 &
  \cellcolor[HTML]{FBB982} &
  FPN &
  \multicolumn{1}{c|}{48.47} &
  \multicolumn{1}{c|}{48.75} &
  \multicolumn{1}{c|}{90.99} &
  \multicolumn{1}{c|}{70.00} &
  \multicolumn{1}{c|}{61.13} &
  \multicolumn{1}{c|}{35.81} &
  \multicolumn{1}{c|}{16.07} &
  \multicolumn{1}{c|}{51.31} &
  \multicolumn{1}{c|}{41.23} &
  \multicolumn{1}{c|}{47.65} &
  53.13 &
  4.73 \\
 &
  \cellcolor[HTML]{FBB982} &
  DeepLab &
  \multicolumn{1}{c|}{46.02} &
  \multicolumn{1}{c|}{47.94} &
  \multicolumn{1}{c|}{93.65} &
  \multicolumn{1}{c|}{77.01} &
  \multicolumn{1}{c|}{57.39} &
  \multicolumn{1}{c|}{32.74} &
  \multicolumn{1}{c|}{13.25} &
  \multicolumn{1}{c|}{48.65} &
  \multicolumn{1}{c|}{37.10} &
  \multicolumn{1}{c|}{43.59} &
  49.82 &
  3.44 \\
\multirow{-12}{*}{HP + IP} &
  \multirow{-4}{*}{\cellcolor[HTML]{FBB982}HP} &
  MA-Net &
  \multicolumn{1}{c|}{53.55} &
  \multicolumn{1}{c|}{54.97} &
  \multicolumn{1}{c|}{93.53} &
  \multicolumn{1}{c|}{71.13} &
  \multicolumn{1}{c|}{71.95} &
  \multicolumn{1}{c|}{38.54} &
  \multicolumn{1}{c|}{15.68} &
  \multicolumn{1}{c|}{54.97} &
  \multicolumn{1}{c|}{43.01} &
  \multicolumn{1}{c|}{48.87} &
  58.62 &
  3.32 \\ \midrule
 &
   &
  U-Net &
  \multicolumn{1}{c|}{50.31} &
  \multicolumn{1}{c|}{54.92} &
  \multicolumn{1}{c|}{95.76} &
  \multicolumn{1}{c|}{79.95} &
  \multicolumn{1}{c|}{65.46} &
  \multicolumn{1}{c|}{37.95} &
  \multicolumn{1}{c|}{21.71} &
  \multicolumn{1}{c|}{60.84} &
  \multicolumn{1}{c|}{42.13} &
  \multicolumn{1}{c|}{56.79} &
  47.64 &
  0.00 \\
 &
   &
  FPN &
  \multicolumn{1}{c|}{55.44} &
  \multicolumn{1}{c|}{55.73} &
  \multicolumn{1}{c|}{96.10} &
  \multicolumn{1}{c|}{83.86} &
  \multicolumn{1}{c|}{63.87} &
  \multicolumn{1}{c|}{41.62} &
  \multicolumn{1}{c|}{21.88} &
  \multicolumn{1}{c|}{63.45} &
  \multicolumn{1}{c|}{46.56} &
  \multicolumn{1}{c|}{66.36} &
  50.1 &
  0.00 \\
 &
   &
  DeepLab &
  \multicolumn{1}{c|}{42.84} &
  \multicolumn{1}{c|}{39.54} &
  \multicolumn{1}{c|}{93.91} &
  \multicolumn{1}{c|}{83.26} &
  \multicolumn{1}{c|}{48.83} &
  \multicolumn{1}{c|}{29.48} &
  \multicolumn{1}{c|}{14.19} &
  \multicolumn{1}{c|}{54.75} &
  \multicolumn{1}{c|}{32.28} &
  \multicolumn{1}{c|}{50.93} &
  36.79 &
  0.00 \\
 &
  \multirow{-4}{*}{IP} &
  MA-Net &
  \multicolumn{1}{c|}{50.74} &
  \multicolumn{1}{c|}{49.80} &
  \multicolumn{1}{c|}{95.24} &
  \multicolumn{1}{c|}{79.57} &
  \multicolumn{1}{c|}{60.62} &
  \multicolumn{1}{c|}{34.37} &
  \multicolumn{1}{c|}{16.71} &
  \multicolumn{1}{c|}{55.95} &
  \multicolumn{1}{c|}{40.93} &
  \multicolumn{1}{c|}{56.61} &
  44.42 &
  0.00 \\\cmidrule{2-15}
 &
  \cellcolor[HTML]{FBB982} &
  U-Net &
  \multicolumn{1}{c|}{35.33} &
  \multicolumn{1}{c|}{40.46} &
  \multicolumn{1}{c|}{94.43} &
  \multicolumn{1}{c|}{76.45} &
  \multicolumn{1}{c|}{49.29} &
  \multicolumn{1}{c|}{31.37} &
  \multicolumn{1}{c|}{17.61} &
  \multicolumn{1}{c|}{43.56} &
  \multicolumn{1}{c|}{35.72} &
  \multicolumn{1}{c|}{39.11} &
  42.68 &
  10.03 \\
 &
  \cellcolor[HTML]{FBB982} &
  FPN &
  \multicolumn{1}{c|}{45.35} &
  \multicolumn{1}{c|}{46.30} &
  \multicolumn{1}{c|}{94.31} &
  \multicolumn{1}{c|}{79.74} &
  \multicolumn{1}{c|}{54.31} &
  \multicolumn{1}{c|}{35.64} &
  \multicolumn{1}{c|}{21.74} &
  \multicolumn{1}{c|}{50.92} &
  \multicolumn{1}{c|}{39.70} &
  \multicolumn{1}{c|}{49.70} &
  49.82 &
  8.70 \\
 &
  \cellcolor[HTML]{FBB982} &
  DeepLab &
  \multicolumn{1}{c|}{48.83} &
  \multicolumn{1}{c|}{48.55} &
  \multicolumn{1}{c|}{93.57} &
  \multicolumn{1}{c|}{72.22} &
  \multicolumn{1}{c|}{64.70} &
  \multicolumn{1}{c|}{44.66} &
  \multicolumn{1}{c|}{18.93} &
  \multicolumn{1}{c|}{60.77} &
  \multicolumn{1}{c|}{49.14} &
  \multicolumn{1}{c|}{57.38} &
  61.01 &
  8.42 \\
 &
  \multirow{-4}{*}{\cellcolor[HTML]{FBB982}AP} &
  MA-Net &
  \multicolumn{1}{c|}{50.52} &
  \multicolumn{1}{c|}{54.94} &
  \multicolumn{1}{c|}{94.55} &
  \multicolumn{1}{c|}{71.33} &
  \multicolumn{1}{c|}{72.74} &
  \multicolumn{1}{c|}{45.68} &
  \multicolumn{1}{c|}{20.74} &
  \multicolumn{1}{c|}{61.35} &
  \multicolumn{1}{c|}{50.54} &
  \multicolumn{1}{c|}{54.11} &
  65.77 &
  7.58 \\\cmidrule{2-15}
 &
  \cellcolor[HTML]{FBB982} &
  U-Net &
  \multicolumn{1}{c|}{48.60} &
  \multicolumn{1}{c|}{49.96} &
  \multicolumn{1}{c|}{93.50} &
  \multicolumn{1}{c|}{72.92} &
  \multicolumn{1}{c|}{62.18} &
  \multicolumn{1}{c|}{36.59} &
  \multicolumn{1}{c|}{18.55} &
  \multicolumn{1}{c|}{52.11} &
  \multicolumn{1}{c|}{41.96} &
  \multicolumn{1}{c|}{48.33} &
  52.05 &
  3.20 \\
 &
  \cellcolor[HTML]{FBB982} &
  FPN &
  \multicolumn{1}{c|}{49.84} &
  \multicolumn{1}{c|}{49.54} &
  \multicolumn{1}{c|}{93.68} &
  \multicolumn{1}{c|}{79.43} &
  \multicolumn{1}{c|}{58.12} &
  \multicolumn{1}{c|}{38.33} &
  \multicolumn{1}{c|}{22.65} &
  \multicolumn{1}{c|}{52.06} &
  \multicolumn{1}{c|}{45.35} &
  \multicolumn{1}{c|}{53.36} &
  54.14 &
  1.58 \\
 &
  \cellcolor[HTML]{FBB982} &
  DeepLab &
  \multicolumn{1}{c|}{50.68} &
  \multicolumn{1}{c|}{51.57} &
  \multicolumn{1}{c|}{93.77} &
  \multicolumn{1}{c|}{74.41} &
  \multicolumn{1}{c|}{64.10} &
  \multicolumn{1}{c|}{37.71} &
  \multicolumn{1}{c|}{19.74} &
  \multicolumn{1}{c|}{53.36} &
  \multicolumn{1}{c|}{42.76} &
  \multicolumn{1}{c|}{49.97} &
  54.71 &
  2.23 \\
\multirow{-12}{*}{AP + HP} &
  \multirow{-4}{*}{\cellcolor[HTML]{FBB982}HP} &
  MA-Net &
  \multicolumn{1}{c|}{50.92} &
  \multicolumn{1}{c|}{51.80} &
  \multicolumn{1}{c|}{93.86} &
  \multicolumn{1}{c|}{72.27} &
  \multicolumn{1}{c|}{64.46} &
  \multicolumn{1}{c|}{38.08} &
  \multicolumn{1}{c|}{17.77} &
  \multicolumn{1}{c|}{53.54} &
  \multicolumn{1}{c|}{42.93} &
  \multicolumn{1}{c|}{49.99} &
  55.24 &
  2.36 \\ \midrule
 &
  \cellcolor[HTML]{FBB982} &
  U-Net &
  \multicolumn{1}{c|}{57.63} &
  \multicolumn{1}{c|}{55.90} &
  \multicolumn{1}{c|}{96.04} &
  \multicolumn{1}{c|}{88.12} &
  \multicolumn{1}{c|}{60.07} &
  \multicolumn{1}{c|}{43.47} &
  \multicolumn{1}{c|}{26.69} &
  \multicolumn{1}{c|}{67.67} &
  \multicolumn{1}{c|}{46.89} &
  \multicolumn{1}{c|}{64.75} &
  52.66 &
  0.00 \\
 &
  \cellcolor[HTML]{FBB982} &
  FPN &
  \multicolumn{1}{c|}{57.13} &
  \multicolumn{1}{c|}{59.66} &
  \multicolumn{1}{c|}{96.01} &
  \multicolumn{1}{c|}{83.72} &
  \multicolumn{1}{c|}{67.59} &
  \multicolumn{1}{c|}{43.78} &
  \multicolumn{1}{c|}{24.63} &
  \multicolumn{1}{c|}{66.97} &
  \multicolumn{1}{c|}{48.26} &
  \multicolumn{1}{c|}{65.74} &
  55.51 &
  0.00 \\
 &
  \cellcolor[HTML]{FBB982} &
  DeepLab &
  \multicolumn{1}{c|}{54.36} &
  \multicolumn{1}{c|}{53.52} &
  \multicolumn{1}{c|}{95.65} &
  \multicolumn{1}{c|}{86.86} &
  \multicolumn{1}{c|}{58.72} &
  \multicolumn{1}{c|}{37.29} &
  \multicolumn{1}{c|}{18.87} &
  \multicolumn{1}{c|}{58.08} &
  \multicolumn{1}{c|}{40.87} &
  \multicolumn{1}{c|}{58.44} &
  45.41 &
  0.00 \\
 &
  \multirow{-4}{*}{\cellcolor[HTML]{FBB982}IP} &
  MA-Net &
  \multicolumn{1}{c|}{55.02} &
  \multicolumn{1}{c|}{55.88} &
  \multicolumn{1}{c|}{95.96} &
  \multicolumn{1}{c|}{86.90} &
  \multicolumn{1}{c|}{61.55} &
  \multicolumn{1}{c|}{42.32} &
  \multicolumn{1}{c|}{22.09} &
  \multicolumn{1}{c|}{64.19} &
  \multicolumn{1}{c|}{45.25} &
  \multicolumn{1}{c|}{60.29} &
  53.33 &
  0.00 \\\cmidrule{2-15}
 &
  \cellcolor[HTML]{FBB982} &
  U-Net &
  \multicolumn{1}{c|}{54.48} &
  \multicolumn{1}{c|}{50.87} &
  \multicolumn{1}{c|}{94.01} &
  \multicolumn{1}{c|}{73.59} &
  \multicolumn{1}{c|}{67.10} &
  \multicolumn{1}{c|}{44.62} &
  \multicolumn{1}{c|}{27.66} &
  \multicolumn{1}{c|}{54.02} &
  \multicolumn{1}{c|}{53.69} &
  \multicolumn{1}{c|}{59.17} &
  61.61 &
  9.45 \\
 &
  \cellcolor[HTML]{FBB982} &
  FPN &
  \multicolumn{1}{c|}{47.84} &
  \multicolumn{1}{c|}{48.94} &
  \multicolumn{1}{c|}{93.87} &
  \multicolumn{1}{c|}{74.32} &
  \multicolumn{1}{c|}{60.52} &
  \multicolumn{1}{c|}{39.56} &
  \multicolumn{1}{c|}{21.61} &
  \multicolumn{1}{c|}{52.21} &
  \multicolumn{1}{c|}{44.11} &
  \multicolumn{1}{c|}{50.89} &
  54.46 &
  10.58 \\
 &
  \cellcolor[HTML]{FBB982} &
  DeepLab &
  \multicolumn{1}{c|}{48.26} &
  \multicolumn{1}{c|}{49.39} &
  \multicolumn{1}{c|}{94.08} &
  \multicolumn{1}{c|}{71.10} &
  \multicolumn{1}{c|}{64.71} &
  \multicolumn{1}{c|}{40.16} &
  \multicolumn{1}{c|}{22.92} &
  \multicolumn{1}{c|}{53.11} &
  \multicolumn{1}{c|}{46.67} &
  \multicolumn{1}{c|}{50.00} &
  60.12 &
  9.73 \\
 &
  \multirow{-4}{*}{\cellcolor[HTML]{FBB982}AP} &
  MA-Net &
  \multicolumn{1}{c|}{53.83} &
  \multicolumn{1}{c|}{57.92} &
  \multicolumn{1}{c|}{96.43} &
  \multicolumn{1}{c|}{76.17} &
  \multicolumn{1}{c|}{69.55} &
  \multicolumn{1}{c|}{47.61} &
  \multicolumn{1}{c|}{23.61} &
  \multicolumn{1}{c|}{66.01} &
  \multicolumn{1}{c|}{49.88} &
  \multicolumn{1}{c|}{56.55} &
  65.18 &
  1.12 \\\cmidrule{2-15}
 &
  \cellcolor[HTML]{FBB982} &
  U-Net &
  \multicolumn{1}{c|}{49.89} &
  \multicolumn{1}{c|}{49.55} &
  \multicolumn{1}{c|}{93.70} &
  \multicolumn{1}{c|}{77.38} &
  \multicolumn{1}{c|}{60.54} &
  \multicolumn{1}{c|}{36.62} &
  \multicolumn{1}{c|}{19.47} &
  \multicolumn{1}{c|}{52.64} &
  \multicolumn{1}{c|}{42.47} &
  \multicolumn{1}{c|}{50.65} &
  53.07 &
  2.14 \\
 &
  \cellcolor[HTML]{FBB982} &
  FPN &
  \multicolumn{1}{c|}{50.17} &
  \multicolumn{1}{c|}{51.84} &
  \multicolumn{1}{c|}{93.60} &
  \multicolumn{1}{c|}{70.81} &
  \multicolumn{1}{c|}{67.14} &
  \multicolumn{1}{c|}{35.99} &
  \multicolumn{1}{c|}{18.53} &
  \multicolumn{1}{c|}{53.19} &
  \multicolumn{1}{c|}{39.18} &
  \multicolumn{1}{c|}{46.19} &
  53.39 &
  3.20 \\
 &
  \cellcolor[HTML]{FBB982} &
  DeepLab &
  \multicolumn{1}{c|}{45.72} &
  \multicolumn{1}{c|}{47.13} &
  \multicolumn{1}{c|}{92.72} &
  \multicolumn{1}{c|}{76.59} &
  \multicolumn{1}{c|}{56.76} &
  \multicolumn{1}{c|}{33.12} &
  \multicolumn{1}{c|}{15.92} &
  \multicolumn{1}{c|}{48.62} &
  \multicolumn{1}{c|}{38.06} &
  \multicolumn{1}{c|}{43.59} &
  50.52 &
  3.96 \\
\multirow{-12}{*}{IP + AP + HP} &
  \multirow{-4}{*}{\cellcolor[HTML]{FBB982}HP} &
  MA-Net &
  \multicolumn{1}{c|}{52.37} &
  \multicolumn{1}{c|}{53.25} &
  \multicolumn{1}{c|}{94.11} &
  \multicolumn{1}{c|}{76.63} &
  \multicolumn{1}{c|}{65.24} &
  \multicolumn{1}{c|}{40.47} &
  \multicolumn{1}{c|}{18.87} &
  \multicolumn{1}{c|}{55.09} &
  \multicolumn{1}{c|}{46.27} &
  \multicolumn{1}{c|}{53.57} &
  56.69 &
  3.44 \\ \bottomrule
\end{tabular}
}
\caption{Results for BM-1 on $\mathcal{D}_{id}$ (\textit{highlighted}) and $\mathcal{D}_{ood}$ for multi-region $\mathcal{T}_{p}$.}
\label{tab:bm_1_multi_negative_all_results}
\end{table*}

\begin{table*}[!ht]
\setlength\tabcolsep{4.0pt}
\setlength{\belowcaptionskip}{-10pt}
\centering
\footnotesize
\resizebox{0.96\linewidth}{!}{
\begin{tabular}{c|c|c|ccccccccccc|c}
\toprule[1.5pt]
\multirow{2}{*}{\textbf{\begin{tabular}[c]{@{}c@{}}Training\\Category\end{tabular}}} &
  \multirow{2}{*}{\textbf{\begin{tabular}[c]{@{}c@{}}Testing\\Category\end{tabular}}} &
  \multirow{2}{*}{\textbf{\begin{tabular}[c]{@{}c@{}}Training\\Model\end{tabular}}} &
  \multicolumn{11}{c|}{\textbf{Cone}} &
  \textbf{Non-cone} \\ \cmidrule{4-15} 
 &
   &
   &
  \multicolumn{1}{c|}{\textbf{\begin{tabular}[c]{@{}c@{}}Mask\\IoU\end{tabular}}} &
  \multicolumn{1}{c|}{\textbf{\begin{tabular}[c]{@{}c@{}}Pixel\\IoU\end{tabular}}} &
  \multicolumn{1}{c|}{\textbf{\begin{tabular}[c]{@{}c@{}}Pixel\\Accuracy\end{tabular}}} &
  \multicolumn{1}{c|}{\textbf{\begin{tabular}[c]{@{}c@{}}Pixel\\Precision\end{tabular}}} &
  \multicolumn{1}{c|}{\textbf{\begin{tabular}[c]{@{}c@{}}Pixel\\Recall\end{tabular}}} &
  \multicolumn{1}{c|}{\textbf{\begin{tabular}[c]{@{}c@{}}Panoptic\\Quality\end{tabular}}} &
  \multicolumn{1}{c|}{\textbf{mAP}} &
  \multicolumn{1}{c|}{\textbf{\begin{tabular}[c]{@{}c@{}}Object\\IoU\end{tabular}}} &
  \multicolumn{1}{c|}{\textbf{\begin{tabular}[c]{@{}c@{}}Object\\Accuracy\end{tabular}}} &
  \multicolumn{1}{c|}{\textbf{\begin{tabular}[c]{@{}c@{}}Object\\Precision\end{tabular}}} &
  \textbf{\begin{tabular}[c]{@{}c@{}}Object\\ Recall\end{tabular}} &
  \textbf{$A_{FP}$} \\ \midrule[1pt]
 &
  \cellcolor[HTML]{FBB982} &
  U-Net &
  \multicolumn{1}{c|}{26.56} &
  \multicolumn{1}{c|}{24.68} &
  \multicolumn{1}{c|}{96.88} &
  \multicolumn{1}{c|}{70.36} &
  \multicolumn{1}{c|}{29.01} &
  \multicolumn{1}{c|}{17.04} &
  \multicolumn{1}{c|}{8.59} &
  \multicolumn{1}{c|}{24.99} &
  \multicolumn{1}{c|}{21.74} &
  \multicolumn{1}{c|}{28.00} &
  23.68 &
  0.30 \\
 &
  \cellcolor[HTML]{FBB982} &
  FPN &
  \multicolumn{1}{c|}{30.00} &
  \multicolumn{1}{c|}{29.05} &
  \multicolumn{1}{c|}{97.08} &
  \multicolumn{1}{c|}{62.67} &
  \multicolumn{1}{c|}{39.30} &
  \multicolumn{1}{c|}{19.10} &
  \multicolumn{1}{c|}{12.20} &
  \multicolumn{1}{c|}{34.10} &
  \multicolumn{1}{c|}{21.76} &
  \multicolumn{1}{c|}{26.54} &
  32.39 &
  0.48 \\
 &
  \cellcolor[HTML]{FBB982} &
  DeepLab &
  \multicolumn{1}{c|}{29.54} &
  \multicolumn{1}{c|}{30.56} &
  \multicolumn{1}{c|}{97.21} &
  \multicolumn{1}{c|}{66.39} &
  \multicolumn{1}{c|}{37.39} &
  \multicolumn{1}{c|}{21.15} &
  \multicolumn{1}{c|}{9.14} &
  \multicolumn{1}{c|}{34.46} &
  \multicolumn{1}{c|}{24.95} &
  \multicolumn{1}{c|}{29.57} &
  32.72 &
  0.26 \\
 &
  \multirow{-4}{*}{\cellcolor[HTML]{FBB982}S} &
  MA-Net &
  \multicolumn{1}{c|}{29.93} &
  \multicolumn{1}{c|}{30.58} &
  \multicolumn{1}{c|}{97.24} &
  \multicolumn{1}{c|}{66.91} &
  \multicolumn{1}{c|}{38.51} &
  \multicolumn{1}{c|}{20.39} &
  \multicolumn{1}{c|}{10.33} &
  \multicolumn{1}{c|}{32.80} &
  \multicolumn{1}{c|}{23.82} &
  \multicolumn{1}{c|}{28.24} &
  30.75 &
  0.32 \\ \cmidrule{2-15}
 &
   &
  U-Net &
  \multicolumn{1}{c|}{36.16} &
  \multicolumn{1}{c|}{34.46} &
  \multicolumn{1}{c|}{95.60} &
  \multicolumn{1}{c|}{86.72} &
  \multicolumn{1}{c|}{37.74} &
  \multicolumn{1}{c|}{21.62} &
  \multicolumn{1}{c|}{11.24} &
  \multicolumn{1}{c|}{39.11} &
  \multicolumn{1}{c|}{26.27} &
  \multicolumn{1}{c|}{33.24} &
  35.99 &
  0.20 \\
 &
   &
  FPN &
  \multicolumn{1}{c|}{53.46} &
  \multicolumn{1}{c|}{48.02} &
  \multicolumn{1}{c|}{96.43} &
  \multicolumn{1}{c|}{86.15} &
  \multicolumn{1}{c|}{52.11} &
  \multicolumn{1}{c|}{30.98} &
  \multicolumn{1}{c|}{13.58} &
  \multicolumn{1}{c|}{49.78} &
  \multicolumn{1}{c|}{34.51} &
  \multicolumn{1}{c|}{47.06} &
  50.56 &
  0.39 \\
 &
   &
  DeepLab &
  \multicolumn{1}{c|}{40.88} &
  \multicolumn{1}{c|}{41.49} &
  \multicolumn{1}{c|}{96.44} &
  \multicolumn{1}{c|}{76.02} &
  \multicolumn{1}{c|}{44.54} &
  \multicolumn{1}{c|}{28.66} &
  \multicolumn{1}{c|}{14.95} &
  \multicolumn{1}{c|}{44.01} &
  \multicolumn{1}{c|}{31.75} &
  \multicolumn{1}{c|}{35.29} &
  47.48 &
  0.17 \\
 &
  \multirow{-4}{*}{M} &
  MA-Net &
  \multicolumn{1}{c|}{41.93} &
  \multicolumn{1}{c|}{39.83} &
  \multicolumn{1}{c|}{96.04} &
  \multicolumn{1}{c|}{82.33} &
  \multicolumn{1}{c|}{43.67} &
  \multicolumn{1}{c|}{28.77} &
  \multicolumn{1}{c|}{10.26} &
  \multicolumn{1}{c|}{42.26} &
  \multicolumn{1}{c|}{33.04} &
  \multicolumn{1}{c|}{38.45} &
  42.86 &
  0.33 \\ \cmidrule{2-15}
 &
   &
  U-Net &
  \multicolumn{1}{c|}{10.63} &
  \multicolumn{1}{c|}{10.91} &
  \multicolumn{1}{c|}{81.04} &
  \multicolumn{1}{c|}{70.44} &
  \multicolumn{1}{c|}{11.15} &
  \multicolumn{1}{c|}{3.98} &
  \multicolumn{1}{c|}{5.08} &
  \multicolumn{1}{c|}{5.42} &
  \multicolumn{1}{c|}{6.06} &
  \multicolumn{1}{c|}{6.82} &
  6.82 &
  0.19 \\
 &
   &
  FPN &
  \multicolumn{1}{c|}{22.80} &
  \multicolumn{1}{c|}{22.28} &
  \multicolumn{1}{c|}{82.91} &
  \multicolumn{1}{c|}{80.77} &
  \multicolumn{1}{c|}{23.51} &
  \multicolumn{1}{c|}{13.28} &
  \multicolumn{1}{c|}{7.15} &
  \multicolumn{1}{c|}{17.47} &
  \multicolumn{1}{c|}{18.94} &
  \multicolumn{1}{c|}{21.97} &
  22.73 &
  0.46 \\
 &
   &
  DeepLab &
  \multicolumn{1}{c|}{18.65} &
  \multicolumn{1}{c|}{19.25} &
  \multicolumn{1}{c|}{81.99} &
  \multicolumn{1}{c|}{80.70} &
  \multicolumn{1}{c|}{19.68} &
  \multicolumn{1}{c|}{13.79} &
  \multicolumn{1}{c|}{7.91} &
  \multicolumn{1}{c|}{18.38} &
  \multicolumn{1}{c|}{17.95} &
  \multicolumn{1}{c|}{18.94} &
  21.97 &
  0.28 \\
\multirow{-12}{*}{S} &
  \multirow{-4}{*}{L} &
  MA-Net &
  \multicolumn{1}{c|}{18.07} &
  \multicolumn{1}{c|}{18.89} &
  \multicolumn{1}{c|}{82.06} &
  \multicolumn{1}{c|}{93.43} &
  \multicolumn{1}{c|}{19.15} &
  \multicolumn{1}{c|}{13.75} &
  \multicolumn{1}{c|}{9.17} &
  \multicolumn{1}{c|}{20.62} &
  \multicolumn{1}{c|}{17.80} &
  \multicolumn{1}{c|}{20.83} &
  27.27 &
  0.45 \\ \midrule
 &
   &
  U-Net &
  \multicolumn{1}{c|}{21.64} &
  \multicolumn{1}{c|}{22.35} &
  \multicolumn{1}{c|}{95.91} &
  \multicolumn{1}{c|}{49.27} &
  \multicolumn{1}{c|}{32.77} &
  \multicolumn{1}{c|}{7.66} &
  \multicolumn{1}{c|}{11.26} &
  \multicolumn{1}{c|}{17.53} &
  \multicolumn{1}{c|}{7.89} &
  \multicolumn{1}{c|}{14.21} &
  12.02 &
  1.70 \\
 &
   &
  FPN &
  \multicolumn{1}{c|}{30.18} &
  \multicolumn{1}{c|}{28.35} &
  \multicolumn{1}{c|}{96.24} &
  \multicolumn{1}{c|}{56.61} &
  \multicolumn{1}{c|}{41.94} &
  \multicolumn{1}{c|}{18.65} &
  \multicolumn{1}{c|}{10.20} &
  \multicolumn{1}{c|}{30.62} &
  \multicolumn{1}{c|}{21.93} &
  \multicolumn{1}{c|}{30.98} &
  26.73 &
  1.17 \\
 &
   &
  DeepLab &
  \multicolumn{1}{c|}{27.76} &
  \multicolumn{1}{c|}{31.39} &
  \multicolumn{1}{c|}{96.54} &
  \multicolumn{1}{c|}{61.04} &
  \multicolumn{1}{c|}{45.49} &
  \multicolumn{1}{c|}{20.27} &
  \multicolumn{1}{c|}{11.12} &
  \multicolumn{1}{c|}{33.31} &
  \multicolumn{1}{c|}{22.67} &
  \multicolumn{1}{c|}{27.88} &
  27.21 &
  1.17 \\
 &
  \multirow{-4}{*}{S} &
  MA-Net &
  \multicolumn{1}{c|}{32.44} &
  \multicolumn{1}{c|}{30.46} &
  \multicolumn{1}{c|}{95.61} &
  \multicolumn{1}{c|}{50.99} &
  \multicolumn{1}{c|}{47.27} &
  \multicolumn{1}{c|}{14.36} &
  \multicolumn{1}{c|}{11.53} &
  \multicolumn{1}{c|}{24.42} &
  \multicolumn{1}{c|}{14.74} &
  \multicolumn{1}{c|}{23.08} &
  22.88 &
  1.73 \\ \cmidrule{2-15}
 &
  \cellcolor[HTML]{FBB982} &
  U-Net &
  \multicolumn{1}{c|}{47.18} &
  \multicolumn{1}{c|}{47.69} &
  \multicolumn{1}{c|}{93.93} &
  \multicolumn{1}{c|}{67.30} &
  \multicolumn{1}{c|}{63.57} &
  \multicolumn{1}{c|}{32.28} &
  \multicolumn{1}{c|}{14.04} &
  \multicolumn{1}{c|}{52.66} &
  \multicolumn{1}{c|}{37.24} &
  \multicolumn{1}{c|}{45.25} &
  53.31 &
  1.19 \\
 &
  \cellcolor[HTML]{FBB982} &
  FPN &
  \multicolumn{1}{c|}{51.33} &
  \multicolumn{1}{c|}{51.73} &
  \multicolumn{1}{c|}{94.97} &
  \multicolumn{1}{c|}{77.50} &
  \multicolumn{1}{c|}{61.96} &
  \multicolumn{1}{c|}{40.38} &
  \multicolumn{1}{c|}{20.35} &
  \multicolumn{1}{c|}{55.67} &
  \multicolumn{1}{c|}{48.29} &
  \multicolumn{1}{c|}{56.54} &
  57.76 &
  0.76 \\
 &
  \cellcolor[HTML]{FBB982} &
  DeepLab &
  \multicolumn{1}{c|}{51.13} &
  \multicolumn{1}{c|}{52.67} &
  \multicolumn{1}{c|}{95.14} &
  \multicolumn{1}{c|}{77.23} &
  \multicolumn{1}{c|}{63.83} &
  \multicolumn{1}{c|}{38.43} &
  \multicolumn{1}{c|}{17.29} &
  \multicolumn{1}{c|}{58.28} &
  \multicolumn{1}{c|}{43.19} &
  \multicolumn{1}{c|}{49.70} &
  60.41 &
  0.76 \\
 &
  \multirow{-4}{*}{\cellcolor[HTML]{FBB982}M} &
  MA-Net &
  \multicolumn{1}{c|}{51.97} &
  \multicolumn{1}{c|}{52.89} &
  \multicolumn{1}{c|}{94.91} &
  \multicolumn{1}{c|}{77.20} &
  \multicolumn{1}{c|}{64.18} &
  \multicolumn{1}{c|}{40.88} &
  \multicolumn{1}{c|}{16.73} &
  \multicolumn{1}{c|}{58.35} &
  \multicolumn{1}{c|}{48.86} &
  \multicolumn{1}{c|}{58.18} &
  59.27 &
  1.48 \\ \cmidrule{2-15}
 &
   &
  U-Net &
  \multicolumn{1}{c|}{41.63} &
  \multicolumn{1}{c|}{43.67} &
  \multicolumn{1}{c|}{86.65} &
  \multicolumn{1}{c|}{83.07} &
  \multicolumn{1}{c|}{49.41} &
  \multicolumn{1}{c|}{26.60} &
  \multicolumn{1}{c|}{10.38} &
  \multicolumn{1}{c|}{42.56} &
  \multicolumn{1}{c|}{32.99} &
  \multicolumn{1}{c|}{38.10} &
  49.24 &
  1.30 \\
 &
   &
  FPN &
  \multicolumn{1}{c|}{38.11} &
  \multicolumn{1}{c|}{37.96} &
  \multicolumn{1}{c|}{86.00} &
  \multicolumn{1}{c|}{83.91} &
  \multicolumn{1}{c|}{40.91} &
  \multicolumn{1}{c|}{29.19} &
  \multicolumn{1}{c|}{15.58} &
  \multicolumn{1}{c|}{38.53} &
  \multicolumn{1}{c|}{37.01} &
  \multicolumn{1}{c|}{43.33} &
  42.42 &
  0.90 \\
 &
   &
  DeepLab &
  \multicolumn{1}{c|}{45.58} &
  \multicolumn{1}{c|}{49.24} &
  \multicolumn{1}{c|}{87.55} &
  \multicolumn{1}{c|}{84.77} &
  \multicolumn{1}{c|}{53.45} &
  \multicolumn{1}{c|}{36.46} &
  \multicolumn{1}{c|}{19.61} &
  \multicolumn{1}{c|}{58.41} &
  \multicolumn{1}{c|}{40.06} &
  \multicolumn{1}{c|}{44.47} &
  54.92 &
  0.97 \\
\multirow{-12}{*}{M} &
  \multirow{-4}{*}{L} &
  MA-Net &
  \multicolumn{1}{c|}{53.00} &
  \multicolumn{1}{c|}{57.38} &
  \multicolumn{1}{c|}{89.35} &
  \multicolumn{1}{c|}{90.03} &
  \multicolumn{1}{c|}{61.66} &
  \multicolumn{1}{c|}{44.45} &
  \multicolumn{1}{c|}{27.76} &
  \multicolumn{1}{c|}{57.72} &
  \multicolumn{1}{c|}{51.94} &
  \multicolumn{1}{c|}{60.38} &
  60.61 &
  1.59 \\ \midrule
 &
   &
  U-Net &
  \multicolumn{1}{c|}{15.59} &
  \multicolumn{1}{c|}{11.93} &
  \multicolumn{1}{c|}{91.97} &
  \multicolumn{1}{c|}{73.41} &
  \multicolumn{1}{c|}{21.34} &
  \multicolumn{1}{c|}{5.73} &
  \multicolumn{1}{c|}{3.05} &
  \multicolumn{1}{c|}{12.70} &
  \multicolumn{1}{c|}{6.76} &
  \multicolumn{1}{c|}{11.54} &
  9.58 &
  3.78 \\
 &
   &
  FPN &
  \multicolumn{1}{c|}{23.91} &
  \multicolumn{1}{c|}{24.14} &
  \multicolumn{1}{c|}{91.23} &
  \multicolumn{1}{c|}{44.07} &
  \multicolumn{1}{c|}{43.74} &
  \multicolumn{1}{c|}{12.12} &
  \multicolumn{1}{c|}{6.57} &
  \multicolumn{1}{c|}{22.01} &
  \multicolumn{1}{c|}{14.06} &
  \multicolumn{1}{c|}{21.70} &
  17.79 &
  5.11 \\
 &
   &
  DeepLab &
  \multicolumn{1}{c|}{23.18} &
  \multicolumn{1}{c|}{23.40} &
  \multicolumn{1}{c|}{92.26} &
  \multicolumn{1}{c|}{56.19} &
  \multicolumn{1}{c|}{42.89} &
  \multicolumn{1}{c|}{9.65} &
  \multicolumn{1}{c|}{6.59} &
  \multicolumn{1}{c|}{18.24} &
  \multicolumn{1}{c|}{10.71} &
  \multicolumn{1}{c|}{17.31} &
  12.02 &
  5.17 \\
 &
  \multirow{-4}{*}{S} &
  MA-Net &
  \multicolumn{1}{c|}{23.72} &
  \multicolumn{1}{c|}{24.60} &
  \multicolumn{1}{c|}{91.88} &
  \multicolumn{1}{c|}{41.44} &
  \multicolumn{1}{c|}{44.99} &
  \multicolumn{1}{c|}{13.39} &
  \multicolumn{1}{c|}{8.02} &
  \multicolumn{1}{c|}{30.55} &
  \multicolumn{1}{c|}{14.56} &
  \multicolumn{1}{c|}{22.12} &
  21.76 &
  4.09 \\ \cmidrule{2-15}
 &
   &
  U-Net &
  \multicolumn{1}{c|}{34.11} &
  \multicolumn{1}{c|}{28.90} &
  \multicolumn{1}{c|}{90.61} &
  \multicolumn{1}{c|}{70.71} &
  \multicolumn{1}{c|}{39.67} &
  \multicolumn{1}{c|}{22.42} &
  \multicolumn{1}{c|}{8.65} &
  \multicolumn{1}{c|}{33.52} &
  \multicolumn{1}{c|}{26.88} &
  \multicolumn{1}{c|}{34.56} &
  32.91 &
  2.91 \\
 &
   &
  FPN &
  \multicolumn{1}{c|}{41.92} &
  \multicolumn{1}{c|}{38.74} &
  \multicolumn{1}{c|}{88.21} &
  \multicolumn{1}{c|}{55.71} &
  \multicolumn{1}{c|}{61.16} &
  \multicolumn{1}{c|}{21.32} &
  \multicolumn{1}{c|}{9.96} &
  \multicolumn{1}{c|}{32.40} &
  \multicolumn{1}{c|}{26.41} &
  \multicolumn{1}{c|}{30.88} &
  35.99 &
  4.14 \\
 &
   &
  DeepLab &
  \multicolumn{1}{c|}{40.98} &
  \multicolumn{1}{c|}{38.53} &
  \multicolumn{1}{c|}{91.62} &
  \multicolumn{1}{c|}{61.08} &
  \multicolumn{1}{c|}{62.69} &
  \multicolumn{1}{c|}{24.85} &
  \multicolumn{1}{c|}{10.65} &
  \multicolumn{1}{c|}{35.97} &
  \multicolumn{1}{c|}{30.69} &
  \multicolumn{1}{c|}{36.27} &
  37.82 &
  3.99 \\
 &
  \multirow{-4}{*}{M} &
  MA-Net &
  \multicolumn{1}{c|}{42.45} &
  \multicolumn{1}{c|}{42.44} &
  \multicolumn{1}{c|}{92.79} &
  \multicolumn{1}{c|}{49.20} &
  \multicolumn{1}{c|}{66.00} &
  \multicolumn{1}{c|}{27.91} &
  \multicolumn{1}{c|}{9.89} &
  \multicolumn{1}{c|}{40.63} &
  \multicolumn{1}{c|}{32.88} &
  \multicolumn{1}{c|}{38.73} &
  44.82 &
  3.39 \\ \cmidrule{2-15}
 &
  \cellcolor[HTML]{FBB982} &
  U-Net &
  \multicolumn{1}{c|}{55.82} &
  \multicolumn{1}{c|}{58.25} &
  \multicolumn{1}{c|}{87.79} &
  \multicolumn{1}{c|}{85.57} &
  \multicolumn{1}{c|}{67.06} &
  \multicolumn{1}{c|}{46.60} &
  \multicolumn{1}{c|}{25.31} &
  \multicolumn{1}{c|}{58.41} &
  \multicolumn{1}{c|}{54.38} &
  \multicolumn{1}{c|}{61.85} &
  63.82 &
  3.79 \\
 &
  \cellcolor[HTML]{FBB982} &
  FPN &
  \multicolumn{1}{c|}{61.67} &
  \multicolumn{1}{c|}{61.98} &
  \multicolumn{1}{c|}{88.45} &
  \multicolumn{1}{c|}{77.24} &
  \multicolumn{1}{c|}{78.36} &
  \multicolumn{1}{c|}{50.62} &
  \multicolumn{1}{c|}{22.26} &
  \multicolumn{1}{c|}{66.47} &
  \multicolumn{1}{c|}{57.63} &
  \multicolumn{1}{c|}{64.3} &
  75.84 &
  5.09 \\
 &
  \cellcolor[HTML]{FBB982} &
  DeepLab &
  \multicolumn{1}{c|}{62.52} &
  \multicolumn{1}{c|}{63.98} &
  \multicolumn{1}{c|}{88.98} &
  \multicolumn{1}{c|}{76.67} &
  \multicolumn{1}{c|}{82.22} &
  \multicolumn{1}{c|}{49.71} &
  \multicolumn{1}{c|}{23.42} &
  \multicolumn{1}{c|}{62.77} &
  \multicolumn{1}{c|}{57.62} &
  \multicolumn{1}{c|}{64.56} &
  70.18 &
  4.28 \\
\multirow{-12}{*}{L} &
  \multirow{-4}{*}{\cellcolor[HTML]{FBB982}L} &
  MA-Net &
  \multicolumn{1}{c|}{64.51} &
  \multicolumn{1}{c|}{68.85} &
  \multicolumn{1}{c|}{91.13} &
  \multicolumn{1}{c|}{80.00} &
  \multicolumn{1}{c|}{84.16} &
  \multicolumn{1}{c|}{50.94} &
  \multicolumn{1}{c|}{21.08} &
  \multicolumn{1}{c|}{68.08} &
  \multicolumn{1}{c|}{55.75} &
  \multicolumn{1}{c|}{62.40} &
  72.97 &
  3.91 \\ \bottomrule
\end{tabular}
}
\caption{Results for BM-2 on $\mathcal{D}_{id}$ (\textit{highlighted}) and $\mathcal{D}_{ood}$ for single-size category $\mathcal{T}_{p}$.}
\label{tab:bm_2_single_negative_all_results}
\end{table*}

\begin{table*}[!ht]
\setlength\tabcolsep{4.0pt}
\setlength{\belowcaptionskip}{-10pt}
\centering
\footnotesize
\resizebox{0.96\linewidth}{!}{
\begin{tabular}{c|c|c|ccccccccccc|c}
\toprule[1.5pt]
\multirow{2}{*}{\textbf{\begin{tabular}[c]{@{}c@{}}Training\\Category\end{tabular}}} &
  \multirow{2}{*}{\textbf{\begin{tabular}[c]{@{}c@{}}Testing\\Category\end{tabular}}} &
  \multirow{2}{*}{\textbf{\begin{tabular}[c]{@{}c@{}}Training\\Model\end{tabular}}} &
  \multicolumn{11}{c|}{\textbf{Cone}} &
  \textbf{Non-cone} \\ \cmidrule{4-15} 
 &
   &
   &
  \multicolumn{1}{c|}{\textbf{\begin{tabular}[c]{@{}c@{}}Mask\\IoU\end{tabular}}} &
  \multicolumn{1}{c|}{\textbf{\begin{tabular}[c]{@{}c@{}}Pixel\\IoU\end{tabular}}} &
  \multicolumn{1}{c|}{\textbf{\begin{tabular}[c]{@{}c@{}}Pixel\\Accuracy\end{tabular}}} &
  \multicolumn{1}{c|}{\textbf{\begin{tabular}[c]{@{}c@{}}Pixel\\Precision\end{tabular}}} &
  \multicolumn{1}{c|}{\textbf{\begin{tabular}[c]{@{}c@{}}Pixel\\Recall\end{tabular}}} &
  \multicolumn{1}{c|}{\textbf{\begin{tabular}[c]{@{}c@{}}Panoptic\\Quality\end{tabular}}} &
  \multicolumn{1}{c|}{\textbf{mAP}} &
  \multicolumn{1}{c|}{\textbf{\begin{tabular}[c]{@{}c@{}}Object\\IoU\end{tabular}}} &
  \multicolumn{1}{c|}{\textbf{\begin{tabular}[c]{@{}c@{}}Object\\Accuracy\end{tabular}}} &
  \multicolumn{1}{c|}{\textbf{\begin{tabular}[c]{@{}c@{}}Object\\Precision\end{tabular}}} &
  \textbf{\begin{tabular}[c]{@{}c@{}}Object\\ Recall\end{tabular}} &
  \textbf{$A_{FP}$} \\ \midrule[1pt]
 &
  \cellcolor[HTML]{FBB982} &
  U-Net &
  \multicolumn{1}{c|}{32.65} &
  \multicolumn{1}{c|}{32.54} &
  \multicolumn{1}{c|}{96.84} &
  \multicolumn{1}{c|}{66.34} &
  \multicolumn{1}{c|}{45.6}0 &
  \multicolumn{1}{c|}{23.83} &
  \multicolumn{1}{c|}{13.45} &
  \multicolumn{1}{c|}{37.43} &
  \multicolumn{1}{c|}{28.21} &
  \multicolumn{1}{c|}{36.64} &
  34.33 &
  1.41 \\
 &
  \cellcolor[HTML]{FBB982} &
  FPN &
  \multicolumn{1}{c|}{21.64} &
  \multicolumn{1}{c|}{19.95} &
  \multicolumn{1}{c|}{96.75} &
  \multicolumn{1}{c|}{77.80} &
  \multicolumn{1}{c|}{23.45} &
  \multicolumn{1}{c|}{11.08} &
  \multicolumn{1}{c|}{10.53} &
  \multicolumn{1}{c|}{21.85} &
  \multicolumn{1}{c|}{13.13} &
  \multicolumn{1}{c|}{19.79} &
  16.83 &
  0.46 \\
 &
  \cellcolor[HTML]{FBB982} &
  DeepLab &
  \multicolumn{1}{c|}{30.66} &
  \multicolumn{1}{c|}{31.66} &
  \multicolumn{1}{c|}{97.29} &
  \multicolumn{1}{c|}{71.83} &
  \multicolumn{1}{c|}{41.11} &
  \multicolumn{1}{c|}{21.60} &
  \multicolumn{1}{c|}{11.69} &
  \multicolumn{1}{c|}{34.84} &
  \multicolumn{1}{c|}{24.49} &
  \multicolumn{1}{c|}{30.59} &
  31.29 &
  1.14 \\
 &
  \multirow{-4}{*}{\cellcolor[HTML]{FBB982}S} &
  MA-Net &
  \multicolumn{1}{c|}{34.60} &
  \multicolumn{1}{c|}{35.01} &
  \multicolumn{1}{c|}{97.25} &
  \multicolumn{1}{c|}{63.54} &
  \multicolumn{1}{c|}{47.05} &
  \multicolumn{1}{c|}{22.53} &
  \multicolumn{1}{c|}{15.60} &
  \multicolumn{1}{c|}{36.01} &
  \multicolumn{1}{c|}{25.65} &
  \multicolumn{1}{c|}{30.99} &
  35.54 &
  1.49 \\\cmidrule{2-15}
 &
  \cellcolor[HTML]{FBB982} &
  U-Net &
  \multicolumn{1}{c|}{53.70} &
  \multicolumn{1}{c|}{54.59} &
  \multicolumn{1}{c|}{94.62} &
  \multicolumn{1}{c|}{70.96} &
  \multicolumn{1}{c|}{73.28} &
  \multicolumn{1}{c|}{38.61} &
  \multicolumn{1}{c|}{16.33} &
  \multicolumn{1}{c|}{59.33} &
  \multicolumn{1}{c|}{44.70} &
  \multicolumn{1}{c|}{51.06} &
  60.91 &
  4.28 \\
 &
  \cellcolor[HTML]{FBB982} &
  FPN &
  \multicolumn{1}{c|}{42.45} &
  \multicolumn{1}{c|}{39.61} &
  \multicolumn{1}{c|}{94.54} &
  \multicolumn{1}{c|}{90.53} &
  \multicolumn{1}{c|}{42.62} &
  \multicolumn{1}{c|}{27.66} &
  \multicolumn{1}{c|}{14.84} &
  \multicolumn{1}{c|}{43.06} &
  \multicolumn{1}{c|}{33.99} &
  \multicolumn{1}{c|}{44.23} &
  40.76 &
  1.01 \\
 &
  \cellcolor[HTML]{FBB982} &
  DeepLab &
  \multicolumn{1}{c|}{51.10} &
  \multicolumn{1}{c|}{51.70} &
  \multicolumn{1}{c|}{95.08} &
  \multicolumn{1}{c|}{77.11} &
  \multicolumn{1}{c|}{63.31} &
  \multicolumn{1}{c|}{37.44} &
  \multicolumn{1}{c|}{18.97} &
  \multicolumn{1}{c|}{53.76} &
  \multicolumn{1}{c|}{42.44} &
  \multicolumn{1}{c|}{47.81} &
  56.31 &
  2.67 \\
 &
  \multirow{-4}{*}{\cellcolor[HTML]{FBB982}M} &
  MA-Net &
  \multicolumn{1}{c|}{53.55} &
  \multicolumn{1}{c|}{55.13} &
  \multicolumn{1}{c|}{95.15} &
  \multicolumn{1}{c|}{77.39} &
  \multicolumn{1}{c|}{66.78} &
  \multicolumn{1}{c|}{37.15} &
  \multicolumn{1}{c|}{16.66} &
  \multicolumn{1}{c|}{56.83} &
  \multicolumn{1}{c|}{40.86} &
  \multicolumn{1}{c|}{45.69} &
  59.90 &
  2.74 \\\cmidrule{2-15}
 &
   &
  U-Net &
  \multicolumn{1}{c|}{46.93} &
  \multicolumn{1}{c|}{45.05} &
  \multicolumn{1}{c|}{87.50} &
  \multicolumn{1}{c|}{82.83} &
  \multicolumn{1}{c|}{48.95} &
  \multicolumn{1}{c|}{37.00} &
  \multicolumn{1}{c|}{17.43} &
  \multicolumn{1}{c|}{51.50} &
  \multicolumn{1}{c|}{41.86} &
  \multicolumn{1}{c|}{52.65} &
  51.89 &
  0.77 \\
 &
   &
  FPN &
  \multicolumn{1}{c|}{33.00} &
  \multicolumn{1}{c|}{31.85} &
  \multicolumn{1}{c|}{84.94} &
  \multicolumn{1}{c|}{89.20} &
  \multicolumn{1}{c|}{32.55} &
  \multicolumn{1}{c|}{25.21} &
  \multicolumn{1}{c|}{11.35} &
  \multicolumn{1}{c|}{34.09} &
  \multicolumn{1}{c|}{32.58} &
  \multicolumn{1}{c|}{39.02} &
  37.88 &
  0.40 \\
 &
   &
  DeepLab &
  \multicolumn{1}{c|}{35.72} &
  \multicolumn{1}{c|}{38.63} &
  \multicolumn{1}{c|}{85.91} &
  \multicolumn{1}{c|}{83.11} &
  \multicolumn{1}{c|}{41.14} &
  \multicolumn{1}{c|}{27.82} &
  \multicolumn{1}{c|}{9.82} &
  \multicolumn{1}{c|}{42.09} &
  \multicolumn{1}{c|}{31.69} &
  \multicolumn{1}{c|}{33.00} &
  43.56 &
  0.75 \\
\multirow{-12}{*}{S + M} &
  \multirow{-4}{*}{L} &
  MA-Net &
  \multicolumn{1}{c|}{51.80} &
  \multicolumn{1}{c|}{54.29} &
  \multicolumn{1}{c|}{88.42} &
  \multicolumn{1}{c|}{89.20} &
  \multicolumn{1}{c|}{59.73} &
  \multicolumn{1}{c|}{38.23} &
  \multicolumn{1}{c|}{19.24} &
  \multicolumn{1}{c|}{56.47} &
  \multicolumn{1}{c|}{41.01} &
  \multicolumn{1}{c|}{44.67} &
  58.33 &
  0.97 \\ \midrule
 &
   &
  U-Net &
  \multicolumn{1}{c|}{29.02} &
  \multicolumn{1}{c|}{26.06} &
  \multicolumn{1}{c|}{95.54} &
  \multicolumn{1}{c|}{71.13} &
  \multicolumn{1}{c|}{35.95} &
  \multicolumn{1}{c|}{18.29} &
  \multicolumn{1}{c|}{8.87} &
  \multicolumn{1}{c|}{30.77} &
  \multicolumn{1}{c|}{20.82} &
  \multicolumn{1}{c|}{28.55} &
  26.25 &
  2.32 \\
 &
   &
  FPN &
  \multicolumn{1}{c|}{21.94} &
  \multicolumn{1}{c|}{19.10} &
  \multicolumn{1}{c|}{95.48} &
  \multicolumn{1}{c|}{62.98} &
  \multicolumn{1}{c|}{28.58} &
  \multicolumn{1}{c|}{11.27} &
  \multicolumn{1}{c|}{7.25} &
  \multicolumn{1}{c|}{21.26} &
  \multicolumn{1}{c|}{12.54} &
  \multicolumn{1}{c|}{18.91} &
  17.44 &
  1.65 \\
 &
   &
  DeepLab &
  \multicolumn{1}{c|}{35.51} &
  \multicolumn{1}{c|}{29.91} &
  \multicolumn{1}{c|}{95.20} &
  \multicolumn{1}{c|}{65.54} &
  \multicolumn{1}{c|}{50.61} &
  \multicolumn{1}{c|}{23.22} &
  \multicolumn{1}{c|}{11.11} &
  \multicolumn{1}{c|}{41.13} &
  \multicolumn{1}{c|}{25.57} &
  \multicolumn{1}{c|}{33.43} &
  38.11 &
  2.45 \\
 &
  \multirow{-4}{*}{S} &
  MA-Net &
  \multicolumn{1}{c|}{27.86} &
  \multicolumn{1}{c|}{28.72} &
  \multicolumn{1}{c|}{94.69} &
  \multicolumn{1}{c|}{46.06} &
  \multicolumn{1}{c|}{46.68} &
  \multicolumn{1}{c|}{16.77} &
  \multicolumn{1}{c|}{7.75} &
  \multicolumn{1}{c|}{31.23} &
  \multicolumn{1}{c|}{19.79} &
  \multicolumn{1}{c|}{23.85} &
  28.37 &
  2.25 \\\cmidrule{2-15}
 &
  \cellcolor[HTML]{FBB982} &
  U-Net &
  \multicolumn{1}{c|}{52.22} &
  \multicolumn{1}{c|}{52.67} &
  \multicolumn{1}{c|}{94.76} &
  \multicolumn{1}{c|}{79.65} &
  \multicolumn{1}{c|}{64.14} &
  \multicolumn{1}{c|}{38.98} &
  \multicolumn{1}{c|}{18.20} &
  \multicolumn{1}{c|}{56.44} &
  \multicolumn{1}{c|}{46.22} &
  \multicolumn{1}{c|}{56.26} &
  54.61 &
  1.99 \\
 &
  \cellcolor[HTML]{FBB982} &
  FPN &
  \multicolumn{1}{c|}{45.55} &
  \multicolumn{1}{c|}{43.95} &
  \multicolumn{1}{c|}{94.32} &
  \multicolumn{1}{c|}{81.91} &
  \multicolumn{1}{c|}{53.34} &
  \multicolumn{1}{c|}{31.62} &
  \multicolumn{1}{c|}{14.57} &
  \multicolumn{1}{c|}{47.73} &
  \multicolumn{1}{c|}{38.69} &
  \multicolumn{1}{c|}{47.52} &
  47.60 &
  1.15 \\
 &
  \cellcolor[HTML]{FBB982} &
  DeepLab &
  \multicolumn{1}{c|}{55.40} &
  \multicolumn{1}{c|}{55.62} &
  \multicolumn{1}{c|}{94.60} &
  \multicolumn{1}{c|}{71.99} &
  \multicolumn{1}{c|}{70.30} &
  \multicolumn{1}{c|}{43.07} &
  \multicolumn{1}{c|}{17.08} &
  \multicolumn{1}{c|}{60.28} &
  \multicolumn{1}{c|}{50.14} &
  \multicolumn{1}{c|}{56.14} &
  63.17 &
  2.09 \\
 &
  \multirow{-4}{*}{\cellcolor[HTML]{FBB982}M} &
  MA-Net &
  \multicolumn{1}{c|}{49.59} &
  \multicolumn{1}{c|}{50.54} &
  \multicolumn{1}{c|}{94.31} &
  \multicolumn{1}{c|}{76.50} &
  \multicolumn{1}{c|}{64.11} &
  \multicolumn{1}{c|}{36.45} &
  \multicolumn{1}{c|}{18.05} &
  \multicolumn{1}{c|}{54.67} &
  \multicolumn{1}{c|}{42.68} &
  \multicolumn{1}{c|}{49.39} &
  55.70 &
  1.87 \\\cmidrule{2-15}
 &
  \cellcolor[HTML]{FBB982} &
  U-Net &
  \multicolumn{1}{c|}{61.63} &
  \multicolumn{1}{c|}{62.74} &
  \multicolumn{1}{c|}{90.89} &
  \multicolumn{1}{c|}{86.69} &
  \multicolumn{1}{c|}{70.60} &
  \multicolumn{1}{c|}{50.47} &
  \multicolumn{1}{c|}{27.10} &
  \multicolumn{1}{c|}{61.61} &
  \multicolumn{1}{c|}{57.98} &
  \multicolumn{1}{c|}{64.60} &
  67.00 &
  3.61 \\
 &
  \cellcolor[HTML]{FBB982} &
  FPN &
  \multicolumn{1}{c|}{51.84} &
  \multicolumn{1}{c|}{54.54} &
  \multicolumn{1}{c|}{89.00} &
  \multicolumn{1}{c|}{88.06} &
  \multicolumn{1}{c|}{59.12} &
  \multicolumn{1}{c|}{44.69} &
  \multicolumn{1}{c|}{22.92} &
  \multicolumn{1}{c|}{58.17} &
  \multicolumn{1}{c|}{53.63} &
  \multicolumn{1}{c|}{59.45} &
  66.02 &
  2.80 \\
 &
  \cellcolor[HTML]{FBB982} &
  DeepLab &
  \multicolumn{1}{c|}{64.36} &
  \multicolumn{1}{c|}{67.11} &
  \multicolumn{1}{c|}{90.97} &
  \multicolumn{1}{c|}{83.98} &
  \multicolumn{1}{c|}{77.64} &
  \multicolumn{1}{c|}{51.65} &
  \multicolumn{1}{c|}{19.56} &
  \multicolumn{1}{c|}{67.81} &
  \multicolumn{1}{c|}{55.80} &
  \multicolumn{1}{c|}{60.63} &
  72.80 &
  4.69 \\
\multirow{-12}{*}{M + L} &
  \multirow{-4}{*}{\cellcolor[HTML]{FBB982}L} &
  MA-Net &
  \multicolumn{1}{c|}{60.98} &
  \multicolumn{1}{c|}{63.59} &
  \multicolumn{1}{c|}{90.44} &
  \multicolumn{1}{c|}{83.96} &
  \multicolumn{1}{c|}{73.04} &
  \multicolumn{1}{c|}{49.26} &
  \multicolumn{1}{c|}{23.85} &
  \multicolumn{1}{c|}{65.65} &
  \multicolumn{1}{c|}{54.97} &
  \multicolumn{1}{c|}{60.66} &
  74.34 &
  4.14 \\ \midrule
 &
  \cellcolor[HTML]{FBB982} &
  U-Net &
  \multicolumn{1}{c|}{31.94} &
  \multicolumn{1}{c|}{30.27} &
  \multicolumn{1}{c|}{95.95} &
  \multicolumn{1}{c|}{59.95} &
  \multicolumn{1}{c|}{44.87} &
  \multicolumn{1}{c|}{20.85} &
  \multicolumn{1}{c|}{12.19} &
  \multicolumn{1}{c|}{36.11} &
  \multicolumn{1}{c|}{23.10} &
  \multicolumn{1}{c|}{29.22} &
  33.25 &
  2.82 \\
 &
  \cellcolor[HTML]{FBB982} &
  FPN &
  \multicolumn{1}{c|}{26.01} &
  \multicolumn{1}{c|}{24.83} &
  \multicolumn{1}{c|}{94.38} &
  \multicolumn{1}{c|}{61.55} &
  \multicolumn{1}{c|}{40.27} &
  \multicolumn{1}{c|}{14.62} &
  \multicolumn{1}{c|}{10.81} &
  \multicolumn{1}{c|}{25.11} &
  \multicolumn{1}{c|}{17.67} &
  \multicolumn{1}{c|}{25.29} &
  21.11 &
  4.06 \\
 &
  \cellcolor[HTML]{FBB982} &
  DeepLab &
  \multicolumn{1}{c|}{26.52} &
  \multicolumn{1}{c|}{26.52} &
  \multicolumn{1}{c|}{96.41} &
  \multicolumn{1}{c|}{68.21} &
  \multicolumn{1}{c|}{36.65} &
  \multicolumn{1}{c|}{15.41} &
  \multicolumn{1}{c|}{9.81} &
  \multicolumn{1}{c|}{28.44} &
  \multicolumn{1}{c|}{18.14} &
  \multicolumn{1}{c|}{26.22} &
  22.74 &
  2.54 \\
 &
  \multirow{-4}{*}{\cellcolor[HTML]{FBB982}S} &
  MA-Net &
  \multicolumn{1}{c|}{33.19} &
  \multicolumn{1}{c|}{32.33} &
  \multicolumn{1}{c|}{96.45} &
  \multicolumn{1}{c|}{57.88} &
  \multicolumn{1}{c|}{47.59} &
  \multicolumn{1}{c|}{21.23} &
  \multicolumn{1}{c|}{12.37} &
  \multicolumn{1}{c|}{34.69} &
  \multicolumn{1}{c|}{23.37} &
  \multicolumn{1}{c|}{29.59} &
  33.28 &
  2.37 \\\cmidrule{2-15}
 &
   &
  U-Net &
  \multicolumn{1}{c|}{52.29} &
  \multicolumn{1}{c|}{49.67} &
  \multicolumn{1}{c|}{96.17} &
  \multicolumn{1}{c|}{70.60} &
  \multicolumn{1}{c|}{63.37} &
  \multicolumn{1}{c|}{31.90} &
  \multicolumn{1}{c|}{15.00} &
  \multicolumn{1}{c|}{50.57} &
  \multicolumn{1}{c|}{34.40} &
  \multicolumn{1}{c|}{39.89} &
  58.12 &
  1.90 \\
 &
   &
  FPN &
  \multicolumn{1}{c|}{40.17} &
  \multicolumn{1}{c|}{39.37} &
  \multicolumn{1}{c|}{93.38} &
  \multicolumn{1}{c|}{65.25} &
  \multicolumn{1}{c|}{63.57} &
  \multicolumn{1}{c|}{23.86} &
  \multicolumn{1}{c|}{12.16} &
  \multicolumn{1}{c|}{34.95} &
  \multicolumn{1}{c|}{30.92} &
  \multicolumn{1}{c|}{37.25} &
  36.97 &
  3.49 \\
 &
   &
  DeepLab &
  \multicolumn{1}{c|}{39.68} &
  \multicolumn{1}{c|}{45.69} &
  \multicolumn{1}{c|}{96.20} &
  \multicolumn{1}{c|}{79.97} &
  \multicolumn{1}{c|}{56.90} &
  \multicolumn{1}{c|}{30.96} &
  \multicolumn{1}{c|}{12.51} &
  \multicolumn{1}{c|}{49.82} &
  \multicolumn{1}{c|}{35.98} &
  \multicolumn{1}{c|}{41.76} &
  52.66 &
  1.63 \\
 &
  \multirow{-4}{*}{M} &
  MA-Net &
  \multicolumn{1}{c|}{49.84} &
  \multicolumn{1}{c|}{49.07} &
  \multicolumn{1}{c|}{95.71} &
  \multicolumn{1}{c|}{72.53} &
  \multicolumn{1}{c|}{60.84} &
  \multicolumn{1}{c|}{31.44} &
  \multicolumn{1}{c|}{12.59} &
  \multicolumn{1}{c|}{52.68} &
  \multicolumn{1}{c|}{35.88} &
  \multicolumn{1}{c|}{42.35} &
  58.54 &
  2.23 \\\cmidrule{2-15}
 &
  \cellcolor[HTML]{FBB982} &
  U-Net &
  \multicolumn{1}{c|}{64.87} &
  \multicolumn{1}{c|}{66.61} &
  \multicolumn{1}{c|}{91.06} &
  \multicolumn{1}{c|}{82.20} &
  \multicolumn{1}{c|}{78.37} &
  \multicolumn{1}{c|}{50.93} &
  \multicolumn{1}{c|}{23.53} &
  \multicolumn{1}{c|}{69.71} &
  \multicolumn{1}{c|}{54.54} &
  \multicolumn{1}{c|}{58.90} &
  74.59 &
  5.95 \\
 &
  \cellcolor[HTML]{FBB982} &
  FPN &
  \multicolumn{1}{c|}{61.10} &
  \multicolumn{1}{c|}{63.40} &
  \multicolumn{1}{c|}{89.52} &
  \multicolumn{1}{c|}{78.31} &
  \multicolumn{1}{c|}{78.76} &
  \multicolumn{1}{c|}{47.69} &
  \multicolumn{1}{c|}{25.40} &
  \multicolumn{1}{c|}{58.50} &
  \multicolumn{1}{c|}{55.44} &
  \multicolumn{1}{c|}{62.77} &
  66.70 &
  8.78 \\
 &
  \cellcolor[HTML]{FBB982} &
  DeepLab &
  \multicolumn{1}{c|}{61.30} &
  \multicolumn{1}{c|}{64.56} &
  \multicolumn{1}{c|}{91.32} &
  \multicolumn{1}{c|}{85.28} &
  \multicolumn{1}{c|}{72.91} &
  \multicolumn{1}{c|}{49.41} &
  \multicolumn{1}{c|}{23.20} &
  \multicolumn{1}{c|}{65.23} &
  \multicolumn{1}{c|}{55.25} &
  \multicolumn{1}{c|}{60.36} &
  72.23 &
  4.65 \\
\multirow{-12}{*}{L + S} &
  \multirow{-4}{*}{\cellcolor[HTML]{FBB982}L} &
  MA-Net &
  \multicolumn{1}{c|}{63.26} &
  \multicolumn{1}{c|}{65.60} &
  \multicolumn{1}{c|}{90.90} &
  \multicolumn{1}{c|}{81.93} &
  \multicolumn{1}{c|}{78.18} &
  \multicolumn{1}{c|}{47.95} &
  \multicolumn{1}{c|}{19.35} &
  \multicolumn{1}{c|}{66.31} &
  \multicolumn{1}{c|}{51.76} &
  \multicolumn{1}{c|}{55.57} &
  75.93 &
  6.45 \\ \bottomrule
\end{tabular}
}
\caption{Results for BM-2 on $\mathcal{D}_{id}$ (\textit{highlighted}) and $\mathcal{D}_{ood}$ for multi-size category $\mathcal{T}_{p}$.}
\label{tab:bm_2_multi_negative_all_results}
\end{table*}

